\documentclass[10pt,twocolumn,letterpaper]{article}

\usepackage{iccv}
\usepackage{times}
\usepackage{epsfig}
\usepackage{graphicx}
\usepackage{amsmath}
\usepackage{amssymb}

\usepackage{multirow}
\usepackage{booktabs}
\usepackage{verbatim}
\usepackage{xcolor}
\usepackage{shortbold}
\usepackage{float}
\usepackage{algorithm}
\usepackage{algorithmicx}
\usepackage{algpseudocode}
\usepackage{subfig}
\usepackage{multirow, makecell}
\usepackage{stfloats}

\definecolor{AccessibleBlue}{rgb}{0.10196, 0.52157, 1.0}
\definecolor{AccessibleRed}{rgb}{0.8314, 0.0667, 0.3490}
\definecolor{BlueCam}{rgb}{0.5490196078431373, 0.7607843137254902, 0.9098039215686274}
\definecolor{RedCam}{rgb}{0.9019607843137255, 0.4470588235294118, 0.6039215686274509}
\definecolor{redbox}{rgb}{0.7098039215686275, 0.10196078431372549, 0.0039215686274509}

\DeclareMathOperator*{\argmin}{arg\,min}
\usepackage[pagebackref=true,breaklinks=true,letterpaper=true,colorlinks,bookmarks=false]{hyperref}

\iccvfinalcopy %

\def\iccvPaperID{4394} %

\makeatletter
\g@addto@macro\@maketitle{
\vspace{-2.5em}
\begin{figure}[H]
   \setlength{\linewidth}{\textwidth}
\setlength{\hsize}{\textwidth}
\centering
\includegraphics[width=\linewidth]{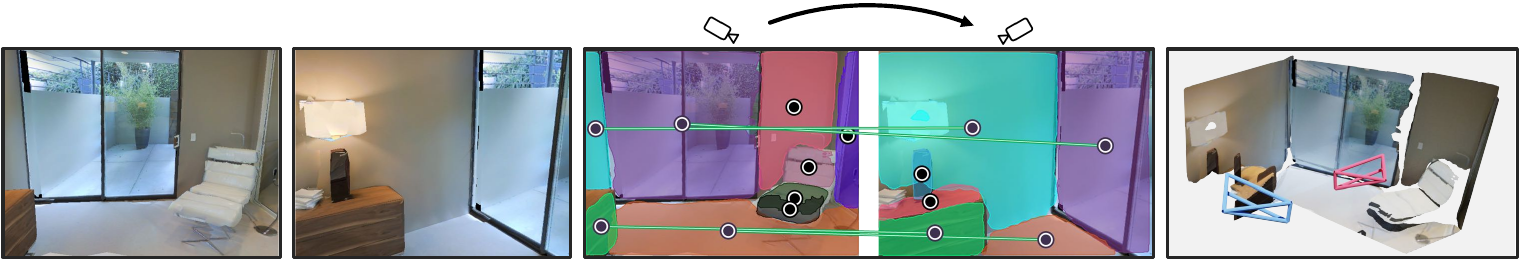}
\caption{Given two RGB images with an unknown relationship, our system produces a single, coherent planar surface reconstruction of the scene in terms
of 3D planes and relative camera poses. Our method succeeds despite the wide baseline and the relatively small amount of overlap in structure (here: $64^\circ$ rotation and $1.7\textrm{m}$ translation and
$27\%$ overlap).
We show this reconstruction 
with the inferred left and right cameras in \textbf{\textcolor{AccessibleBlue}{Blue}} and
\textbf{\textcolor{AccessibleRed}{Red}}.}
\label{fig:teaser}
\end{figure}
}
\makeatother

\ificcvfinal\fi

\begin{document}

\title{Planar Surface Reconstruction from Sparse Views}

\author{
Linyi Jin \kern15pt Shengyi Qian \kern15pt Andrew Owens \kern15pt David F. Fouhey\\
   University of Michigan\\
	{\tt\small \{jinlinyi,syqian,ahowens,fouhey\}@umich.edu}\\
   {\small \url{https://jinlinyi.github.io/SparsePlanes}}
}

\maketitle
\ificcvfinal\thispagestyle{empty}\fi

\begin{abstract}
    The paper studies planar surface reconstruction of indoor scenes from two views with unknown camera poses.
    While prior approaches have successfully created {\em object-centric} reconstructions of many scenes, 
    they fail to exploit other structures, such as planes, which are typically the dominant components of indoor scenes.
    In this paper, we reconstruct planar surfaces from multiple views, while jointly estimating camera pose. 
    Our experiments demonstrate that our method is able to advance the state of the art of reconstruction from sparse views, on challenging scenes from Matterport3D.
\end{abstract}

\vspace{-0.2in}
\section{Introduction}
\label{sec:introduction}
Consider the two photos in Figure~\ref{fig:teaser},
as humans, we can infer that they were taken from the same scene: there is a chair on one side, a bedside
table on the other, and a large glass wall and floor in the middle.
We perceive the scene correctly despite the fact that they are {\em sparse views}~\cite{Qian2020}: they were taken from
very different camera poses, and very little of the scene structure within them overlaps.
There are also challenges in grouping: while each photo might, at first glance, seem to contain its own glass wall and floor, they
are each ``slices'' of the same planar objects. Despite these challenges, humans readily understand spaces like these 
from only a few ordinary photos, such as when they share collections of photos from the same event or look for housing.

Yet this setting poses challenges for today's computer vision methods. Traditional tools from
multi-view geometry~\cite{Hartley04, bao2012semantic} largely rely on 
correspondence for reconstruction and are fundamentally limited to the small part of the scene that directly overlaps, even when the camera pose is known. 
Learning-based single view 3D~\cite{yang2018recovering,liu2019planercnn}, offers ways of reconstructing
each image, but produces two messy piles of partial reconstructions
due to the unknown viewpoint change. In Figure~\ref{fig:teaser}, the floor
is fragmented across both views and the back of the chair is present in only one view. While humans can
associate these pieces and infer their relative positions to produce a
{\it coherent} reconstruction, it is not trivial for today's reconstruction algorithms.

We believe the ease with which humans solve the two unknown camera
reconstruction problem, coupled with the difficulty it poses for computers
marks it as an important task on the path to human-level 3D perception.
Indeed, it poses challenges to
existing work in deep learning for multiview reconstruction, which typically requires known 
camera poses~\cite{kar2017learning,mildenhall2020nerf} as opposed to unknown poses, 
many views~\cite{huang2018deepmvs} as opposed to two, additional depth
information at test time~\cite{yang2020extreme,xiao2013sun3d} as opposed to RGB
images, or works only on synthetic data~\cite{Qian2020}. Typically the extra information used is fundamental
to the algorithm (e.g., using poses for triangulation) and cannot be removed to produce a method that
works in the two unknown view reconstruction settings.

We propose a learning-based approach that constructs a coherent 3D
reconstruction from two views with an unknown relationship. Our insight,
supported empirically, is that progress can be made by jointly tackling three related challenges:
per-view reconstruction, inter-view correspondence, and inter-view 6DOF (rotation and translation) pose. 
Throughout, we use plane segments as our representation since they
have simple parameters, are often good approximations~\cite{furukawa2009manhattan}, and there is a strong
line of work for estimating them ~\cite{liu2018planenet,liu2019planercnn} or related properties \cite{Fouhey13,Eigen15}
from images. 

Our method, described in Section~\ref{sec:approach}, combines a deep neural
network architecture and an optimization problem to jointly estimate
planes, their relationships, and camera transformations. Our architecture builds on PlaneRCNN~\cite{liu2019planercnn}
to produce, per-input, plane segments and parameters, per-plane embeddings for correspondences, as
well as a probability distribution over relative cameras. This information is used in
a discrete-continuous optimization problem (along with optional point features) to produce
a coherent reconstruction across views. This reasoning across views enables our approach to, for instance
in Figure~\ref{fig:teaser}, produce a single floor rather than a set of inconsistent floor fragments, and 
jointly infer the distance from the images to the scene
boundaries as well as the relative camera pose.

We validate our approach on realistic renderings from the
Matterport3D~\cite{chang2017matterport3d} dataset using 
pairs with limited overlap (average $53^\circ$ rotation, $2.3$m translation, $21\%$ overlap).
We report experimental results in Section~\ref{sec:experiments} for three tasks: producing a single
coherent reconstruction from the two views, matching planes across
views, and estimating the full 6DOF relative camera pose.
We compare extensively with a variety of baselines (e.g., adding an independent
network to estimate relative pose followed by fusion of the two scene
layouts) and ablations that test the contributions of our method and design
choices. Our results demonstrate the value of joint consideration of the
interrelated problems: our approach substantially outperforms the fusion of
existing approaches to the independent problems of camera pose estimation
and scene structure estimation.

\section{Related Work}
\label{sec:related}
The goal of this paper is to produce a coherent 3D plane surface reconstruction
of the scene given two images with an unknown relationship between the cameras.
To solve the problem, our approach needs to both reconstruct 3D
shapes from 2D images and establish correspondence and identify the relative camera
pose between views. Our work therefore touches on many topics
in 3D computer vision, ranging from single-view reconstruction to correspondence to 
relative camera view prediction to two-view stereo. 

Much of our signal comes from reconstruction methods that map 2D images
to 3D structure. This has long been a goal of computer vision with methods
that aim to extract normals~\cite{Eigen15,Wang15}, voxels~\cite{Choy20163d,Girdhar16b}, and depth~\cite{Eigen15,Ranftl2020}
from 2D images. Our approach builds most heavily on a
work aiming to produce a planar
reconstruction~\cite{liu2018planenet,yang2018recovering,liu2019planercnn,YuZLZG19,chen2020oasis,jiang2020peek}.
We build upon the work in this area, in particular PlaneRCNN \cite{liu2019planercnn}, but we use it to build a planar
reconstruction from two perspective images (i.e., what an ordinary person's
cell phone might capture casually). This focus on perspective images separates
our work from approaches that use panorama images~\cite{zou20193d, zou2018layoutnet, SunHSC19, Yang:2019:DuLa-Net}.

In the process, we find correspondence between planar regions of the scene.
Correspondence is, of course, one of the long-standing problems in computer
vision. A great deal of work aims to describe patch-based regions, ranging
from classic SIFT descriptors~\cite{lowe2004distinctive} to learned
descriptors~\cite{HardNet2017,BarrosoLaguna2019ICCV,sarlin2020superglue,dusmanu2019d2},
which are often paired with projective geometry~\cite{Hartley04}; other work
finds matches across object-level correspondence~\cite{cai2020messytable} or planar-level correspondence~\cite{raposo2016pi}.  
We see our work of jointly reconstructing and identifying correspondence as
complementary to this work; a core component is an embedding network that aims
to describe the plane, like~\cite{cai2020messytable} and the last component
of our system uses VIP-like features~\cite{wu20083d} to constrain 
plane parameters. Our approach, however, solves a superset of these problems,
since it produces a reconstruction as well.

We additionally predict the relative transformation between
the cameras. This has been studied as an output of correspondence methods
from both a classical and learning-based perspective. There are, 
however, methods that directly try to predict these transformations, including
by deep networks~\cite{Qian2020,en2018rpnet}, aligning RGBD data \cite{raposo2013plane,yang2020extreme,banani2020unsupervisedrr},
or learning to predict the fundamental matrix \cite{ranftl2018deep,poursaeed2018deep,ummenhofer2017demon}.
Like these approaches, we aim to estimate the relative transformation
between the images as a component (although unlike some, we assume only ordinary
RGB images); improving this component is complementary
to our goals.

The relationship between reconstruction, correspondence, and camera pose,
and the value of planes for inference has long been well-understood 
in the stereo community. While we use two views, these are well-separated
and unknown, so our approach is different compared to standard
two-view stereo~\cite{Scharstein02} or visual SLAM~\cite{raposo2016pi,whelan2013robust,schops2019bad} and
more similar to wide-baseline stereo~\cite{Pritchett98a,mishkin2015wxbs}.
Unlike wide-baseline stereo systems, however, we also produce reconstructions
for portions of the scene that are seen in only one camera. 
Nonetheless, we draw inspiration from works in this community that 
use planes as a useful unit of inference~\cite{sinha2009piecewise,furukawa2009manhattan,gallup2010piecewise,Hadfield15}.

The most similar work in this direction is Associative3D~\cite{Qian2020}, which
solves a related problem of reconstructing volumetric objects. Our
approach is inspired by ~\cite{Qian2020}, but overcomes several methodological
limitations, which we experimentally show.
Associative3D uses one network for detecting objects
and another for relative camera pose, and fuses the two
via a heuristic RANSAC-like scheme. While effective on a six-object
subset of the clean synthetic SUNCG dataset~\cite{song2017semantic},
these components fall short on the realistic Matterport3D
dataset~\cite{chang2017matterport3d} that our approach uses. Meeting the challenge
of handling with planar segments (which cover objects {\it and} layouts)
in realistic data requires a more principled optimization strategy and better
signal for relative camera estimation. As another benefit, our approach
has one backbone network forward pass per image.

\begin{figure*}[t]
   \centering
   \includegraphics[width=\textwidth]{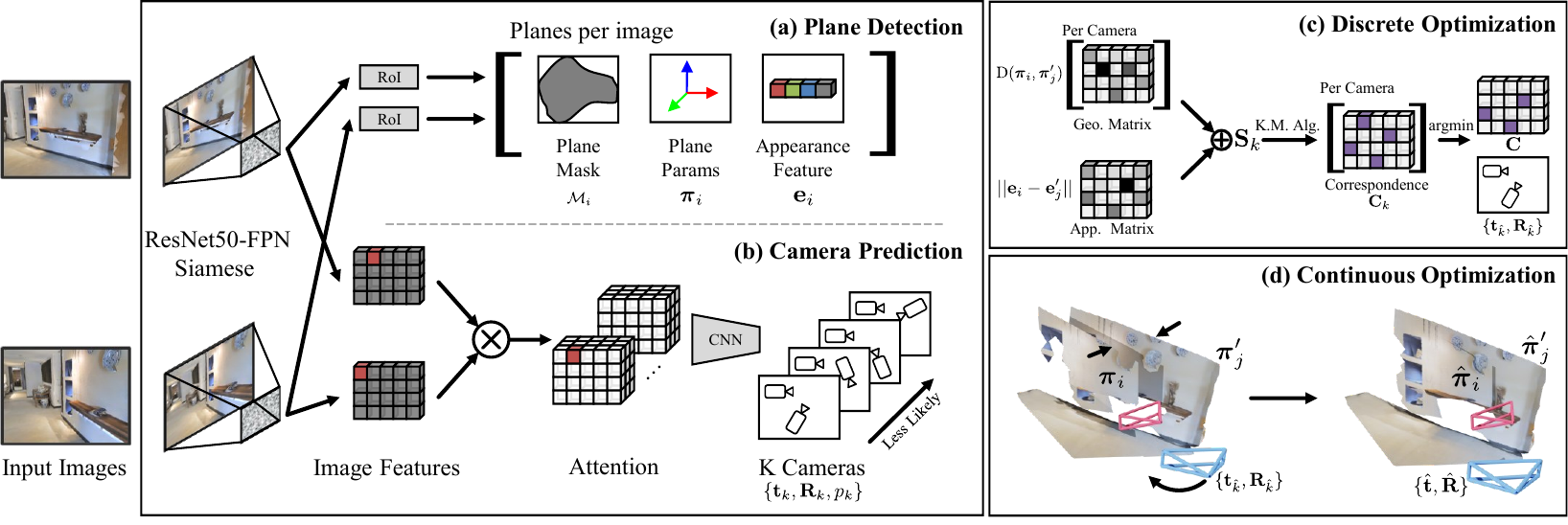}
   \caption{\textbf{Our Approach.} 
   Given a pair of images, we use a ResNet50-FPN to detect planes and predict probabilities of relative camera poses, 
   and use a two-step optimization to generate a coherent planar reconstruction. 
   (a) For each plane, we predict a segmentation mask, plane parameters, and an appearance feature. 
   (b) Concurrently, we pass image features from the detection backbone
   through the attention layer and predict the camera transformation between views. 
   (c) Our discrete optimization fuses the prediction of the separate heads to select 
   the best camera pose and plane correspondence.
   (d) Finally, we use continuous optimization to update the camera and plane parameters.}
   \label{fig:approach}
   \vspace{-0.1in}
\end{figure*}

\section{Approach}
\label{sec:approach}
Our approach, depicted in Figure~\ref{fig:approach}, aims to map
two images with an unknown relationship to a set of globally consistent planes and relative camera
poses. This task requires estimating plane parameters, reasoning about their
relationship (to avoid, e.g., a reconstruction with two floors), and inferring the
relative pose. These subproblems are related since, for instance, plane parameters and correspondence
constrain relative camera pose.
We approach this with a network that predicts parameters
and embeddings for planes (Sec.~\ref{sec:planeprediction}) and a distribution over relative
camera pose (Sec.~\ref{sec:camerapose}), followed by a joint optimization 
(Sec.~\ref{sec:optimization}) to produce a final coherent scene via joint reasoning. 
At training time, our system depends on RGBD for supervision, but can run on ordinary RGB images at test time.

\subsection{Plane Prediction Module}
\label{sec:planeprediction}
Our plane prediction module produces, per-image, a set
of plane segments that also have an embedding
for cross-view matching. Each plane has: a segment $\mathcal{M}_i$; 
plane parameters $\piB_i = [\nB_i, o_i]$ (where $\nB_i$ is a unit vector normal and $o_i$ the offset)
giving the plane equation $\piB_i^T [x,y,z,-1] = 0$;
and a unit-norm embedding $\eB_i$ for cross-view matching. 
Throughout, we denote
planes in view 2 as $\mathcal{M}_j', \piB_j', \eB_j'$, etc.

\par \noindent {\bf Plane detection.} We 
adopt PlaneRCNN \cite{liu2019planercnn} to detect planes in each image. 
During inference, the system produces backbone features using ResNet50-FPN~\cite{lin2017feature}. 
We use a region proposal network to propose boxes and then infer plane masks and normals from features
from RoIAlign~\cite{he2017mask}. At the same time, a decoder maps the backbone feature to a depthmap, which is
used to compute the plane offset.

\par \noindent {\bf Appearance embedding.} We additionally predict
a cross-view embedding by training the network on {\it pairs} of images via triplet loss. Given
plane correspondence, we form cross-view triplets $\eB_a, \eB_p', \eB_n'$ where 
{\bf a}nchor $\eB_a$ corresponds with the {\bf p}ositive match $\eB_p'$ and not the {\bf n}egative match
$\eB_n'$. We minimize a standard triplet loss~\cite{schroff2015facenet} 
$\max(||\eB_a-\eB_n'||_2 - ||\eB_a-\eB_p'||_2 + \alpha,0)$,
which gives a loss if the anchor and positive are not closer than the anchor and the negative by a margin $\alpha=0.2$.
We use online triplet mining and randomly pick negative matches with positive loss.

\subsection{Camera Pose Module}
\label{sec:camerapose}

Our camera pose module estimates a distribution over the relative camera pose
between views. This enables joint reasoning between a holistic 
estimate of the camera as well as the valuable cues in the estimated geometry:
plane correspondences and parameters can
constrain relative poses. We produce a distribution $p_k>0$, $\sum_k p_k =1$ over a set of discrete pairs of rotations and translations $\{(\tB_k, \RB_k)\}$).

Our network outputs two independent multinomial distributions over
translation and rotation using attention-style features. 
The joint translation/rotation distribution is then their product. 
Our attention features follow recent literature that 
find similarity between images using attention~\cite{wang2019learning, 
vaswani2017attention, rocco2018neighbourhood}
and capture, at each pixel in a feature map, the relative
similarity between that pixel and the pixels in the other image.
We use the backbone network to extract
$c$-dimensional feature maps $F_1, F_2$ of size
$c \times h_1 \times w_1$ and $c \times h_2 \times w_2$. Suppose $p_1$ and $p_2$ index pixels in feature $F_1$ and $F_2$ over both
rows and columns (i.e., $F_1(p_1)$ is a $c$-dimensional column), we
then compute the $(h_2w_2) \times (h_1 w_1)$ attended feature $A$:
\begin{equation}
A(p_1, p_2) =  \frac{\exp(F_2(p_2)^\intercal F_1(p_1))}{\Sigma_{p_2}\exp(F_2(p_2)^\intercal F_1(p_1))}.
\end{equation}
We reshape this to a $h_2w_2$-channel $h_1 \times w_1$ feature map where each pixel represents
the normalized correlation with each of the other pixels in feature map 2.
We apply a convolutional network to the reshaped $A$ with six layers of
$3\times3$ convolutions and two fully connected layers that predicts the camera
distribution. We found this attention to be superior compared to the
strategy of concatenating average-pooled vector outputs in \cite{Qian2020,en2018rpnet}.
as well as other alternate strategies in experiments in Section \ref{sec:experiments}.
We train the camera pose module on top of the ResNet50-FPN's $P_3$ feature~\cite{lin2017feature}.

\subsection{Optimization}
\label{sec:optimization}
Our final step is an optimization that serves two purposes. First, it
propagates information between the related problems: the parameters of a
mutually visible plane can inform relative camera poses, and a better pose of a
camera can help disambiguate matches. Second, it ensures the coherence: naively
concatenating two single-view parses of the scene (as we show empirically)
yields a collection of possibly overlapping, inconsistent planar fragments
showing the same object from different vantage points.

We cast this as an optimization
problem over camera-to-camera transformation $\hat{\RB}, \hat{\tB}$ and
set of plane parameters $\hat{\piB}_i, \hat{\piB}_j'$ as well as 
a plane correspondence matrix $\CB \in \{0,1\}^{m \times n}$ where
$\CB_{i,j}$ is $1$ if and only if plane $i$ corresponds to plane $j$.
As input, the optimization has
a series of $m$ planes for the view 1 $\{\mathcal{M}_i, \piB_i, \eB_i\}$ and 
$n$ planes for view 2 $\{\mathcal{M}_j', \piB_j', \eB_j'\}$, and a distribution over
relative camera transformations $\{\tB_k, \RB_k, p_k\}$. The optimization
tightly couples discrete and continuous variables and has many degenerate
solutions. We thus follow a two stage approach: we hold plane parameters
fixed and solve for plane correspondence and rough camera location; we then continuously
optimize camera and plane parameters.

\begin{figure*}[!t]
    \centering
    \scriptsize
    \begin{tabular}{c@{\hskip4pt}c@{\hskip4pt}c@{\hskip4pt}c@{\hskip4pt}c@{\hskip4pt}c}
    \toprule

    Image 1 & Image 2 & Prediction & Ground Truth  & Prediction & Ground Truth \\
    \midrule
    \frame{\includegraphics[width=0.15\textwidth]{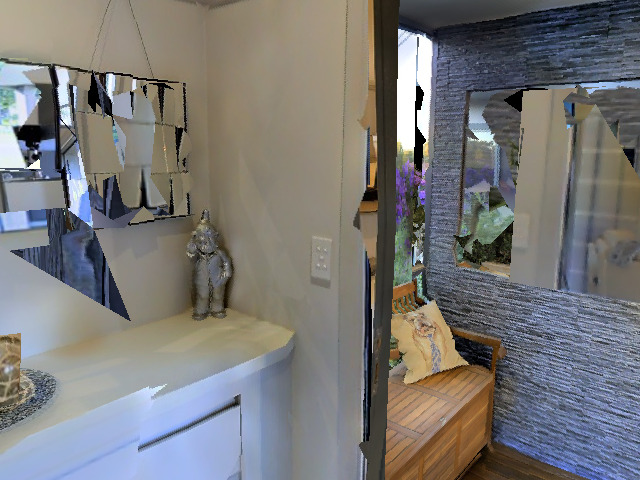}}
    & \frame{\includegraphics[width=0.15\textwidth]{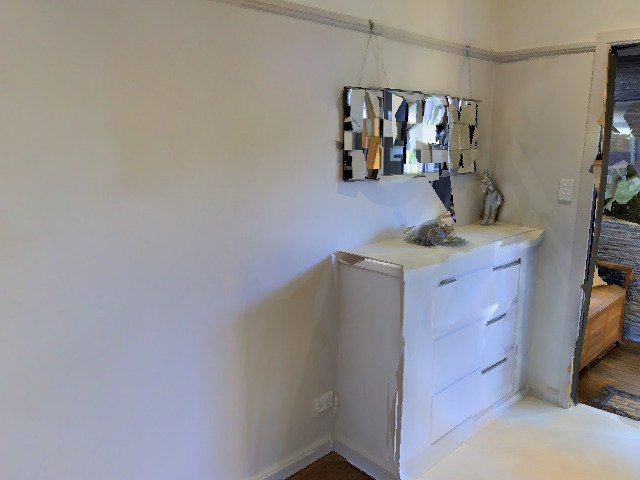}}
    & \frame{\includegraphics[width=0.15\textwidth]{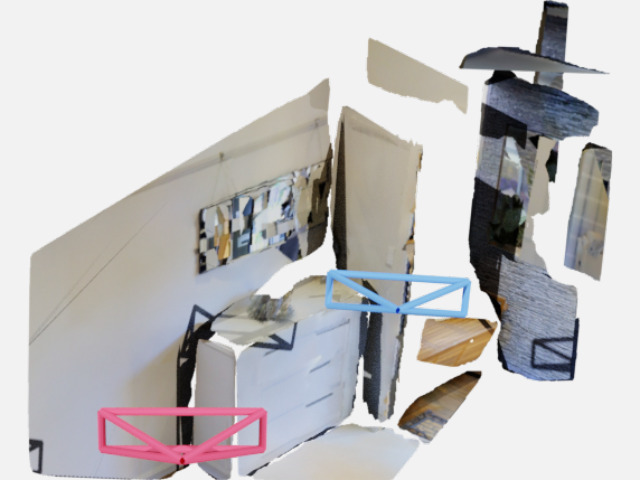}}
    & \frame{\includegraphics[width=0.15\textwidth]{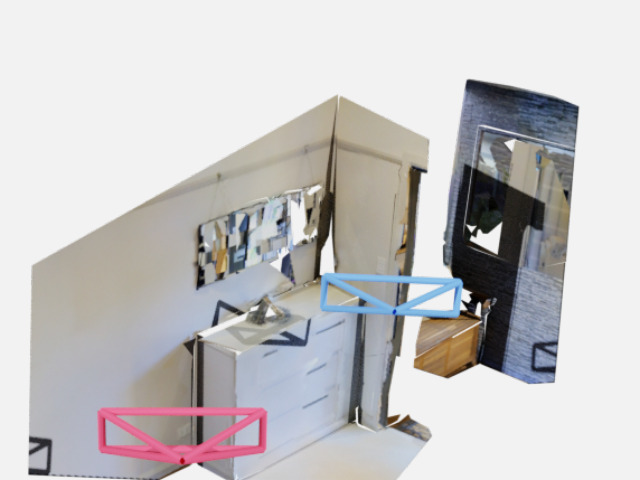}}
    & \frame{\includegraphics[width=0.15\textwidth]{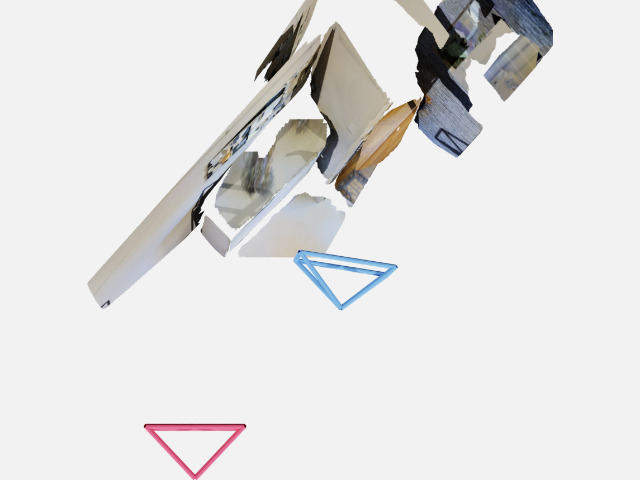}}
    & \frame{\includegraphics[width=0.15\textwidth]{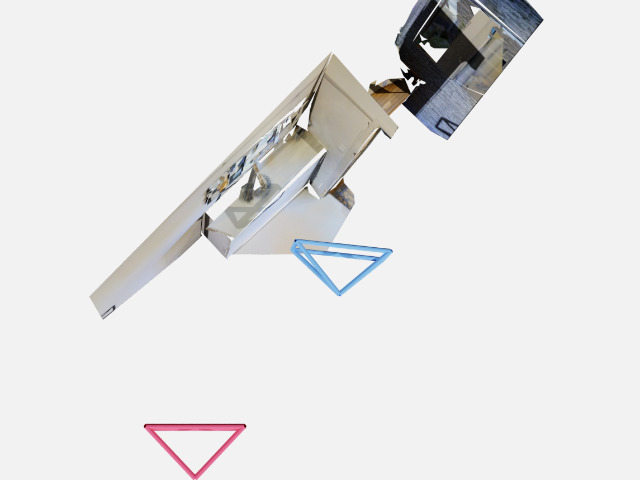}}\\

    \frame{\includegraphics[width=0.15\textwidth]{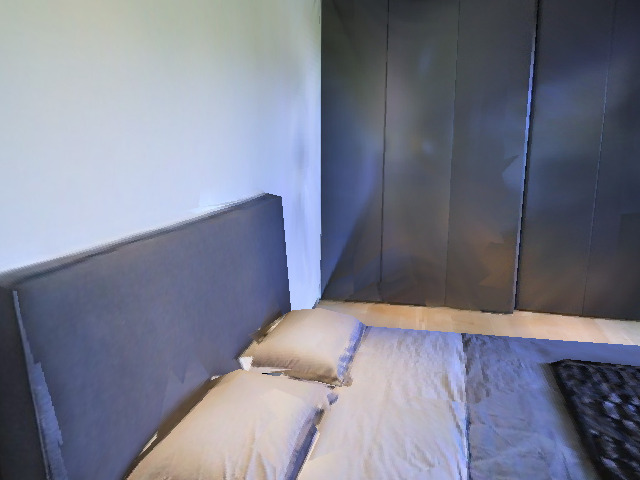}}
    & \frame{\includegraphics[width=0.15\textwidth]{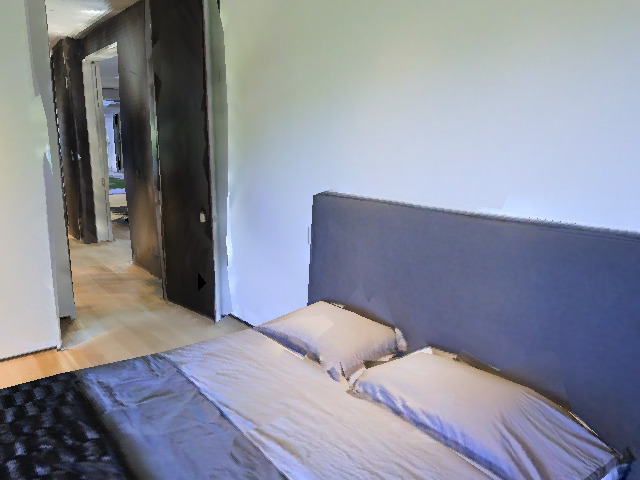}}
    & \frame{\includegraphics[width=0.15\textwidth]{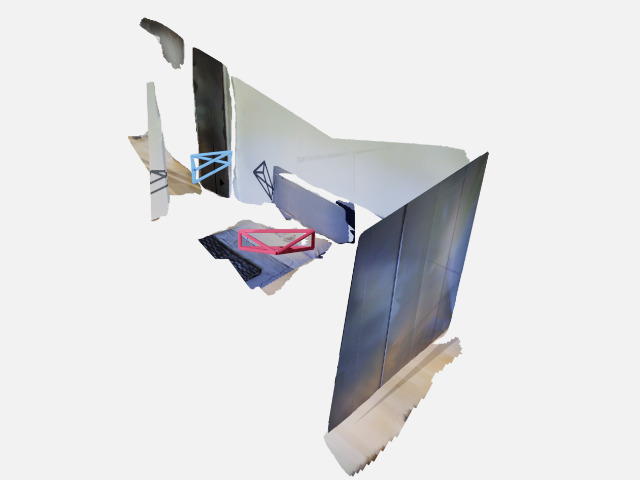}}
    & \frame{\includegraphics[width=0.15\textwidth]{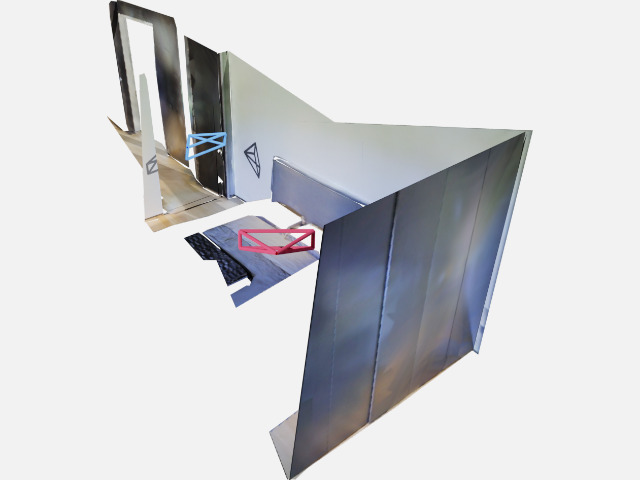}}
    & \frame{\includegraphics[width=0.15\textwidth]{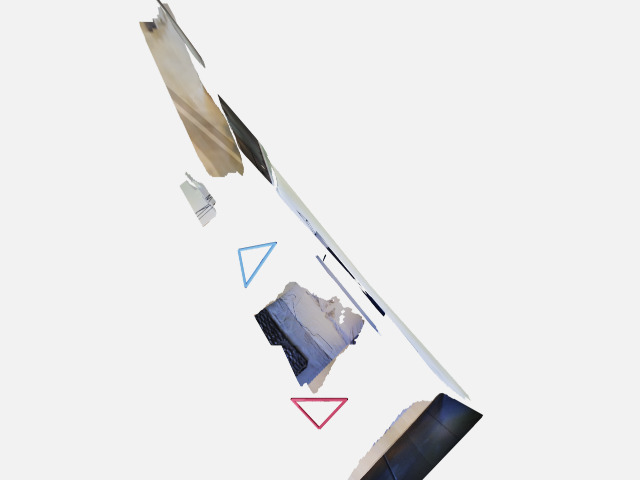}}
    & \frame{\includegraphics[width=0.15\textwidth]{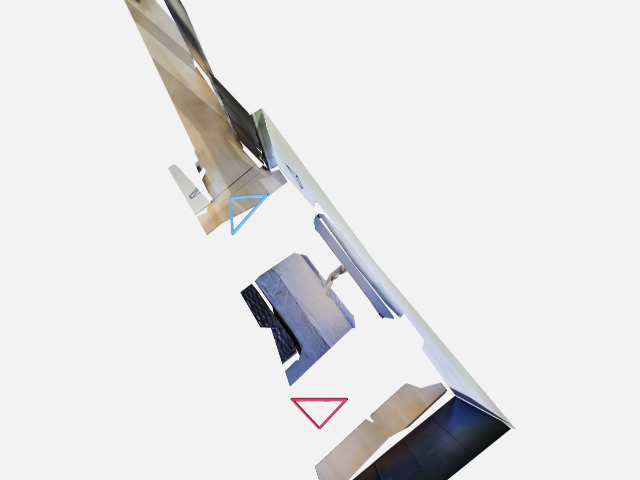}}\\

    \frame{\includegraphics[width=0.15\textwidth]{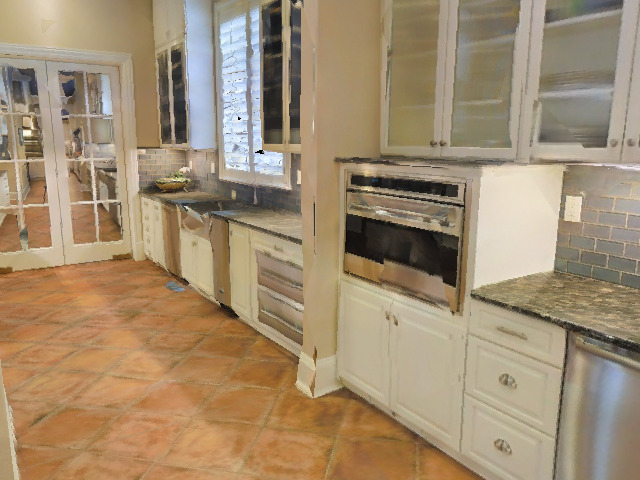}}
    & \frame{\includegraphics[width=0.15\textwidth]{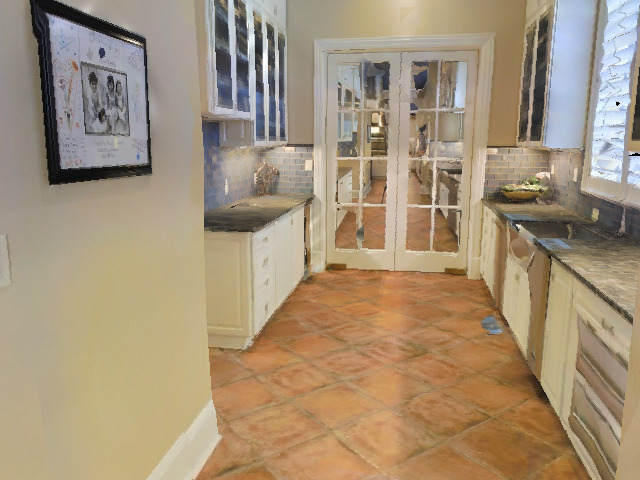}}
    & \frame{\includegraphics[width=0.15\textwidth]{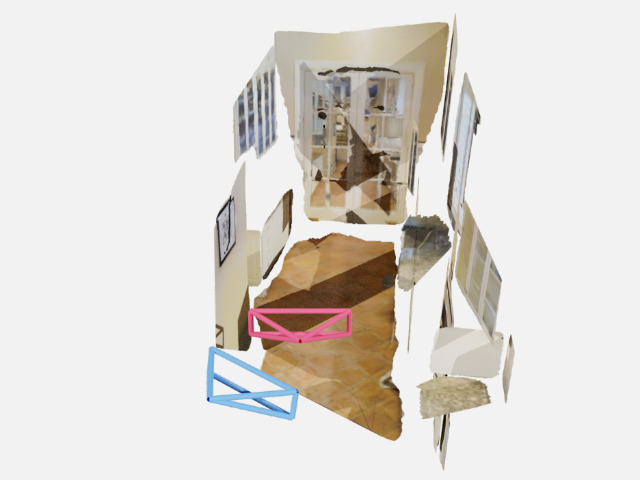}}
    & \frame{\includegraphics[width=0.15\textwidth]{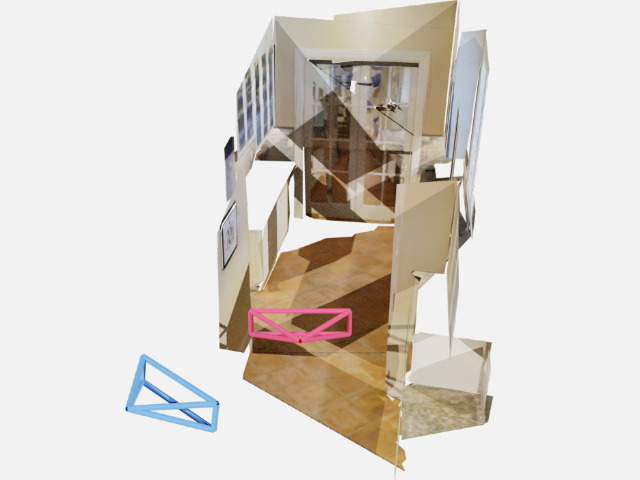}}
    & \frame{\includegraphics[width=0.15\textwidth]{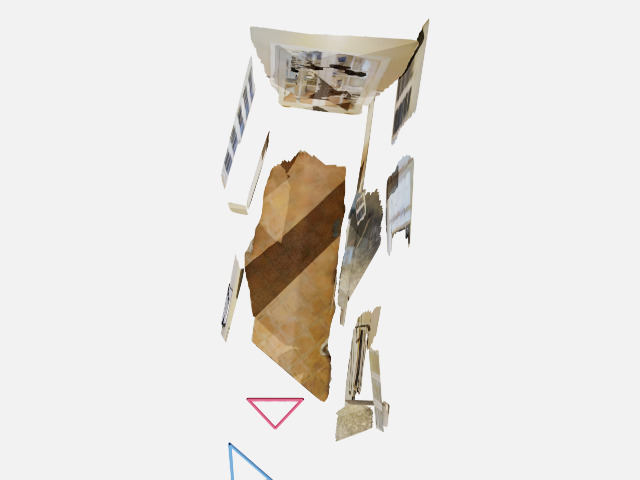}}
    & \frame{\includegraphics[width=0.15\textwidth]{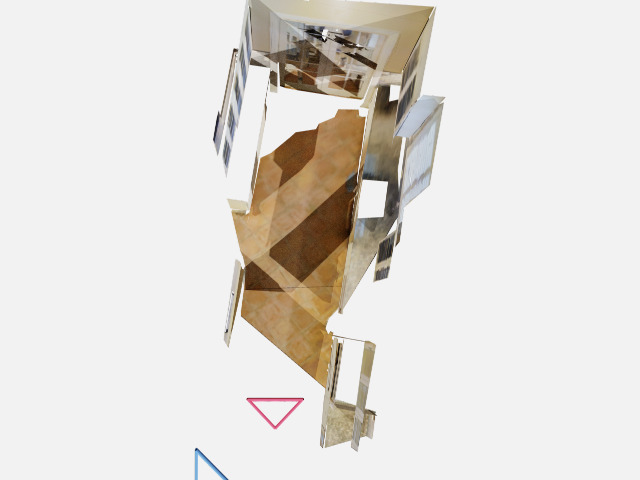}}\\

    \frame{\includegraphics[width=0.15\textwidth]{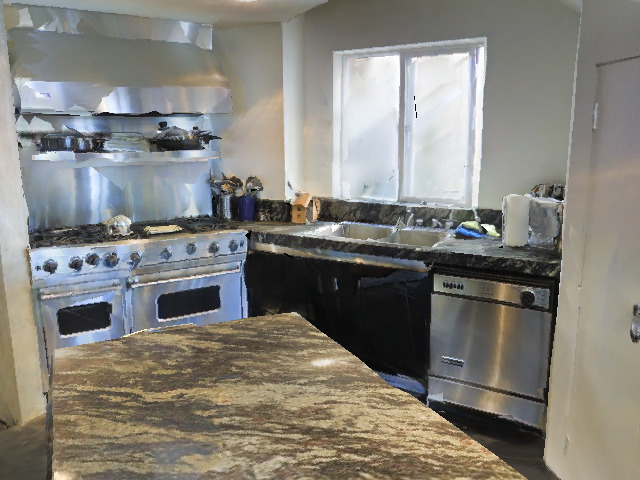}}
    & \frame{\includegraphics[width=0.15\textwidth]{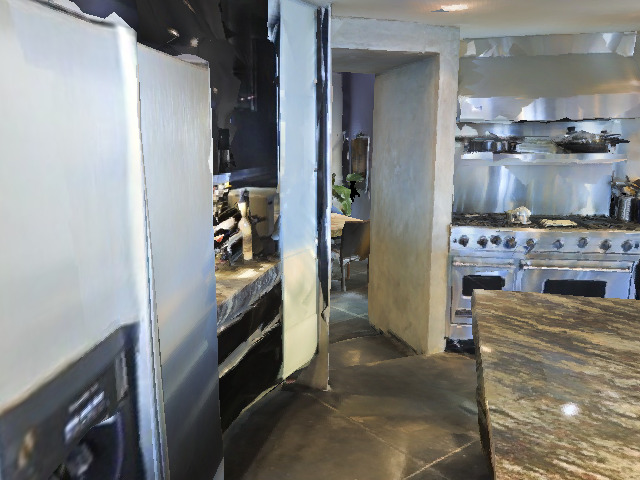}}
    & \frame{\includegraphics[width=0.15\textwidth]{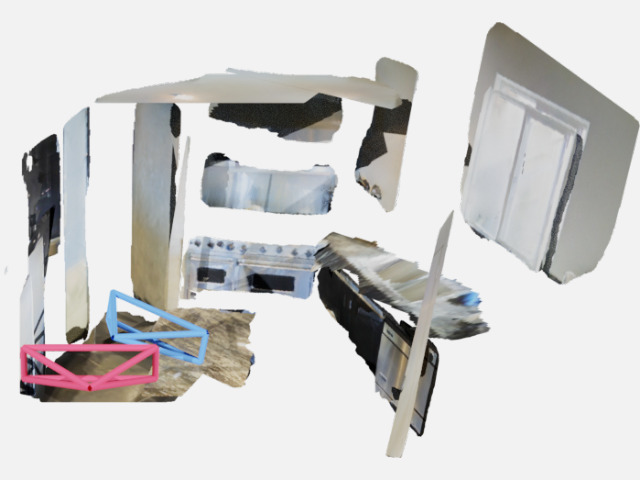}}
    & \frame{\includegraphics[width=0.15\textwidth]{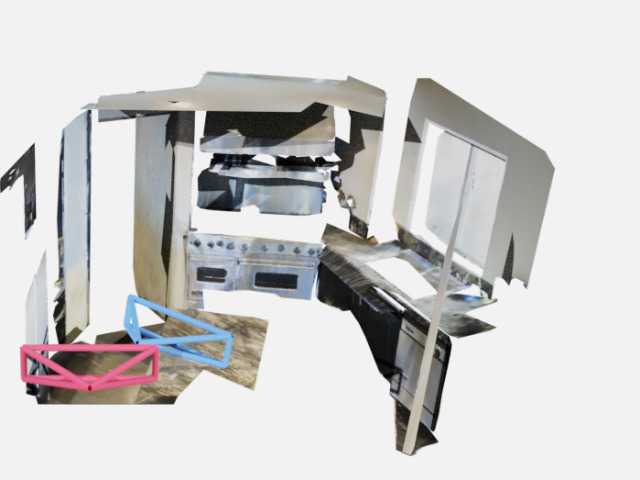}}
    & \frame{\includegraphics[width=0.15\textwidth]{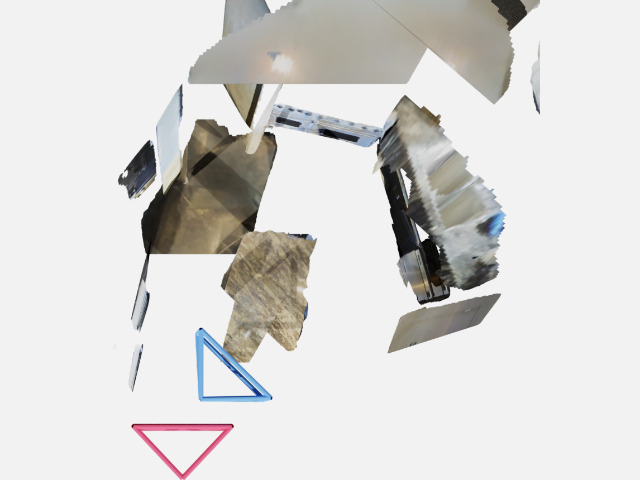}}
    & \frame{\includegraphics[width=0.15\textwidth]{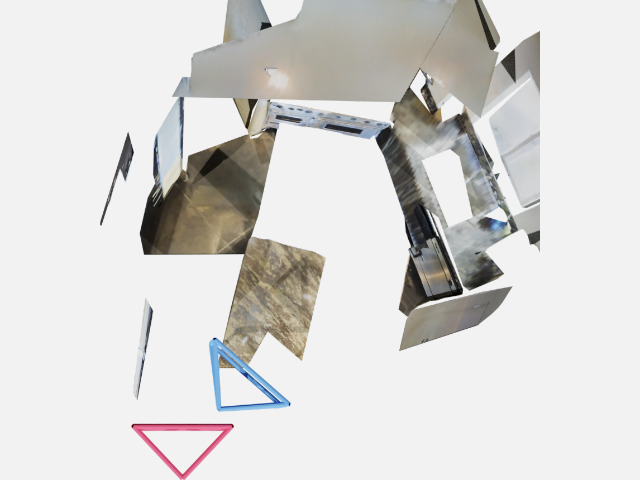}}\\

   \frame{\includegraphics[width=0.15\textwidth]{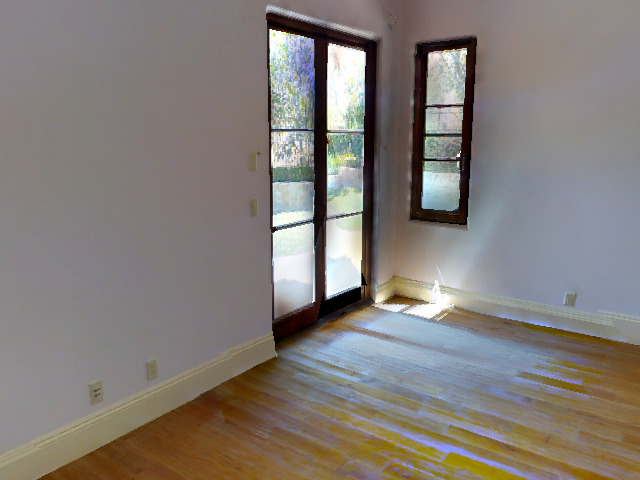}}
   & \frame{\includegraphics[width=0.15\textwidth]{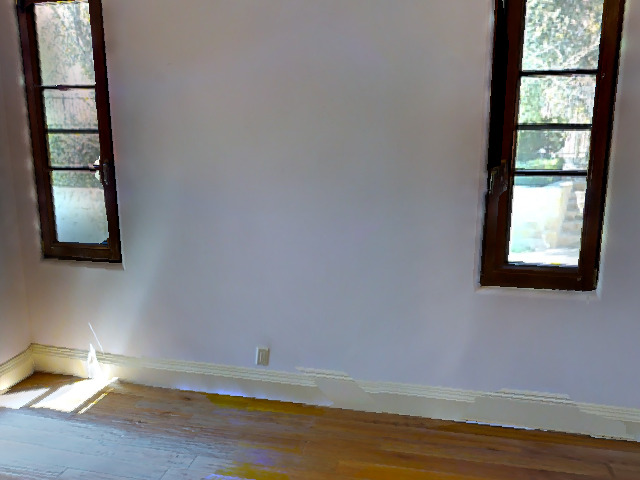}}
   & \frame{\includegraphics[width=0.15\textwidth]{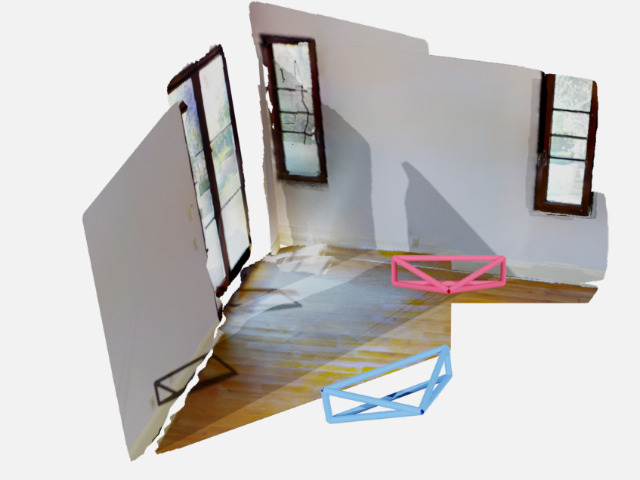}}
   & \frame{\includegraphics[width=0.15\textwidth]{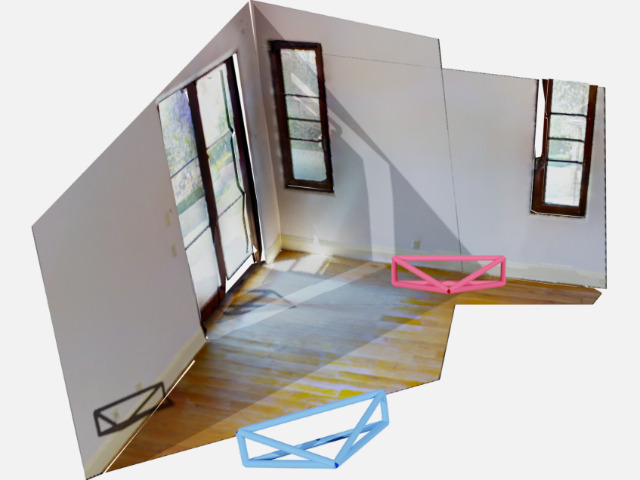}}
   & \frame{\includegraphics[width=0.15\textwidth]{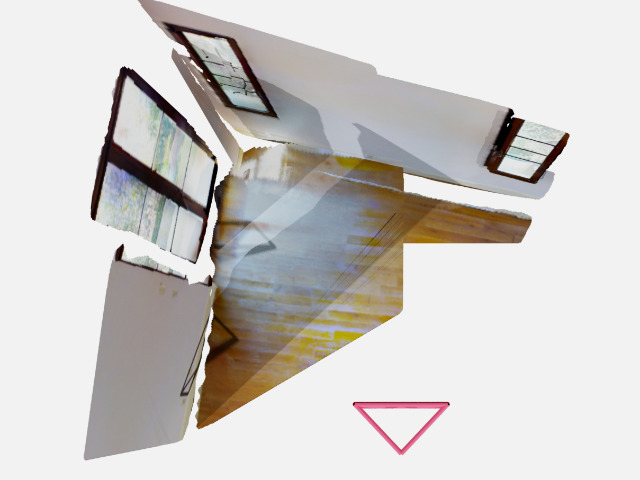}}
   & \frame{\includegraphics[width=0.15\textwidth]{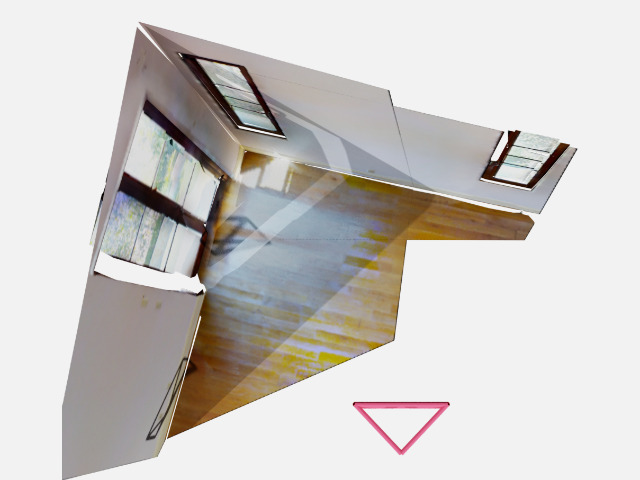}}\\

    \bottomrule
    \end{tabular}
    \caption{Qualitative results.
    \textbf{\textcolor{AccessibleBlue}{Blue}} and \textbf{\textcolor{AccessibleRed}{Red}} frustums show
    cameras for image 1 and 2. See supplemental for more results.
    }
    \label{fig:example-wall}
    \vspace{-0.1in}
\end{figure*}

\noindent {\bf Discrete Problem:} We first solve a discrete problem that selects a camera from the $K$ options 
and the plane correspondence
matrix $\CB$. The most important term is
expressed via a $m \times n$ cost matrix $\SB_k$ encoding the quality of a plane
correspondence $\CB$ assuming camera $k$ has been selected.
When the second view's plane parameters have been transformed
into the first view by camera $k$ (omitted for clarity), the cost matrix $\SB_k$ (with trade-off parameters $\lambda$) is
\begin{equation}
\label{eqn:costmatrix}
(\SB_k)_{i,j} = \lambda_e ||\eB_i - \eB_j'|| + \lambda_n \textrm{acos}(|\nB_i^\intercal \nB_j'|) + \lambda_o |o_i-o_j'|
\end{equation}
which includes terms for the embedding, normal, and offset which are $0$ with perfect matches.

This cost term is combined with two regularizing terms.
One, $-\log(p_k)$, penalizes unlikely cameras via the negative log-likelihood of camera $k$; the other
rewards matching objects: $-\sum_{i,j} \CB_{i,j}$. We pick the best correspondence and camera that optimize the objective
\begin{equation}
\arg\min_{\CB,k} -\lambda_c \log(p_k) + \sum_{i,j} (\CB\circ\SB_k)_{i,j} - \sum_{i,j}\CB_{i,j},
\label{eqn:discrete objective}
\end{equation}
which encourages selecting a likely camera, as many planes as possible, and correspondence that is
consistent in both appearance and geometry assuming the camera is correct. 

With a fixed camera $k$, the objective $\sum_{i,j} (\CB\circ\SB_k)_{i,j}$ can be efficiently 
solved using the Hungarian algorithm, and the number of matches handled via
thresholding. Since there are a finite number ($K$) of camera hypotheses, this
amounts to solving $K$ independent matching problems.

\noindent {\bf Continuous Problem:}
Having selected a camera $\hat{k}$ and planar correspondences $\CB$, we can then 
refine the predictions from the deep networks for the camera and planes. We optimize
over camera transformations $\hat{\RB},\hat{\tB}$ and plane parameters
$\hat{\piB}_i, \hat{\piB}_j'$ to minimize 
geometric distance between the corresponding planes 
(assuming the same coordinate frame) and pixel alignment error based on pixel-level features: 
\begin{equation}
\label{eqn:continuous}
\argmin_{\hat{\RB},\hat{\tB},\hat{\piB}_i, \hat{\piB}_j'} 
\sum_{i,j} \CB_{i,j} (||\hat{\piB}_i - \hat{\piB}_j'|| + d_{\textrm{pixel}}(\hat{\piB}_i,\hat{\piB}_j')) + d_{\textrm{cam}}(\hat{\RB},\hat{\tB})
\end{equation}
where $d_{\textrm{pixel}}$ measures
the Euclidean distance of back-projected points that match across corresponding planes.
Inspired by~\cite{wu20083d}, we warp texture to viewpoint normalized cameras 
using the plane parameters to extract viewpoint-invariant SIFT~\cite{lowe2004distinctive} features. 
The term $d_{\textrm{cam}}$ regulates the deviation from the selected camera bin. 
We initialize the optimization at $\RB_{\hat{k}}, \tB_{\hat{k}}, \piB_i, \piB_j'$,
and optimize with a trust-region reflective minimizer.
We parameterize rotations as 6D vectors following~\cite{zhou2019continuity}. 

\noindent {\bf Merging Planes:}
Given planes in correspondence and camera transformations, we merge them in the global frame. We merge offsets by averaging 
and normals $\{\nB_i\}$ by solving for the $\hat{\nB}$ maximizing $\sum_{i}(\hat{\nB}^\intercal \nB_i)^2$
via an eigenvalue problem.

\subsection{Implementation Details}

A full detailed description appears in the supplemental.
We use Detectron2 \cite{wu2019detectron2} to implement our network.
The plane detection backbone uses ResNet50-FPN pretrained on COCO~\cite{lin2014microsoft}. 
We directly regress normals instead of using classification in~\cite{liu2019planercnn}.
The camera branch and the plane embedding are trained using a Siamese network whose backbone is the plane detection
backbone. During training, we first train the plane prediction backbone on single images and then freeze the network. 
We train the plane appearance embedding and camera pose module on the frozen backbone. 
We fit all trade-off parameters (\eg, $\lambda_n$) on the validation set using randomized search.
We run $k$-means clustering and spherical $k$-means clustering on the training set to produce 32 bins for translation and rotation respectively.

\section{Experiments}
\label{sec:experiments}
We evaluate our approach using renderings of real-world scenes.
Our approach generates a new, rich and coherent output in terms of a planar reconstruction
plus a geometric relationship between the two views.
In the process of producing this reconstruction, it solves two
other problems: predicting object correspondences and relative camera pose estimation.
We therefore evaluate our model in three ways, which each naturally have their own metrics and baselines: the full model (Section~\ref{sec:exp_full}),
the plane correspondence (Section~\ref{sec:exp_corr}) and the relative camera poses
(Section~\ref{sec:exp_camera}). %

\begin{figure*}[!t]
    \centering
    \scriptsize
    \resizebox{\textwidth}{!}{
    \begin{tabular}{c@{\hskip4pt}c@{\hskip4pt}c@{\hskip4pt}c@{\hskip4pt}c@{\hskip4pt}c@{\hskip4pt}c@{\hskip4pt}c}
    \toprule
    Image 1 & Image 2 & Odometry + MWS & Odometry + MWS-G & SuperGlue-G + PlaneRCNN & No Conti. Optimization & Proposed & Ground Truth \\
    \midrule
    \frame{\includegraphics[width=0.15\textwidth]{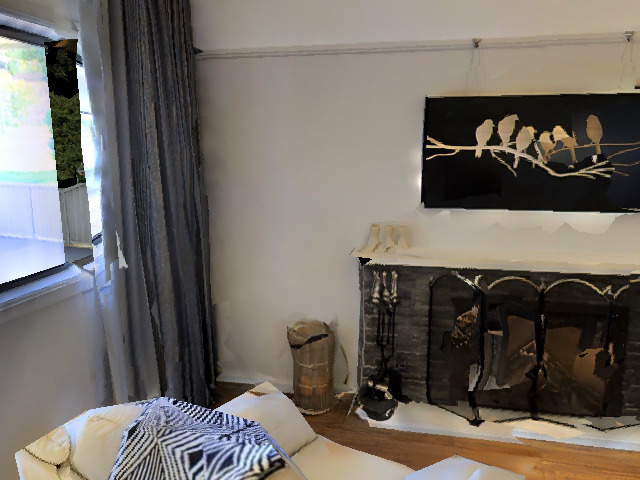}}
    & \frame{\includegraphics[width=0.15\textwidth]{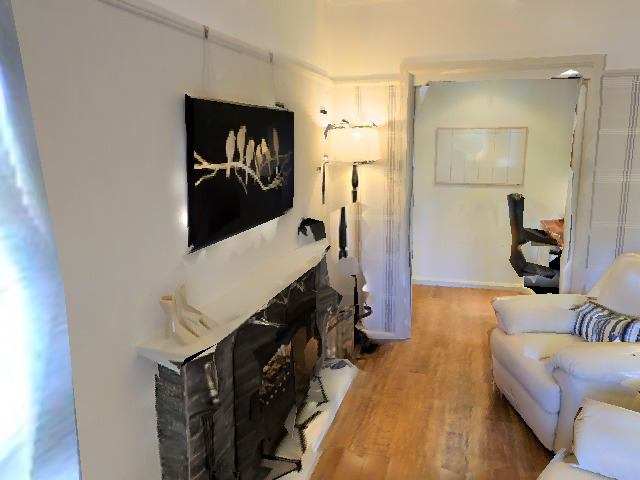}}
    & \frame{\includegraphics[width=0.15\textwidth]{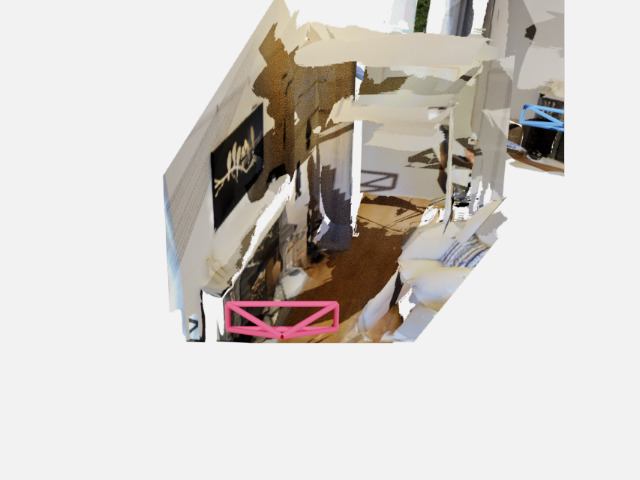}}
    & \frame{\includegraphics[width=0.15\textwidth]{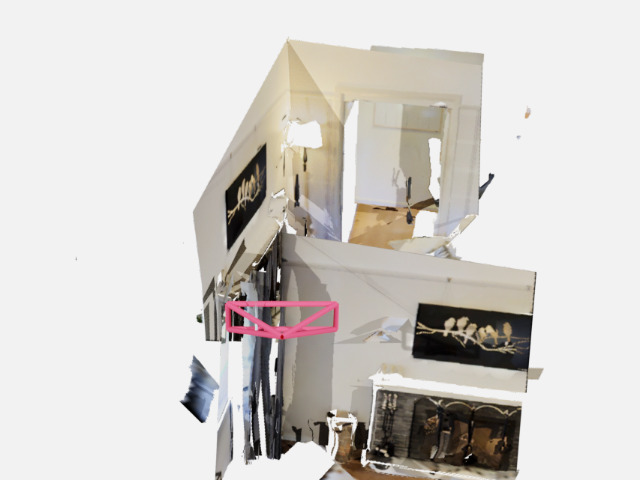}}
    & \frame{\includegraphics[width=0.15\textwidth]{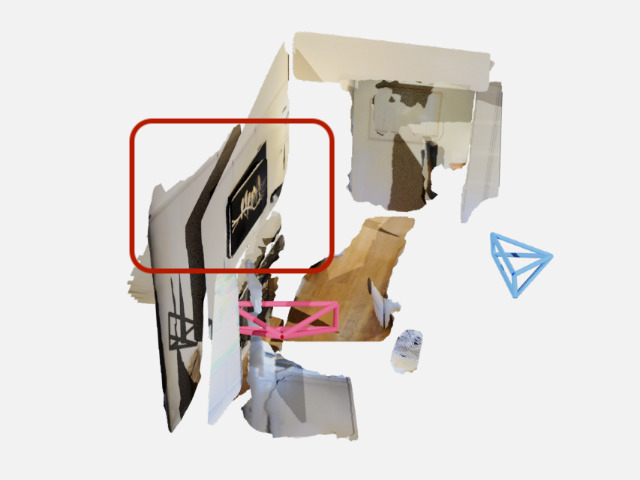}}
    & \frame{\includegraphics[width=0.15\textwidth]{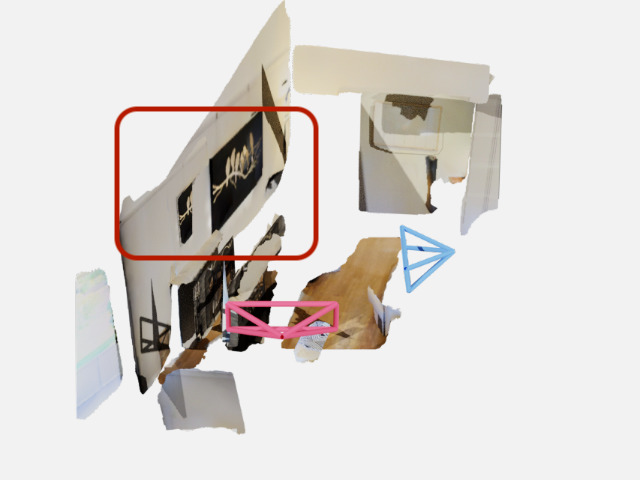}}
    & \frame{\includegraphics[width=0.15\textwidth]{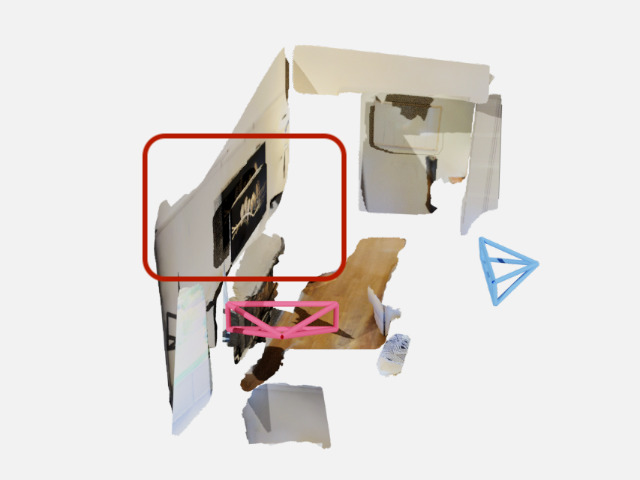}}
    & \frame{\includegraphics[width=0.15\textwidth]{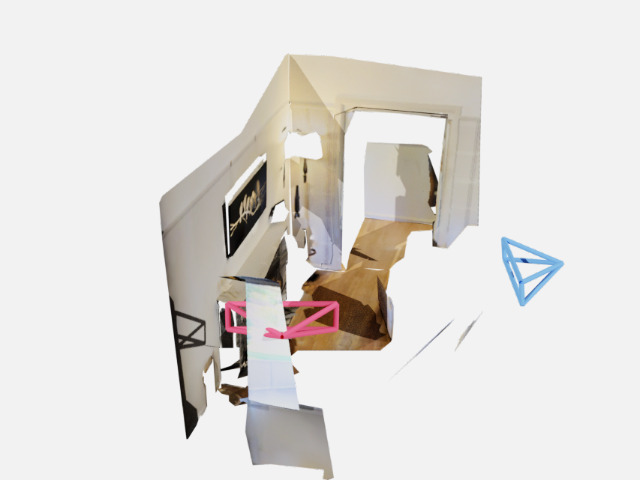}}\\
    \frame{\includegraphics[width=0.15\textwidth]{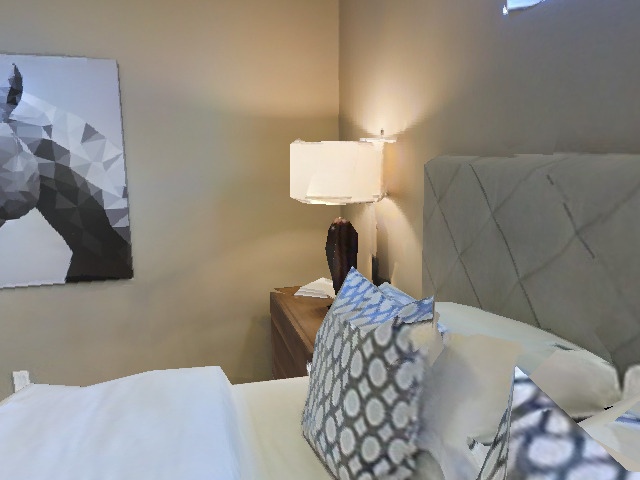}}
    & \frame{\includegraphics[width=0.15\textwidth]{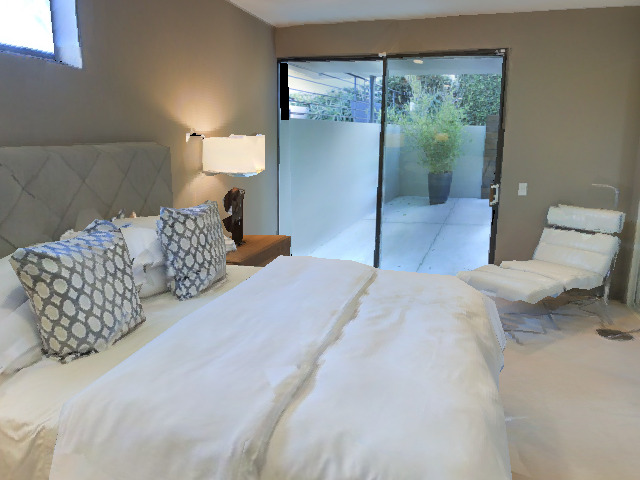}}
    & \frame{\includegraphics[width=0.15\textwidth]{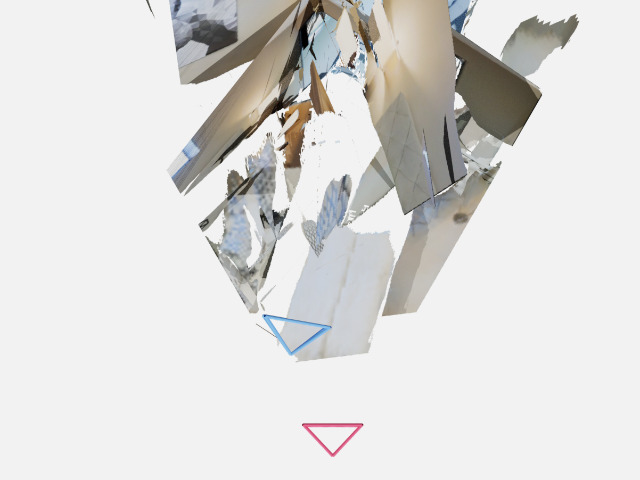}}
    & \frame{\includegraphics[width=0.15\textwidth]{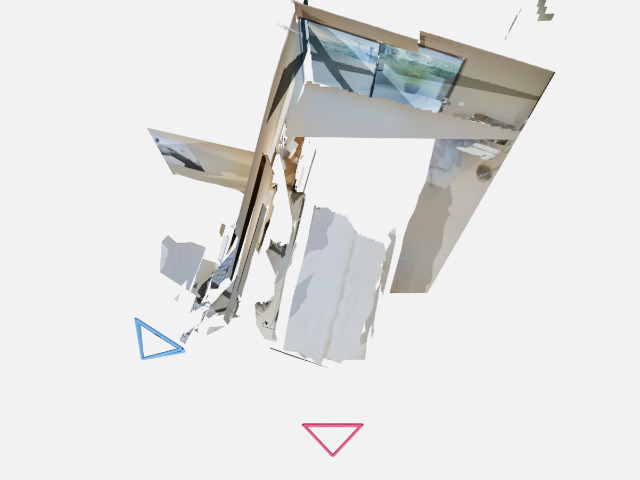}}
    & \frame{\includegraphics[width=0.15\textwidth]{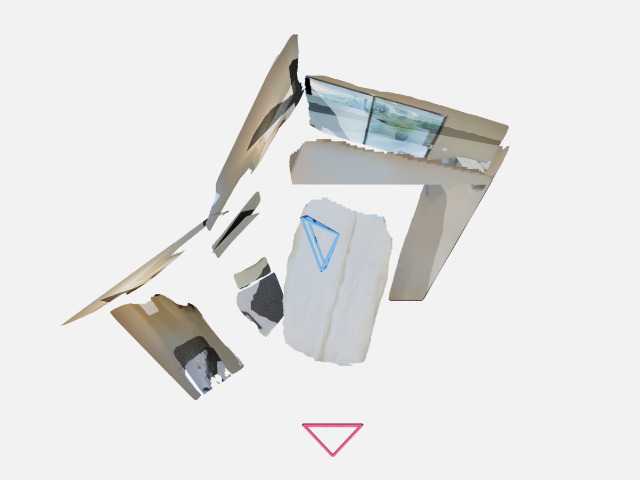}}
    & \frame{\includegraphics[width=0.15\textwidth]{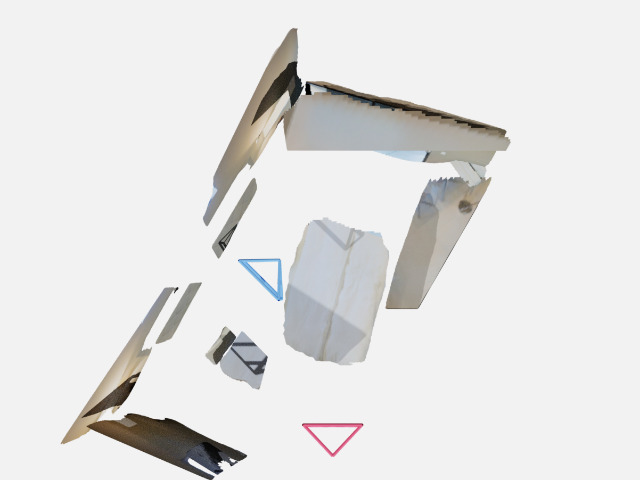}}
    & \frame{\includegraphics[width=0.15\textwidth]{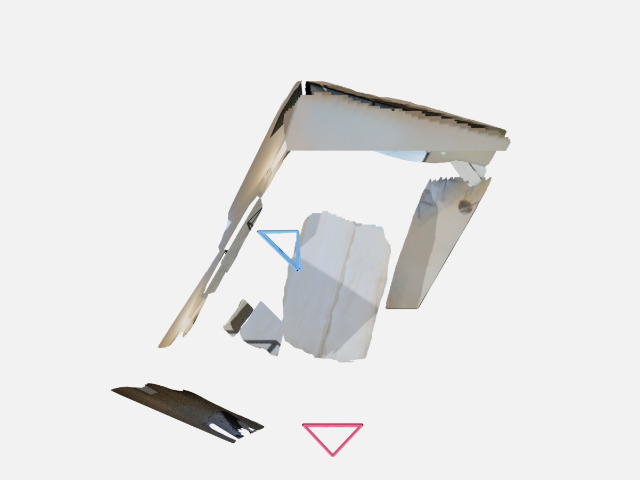}}
    & \frame{\includegraphics[width=0.15\textwidth]{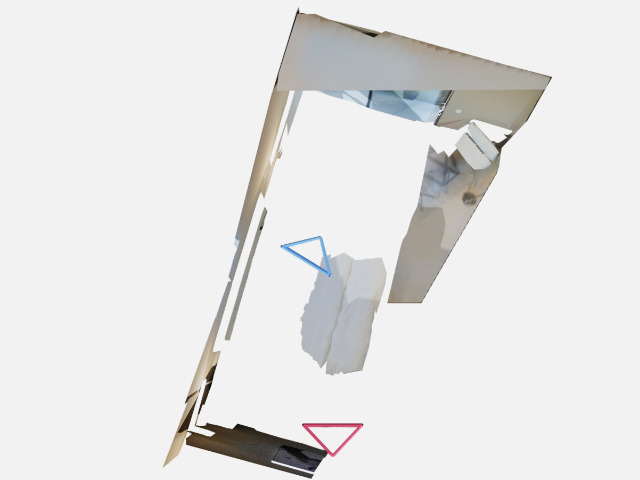}}\\
    \bottomrule
    \end{tabular}
    }
    \caption{Comparison between baselines. 
    \textbf{Row 1:} Shown in \textbf{\textcolor{redbox}{red}} boxes, SuperGlue + PlaneRCNN does not predict the plane correspondence; 
    our discrete optimization merges the same wall from two views into one plane; our proposed method further aligns the texture on the wall.
    \textbf{Row 2: } Plane Odometry + MWS-G aligns plane normals but with incorrect correspondence.}
    \label{fig:comparison-wall}
    \vspace{-0.1in}
\end{figure*}

\subsection{Experimental Setup}
\label{sec:exp_setup}

We use renderings of real scenes from Habitat \cite{savva2019habitat} and
Matterport3D \cite{chang2017matterport3d}. Habitat
enables the rendering of realistic images and provides ground truth pose and depth that can be used for
evaluation. We stress that our image pairs have far less
overlap compared to other settings with an average overlap of
just $21\%$ vs 
$64\%$ used by DeMoN \cite{ummenhofer2017demon} on Sun3D \cite{xiao2013sun3d} or 80\% by the 
plane odometry method \cite{raposo2013plane} (see supplement for more statistics).

\vspace{0mm}
\noindent
{\bf Dataset:}
Our training, validation, and test set consist of 31932, 4707, and 7996 image pairs respectively.
We fit our ground truth planes on Matterport3D using a standard RANSAC approach, using
semantic labels to constrain the planes, following~\cite{liu2018planenet}.
We fit planes on the 3D meshes and
render them to 2D, which provides us with plane correspondence even when the planes have
limited overlap. We generate camera poses following~\cite{zhang2017physically}
and select pairs with $\ge 3$ common planes and $\ge 3$ unique planes like~\cite{Qian2020}.

\subsection{Full Scene Evaluation}
\label{sec:exp_full} 
We begin by evaluating our approach's ability to produce full scene reconstruction in
terms of a set of 3D plane segments in the scene in a common coordinate frame.

\vspace{0mm}
\par \noindent {\bf Metrics:} We follow other approaches that reconstruct the scene
factored into components
\cite{tulsiani2018factoring,kulkarni2019relnet,LiSilhouette19,nie2020total3dunderstanding,Qian2020}
and treat the full
problem like a detection problem, evaluated using average precision (AP). Given a ground-truth
decomposition into components (in this case, planes), we evaluate how well the approach
detects and successfully reconstructs each one. We define a true positive
as a detection satisfying three criteria: (i) ({\it Mask}) mask intersection-over-union 
$\geq 0.5$; (ii) ({\it Normal}) surface normal distance, $\textrm{acos}(|\xB_1^\intercal \xB_2|) \leq 30^\circ$;
and (iii) ({\it Offset}) offset distance, $|o_1 - o_2| \leq 1$m.

\vspace{0mm}

\par \noindent {\bf Baselines:} We compare with alternate approaches and ablations that test the system's contributions.
To the best of our knowledge, no existing work solves our task, so we test
fusions of existing systems. These baselines estimate a relative camera pose
and per-image planes.

\par \noindent {\it Plane Odometry \cite{raposo2013plane} + MWS~\cite{furukawa2009manhattan} / PlaneRCNN~\cite{liu2019planercnn}:} 
\cite{raposo2013plane} extracts plane-primitives from successive RGB-D video frames and uses plane-to-plane registration to solve for the camera pose.
For the 3D representation, we either predict planes using PlaneRCNN in our plane prediction module or
run Manhattan-world Stereo~\cite{furukawa2009manhattan} 
on predicted depth (MWS) or ground truth depth (MWS-G).
To provide~\cite{raposo2013plane} and ~\cite{furukawa2009manhattan} with predicted depthmaps, we use MiDaS~\cite{Ranftl2020}, a state of the art system trained on ${\approx}2$M images. We found that
MWS gave better plane fits on depthmaps without semantic labels compared to our RANSAC approach,
also shown in~\cite{liu2019planercnn,liu2018planenet}. 
Plane Odometry mainly reasons with geometric information when producing its final output and tests a simple fusion of 
visual odometry and plane extraction systems.

\par \noindent {\it SuperGlue-G \cite{sarlin2020superglue} + MWS~\cite{furukawa2009manhattan} / PlaneRCNN~\cite{liu2019planercnn}:} 
\cite{sarlin2020superglue} is a point-based matching algorithm that uses deep features to find pixel correspondence across images
to estimate an essential matrix. The scale of the translation is not determined by the essential matrix so
we use the {\it ground truth} translation scale for this method. Approaches like \cite{sarlin2020superglue} are complementary to our 
camera branch since they have finer resolution but no object-size prior;
our approach could integrate \cite{sarlin2020superglue} in its continuous
optimization. Like the previous baseline, we try each of MWS/MWS-G/Plane-RCNN.

\par \noindent {\it RPNet \cite{en2018rpnet} + PlaneRCNN~\cite{liu2019planercnn}:} 
We apply the PlaneRCNN module used by our system and
our improved variant of RPNet \cite{en2018rpnet} -- see Sec.~\ref{sec:exp_camera} for more details. This is equivalent to our
system, but ignoring our optimization or reasoning over plane parameters. This tests the value of joint reasoning while holding 
base networks fixed.

\par \noindent {\it Appearance Embedding Only:}
We run the optimization with only the appearance embedding $\eB_i$
predicted for each plane. This gives a fusion of planes based on learned
embeddings, which tests the value of our geometric reasoning.

\par \noindent {\it Associative3D \cite{Qian2020} Optimization:} We apply a heuristic RANSAC-style optimization scheme from the most closely
related paper~\cite{Qian2020}. We adapt it to our setting for a fair comparison (see supplemental for detail). This tests the value of
our optimization, which casts camera selection and matching in terms that are solvable via principled algorithms.

\par \noindent {\it No Continuous Optimization:} We perform the full method without the continuous optimization in Section~\ref{sec:optimization}. 

\par \noindent {\it Proposed:} We apply the full proposed method.

\vspace{0mm}
\par \noindent {\bf Qualitative Results:} 
We show qualitative results on Matterport3D test set in Figure~\ref{fig:example-wall}. Prediction and ground truth 
are shown in two novel views to see all planes and camera poses in the scene.
Our method predicts accurate camera poses from inputs with small overlap, 
 merges planes across different views, and aligns their textures to produce a coherent reconstruction. More
examples are in the supplement.

We show comparison between our proposed approach and baselines in Figure \ref{fig:comparison-wall}.
{\em Plane Odometry} fails on predicted depth due to distortion and rarely finds the correct correspondences between planes, 
\eg incorrect walls are aligned. 
{\em SuperGlue-G} can usually obtain a reasonable camera pose given oracle translation scale
but does not merge planes across views, \eg, non-parallel walls. 
{\em No Continuous Optimization} finds and merges the corresponding planes using the Hungarian algorithm in our discrete optimization. 
However, simply merging the corresponding planes still generates erroneous results, \eg, unaligned texture of pictures on the wall in row 1, 
disconnected walls in row 2, which are caused by discretization bins of the camera transformation and inaccurate single-view prediction. 
Our proposed approach refines the camera and planes further, so that it fixes the inconsistency of plane detection and camera modules and
generates reconstructions that are significantly closer to the ground truth.

\begin{table}[t]
    \centering
    \caption{Average Precision, treating reconstruction as 3D plane detection with different definitions
        of true positive. {\it All:} Mask IoU $\geq 0.5$, Normal error $\leq 30^\circ$, Offset error $\leq 1$m.
        {\it -Offset} and {\it -Normal} ignore the offset and normal condition respectively.
    }
    \resizebox{\columnwidth}{!}{
        \begin{tabular}{@{~~}l@{~~}c@{~~}c@{~~}c@{~~}}
            \toprule
            Methods & All   & -Offset & -Normal \\
            \midrule
            Odometry~\cite{raposo2013plane} + MWS~\cite{furukawa2009manhattan, Ranftl2020}       & 4.80 & 6.72 & 6.12 \\
            Odometry~\cite{raposo2013plane} + MWS-G~\cite{furukawa2009manhattan}          & 16.42 & 19.57 & 18.81 \\
            Odometry~\cite{raposo2013plane} + PlaneRCNN~\cite{liu2019planercnn}           & 21.33 & 27.08 & 24.99 \\
            SuperGlue-G~\cite{sarlin2020superglue} + MWS~\cite{furukawa2009manhattan, Ranftl2020}          & 7.56 & 9.06 & 8.64 \\
            SuperGlue-G~\cite{sarlin2020superglue} + MWS-G~\cite{furukawa2009manhattan}          & 23.01 & 23.89 & 25.22 \\
            SuperGlue-G~\cite{sarlin2020superglue} + PlaneRCNN~\cite{liu2019planercnn}         & 30.06 & 33.24 & 33.52 \\
            RPNet~\cite{en2018rpnet} + PlaneRCNN~\cite{liu2019planercnn}       &  29.44		& 35.25	&	31.67  \\
            \midrule
            Appearance Embedding Only 			                                    &  33.04		& 39.78	&	36.85   \\
            Associative3D \cite{Qian2020} Optimization                                  &  33.01 		& 39.43 & 35.76  \\
            No Continuous Optimization                                                  &  35.87		& \textbf{42.13}	&	38.80   \\
            \textbf{Proposed} 					                                        &  \textbf{36.02}	&	42.01	&	\textbf{39.04} \\
            \bottomrule
        \end{tabular}
    }
    \label{tab:detection}
    \vspace{-0.1in}
\end{table}

\vspace{0mm}
\par \noindent {\bf Quantitative Results:} We show results in 
Table~\ref{tab:detection}. 
Fitting planes on predicted depthmaps ({\it MWS}) fails poorly due to distorted predictions.
This is a known challenge (\eg, addressed by concurrent work~\cite{Wei2021CVPR} without
public code at time of submission). Methods using {\it PlaneRCNN} outperform {\it MWS} even with 
ground truth (GT) depth, also found in~\cite{liu2018planenet,liu2019planercnn}; this is because of
PlaneRCNN's stronger plane labels improved by semantics.
{\it Plane Odometry} finds pose primarily with plane normals. Its correspondence is often wrong 
because our low-overlap setting limits the information in geometric cues alone (in contrast to the combined
normal, offset, and texture used by our method). SuperGlue with GT translation scale ({\it SuperGlue-G}) slightly
outperforms the {\it RPNet} with PlaneRCNN, but falls short of the proposed
full method by 6 AP. Some of this gap is because there are duplicate copies of planes found in both images. One 
contribution of our method is preventing this case. While merging based on appearance only partially solves the problem,
the full method, which incorporates geometry, does the best. The particular optimization for this merging is moreover important: 
the heuristic randomized optimization of~\cite{Qian2020} does worse. The continuous
optimization makes little difference in planes but improves the camera considerably (Sec.~\ref{sec:exp_camera}).
Additional experiments on AP vs. overlap are in the supplement.

\subsection{Plane Correspondence}
\label{sec:exp_corr}

One core challenge for generating a single coherent reconstruction from two images
is ensuring that each visible plane appears exactly once rather than being split.
We therefore evaluate how well we can identify correspondences between planes across
views.

\vspace{0mm}
\par \noindent {\bf Baselines:} 
We compare the proposed approach with three baselines that 
test the contribution of each component of our method. 
We measure performance using ground-truth bounding boxes during evaluation
so that we measure only errors due to correspondence and not detection.
\par \noindent {\it Appearance Only:} We apply the Hungarian algorithm
to the appearance embedding and do not use geometry. This tests how important
geometry is to identifying correspondence.
\par \noindent {\it MessyTable~\cite{cai2020messytable} ASNet on ROI:} We 
adapt the ASNet~\cite{cai2020messytable} framework to our setting
and architecture for a fair comparison (see supplemental for details). This network
improves matching by adding an additional pathway that has more context.
This tests whether our matching is constrained by looking only inside a proposal
region.

\par \noindent {\it Associative3D \cite{Qian2020} Optimization:} 
We again apply the optimization method of \cite{Qian2020}, following Section~\ref{sec:exp_full}.
\par \noindent {\it Proposed:} We apply the full approach.

\vspace{0mm}
\par \noindent {\bf Metrics:}
We aim to measure how well methods associate planes in a pair of images.
We therefore evaluate performance directly with Image Pair Association Accuracy (IPAA) metric by
Cai \etal~\cite{cai2020messytable}, which represents the fraction
of image pairs with no less than X\% of the planes associated correctly (written as IPAA-X). 
We use ground truth bounding boxes in this section for evaluation.

\vspace{0mm}
\par \noindent {\bf Quantitative Results:}
Table \ref{tab:correspondence} shows that our method outperforms all other benchmarks on IPAA-100 and IPAA-90 and 
has a slight drop of 0.4\% on IPAA-80. Our full optimization improves IPAA-100 by a large margin (9.4\%) compared to using
the appearance embedding matrix only. We hypothesize our approach outperforms \cite{Qian2020} due to the
factorial growth of the search space, which the heuristic approach struggles with. We suspect that 
\cite{cai2020messytable} struggles since it is designed for table-top matching, where the planar
table constrains how much context can change from view to view (in contrast to our strongly non-planar scenes).

\begin{table}[t]    
    \centering
    \caption{Plane correspondence using ground truth bounding boxes. IPAA-X~\cite{cai2020messytable} measures the fraction
    of pairs with no less than X\% of planes associated correctly. }
    \label{tab:plane}
        \begin{tabular}{@{\hskip5pt}l@{\hskip5pt}c@{\hskip5pt}c@{\hskip5pt}c@{\hskip5pt}c@{\hskip5pt}}
        \toprule
        & IPAA-100 & IPAA-90 & IPAA-80 \\
        \midrule
        Appearance Only & 6.8 & 23.5 & \textbf{55.7} \\
        ASNet \cite{cai2020messytable} on ROI & 6.9 & 21.7  & 52.1 \\
        Associative3D \cite{Qian2020} Opt. &  0.9 & 10.7 & 44.0 \\
        \textbf{Proposed}  &  \textbf{16.2} & \textbf{28.1} & 55.3 \\
        \bottomrule
        \end{tabular}
        \label{tab:correspondence}
        \vspace{-0.1in}
    \end{table}

\begin{figure}[!t]
    \centering
    \scriptsize
    \begin{tabular}{c@{\hskip4pt}c@{\hskip4pt}c@{\hskip4pt}c@{\hskip4pt}c}
    \toprule
    
    Inputs & Appearance Only & ASNet~\cite{cai2020messytable} & Associative3D~\cite{Qian2020} & \textbf{Proposed} \\
    \midrule
    \frame{\includegraphics[width=0.08\textwidth]{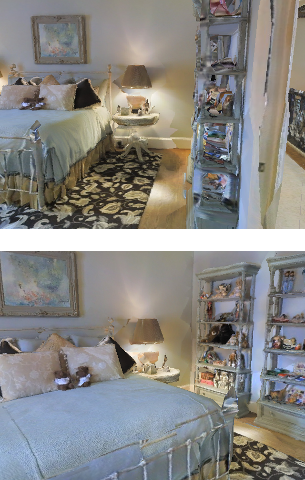}}
    & \frame{\includegraphics[width=0.08\textwidth]{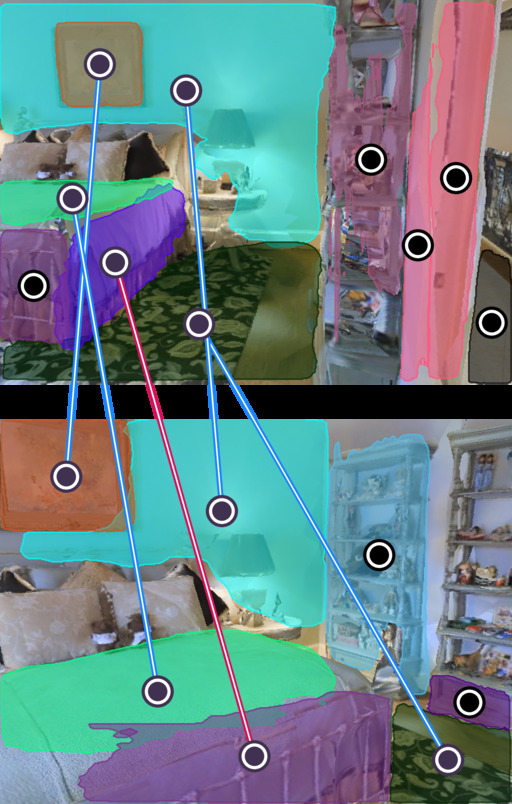}}
    & \frame{\includegraphics[width=0.08\textwidth]{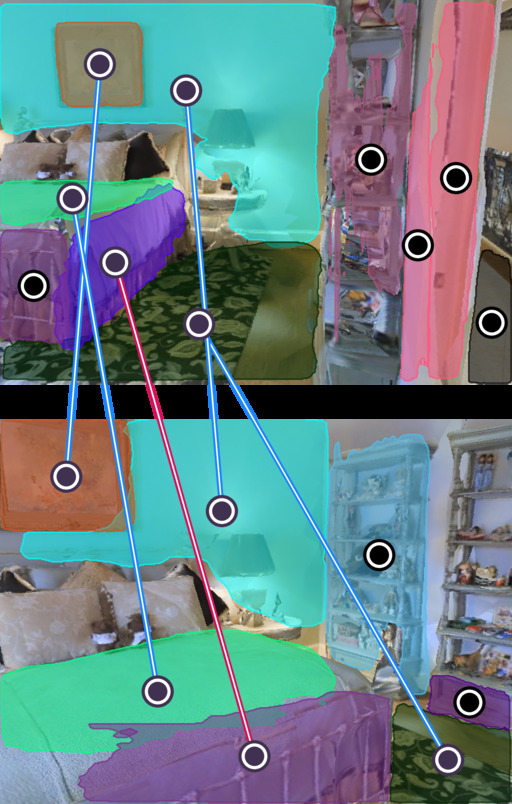}}
    & \frame{\includegraphics[width=0.08\textwidth]{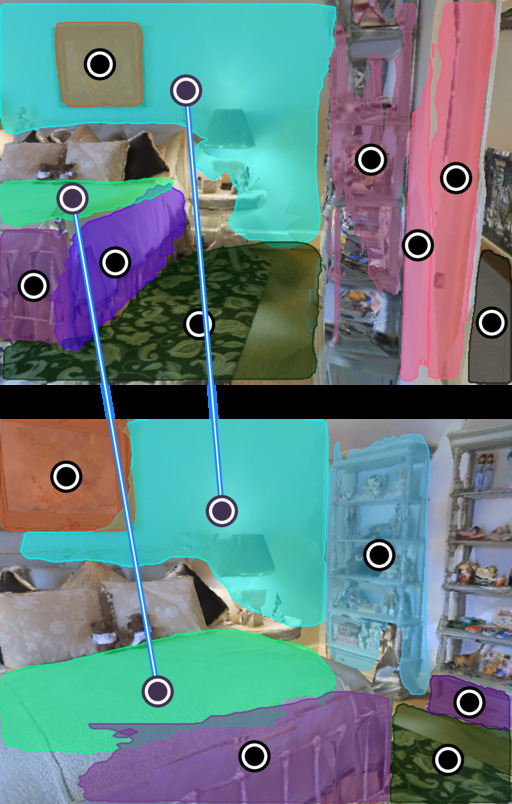}}
    & \frame{\includegraphics[width=0.08\textwidth]{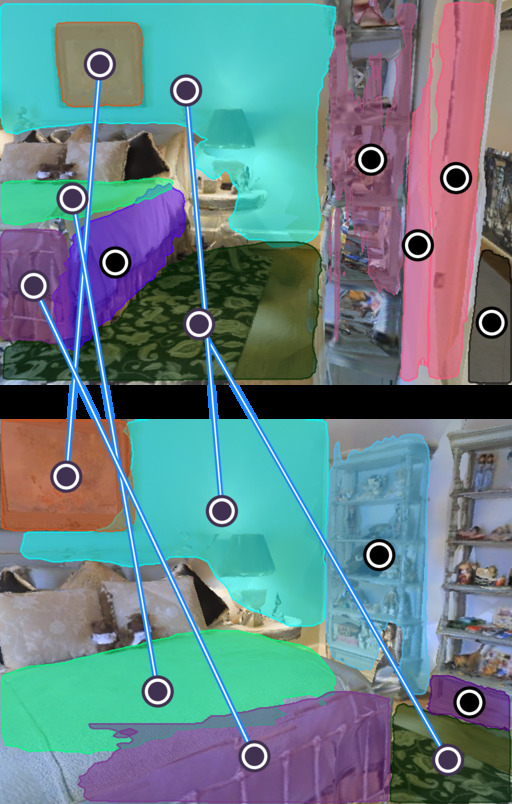}}\\

    \bottomrule
    \end{tabular}
	\caption{Representative correspondence prediction, showing true positive matches in 
\textbf{\textcolor{AccessibleBlue}{Blue}} and false positives in 
\textbf{\textcolor{AccessibleRed}{Red}}.
\vspace{-0.3in} }
    \label{fig:corr_wall}
\end{figure}

\par \noindent {\bf Qualitative Results:} {\it Appearance Only} and {\it ASNet} do not consider 
geometric information; therefore, they have difficulty distinguishing planes of similar texture, \eg, two sides of the bed in Figure~\ref{fig:corr_wall}.
{\it Associative3D} does not find all the correspondence due to its random search scheme. Our method detects all the correspondence
and distinguishes similar texture planes using geometric information.

\subsection{Relative Camera Pose Estimation}
\label{sec:exp_camera}
Our final experiment tests the effectiveness of the predicted relative camera
pose. We compare our method with a number of alternate approaches and baselines. 
Additional experiments testing the use of attention features rather than concatenation features
appear in the supplement.

\begin{table}[t]    
    \centering
    
    \caption{Evaluation of relative camera pose. Rows 1-4 compare with other baselines; Rows 5-6 analyze
	increasing amounts of optimization. We show median, mean error and $\%$ error {$\leq$} 1m or $30^{\circ}$ for translation and rotation.
    }
    \label{tab:camera}
    \resizebox{\columnwidth}{!}{
        \begin{tabular}{@{\hskip5pt}l@{~}c@{~}c@{~}c@{~~~~~~}c@{~}c@{~}c@{\hskip5pt}}
        \toprule
        & \multicolumn{3}{c}{Trans.} & \multicolumn{3}{c}{Rot.} \\
        Method & Med. & Mean & (${\leq}1$m) & Med. & Mean & (${\leq}30^{\circ}$) \\
        \midrule
        \cite{raposo2013plane} + GT Depth & 3.20 & 3.87 & 16.0 & 50.43 & 55.10 & 40.9 \\
        \cite{raposo2013plane} + \cite{Ranftl2020} & 3.34 & 4.00 & 8.30 & 50.98 & 57.92 & 29.89 \\
        Assoc.3D~\cite{Qian2020} & 2.17 & 2.5 & 14.8  & 42.09 & 52.97 & 38.1 \\  
        \midrule
        SuperGlue~\cite{sarlin2020superglue} & - & - & - & \textbf{3.88} & 24.17 & 77.84 \\
        \midrule			
        No Optimization     & 0.90 & 1.40 & 55.5  & 7.65 & 24.57 & 81.9 \\
        No Continuous       & 0.88 & 1.36 & 56.5  & 7.58 & 22.84 & \textbf{83.7} \\
        \textbf{Proposed}    & \textbf{0.63} & \textbf{1.25} & \textbf{66.6}  & 7.33 & \textbf{22.78} & 83.4 \\                           
        \bottomrule
        \end{tabular}
    }
        \label{tab:relpose}
        \vspace{-0.1in}
\end{table}

\vspace{0mm}
\par \noindent {\bf Metrics:} 
Following prior works \cite{su2015render,tulsiani2015viewpoints,Qian2020,banani2020novel},
we measure the camera rotation and translation by geodesic rotation distance and Euclidean distance separately.
We report the mean and median error, as well as the fraction
below a fixed threshold ($30^\circ$ and $1$m following \cite{Qian2020}). Evaluation on the translation 
angle error is in the supplement.

\vspace{0mm}
\par \noindent {\bf Baselines:} We compare our full system with four baselines and two ablations.

\par \noindent {\it Plane Odometry \cite{raposo2013plane} + GT Depth:} We apply the RGBD plane odometry system 
from \cite{raposo2013plane} to the ground-truth depth. As described in Section~\ref{sec:exp_full},
this approach mainly reasons geometrically, without learned priors or deep features.

\par \noindent {\it Plane Odometry \cite{raposo2013plane} + \cite{Ranftl2020}:} We apply the same
approach, using estimated depth from \cite{Ranftl2020} for a more fair comparison.

\par \noindent {\it Associative3D \cite{Qian2020} camera branch:} We apply an
improved version of the camera branch of Associative3D, which is an improved
version of the RPNet~\cite{en2018rpnet}. This method average pools
ResetNet50 features~\cite{he2016deep} followed by a MLP. For a fair comparison, we
upgrade the MLP to match the number of layers and parameters of our approach. This tests the value
of our improved attention features.

\par \noindent {\it \cite{sarlin2020superglue}:} We apply the SuperGlue system. 
Although as a pixel-correspondence-based method, we see this as a complementary
line of work to our method. SuperGlue estimates camera poses via the essential matrix
and thus cannot estimate the translation's scale. We note that the approach in
Section~\ref{sec:exp_full} used the ground-truth camera translation scale to give an
upper bound if the true translation scale were used.  The approach fails on
$11.6\%$ of our pairs due to insufficient correspondence; in this case, we
assign the identity matrix for rotation.
We use the only available model, which was trained on ScanNet~\cite{dai2017scannet}.

\par \noindent {\it No Optimization:} We apply our camera branch and take the most likely camera
without any joint reasoning.
\par \noindent {\it No Continuous:} We apply our approach but do not apply continuous optimization.
\par \noindent {\it Proposed:} We apply the full approach.

\vspace{0mm}
\par \noindent {\bf Quantitative Results:} 
We report results in Table~\ref{tab:camera}.
The full method has low median errors of $0.63$m/$7.33^\circ$.
Comparing with other approaches, we find that
our approach also substantially outperforms the Plane Odometry
\cite{raposo2013plane} method regardless of whether ground truth or predicted
depth is used. Just as described in Section~\ref{sec:exp_full}, this is due 
to the limited amount of information that is available in geometry alone. 
SuperGlue can be extraordinarily accurate in rotation when textures provide strong correspondences across views, 
but also can fail to make inferences on hard scenes.

The most likely camera from our approach ({\it No Optimization}) substantially
outperforms the previous method of \cite{Qian2020}, which follows a simpler
average pooling. This is largely due to our use
of attention features; additional more detailed ablations are in the supplement. 
Discrete optimization consistently slightly improves results,
but continuous optimization substantially improves translation with marginal
gains and loses in rotation performance.

\begin{figure}[!t]
   \centering
   \scriptsize
   \begin{tabular}{c@{\hskip4pt}c@{\hskip4pt}c@{\hskip4pt}c}
   \toprule
   
   Image 1 & Image 2 & Prediction & Ground Truth \\
   \midrule
   \frame{\includegraphics[width=0.11\textwidth]{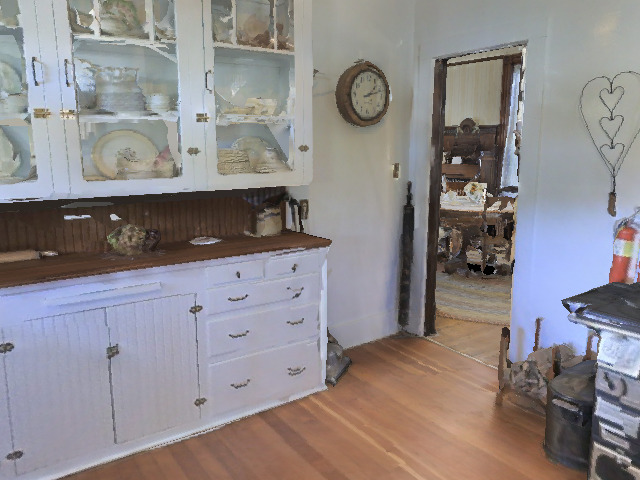}}
   & \frame{\includegraphics[width=0.11\textwidth]{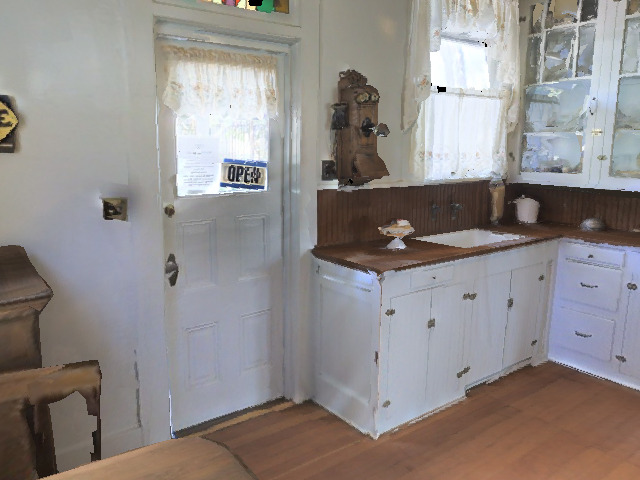}}
   & \frame{\includegraphics[width=0.11\textwidth]{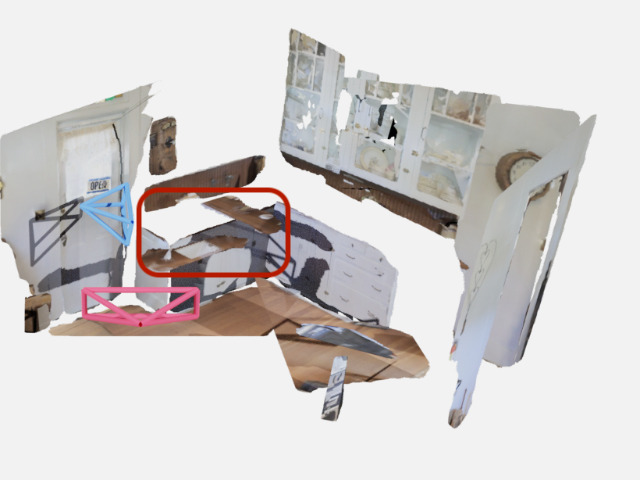}}
   & \frame{\includegraphics[width=0.11\textwidth]{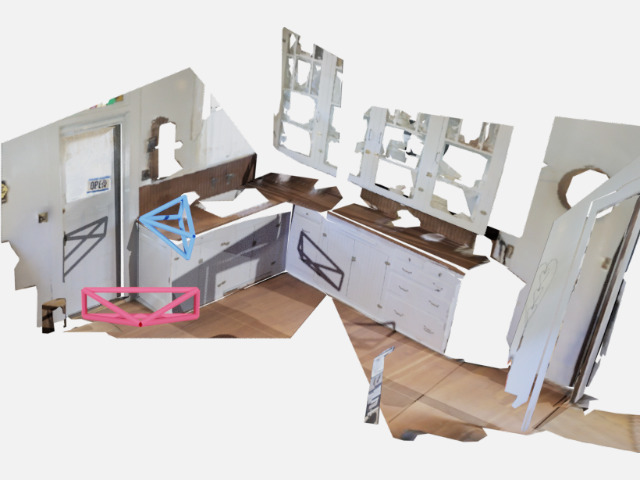}}\\

   \frame{\includegraphics[width=0.11\textwidth]{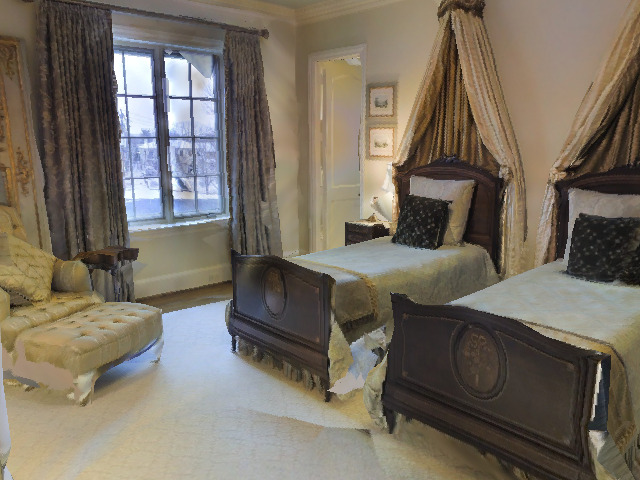}}
   & \frame{\includegraphics[width=0.11\textwidth]{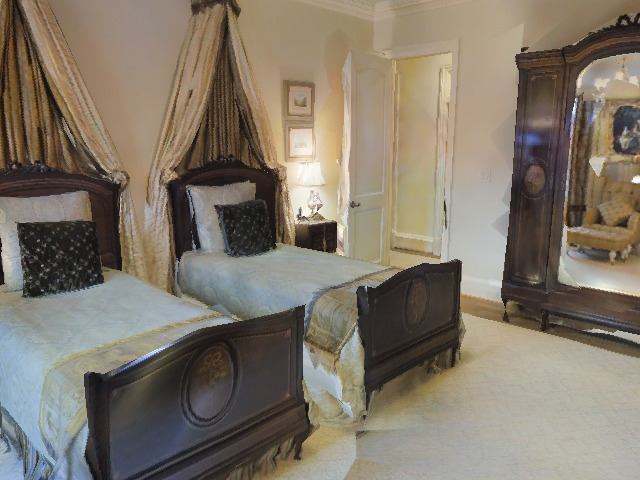}}
   & \frame{\includegraphics[width=0.11\textwidth]{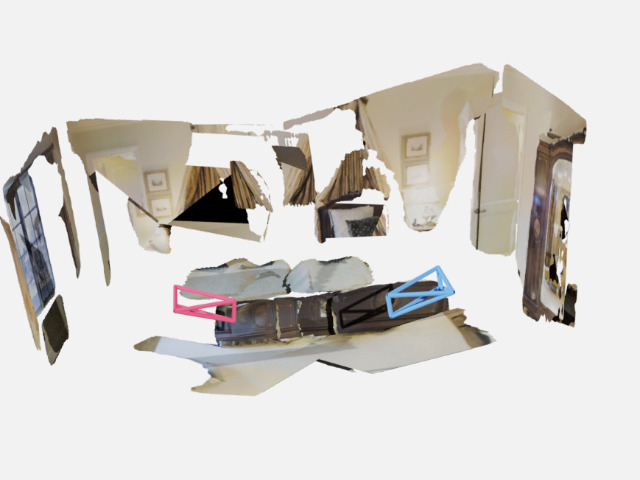}}
   & \frame{\includegraphics[width=0.11\textwidth]{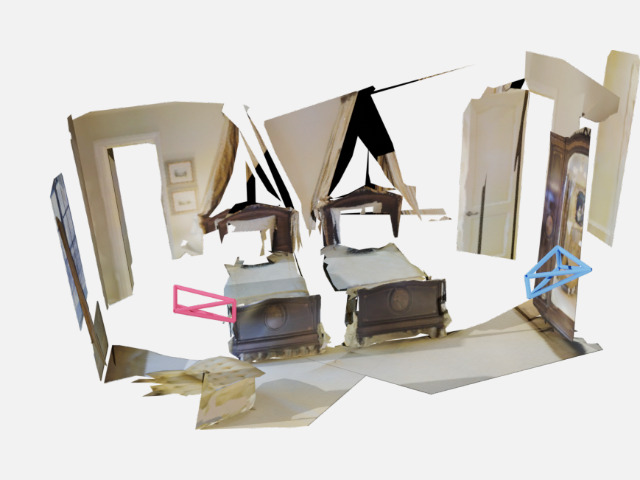}}\\

   \bottomrule
   \end{tabular}
   \caption{Representative failure modes. 
   \textbf{Row 1:} The kitchen countertop has little overlap across views. Though our method 
   stitches them, there is no correct pixel correspondence to constrain the alignment, shown in the \textbf{\textcolor{redbox}{red}} box.
   \textbf{Row 2:} The system fails on multiple coplanar segments with very similar texture.
   }
   \label{fig:failure-mode}
   \vspace{-0.1in}
\end{figure}

\begin{figure}[!t]
   \centering
   \scriptsize
   \begin{tabular}{c@{\hskip4pt}c@{\hskip4pt}c@{\hskip4pt}c}
   \toprule
   
   View 1 & View 2 & View 3 & Prediction \\
   \midrule
   \frame{\includegraphics[width=0.11\textwidth]{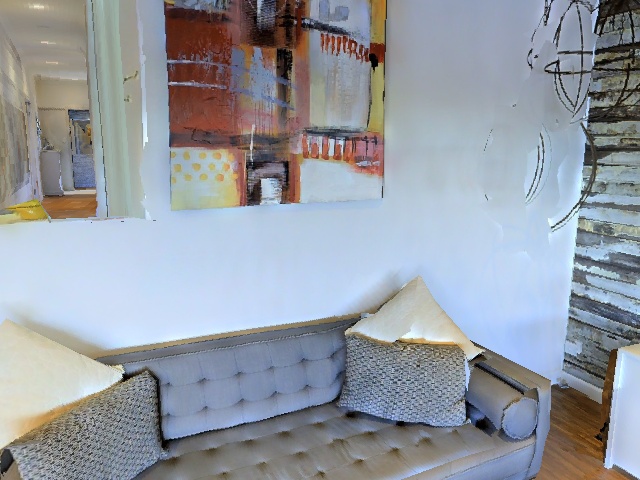}}
   & \frame{\includegraphics[width=0.11\textwidth]{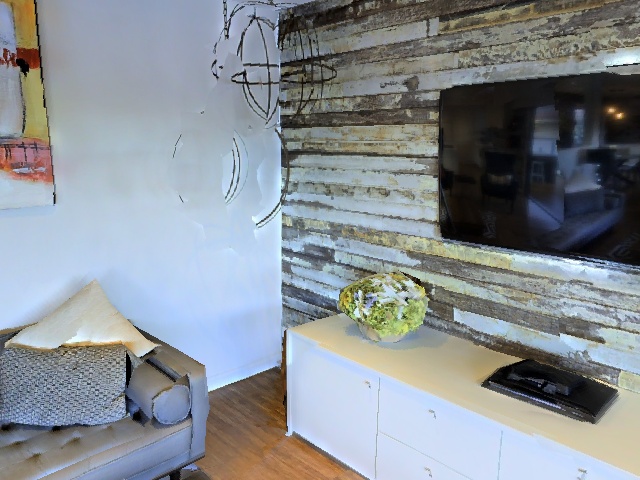}}
   & \frame{\includegraphics[width=0.11\textwidth]{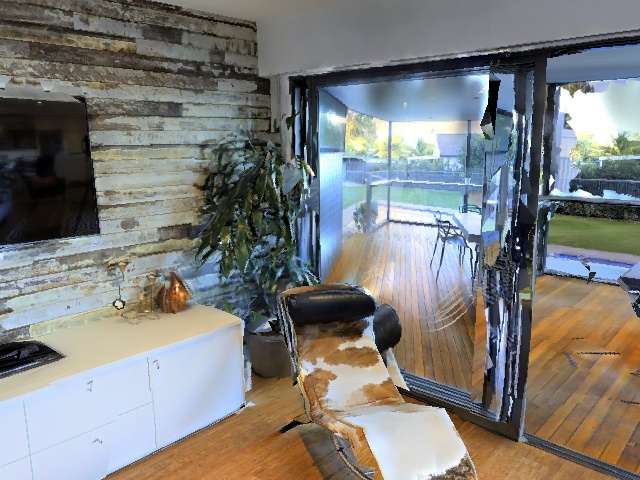}}
   & \frame{\includegraphics[width=0.11\textwidth]{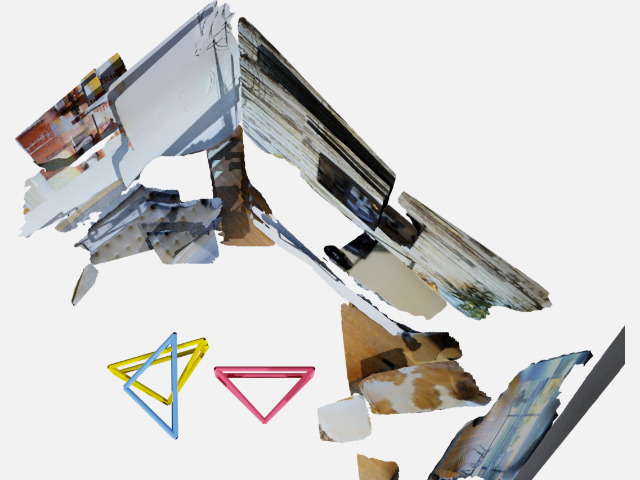}}\\

   \frame{\includegraphics[width=0.11\textwidth]{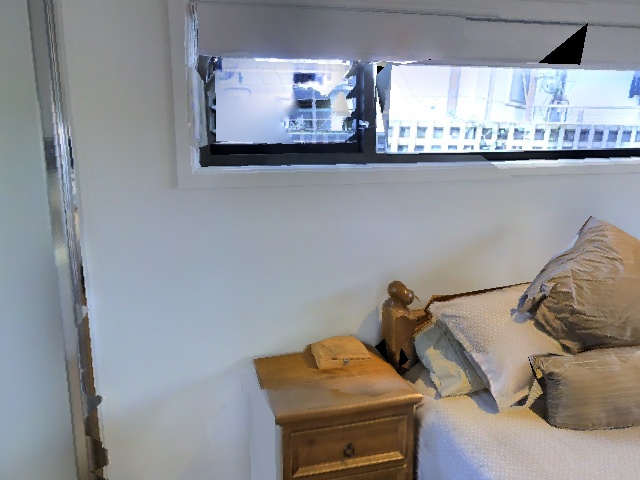}}
   & \frame{\includegraphics[width=0.11\textwidth]{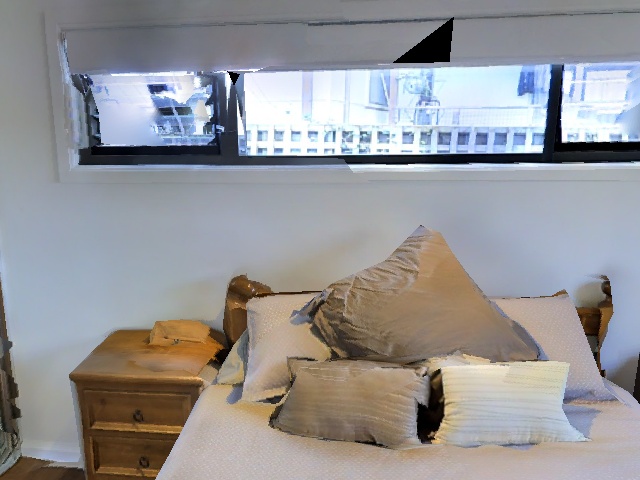}}
   & \frame{\includegraphics[width=0.11\textwidth]{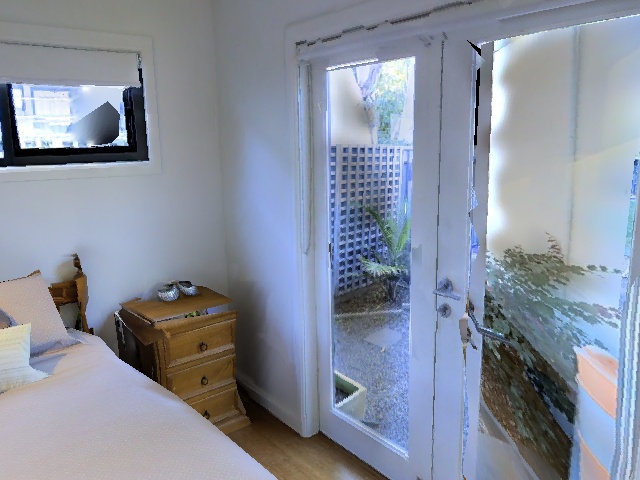}}
   & \frame{\includegraphics[width=0.11\textwidth]{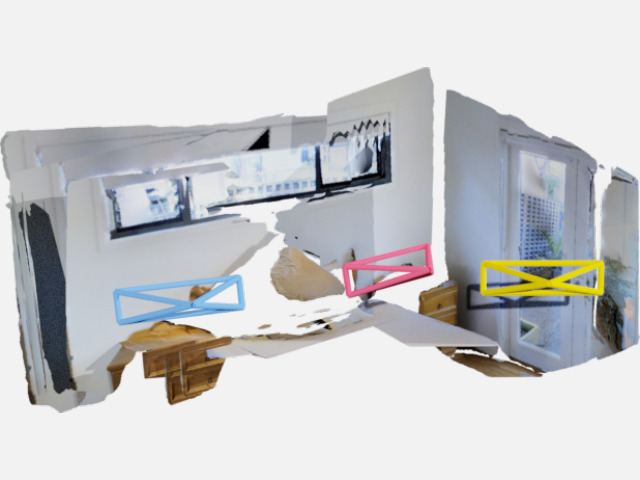}}\\

   \bottomrule
   \end{tabular}
   \caption{Extending the system to $>2$ views. We stitch two views together and stitch the third view incrementally.}
   \vspace{-0.2in}
   \label{fig:incremental}
\end{figure}
\section{Conclusion}
We've presented a learning-based system to produce a coherent planar surface
reconstruction from two unknown views. Our results suggest that jointly considering
correspondence and reconstruction improves both, although our experiments suggest
room for growth (\eg, see typical failure modes in Figure~\ref{fig:failure-mode},
including low-overlap planes and repeated coplanar and similar segments). 
Future directions include using our framework to reconstruct 
a more complete scene with fewer frames than traditional SfM~\cite{schonberger2016structure}. 
Figure~\ref{fig:incremental} shows examples that extend our system to $>2$ views by incrementally stitching new 
views on reconstructed views.

\vspace{0.75em}
\noindent
{\bf Acknowledgments} 
We thank Dandan Shan, Mohamed El Banani, Nilesh Kulkarni, Richard Higgins for helpful discussions.
Toyota Research Institute (``TRI'') provided funds to assist the authors with their research but 
this article solely reflects the opinions and conclusions of its authors and not TRI or any other Toyota entity.

{\small
\bibliographystyle{ieee_fullname}
\bibliography{local}

\begin{thebibliography}{10}\itemsep=-1pt

\bibitem{bao2012semantic}
Sid~Yingze Bao, Mohit Bagra, Yu-Wei Chao, and Silvio Savarese.
\newblock Semantic structure from motion with points, regions, and objects.
\newblock In {\em CVPR}, 2012.

\bibitem{BarrosoLaguna2019ICCV}
Axel Barroso-Laguna, Edgar Riba, Daniel Ponsa, and Krystian Mikolajczyk.
\newblock Key.net: Keypoint detection by handcrafted and learned cnn filters.
\newblock In {\em ICCV}, 2019.

\bibitem{cai2020messytable}
Zhongang Cai, Junzhe Zhang, Daxuan Ren, Cunjun Yu, Haiyu Zhao, Shuai Yi,
  Chai~Kiat Yeo, and Chen~Change Loy.
\newblock Messytable: Instance association in multiple camera views.
\newblock In {\em ECCV}, 2020.

\bibitem{chang2017matterport3d}
Angel Chang, Angela Dai, Thomas Funkhouser, Maciej Halber, Matthias Niessner,
  Manolis Savva, Shuran Song, Andy Zeng, and Yinda Zhang.
\newblock Matterport3d: Learning from rgb-d data in indoor environments.
\newblock In {\em 3DV}, 2017.

\bibitem{chen2020oasis}
Weifeng Chen, Shengyi Qian, David Fan, Noriyuki Kojima, Max Hamilton, and Jia
  Deng.
\newblock Oasis: A large-scale dataset for single image 3d in the wild.
\newblock In {\em CVPR}, 2020.

\bibitem{Choy20163d}
Christopher~B Choy, Danfei Xu, JunYoung Gwak, Kevin Chen, and Silvio Savarese.
\newblock 3d-r2n2: A unified approach for single and multi-view 3d object
  reconstruction.
\newblock In {\em ECCV}, 2016.

\bibitem{dai2017scannet}
Angela Dai, Angel~X Chang, Manolis Savva, Maciej Halber, Thomas Funkhouser, and
  Matthias Nie{\ss}ner.
\newblock Scannet: Richly-annotated 3d reconstructions of indoor scenes.
\newblock In {\em CVPR}, 2017.

\bibitem{dusmanu2019d2}
Mihai Dusmanu, Ignacio Rocco, Tomas Pajdla, Marc Pollefeys, Josef Sivic,
  Akihiko Torii, and Torsten Sattler.
\newblock D2-net: A trainable cnn for joint detection and description of local
  features.
\newblock {\em CVPR}, 2019.

\bibitem{Eigen15}
David Eigen and Rob Fergus.
\newblock Predicting depth, surface normals and semantic labels with a common
  multi-scale convolutional architecture.
\newblock In {\em ICCV}, 2015.

\bibitem{banani2020novel}
Mohamed El~Banani, Jason~J Corso, and David~F Fouhey.
\newblock Novel object viewpoint estimation through reconstruction alignment.
\newblock In {\em CVPR}, 2020.

\bibitem{banani2020unsupervisedrr}
Mohamed El~Banani, Luya Gao, and Justin Johnson.
\newblock Unsupervisedr\&r: Unsupervised point cloud registration via
  differentiable rendering.
\newblock In {\em CVPR}, 2021.

\bibitem{en2018rpnet}
Sovann En, Alexis Lechervy, and Fr{\'e}d{\'e}ric Jurie.
\newblock Rpnet: An end-to-end network for relative camera pose estimation.
\newblock In {\em ECCV}, 2018.

\bibitem{Fouhey13}
David~F. Fouhey, Abhinav Gupta, and Martial Hebert.
\newblock Data-driven {3D} primitives for single image understanding.
\newblock In {\em ICCV}, 2013.

\bibitem{furukawa2009manhattan}
Yasutaka Furukawa, Brian Curless, Steven~M Seitz, and Richard Szeliski.
\newblock Manhattan-world stereo.
\newblock In {\em CVPR}, 2009.

\bibitem{gallup2010piecewise}
David Gallup, Jan-Michael Frahm, and Marc Pollefeys.
\newblock Piecewise planar and non-planar stereo for urban scene
  reconstruction.
\newblock In {\em CVPR}, 2010.

\bibitem{Girdhar16b}
R. Girdhar, D.F. Fouhey, M. Rodriguez, and A. Gupta.
\newblock Learning a predictable and generative vector representation for
  objects.
\newblock In {\em ECCV}, 2016.

\bibitem{Hadfield15}
Simon Hadfield and Richard Bowden.
\newblock Exploiting high level scene cues in stereo reconstruction.
\newblock In {\em ICCV}, 2015.

\bibitem{Hartley04}
R.~I. Hartley and A. Zisserman.
\newblock {\em Multiple View Geometry in Computer Vision}.
\newblock Cambridge University Press, ISBN: 0521540518, 2004.

\bibitem{he2017mask}
Kaiming He, Georgia Gkioxari, Piotr Doll{\'a}r, and Ross Girshick.
\newblock Mask r-cnn.
\newblock In {\em ICCV}, 2017.

\bibitem{he2016deep}
Kaiming He, Xiangyu Zhang, Shaoqing Ren, and Jian Sun.
\newblock Deep residual learning for image recognition.
\newblock In {\em CVPR}, 2016.

\bibitem{huang2018deepmvs}
Po-Han Huang, Kevin Matzen, Johannes Kopf, Narendra Ahuja, and Jia-Bin Huang.
\newblock Deepmvs: Learning multi-view stereopsis.
\newblock In {\em CVPR}, 2018.

\bibitem{jiang2020peek}
Ziyu Jiang, Buyu Liu, Samuel Schulter, Zhangyang Wang, and Manmohan Chandraker.
\newblock Peek-a-boo: Occlusion reasoning in indoor scenes with plane
  representations.
\newblock In {\em CVPR}, 2020.

\bibitem{kar2017learning}
Abhishek Kar, Christian H{\"a}ne, and Jitendra Malik.
\newblock Learning a multi-view stereo machine.
\newblock In {\em NeurIPS}, 2017.

\bibitem{kulkarni2019relnet}
Nilesh Kulkarni, Ishan Misra, Shubham Tulsiani, and Abhinav Gupta.
\newblock 3d-relnet: Joint object and relational network for 3d prediction.
\newblock In {\em ICCV}, 2019.

\bibitem{LiSilhouette19}
Lin Li, Salman Khan, and Nick Barnes.
\newblock Silhouette-assisted 3d object instance reconstruction from a
  cluttered scene.
\newblock In {\em ICCV Workshops}, 2019.

\bibitem{lin2017feature}
Tsung-Yi Lin, Piotr Doll{\'a}r, Ross Girshick, Kaiming He, Bharath Hariharan,
  and Serge Belongie.
\newblock Feature pyramid networks for object detection.
\newblock In {\em CVPR}, 2017.

\bibitem{lin2014microsoft}
Tsung-Yi Lin, Michael Maire, Serge Belongie, James Hays, Pietro Perona, Deva
  Ramanan, Piotr Doll{\'a}r, and C~Lawrence Zitnick.
\newblock Microsoft coco: Common objects in context.
\newblock In {\em ECCV}, 2014.

\bibitem{liu2019planercnn}
Chen Liu, Kihwan Kim, Jinwei Gu, Yasutaka Furukawa, and Jan Kautz.
\newblock Planercnn: 3d plane detection and reconstruction from a single image.
\newblock In {\em CVPR}, 2019.

\bibitem{liu2018planenet}
Chen Liu, Jimei Yang, Duygu Ceylan, Ersin Yumer, and Yasutaka Furukawa.
\newblock Planenet: Piece-wise planar reconstruction from a single rgb image.
\newblock In {\em CVPR}, 2018.

\bibitem{lowe2004distinctive}
David~G Lowe.
\newblock Distinctive image features from scale-invariant keypoints.
\newblock {\em IJCV}, 2004.

\bibitem{mildenhall2020nerf}
Ben Mildenhall, Pratul~P. Srinivasan, Matthew Tancik, Jonathan~T. Barron, Ravi
  Ramamoorthi, and Ren Ng.
\newblock Nerf: Representing scenes as neural radiance fields for view
  synthesis.
\newblock In {\em ECCV}, 2020.

\bibitem{HardNet2017}
Anastasiya Mishchuk, Dmytro Mishkin, Filip Radenovic, and Jiri Matas.
\newblock Working hard to know your neighbor's margins: Local descriptor
  learning loss.
\newblock In {\em NeurIPS}, 2017.

\bibitem{mishkin2015wxbs}
Dmytro Mishkin, Jiri Matas, Michal Perdoch, and Karel Lenc.
\newblock Wxbs: Wide baseline stereo generalizations.
\newblock {\em BMVC}, 2015.

\bibitem{nie2020total3dunderstanding}
Yinyu Nie, Xiaoguang Han, Shihui Guo, Yujian Zheng, Jian Chang, and Jian~Jun
  Zhang.
\newblock Total3dunderstanding: Joint layout, object pose and mesh
  reconstruction for indoor scenes from a single image.
\newblock In {\em CVPR}, 2020.

\bibitem{poursaeed2018deep}
Omid Poursaeed, Guandao Yang, Aditya Prakash, Qiuren Fang, Hanqing Jiang,
  Bharath Hariharan, and Serge Belongie.
\newblock Deep fundamental matrix estimation without correspondences.
\newblock In {\em ECCV}, 2018.

\bibitem{Pritchett98a}
Phil Pritchett and Andrew Zisserman.
\newblock Wide baseline stereo matching.
\newblock In {\em ICCV}, 1998.

\bibitem{Qian2020}
Shengyi Qian, Linyi Jin, and David~F. Fouhey.
\newblock Associative3d: Volumetric reconstruction from sparse views.
\newblock In {\em ECCV}, 2020.

\bibitem{ranftl2018deep}
Ren{\'e} Ranftl and Vladlen Koltun.
\newblock Deep fundamental matrix estimation.
\newblock In {\em ECCV}, 2018.

\bibitem{Ranftl2020}
Ren\'{e} Ranftl, Katrin Lasinger, David Hafner, Konrad Schindler, and Vladlen
  Koltun.
\newblock Towards robust monocular depth estimation: Mixing datasets for
  zero-shot cross-dataset transfer.
\newblock {\em TPAMI}, 2020.

\bibitem{raposo2016pi}
Carolina Raposo and Joao~P Barreto.
\newblock {$\pi$Match}: Monocular vslam and piecewise planar reconstruction
  using fast plane correspondences.
\newblock In {\em ECCV}, 2016.

\bibitem{raposo2013plane}
Carolina Raposo, Miguel Louren{\c{c}}o, Michel Antunes, and Jo{\~a}o~Pedro
  Barreto.
\newblock Plane-based odometry using an rgb-d camera.
\newblock In {\em BMVC}, 2013.

\bibitem{rocco2018neighbourhood}
Ignacio Rocco, Mircea Cimpoi, Relja Arandjelovi{\'c}, Akihiko Torii, Tomas
  Pajdla, and Josef Sivic.
\newblock Neighbourhood consensus networks.
\newblock In {\em NeurIPS}, 2018.

\bibitem{sarlin2020superglue}
Paul-Edouard Sarlin, Daniel DeTone, Tomasz Malisiewicz, and Andrew Rabinovich.
\newblock Superglue: Learning feature matching with graph neural networks.
\newblock In {\em CVPR}, 2020.

\bibitem{savva2019habitat}
Manolis Savva, Abhishek Kadian, Oleksandr Maksymets, Yili Zhao, Erik Wijmans,
  Bhavana Jain, Julian Straub, Jia Liu, Vladlen Koltun, Jitendra Malik, et~al.
\newblock Habitat: A platform for embodied ai research.
\newblock In {\em ICCV}, 2019.

\bibitem{Scharstein02}
Daniel Scharstein and Richard Szeliski.
\newblock A taxonomy and evaluation of dense two-frame stereo correspondence
  algorithms.
\newblock {\em IJCV}, 2002.

\bibitem{schonberger2016structure}
Johannes~L Schonberger and Jan-Michael Frahm.
\newblock Structure-from-motion revisited.
\newblock In {\em CVPR}, 2016.

\bibitem{schops2019bad}
Thomas Schops, Torsten Sattler, and Marc Pollefeys.
\newblock Bad slam: Bundle adjusted direct rgb-d slam.
\newblock In {\em CVPR}, 2019.

\bibitem{schroff2015facenet}
Florian Schroff, Dmitry Kalenichenko, and James Philbin.
\newblock Facenet: A unified embedding for face recognition and clustering.
\newblock In {\em CVPR}, 2015.

\bibitem{sinha2009piecewise}
Sudipta~N Sinha, Drew Steedly, and Richard Szeliski.
\newblock Piecewise planar stereo for image-based rendering.
\newblock In {\em ICCV}, 2009.

\bibitem{song2017semantic}
Shuran Song, Fisher Yu, Andy Zeng, Angel~X Chang, Manolis Savva, and Thomas
  Funkhouser.
\newblock Semantic scene completion from a single depth image.
\newblock In {\em CVPR}, 2017.

\bibitem{straub2019replica}
Julian Straub, Thomas Whelan, Lingni Ma, Yufan Chen, Erik Wijmans, Simon Green,
  Jakob~J Engel, Raul Mur-Artal, Carl Ren, Shobhit Verma, et~al.
\newblock The replica dataset: A digital replica of indoor spaces.
\newblock {\em arXiv:1906.05797}, 2019.

\bibitem{su2015render}
Hao Su, Charles~R Qi, Yangyan Li, and Leonidas~J Guibas.
\newblock Render for cnn: Viewpoint estimation in images using cnns trained
  with rendered 3d model views.
\newblock In {\em ICCV}, 2015.

\bibitem{SunHSC19}
Cheng Sun, Chi{-}Wei Hsiao, Min Sun, and Hwann{-}Tzong Chen.
\newblock Horizonnet: Learning room layout with 1d representation and pano
  stretch data augmentation.
\newblock In {\em CVPR}, 2019.

\bibitem{tulsiani2018factoring}
Shubham Tulsiani, Saurabh Gupta, David~F Fouhey, Alexei~A Efros, and Jitendra
  Malik.
\newblock Factoring shape, pose, and layout from the 2d image of a 3d scene.
\newblock In {\em CVPR}, 2018.

\bibitem{tulsiani2015viewpoints}
Shubham Tulsiani and Jitendra Malik.
\newblock Viewpoints and keypoints.
\newblock In {\em CVPR}, 2015.

\bibitem{ummenhofer2017demon}
Benjamin Ummenhofer, Huizhong Zhou, Jonas Uhrig, Nikolaus Mayer, Eddy Ilg,
  Alexey Dosovitskiy, and Thomas Brox.
\newblock Demon: Depth and motion network for learning monocular stereo.
\newblock In {\em CVPR}, 2017.

\bibitem{vaswani2017attention}
Ashish Vaswani, Noam Shazeer, Niki Parmar, Jakob Uszkoreit, Llion Jones,
  Aidan~N Gomez, {\L}ukasz Kaiser, and Illia Polosukhin.
\newblock Attention is all you need.
\newblock In {\em NeurIPS}, 2017.

\bibitem{Wang15}
Xiaolong Wang, David~F. Fouhey, and Abhinav Gupta.
\newblock Designing deep networks for surface normal estimation.
\newblock In {\em CVPR}, 2015.

\bibitem{wang2019learning}
Xiaolong Wang, Allan Jabri, and Alexei~A Efros.
\newblock Learning correspondence from the cycle-consistency of time.
\newblock In {\em CVPR}, 2019.

\bibitem{whelan2013robust}
Thomas Whelan, Hordur Johannsson, Michael Kaess, John~J Leonard, and John
  McDonald.
\newblock Robust real-time visual odometry for dense rgb-d mapping.
\newblock In {\em ICRA}, 2013.

\bibitem{wu20083d}
Changchang Wu, Brian Clipp, Xiaowei Li, Jan-Michael Frahm, and Marc Pollefeys.
\newblock 3d model matching with viewpoint-invariant patches (vip).
\newblock In {\em CVPR}, 2008.

\bibitem{wu2019detectron2}
Yuxin Wu, Alexander Kirillov, Francisco Massa, Wan-Yen Lo, and Ross Girshick.
\newblock Detectron2.
\newblock \url{https://github.com/facebookresearch/detectron2}, 2019.

\bibitem{xiao2013sun3d}
Jianxiong Xiao, Andrew Owens, and Antonio Torralba.
\newblock Sun3d: A database of big spaces reconstructed using sfm and object
  labels.
\newblock In {\em ICCV}, 2013.

\bibitem{yang2018recovering}
Fengting Yang and Zihan Zhou.
\newblock Recovering 3d planes from a single image via convolutional neural
  networks.
\newblock In {\em ECCV}, 2018.

\bibitem{Yang:2019:DuLa-Net}
Shang-Ta Yang, Fu-En Wang, Chi-Han Peng, Peter Wonka, Min Sun, and Hung-Kuo
  Chu.
\newblock Dula-net: {A} dual-projection network for estimating room layouts
  from a single {RGB} panorama.
\newblock In {\em CVPR}, 2019.

\bibitem{yang2020extreme}
Zhenpei Yang, Siming Yan, and Qixing Huang.
\newblock Extreme relative pose network under hybrid representations.
\newblock In {\em CVPR}, 2020.

\bibitem{Wei2021CVPR}
Wei Yin, Jianming Zhang, Oliver Wang, Simon Niklaus, Long Mai, Simon Chen, and
  Chunhua Shen.
\newblock Learning to recover 3d scene shape from a single image.
\newblock In {\em CVPR}, 2021.

\bibitem{YuZLZG19}
Zehao Yu, Jia Zheng, Dongze Lian, Zihan Zhou, and Shenghua Gao.
\newblock Single-image piece-wise planar 3d reconstruction via associative
  embedding.
\newblock In {\em CVPR}, 2019.

\bibitem{zhang2017physically}
Yinda Zhang, Shuran Song, Ersin Yumer, Manolis Savva, Joon-Young Lee, Hailin
  Jin, and Thomas Funkhouser.
\newblock Physically-based rendering for indoor scene understanding using
  convolutional neural networks.
\newblock In {\em CVPR}, 2017.

\bibitem{zhou2019continuity}
Yi Zhou, Connelly Barnes, Jingwan Lu, Jimei Yang, and Hao Li.
\newblock On the continuity of rotation representations in neural networks.
\newblock In {\em CVPR}, 2019.

\bibitem{zou2018layoutnet}
Chuhang Zou, Alex Colburn, Qi Shan, and Derek Hoiem.
\newblock Layoutnet: Reconstructing the 3d room layout from a single rgb image.
\newblock In {\em CVPR}, 2018.

\bibitem{zou20193d}
Chuhang Zou, Jheng-Wei Su, Chi-Han Peng, Alex Colburn, Qi Shan, Peter Wonka,
  Hung-Kuo Chu, and Derek Hoiem.
\newblock Manhattan room layout reconstruction from a single 360 image: A
  comparative study of state-of-the-art methods.
\newblock {\em IJCV}, 2021.

\end{thebibliography}
}
\clearpage
\appendix
\section{Additional Results}
\subsection{Dataset}

\begin{figure*}[ht]
    \vspace{-3em}
    \centering
    \subfloat[Ours.]{\includegraphics[width=0.3\textwidth]{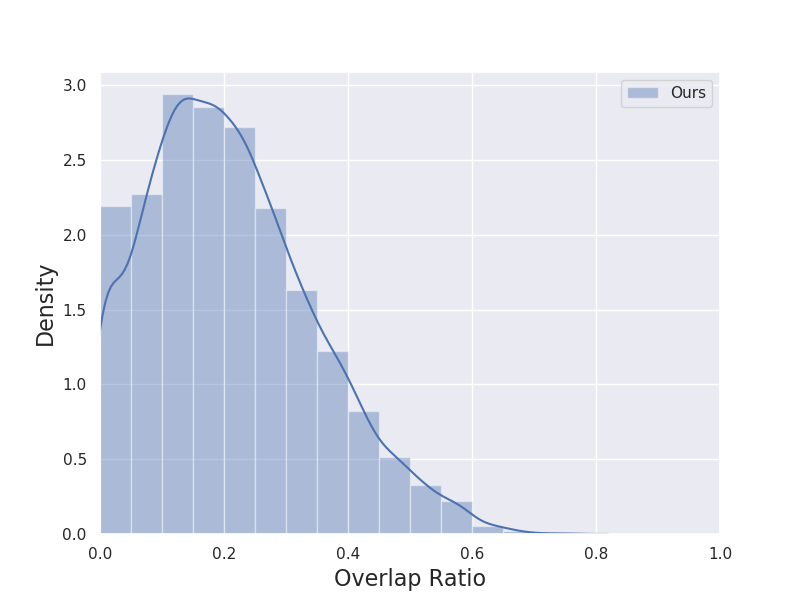}}
    \hfill
    \subfloat[DeMoN.]{\includegraphics[width=0.3\textwidth]{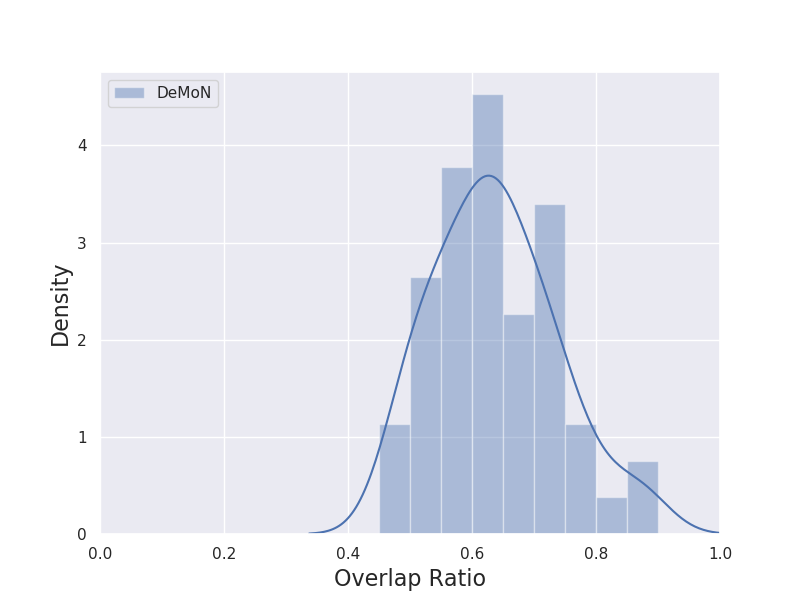}}
    \hfill
    \subfloat[PlaneOdometry.]{\includegraphics[width=0.3\textwidth]{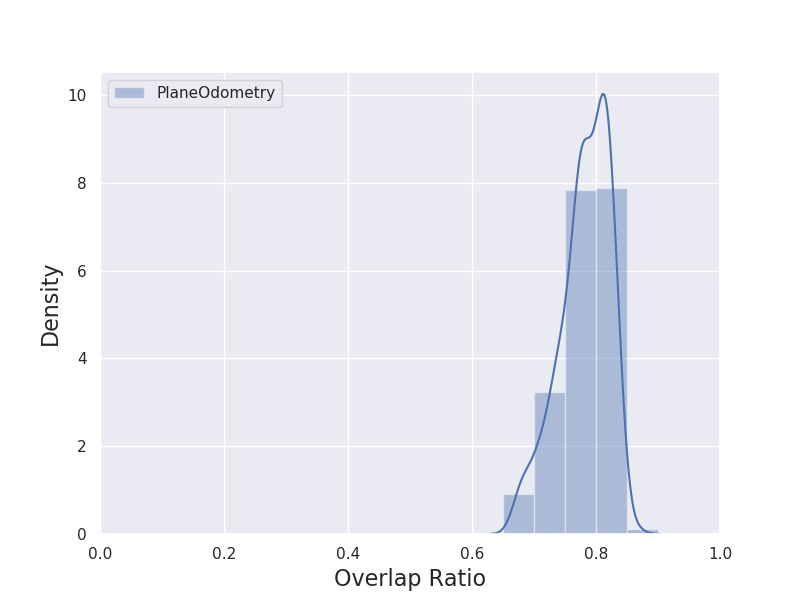}}
    \caption{Histograms of overlap ratio for different datasets.}
    \label{fig:overlap-histogram}
\end{figure*}

To give additional context for the dataset, we transform the point cloud of view 1 to view 2. 
The average percentage of overlapping points of our dataset is $21\%$. For context,
Sun3D \cite{xiao2013sun3d} image pairs used by DeMoN \cite{ummenhofer2017demon} had
an average overlap of $64\%$; FR2\_desk and FR3\_structure used by Plane Odometry~\cite{raposo2013plane} had an overlap of $78\%$.
Figure~\ref{fig:overlap-histogram} shows the histogram of overlap ratio of our dataset versus the other datasets.

\begin{figure*}[ht]
    \centering
    \begin{tabular}{cc}
    \toprule

    Dataset & 
    \begin{tabular}{m{2.3cm}m{2.3cm}m{2.3cm}m{2.3cm}m{2.3cm}m{2.3cm}}
        Image 1 & Image 2 & Overlapping & Image 1  & Image 2 & Overlapping 
    \end{tabular}\\
    \midrule
    Ours &

    \begin{tabular}{m{2.3cm}m{2.3cm}m{2.3cm}m{2.3cm}m{2.3cm}m{2.3cm}}
    \frame{\includegraphics[width=0.15\textwidth]{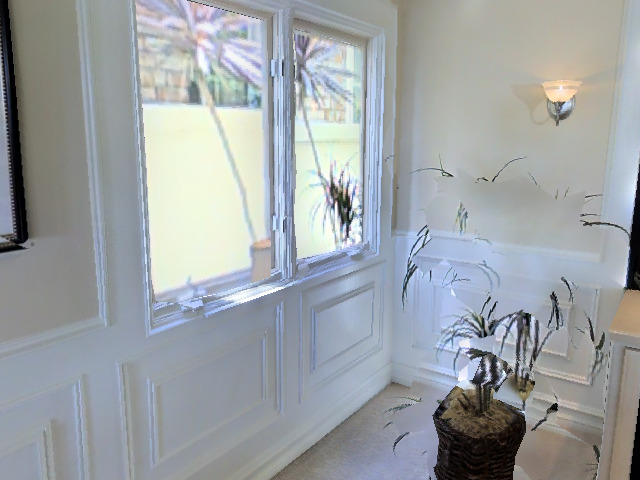}}
    & \frame{\includegraphics[width=0.15\textwidth]{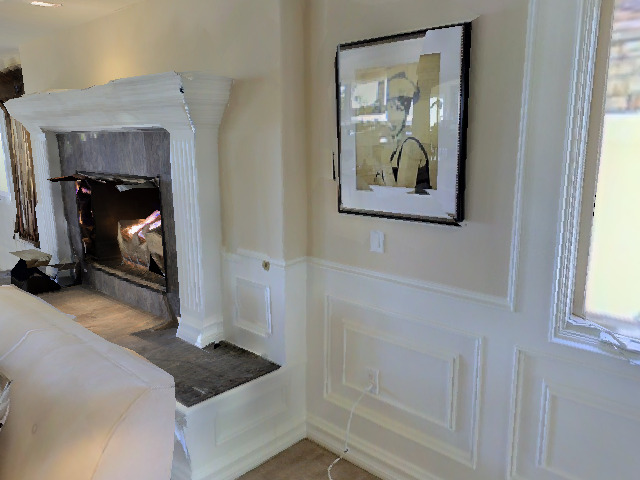}}
    & \frame{\includegraphics[width=0.15\textwidth]{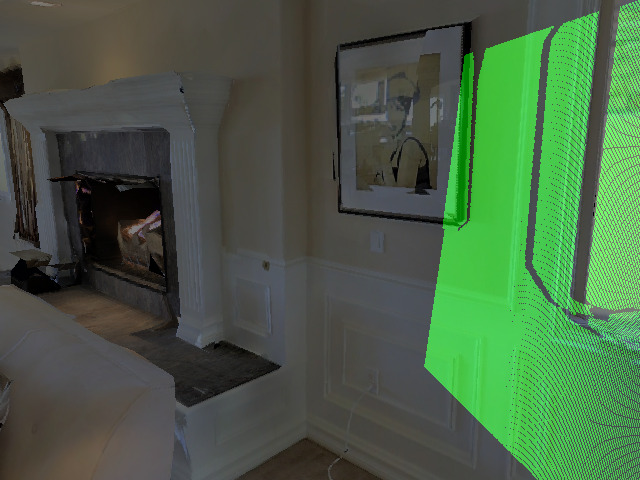}}
    & \frame{\includegraphics[width=0.15\textwidth]{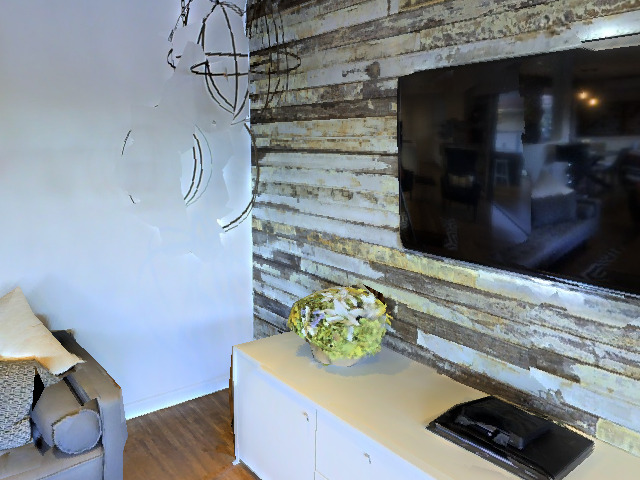}}
    & \frame{\includegraphics[width=0.15\textwidth]{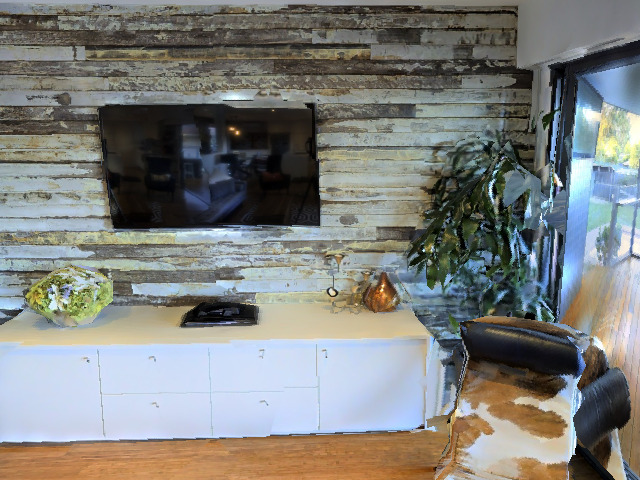}}
    & \frame{\includegraphics[width=0.15\textwidth]{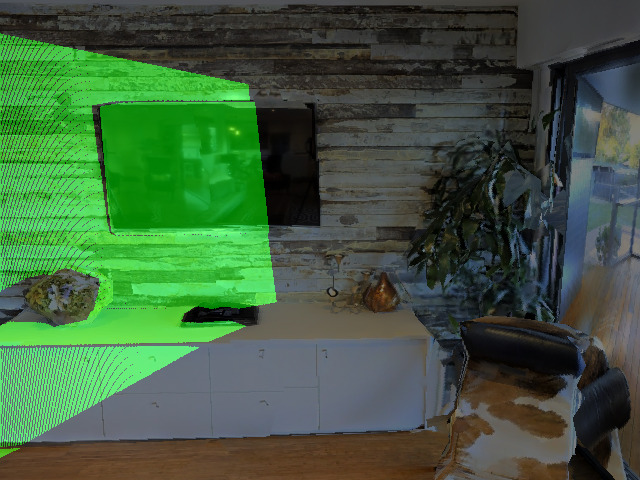}}\\

    \frame{\includegraphics[width=0.15\textwidth]{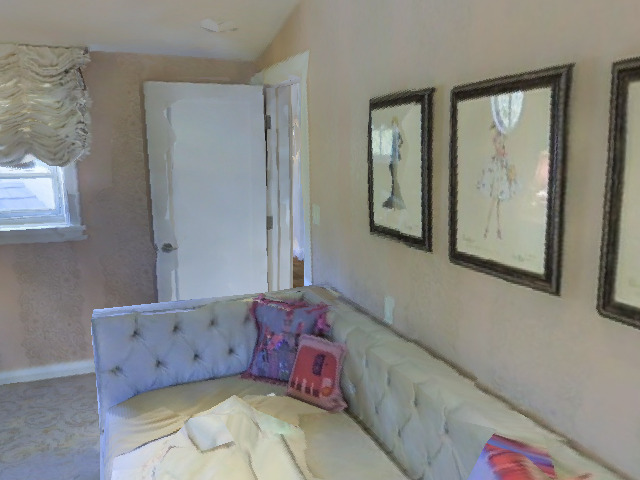}}
    & \frame{\includegraphics[width=0.15\textwidth]{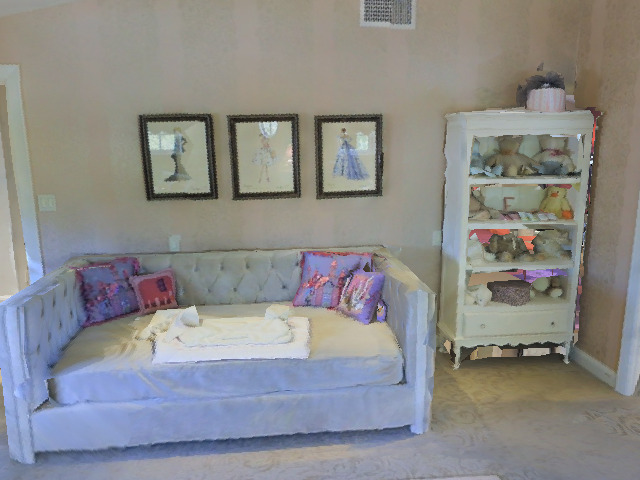}}
    & \frame{\includegraphics[width=0.15\textwidth]{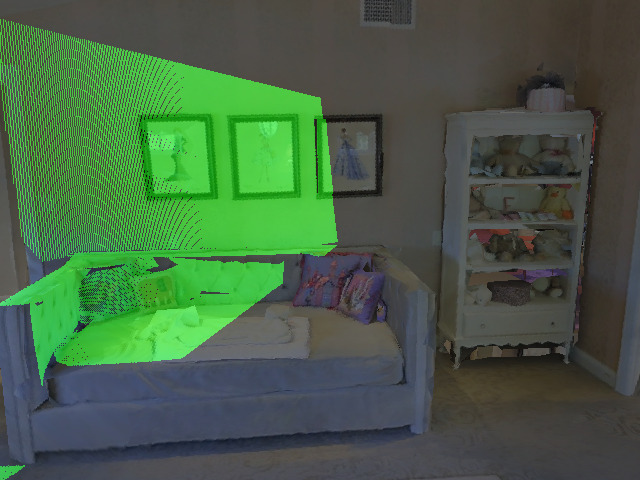}}
    & \frame{\includegraphics[width=0.15\textwidth]{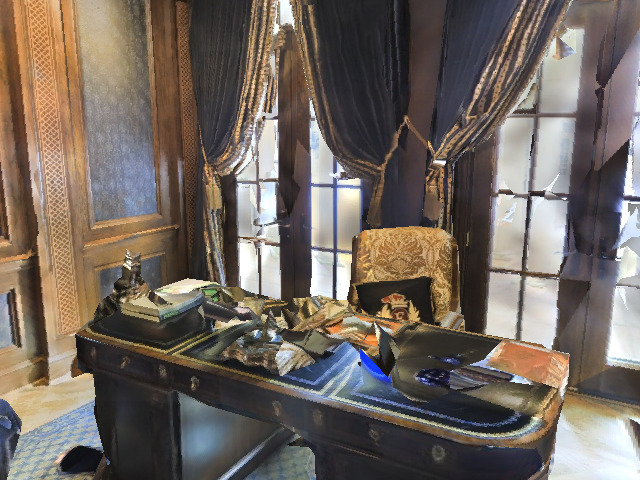}}
    & \frame{\includegraphics[width=0.15\textwidth]{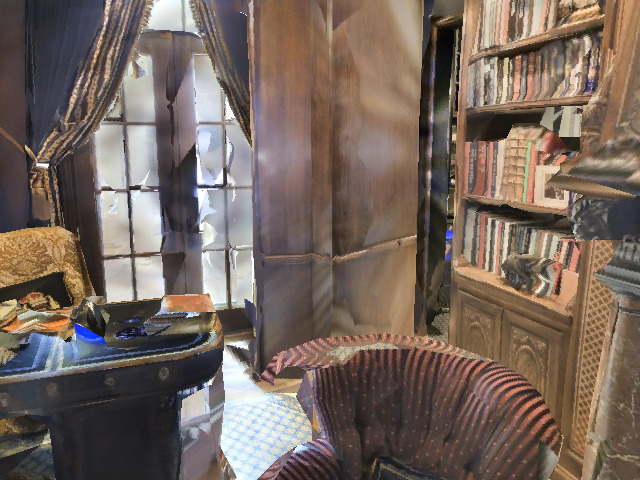}}
    & \frame{\includegraphics[width=0.15\textwidth]{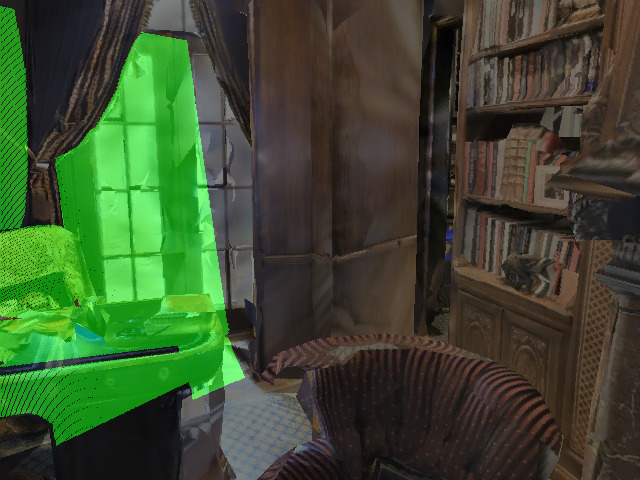}}\\

    \frame{\includegraphics[width=0.15\textwidth]{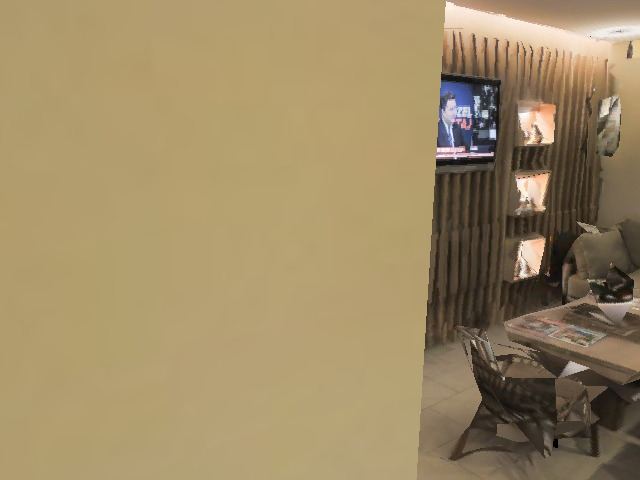}}
    & \frame{\includegraphics[width=0.15\textwidth]{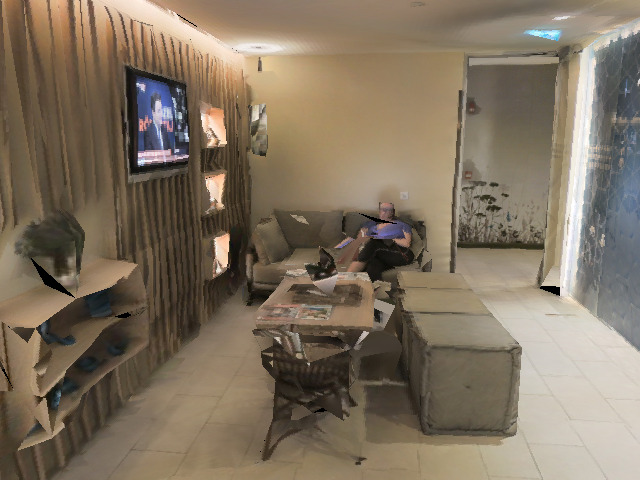}}
    & \frame{\includegraphics[width=0.15\textwidth]{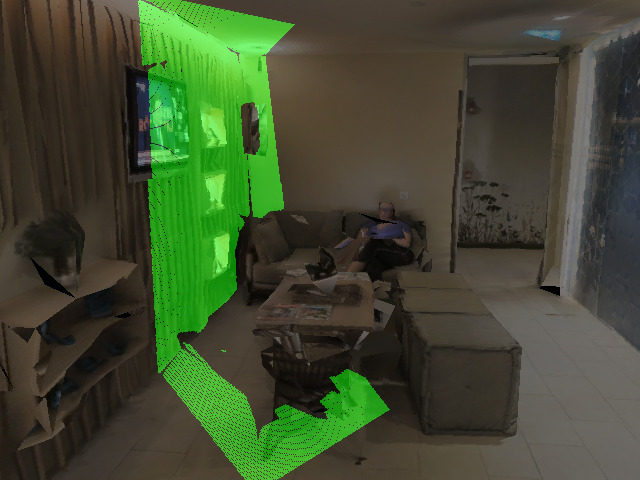}}
    & \frame{\includegraphics[width=0.15\textwidth]{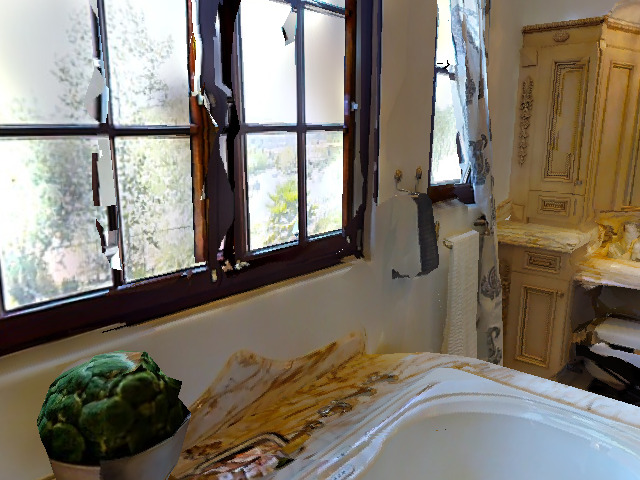}}
    & \frame{\includegraphics[width=0.15\textwidth]{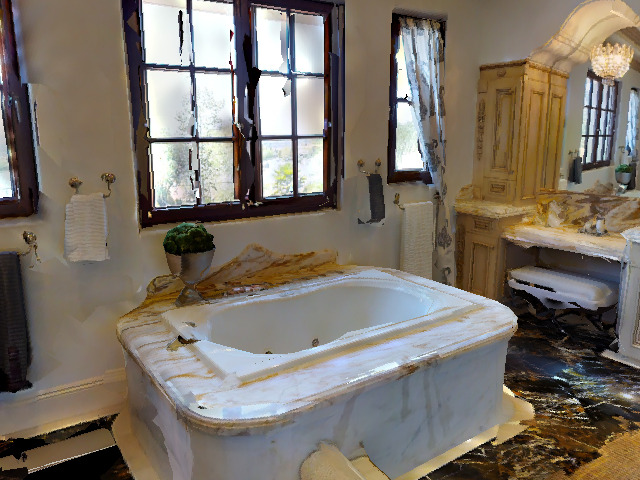}}
    & \frame{\includegraphics[width=0.15\textwidth]{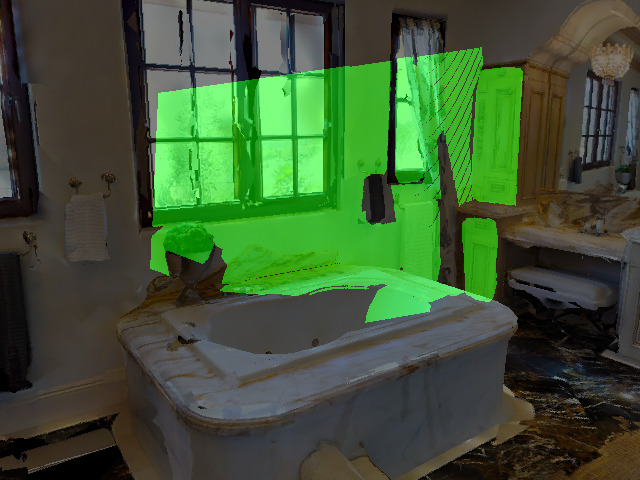}}\\

    \frame{\includegraphics[width=0.15\textwidth]{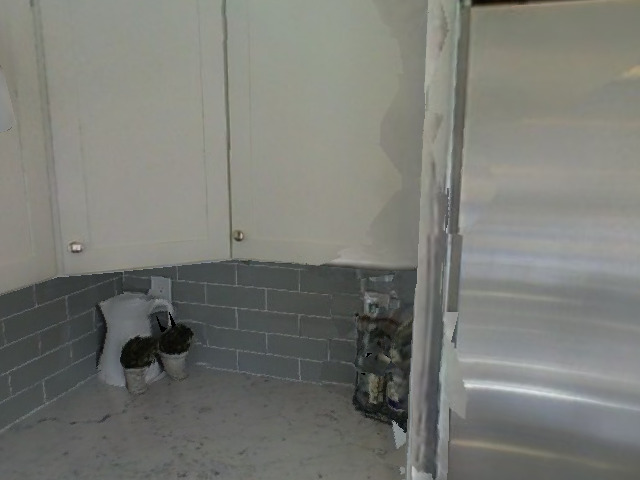}}
    & \frame{\includegraphics[width=0.15\textwidth]{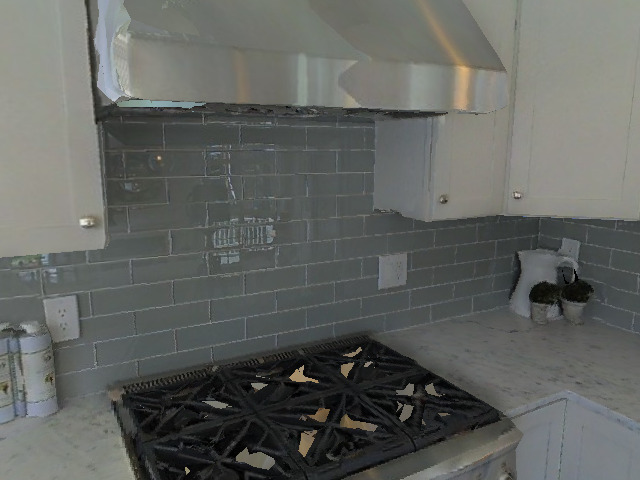}}
    & \frame{\includegraphics[width=0.15\textwidth]{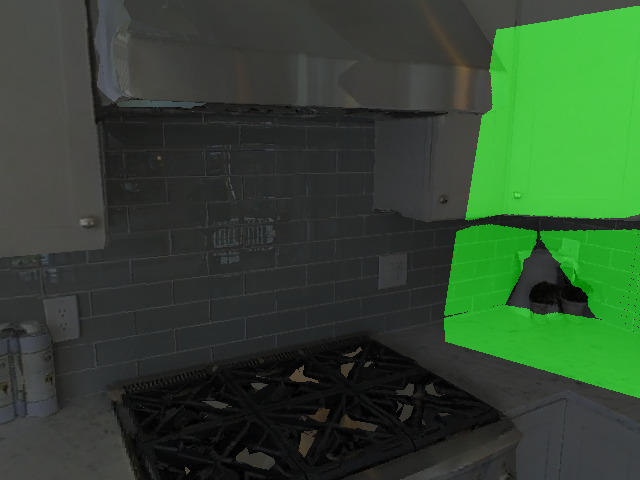}}
    & \frame{\includegraphics[width=0.15\textwidth]{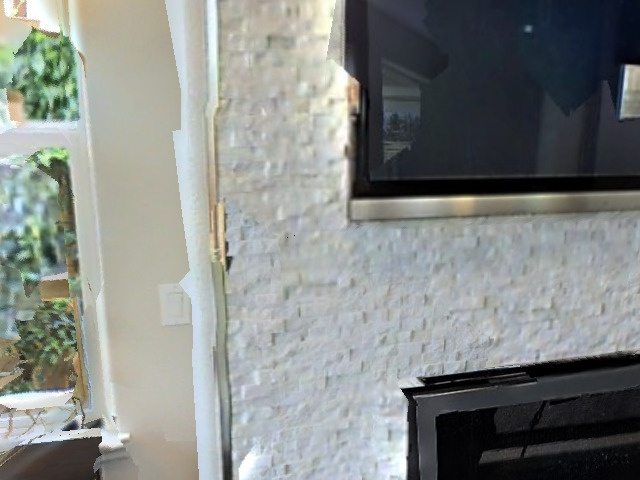}}
    & \frame{\includegraphics[width=0.15\textwidth]{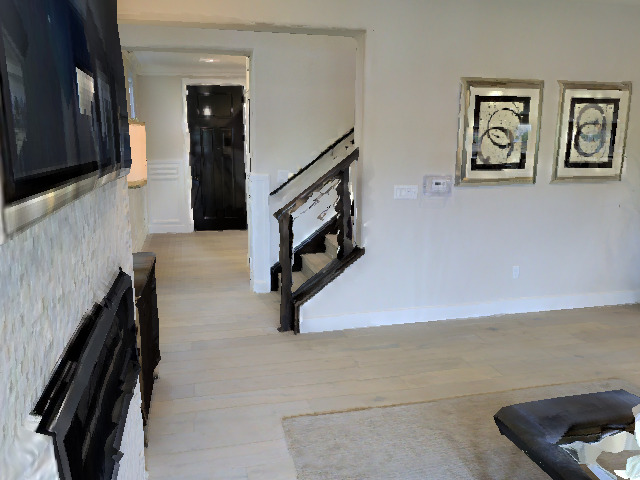}}
    & \frame{\includegraphics[width=0.15\textwidth]{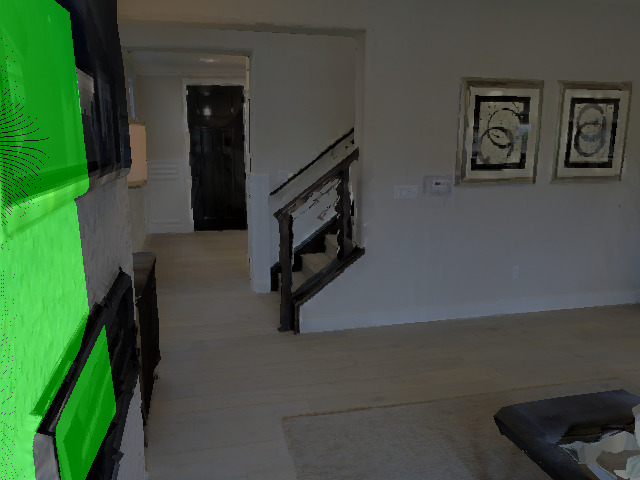}}
    \end{tabular}\\

    \midrule
    \cite{ummenhofer2017demon} &
    \begin{tabular}{m{2.3cm}m{2.3cm}m{2.3cm}m{2.3cm}m{2.3cm}m{2.3cm}}
    \frame{\includegraphics[width=0.15\textwidth]{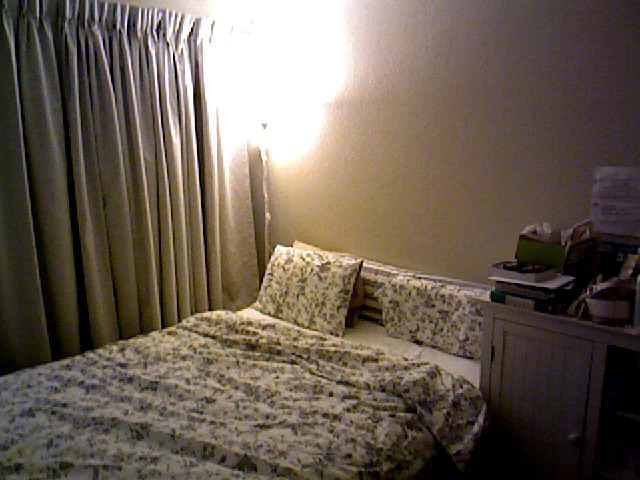}}
    & \frame{\includegraphics[width=0.15\textwidth]{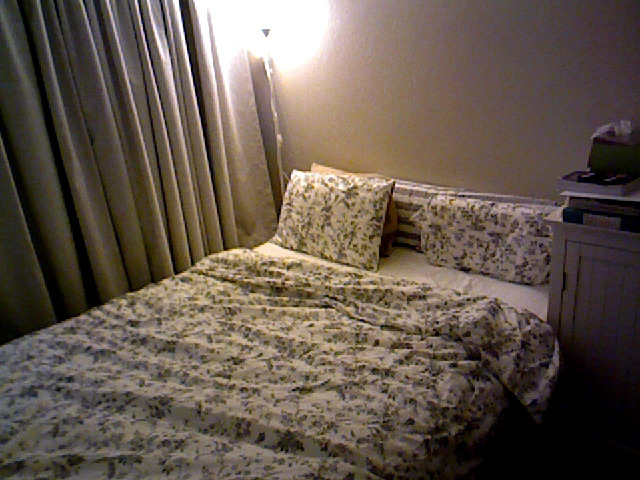}}
    & \frame{\includegraphics[width=0.15\textwidth]{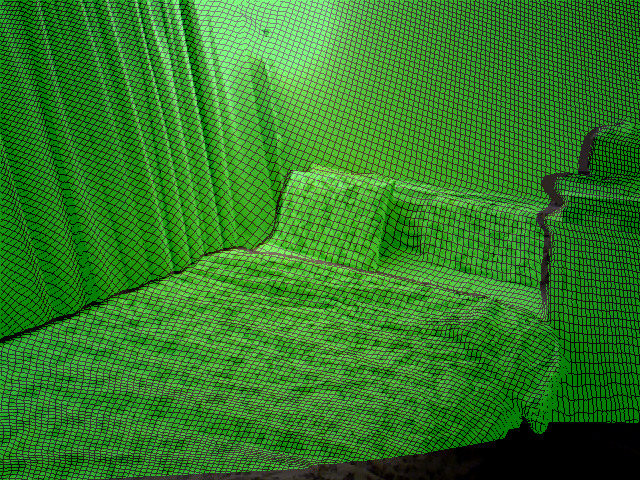}}
    & \frame{\includegraphics[width=0.15\textwidth]{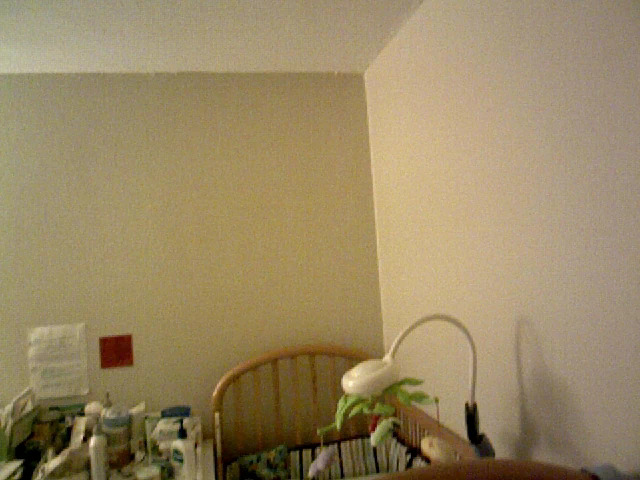}}
    & \frame{\includegraphics[width=0.15\textwidth]{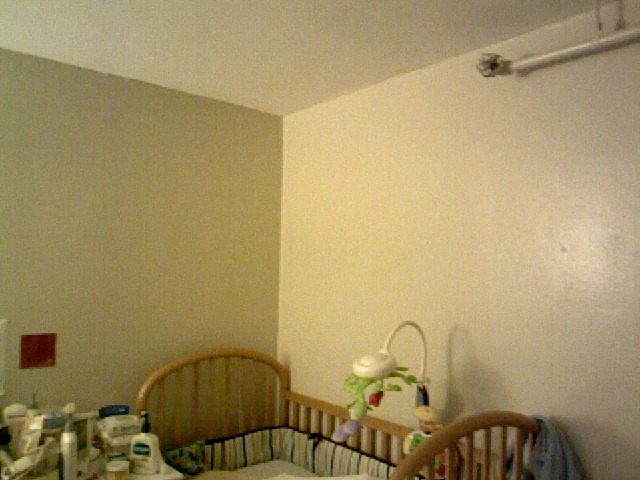}}
    & \frame{\includegraphics[width=0.15\textwidth]{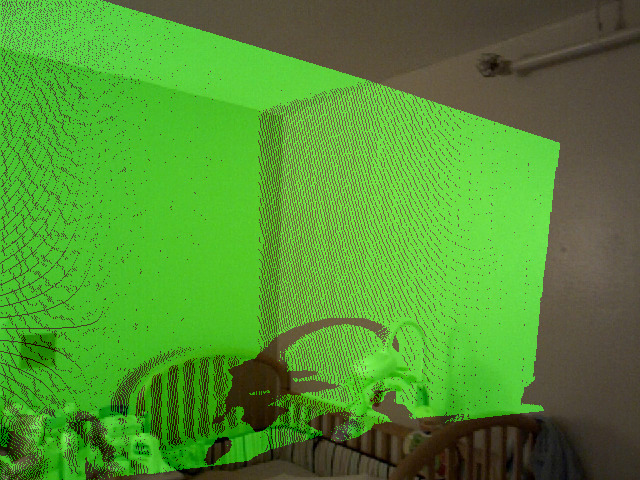}}\\

    \frame{\includegraphics[width=0.15\textwidth]{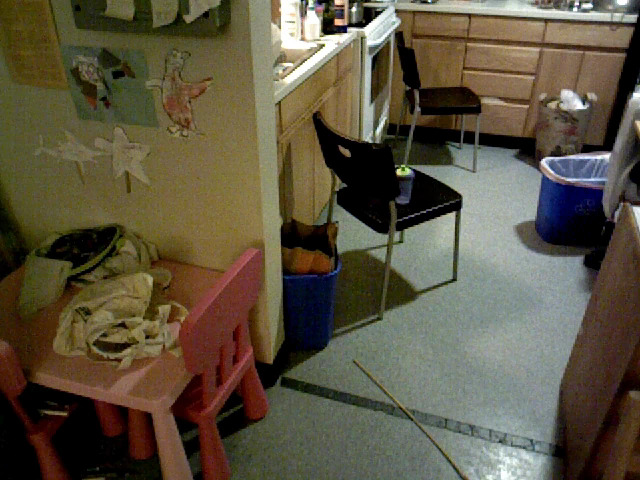}}
    & \frame{\includegraphics[width=0.15\textwidth]{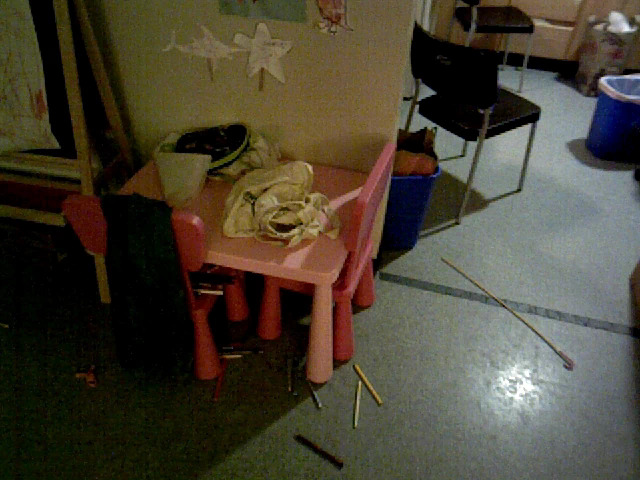}}
    & \frame{\includegraphics[width=0.15\textwidth]{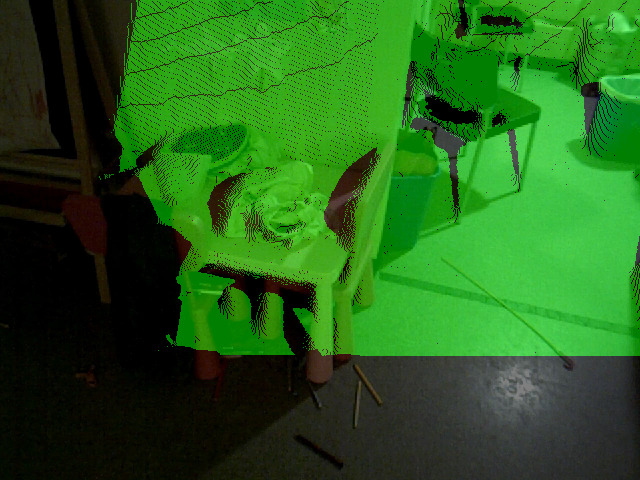}}
    & \frame{\includegraphics[width=0.15\textwidth]{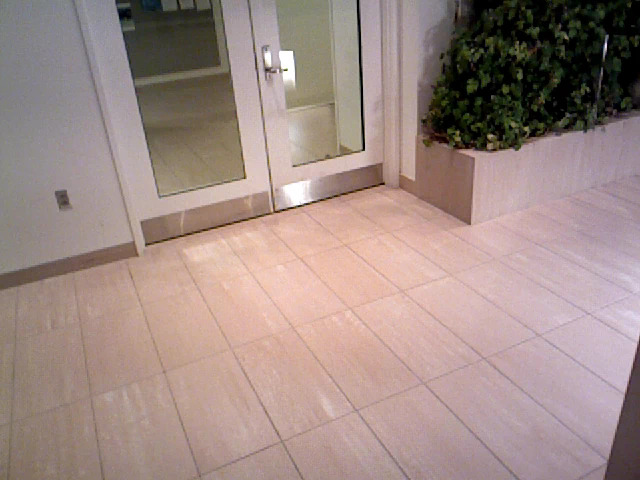}}
    & \frame{\includegraphics[width=0.15\textwidth]{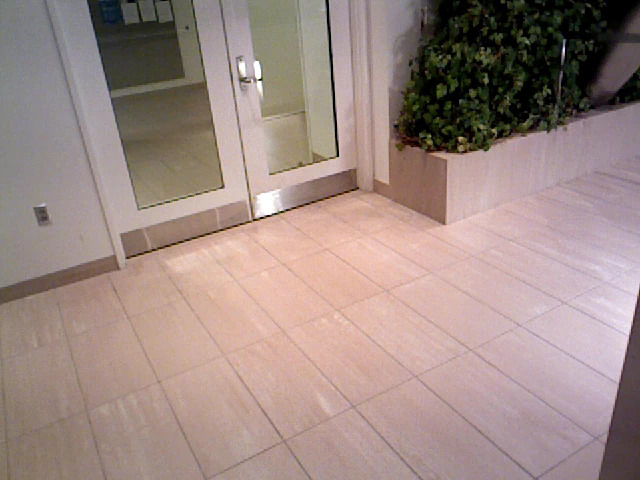}}
    & \frame{\includegraphics[width=0.15\textwidth]{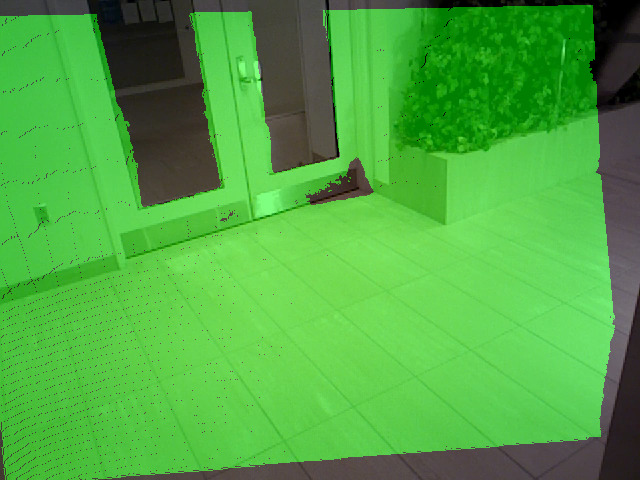}}
    \end{tabular}\\

    \midrule

    \cite{raposo2013plane} &
    \begin{tabular}{m{2.3cm}m{2.3cm}m{2.3cm}m{2.3cm}m{2.3cm}m{2.3cm}}
    \frame{\includegraphics[width=0.15\textwidth]{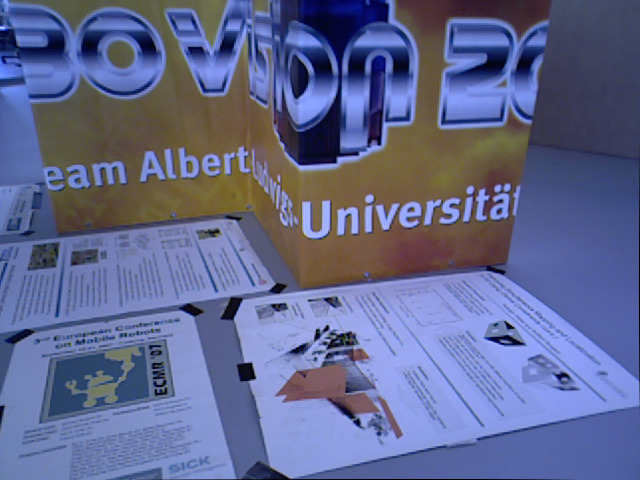}}
    & \frame{\includegraphics[width=0.15\textwidth]{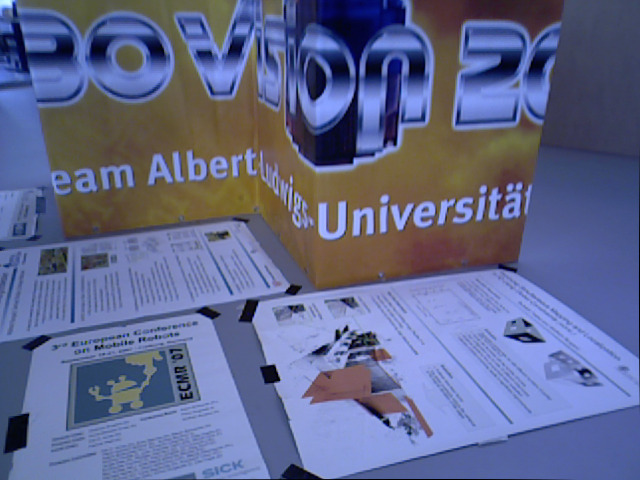}}
    & \frame{\includegraphics[width=0.15\textwidth]{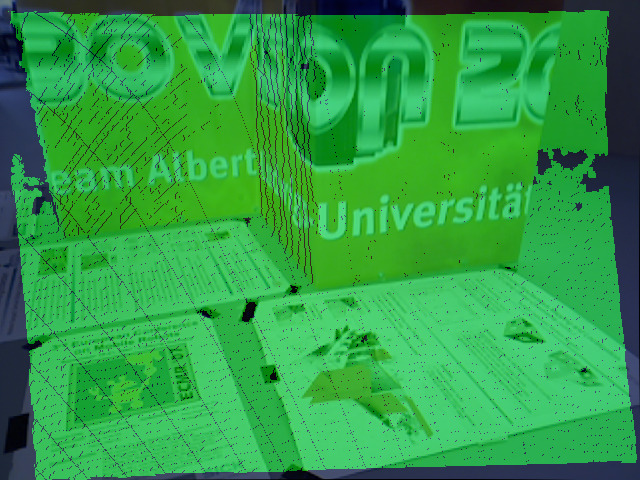}}
    & \frame{\includegraphics[width=0.15\textwidth]{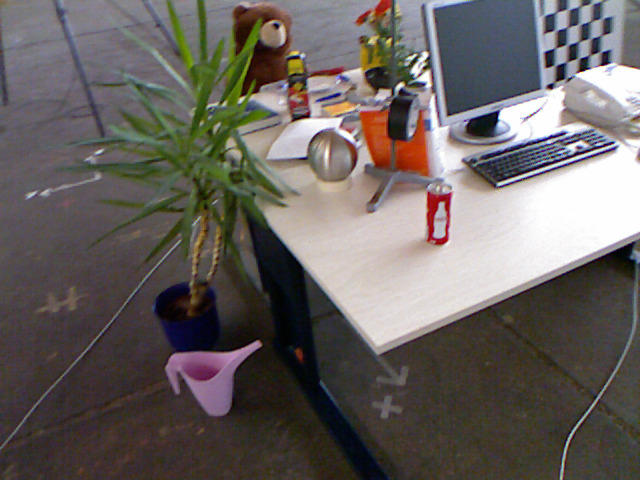}}
    & \frame{\includegraphics[width=0.15\textwidth]{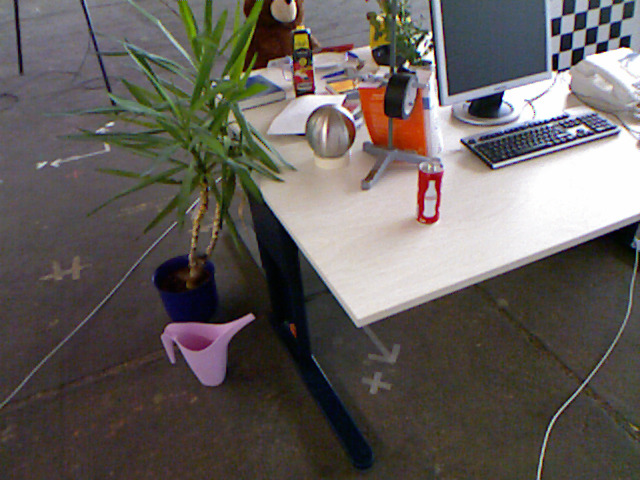}}
    & \frame{\includegraphics[width=0.15\textwidth]{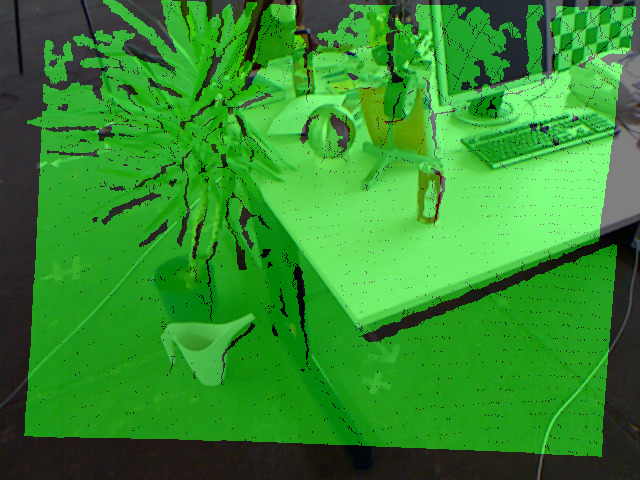}}
    \end{tabular}\\

    \bottomrule
    \end{tabular}
    \caption{Overlapping regions of random examples of Our dataset (row 1-4), DeMoN \cite{ummenhofer2017demon} (row 5-6) and 
    Plane Odometry \cite{raposo2013plane} (row 7). We transform the point cloud of image 1 to image 2 and show it as a green mask.}
    \label{fig:overlap-wall-compare}
\end{figure*}

Random examples are shown in Figure \ref{fig:overlap-wall-compare}. 
Our dataset is much more sparse than datasets used in DeMoN and Plane Odometry, 
because they sample adjacent frames from a video, while we render a sparse view dataset ourselves.

\subsection{Performance vs. Overlapping Ratio}
\label{sec:ap-overlap}
\begin{figure}[t!]
    \vspace{-1.5em}
    \centering
    \includegraphics[width=\columnwidth]{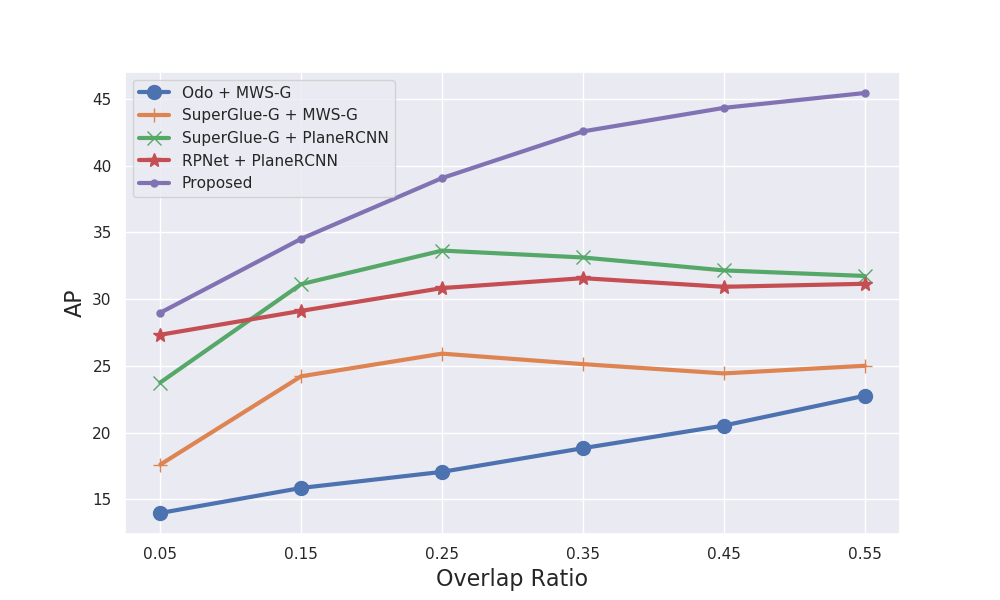}
    \caption{AP vs. different overlap ratios.}
    \label{fig:ap-overlap}
\end{figure}

Figure~\ref{fig:ap-overlap} shows the performance of our method and other baselines 
with images of different overlap ratios. 
Our method has higher AP over other baselines across all overlap ratios in our dataset. 
Overlap ratio of around 0.05 is an extremely hard setting, our method (29 AP) and improved RPNet (27 AP) perform 
better than matching based methods ($<$24 AP) because SuperGlue~\cite{sarlin2020superglue} 
rarely finds enough correspondence. Deep networks, on the other hand, 
can use other cues and prior to make predictions.
When the overlap ratio is higher, our method can also outperform other baselines (increase by over 5 AP when overlap
ratio is around 0.25 and 11 AP when overlap is about 0.55). We can merge planes based on correspondence
to produce a coherent reconstruction while other baselines produce inconsistent and duplicated planes.

\subsection{Rank by Single-view Prediction}
\label{sec:single-view}
We further investigate how the quality of single-view prediction (outputs from plane prediction module on each image) will affect our results. 
Table~\ref{tab:detection-ranked-all} shows results on top 25\%, 50\%, 75\% and 100\% of the test examples ranked by single-view AP.
On top 25\% examples where the single-view prediction are more accurate,
our proposed methods have a higher AP (increase by over 10 points compared to 100\% data).
Moreover, our proposed method has a higher gain over ablation baselines {\it Associative3D optimization} and {\it Appearance Embedding only} by above 4.9 points
on top 25\% examples and even higher ($>10$ points) over the other external baselines. It is still significant but slightly lower when we have worse single-view predictions. 

\begin{table}[!t]
    \centering
    \caption{We further investigate how the quality of single-view prediction will affect our results. 
    We include results on top 25\%, 50\%, 75\% and 100\% of the test examples ranked by single-view prediction AP.
    Similar to Table~\ref{tab:detection} in our paper, a prediction is a true positive only if Mask IoU $\geq 0.5$, Normal error $\leq 30^\circ$ and Offset error $\leq 1$m. }
    \resizebox{\columnwidth}{!}{
        \begin{tabular}{l@{\hskip15pt}c@{\hskip15pt}c@{\hskip15pt}c@{\hskip15pt}c}
            \toprule
            Methods                                                         &  Top 25\%     & Top 50\%      & Top 75\%  & 100\% \\
            \midrule
            PlaneOdometry + MWS     			                            &  	7.26	    & 	 6.22      & 5.56	    &  4.80 \\
            PlaneOdometry + MWS-G     			                            &  	20.84	    & 	 18.94      & 17.79	    &  16.42 \\
            PlaneOdometry + PlaneRCNN     			                        &  	30.33	    & 	 27.12      & 24.48	    &  21.33 \\
            SuperGlue-G + MWS     			                                &  	11.46	    & 	 9.76      & 8.73	    &  7.56 \\
            SuperGlue-G + MWS-G     			                            &  	28.80	    & 	 26.33      & 24.81	    &  23.01 \\
            SuperGlue-G + PlaneRCNN     			                        &  	41.84	    & 	 37.58      & 34.31	    &  30.06 \\                                           
            RPNet + PlaneRCNN				                                &  	41.64	    & 	 37.04      & 33.71	    &  29.44 \\
            \midrule
            Appearance Embedding only 			                            &  	47.83	    & 42.17	        &	38.06   & 33.04  \\
            Associative3D Optimization 		                &   47.51		& 41.93         &  38.01    & 33.01 \\
            No Continuous Optimization                   	                &  52.15		& 46.07	        &	41.47   & 35.87   \\
            \textbf{Proposed} 					                            &  \textbf{52.75}	    &	\textbf{46.45}	    &	\textbf{41.76}   & \textbf{36.02} \\
            \bottomrule
        \end{tabular}
    }
    \label{tab:detection-ranked-all}
\end{table}

\subsection{Translation Direction Evaluation}
In additional to Table~\ref{tab:relpose}, we report a scale-free angle between 
the predicted and GT vectors in Table~\ref{tab:relpose-T-angle}, following SuperGlue~\cite{sarlin2020superglue}. 
When ~\cite{sarlin2020superglue} fails (11.6\% of the data), we use a vertical vector (equidistant to all horizontal directions). 
The results are similar to rotation: ~\cite{sarlin2020superglue} does better if it has good correspondences, but can fail entirely. 
So its median error is lower, but other errors are comparable. 
The ablation shows our continuous optimization improves on using discrete cameras. 
We stress that our system recovers scale, which is not evaluated here, and that we see pixel-based approaches as {\it complementary} -- 
one could also train a SuperPoint-like system on early layers in our network and use it in the optimization.
\begin{table}[t!]    
    \centering
    \caption{Camera translation direction error. 
    \cite{raposo2013plane} + GT / \cite{Ranftl2020} perform worse than Assoc.3D~\cite{Qian2020} and are omitted.}
    \resizebox{0.95\columnwidth}{!}{
        \begin{tabular}{cccc}
        \toprule
        Method & T Med$^{\circ}$. & T Mean$^{\circ}$ & T(${\leq}30^{\circ}$) \\
        \midrule
        Assoc.3D~[37] & 41.9 & 43.2 & 36.5 \\
        \midrule
        SuperGlue~[43] &  \textbf{4.1} & \textbf{19.7} & 77.0 \\
        \midrule
        No Optimization & 16.8 & 23.9 & 71.9 \\
        No Continuous & 16.4 & 23.6 & 72.5 \\
        \textbf{Proposed} & 11.1 & \textbf{19.7} & \textbf{78.0} \\ 
        \bottomrule
        \end{tabular}
    }
    \label{tab:relpose-T-angle}
\end{table}

\clearpage


\clearpage

\begin{table*}[t]  
      
        \centering
        \caption{Ablation study of our camera pose module.}
        \begin{tabular}{@{\hskip4pt}c@{\hskip20pt}c@{\hskip9pt}c@{\hskip9pt}c@{\hskip20pt}c@{\hskip9pt}c@{\hskip9pt}c@{\hskip4pt}}
        \toprule
        & \multicolumn{3}{c}{Translation (meters)} & \multicolumn{3}{c}{Rotation  (degrees)} \\
        Method & Median $\downarrow$ & Mean $\downarrow$ & (Err $\leq$ 1m)\% $\uparrow$  & Median $\downarrow$   & Mean $\downarrow$ & (Err $\leq$ 30$^{ \circ }$)\% $\uparrow$ \\
        \midrule			
        Assoc.3D~\cite{Qian2020} cam.      & 2.17 & 2.46 & 14.8  & 42.09 & 52.97 & 38.1 \\
        ResNet50-CatConv   & 1.55 & 1.98 & 29.4  & 29.33 & 46.08 & 50.6 \\
        ResNet50-Attention & 0.98 & 1.41 & 51.1  & 12.78 & 24.23 & 77.3 \\
        \textbf{Proposed}    & \textbf{0.63} & \textbf{1.15} & \textbf{66.6}  & \textbf{7.33} & \textbf{22.78} & \textbf{83.4} \\                           
        \bottomrule
        \end{tabular}
        \label{tab:relpose_supp}

\end{table*}

\subsection{Ablation Study for Camera Pose Module}

We include detailed architectures for baselines in Section 4.4 of our paper in Table~\ref{tab:resnet50_fc}, \ref{tab:resnet50_catconv} and \ref{tab:resnet50_cc}. 
ReLU is used between all Linear and Conv layers.
We use ResNet50 pretrained on COCO as the backbone to predict camera pose. 
We do not freeze the backbone for the baselines since they are standalone networks. 
Table~\ref{tab:relpose_supp} shows the ablation experiments. With the ability to explicitly calculate the relationship of features across
views, our proposed attention module (ResNet50-Attention) outperforms other standalone architectures by a large margin on all the metrics. 
Running the attention module on the plane detection backbone further improves results, but due to switching to a FPN~\cite{lin2017feature},
it is difficult to directly ascribe performance changes.

\subsection{Qualitative Results}
\begin{figure*}[!t]
    \centering
    \scriptsize
    \begin{tabular}{c@{\hskip4pt}c@{\hskip4pt}c@{\hskip4pt}c@{\hskip4pt}c@{\hskip4pt}c}
    \toprule

    Image 1 & Image 2 & Prediction & Ground Truth  & Prediction & Ground Truth \\
    \midrule
    \frame{\includegraphics[width=0.148\textwidth]{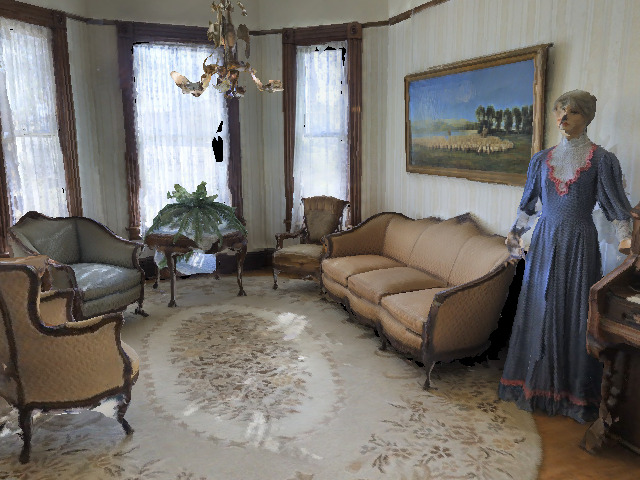}}
    & \frame{\includegraphics[width=0.148\textwidth]{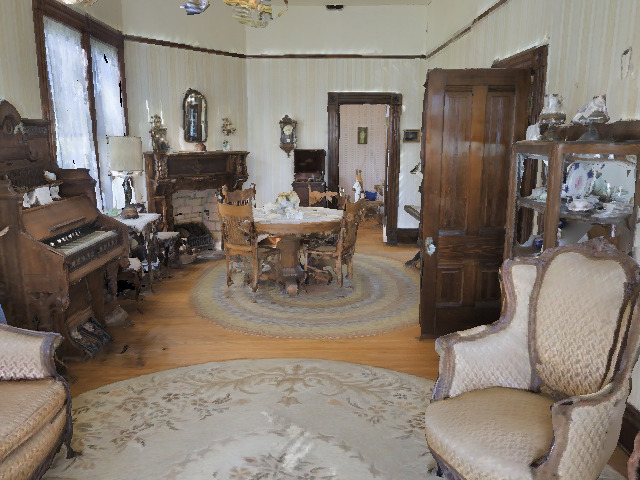}}
    & \frame{\includegraphics[width=0.148\textwidth]{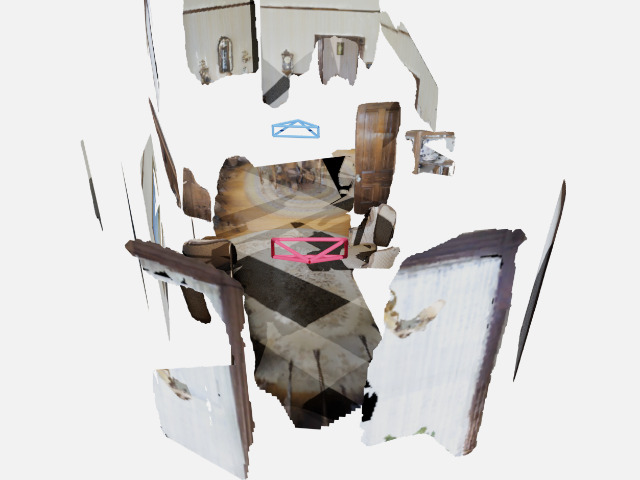}}
    & \frame{\includegraphics[width=0.148\textwidth]{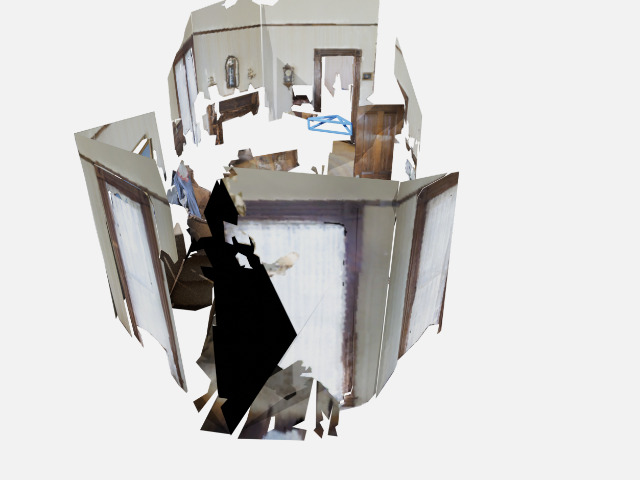}}
    & \frame{\includegraphics[width=0.148\textwidth]{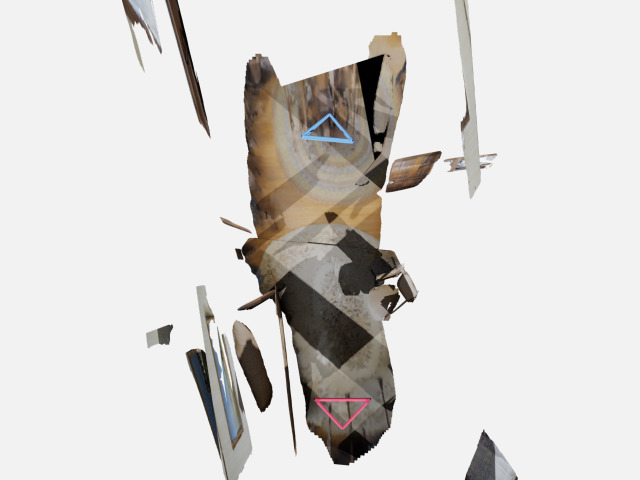}}
    & \frame{\includegraphics[width=0.148\textwidth]{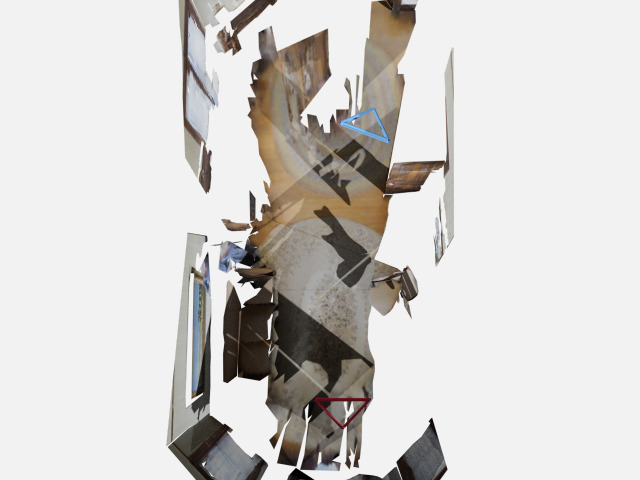}}\\

   \frame{\includegraphics[width=0.148\textwidth]{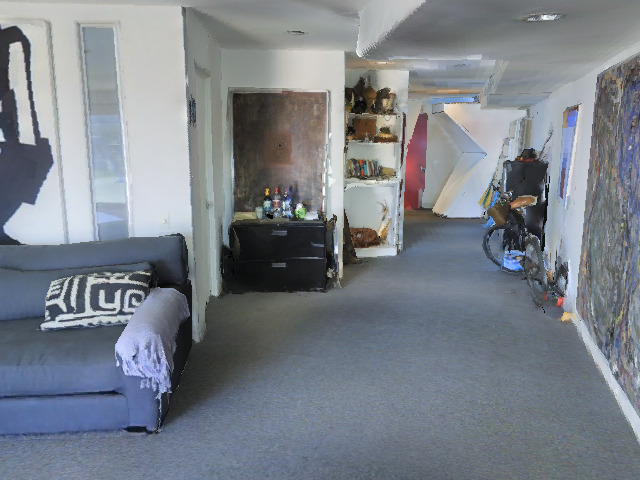}}
   & \frame{\includegraphics[width=0.148\textwidth]{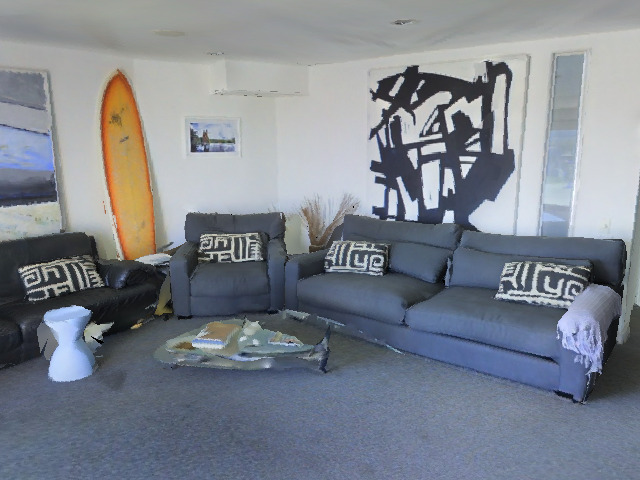}}
   & \frame{\includegraphics[width=0.148\textwidth]{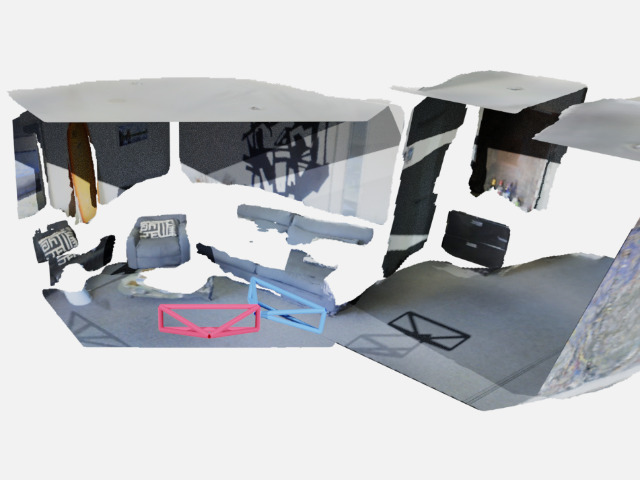}}
   & \frame{\includegraphics[width=0.148\textwidth]{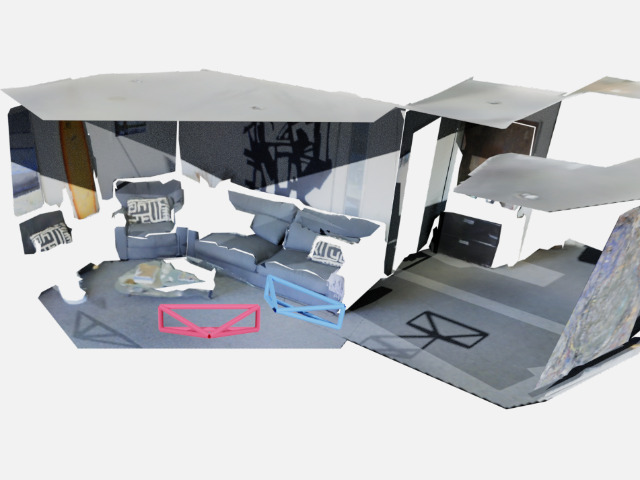}}
   & \frame{\includegraphics[width=0.148\textwidth]{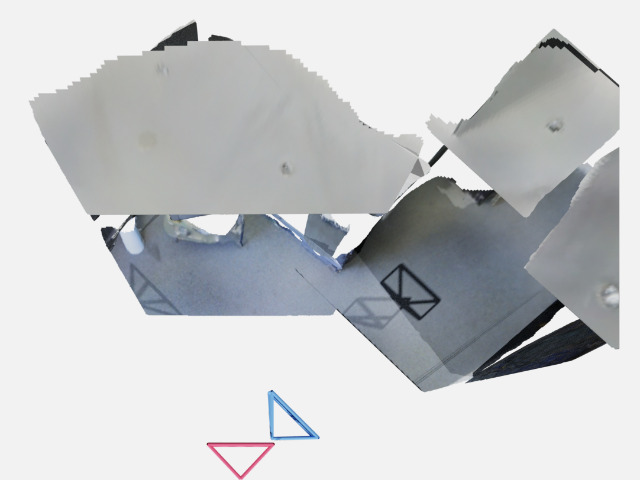}}
   & \frame{\includegraphics[width=0.148\textwidth]{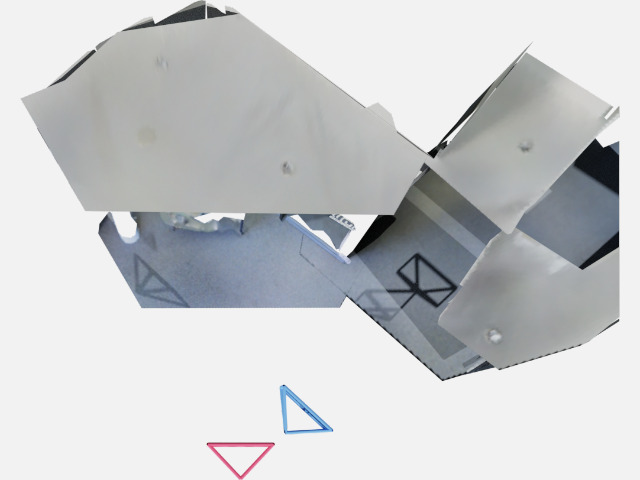}}\\

   \frame{\includegraphics[width=0.148\textwidth]{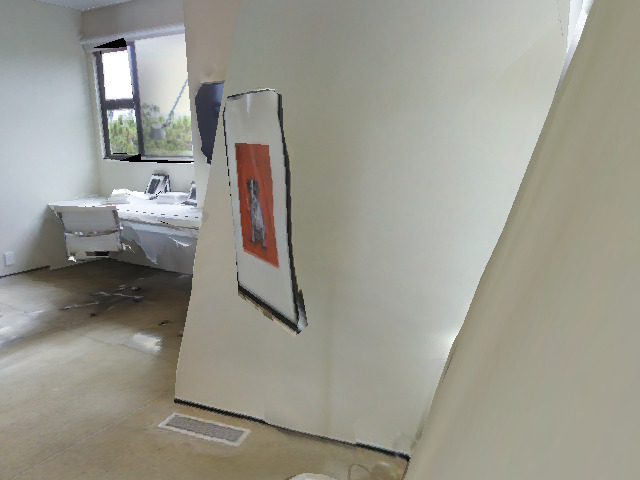}}
   & \frame{\includegraphics[width=0.148\textwidth]{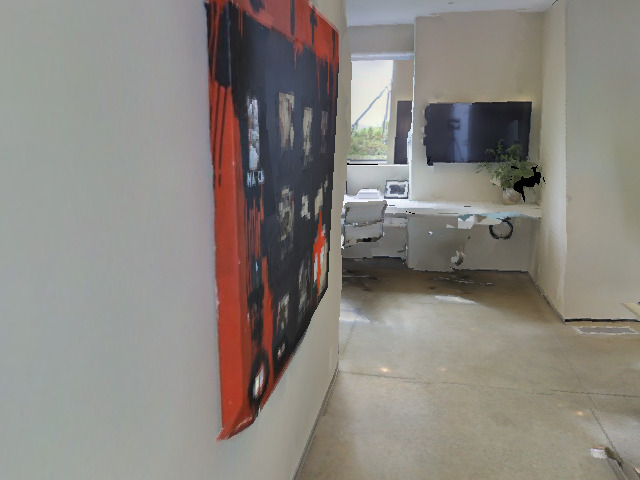}}
   & \frame{\includegraphics[width=0.148\textwidth]{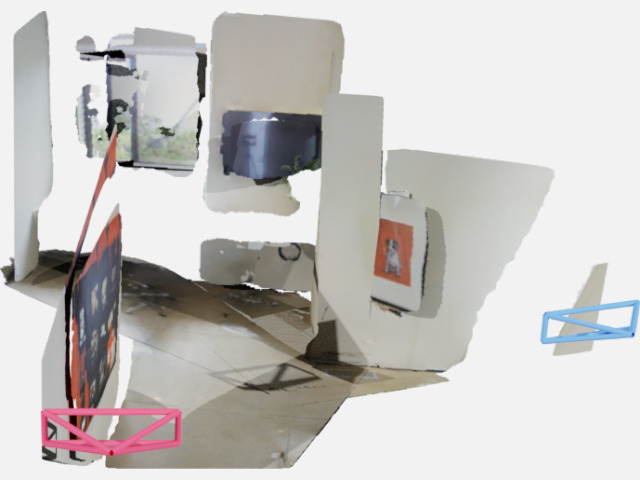}}
   & \frame{\includegraphics[width=0.148\textwidth]{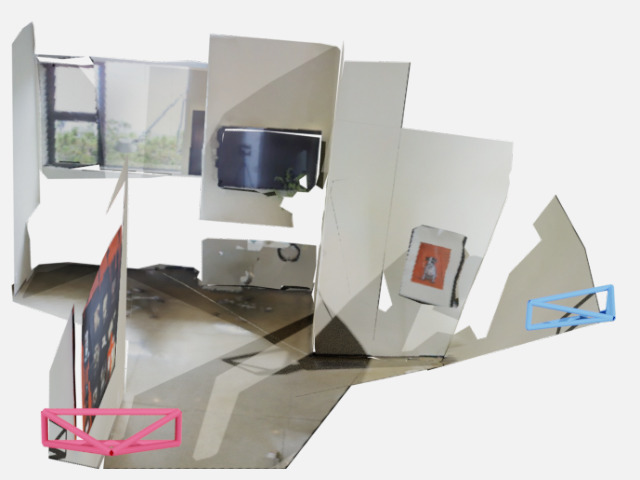}}
   & \frame{\includegraphics[width=0.148\textwidth]{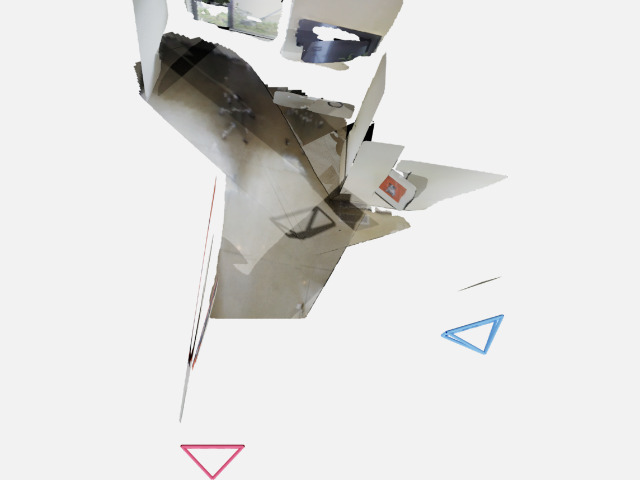}}
   & \frame{\includegraphics[width=0.148\textwidth]{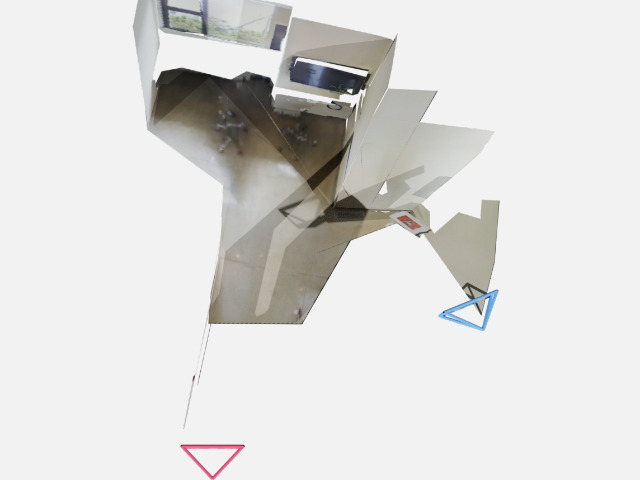}}\\

   \frame{\includegraphics[width=0.148\textwidth]{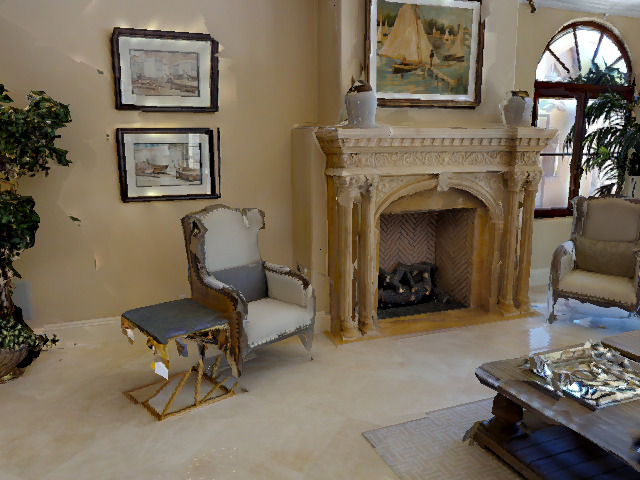}}
   & \frame{\includegraphics[width=0.148\textwidth]{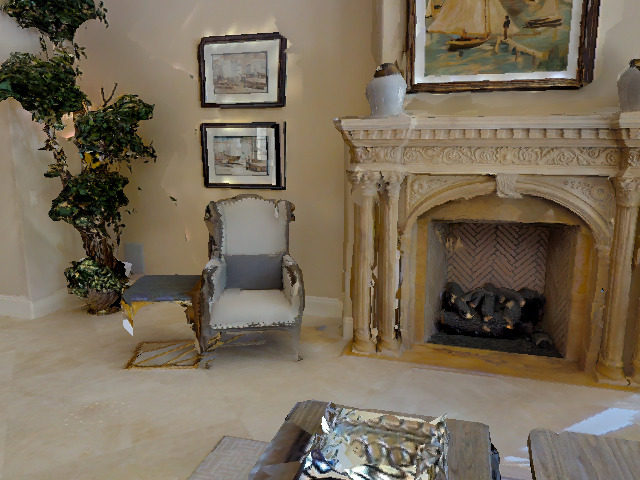}}
   & \frame{\includegraphics[width=0.148\textwidth]{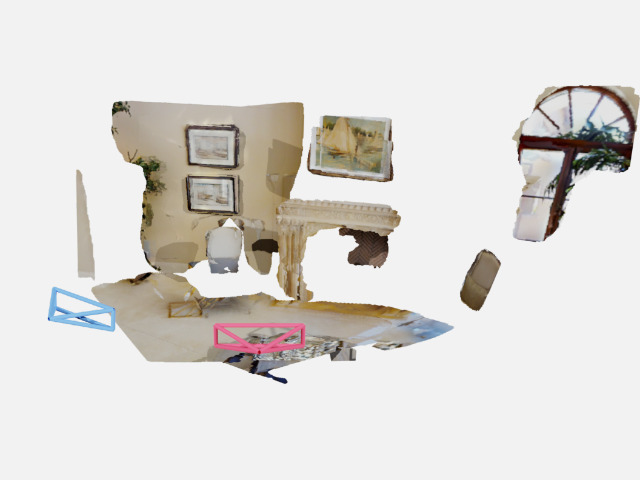}}
   & \frame{\includegraphics[width=0.148\textwidth]{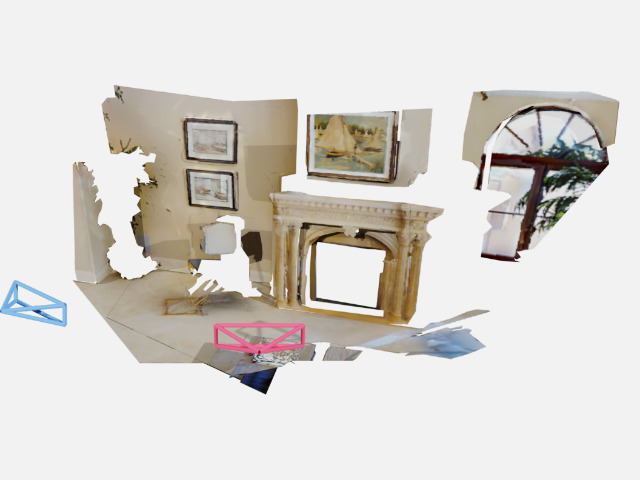}}
   & \frame{\includegraphics[width=0.148\textwidth]{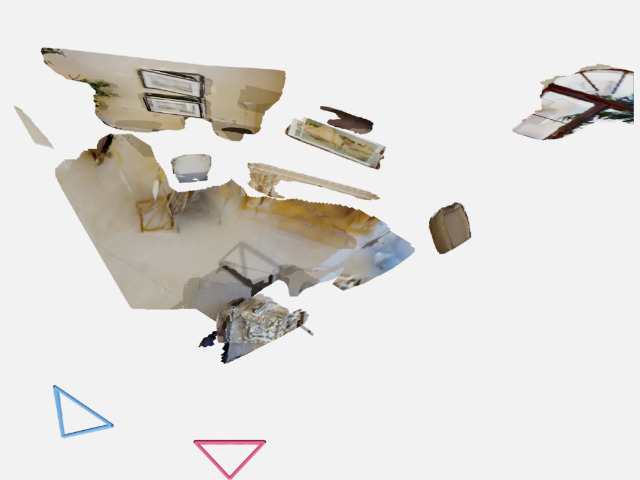}}
   & \frame{\includegraphics[width=0.148\textwidth]{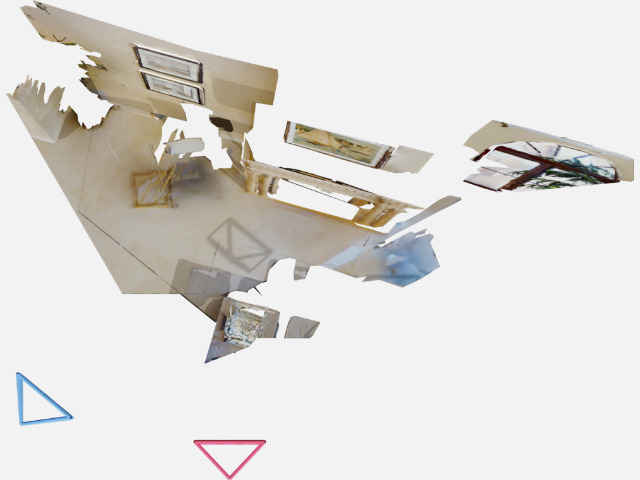}}\\

   \frame{\includegraphics[width=0.148\textwidth]{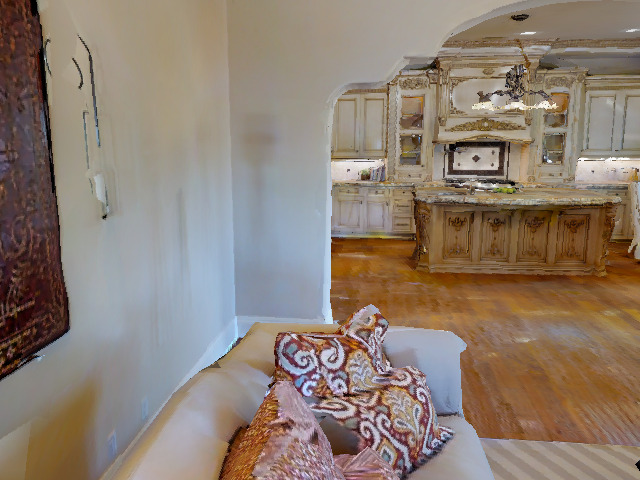}}
   & \frame{\includegraphics[width=0.148\textwidth]{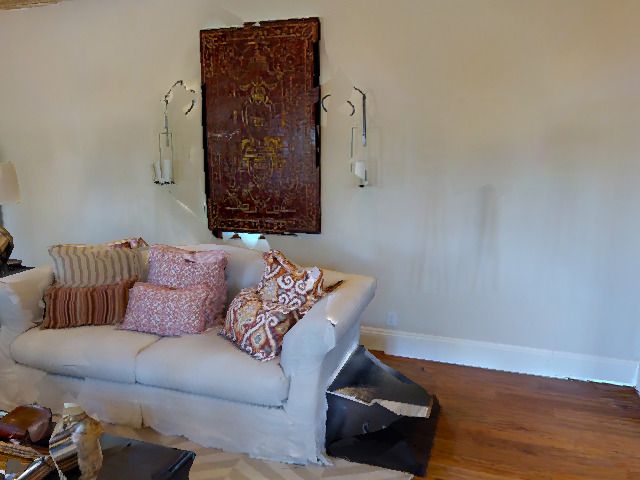}}
   & \frame{\includegraphics[width=0.148\textwidth]{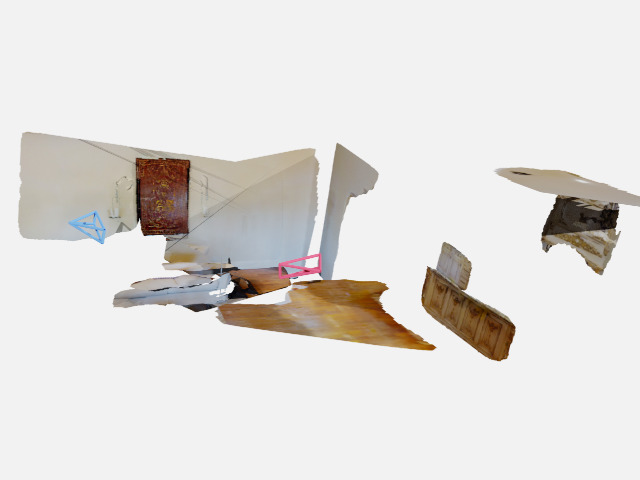}}
   & \frame{\includegraphics[width=0.148\textwidth]{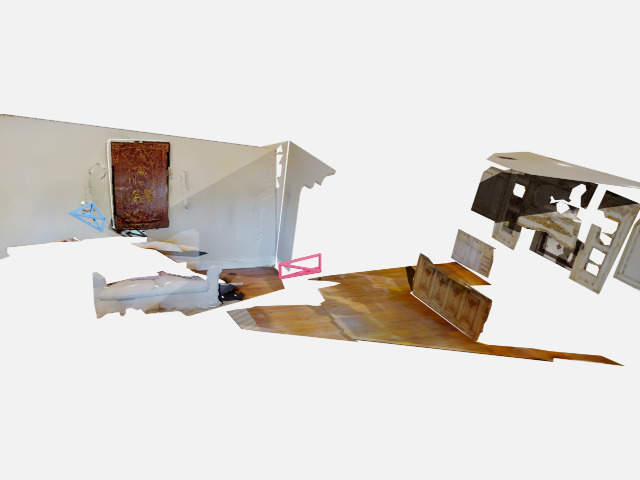}}
   & \frame{\includegraphics[width=0.148\textwidth]{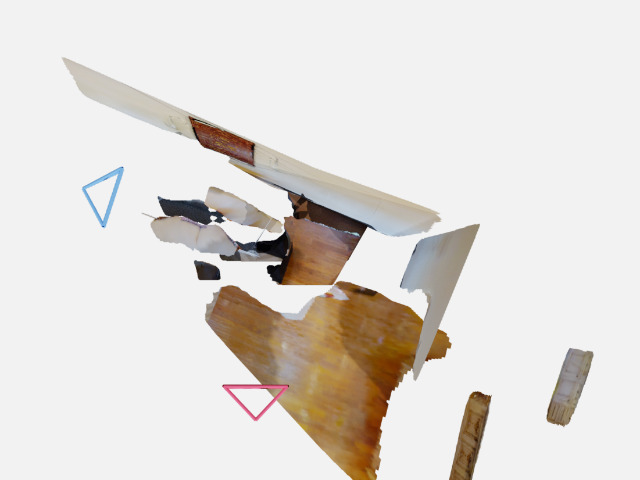}}
   & \frame{\includegraphics[width=0.148\textwidth]{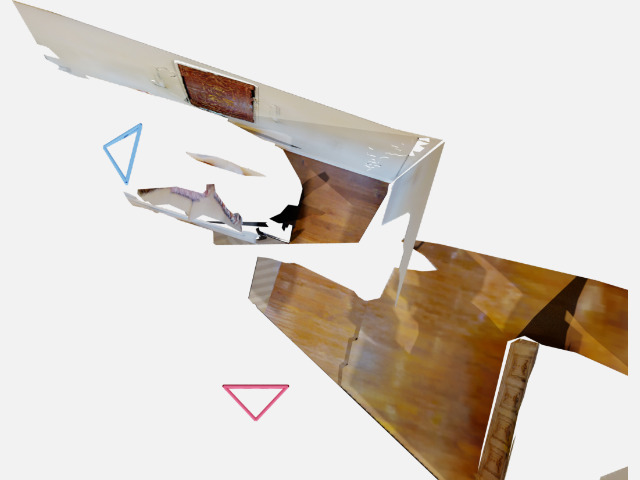}}\\

   \frame{\includegraphics[width=0.148\textwidth]{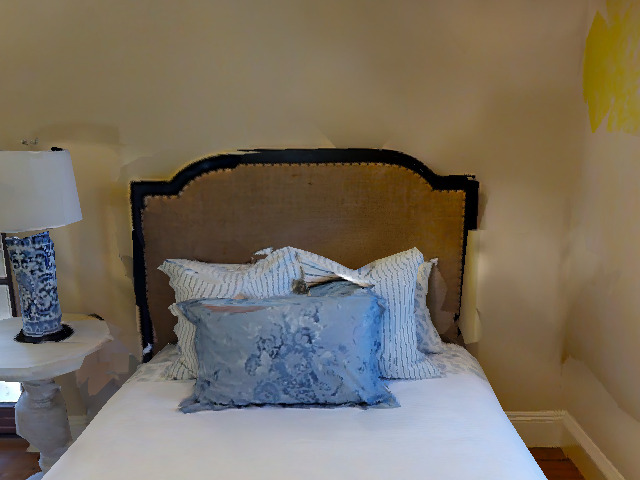}}
   & \frame{\includegraphics[width=0.148\textwidth]{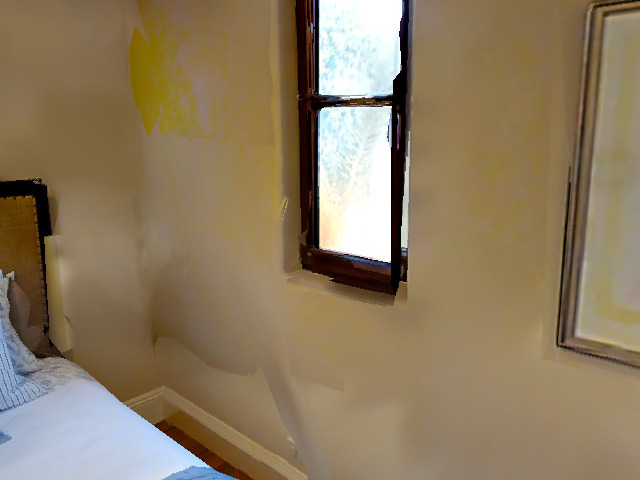}}
   & \frame{\includegraphics[width=0.148\textwidth]{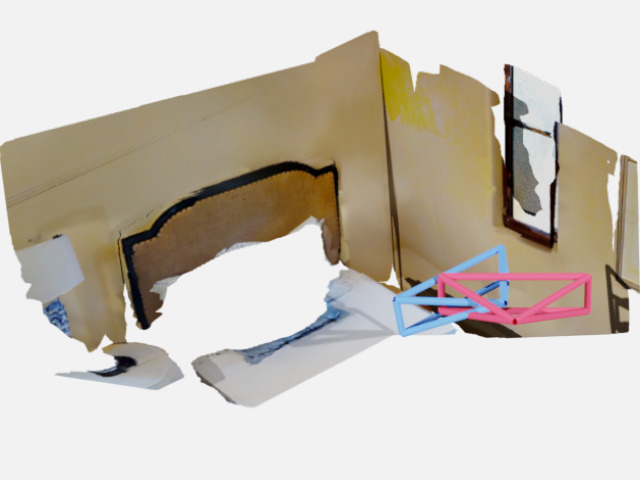}}
   & \frame{\includegraphics[width=0.148\textwidth]{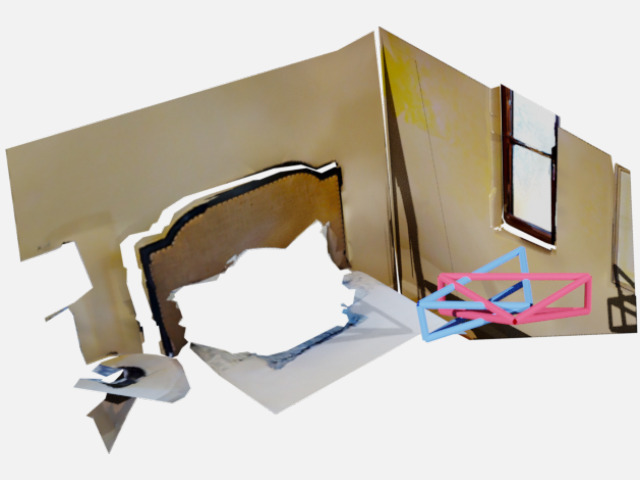}}
   & \frame{\includegraphics[width=0.148\textwidth]{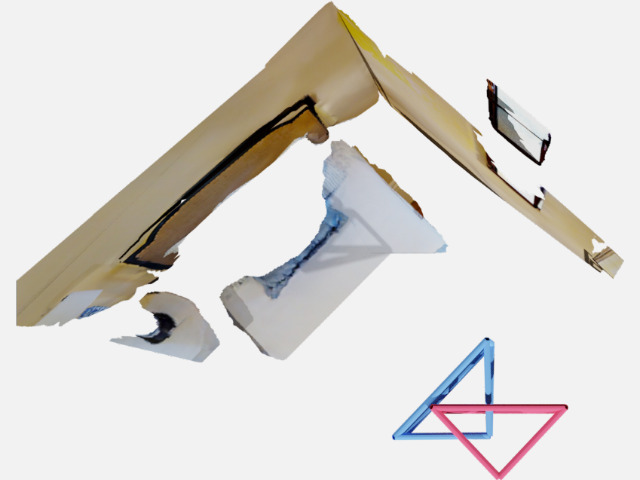}}
   & \frame{\includegraphics[width=0.148\textwidth]{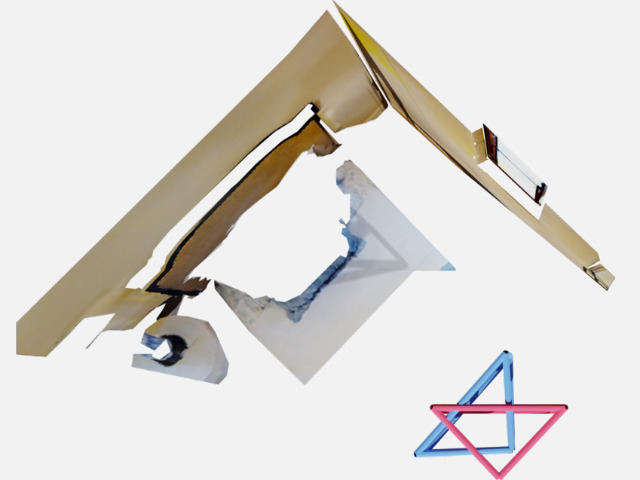}}\\

    \frame{\includegraphics[width=0.148\textwidth]{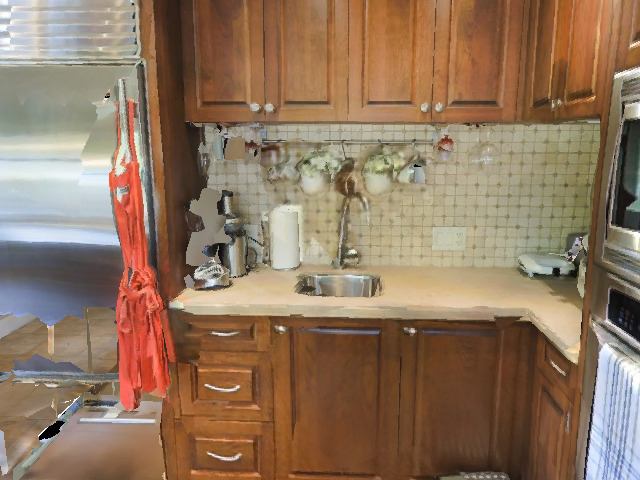}}
    & \frame{\includegraphics[width=0.148\textwidth]{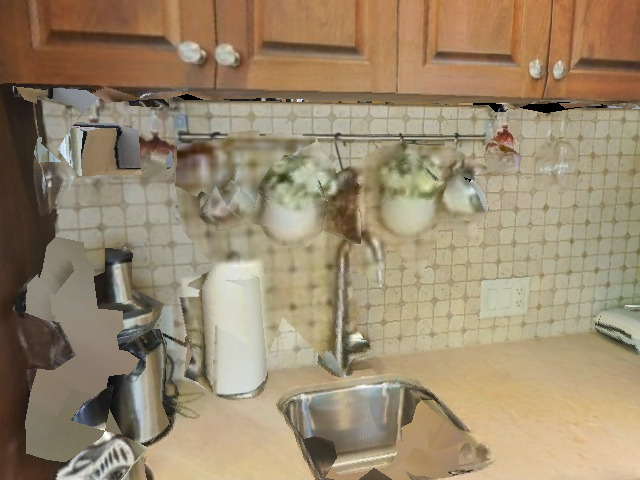}}
    & \frame{\includegraphics[width=0.148\textwidth]{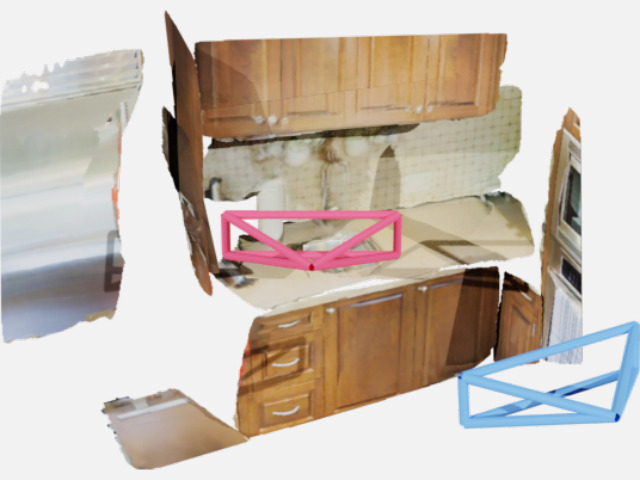}}
    & \frame{\includegraphics[width=0.148\textwidth]{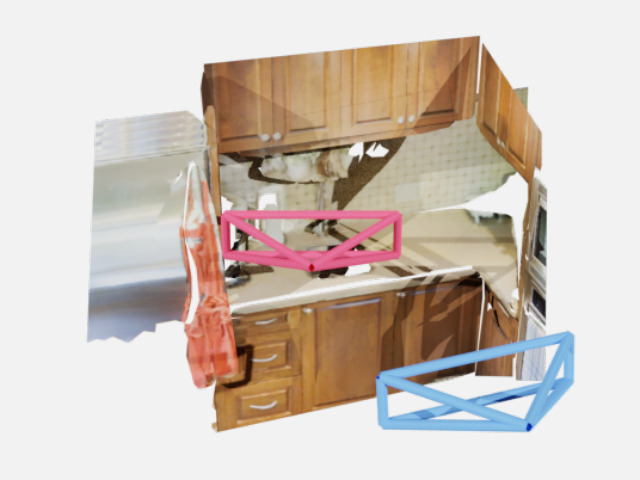}}
    & \frame{\includegraphics[width=0.148\textwidth]{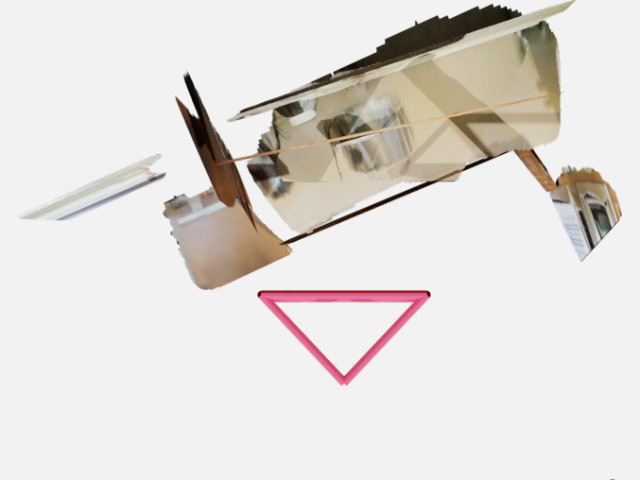}}
    & \frame{\includegraphics[width=0.148\textwidth]{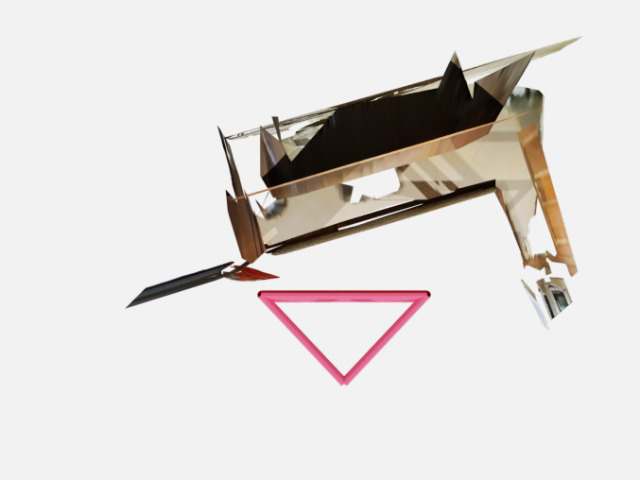}}\\

    \frame{\includegraphics[width=0.148\textwidth]{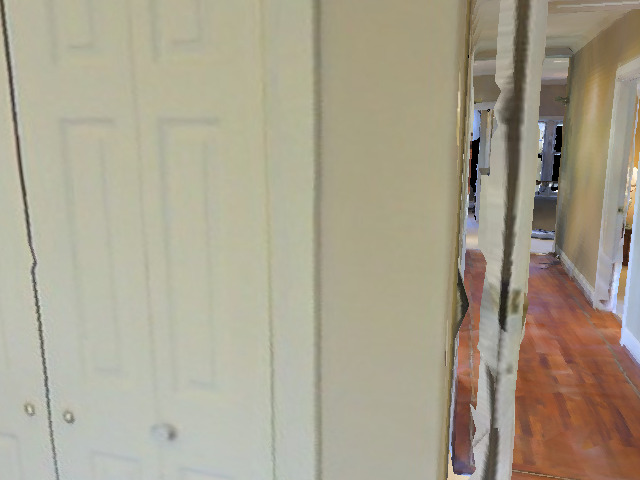}}
    & \frame{\includegraphics[width=0.148\textwidth]{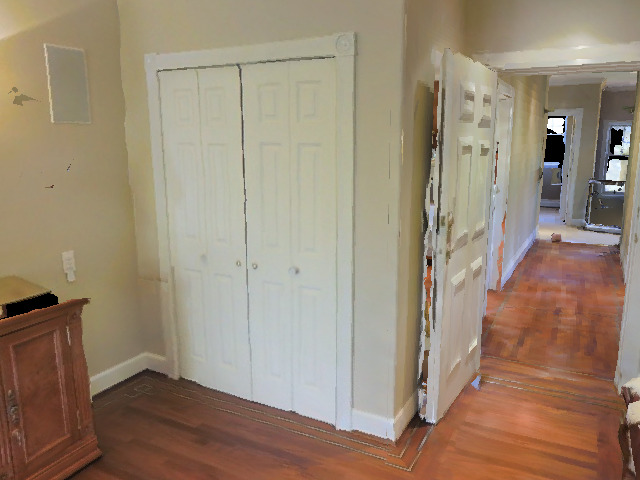}}
    & \frame{\includegraphics[width=0.148\textwidth]{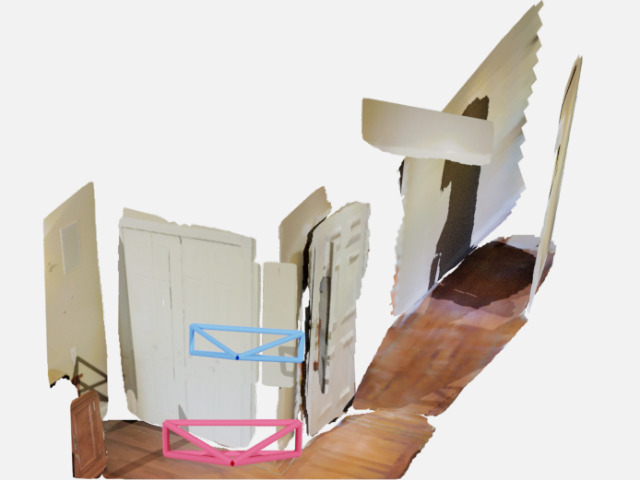}}
    & \frame{\includegraphics[width=0.148\textwidth]{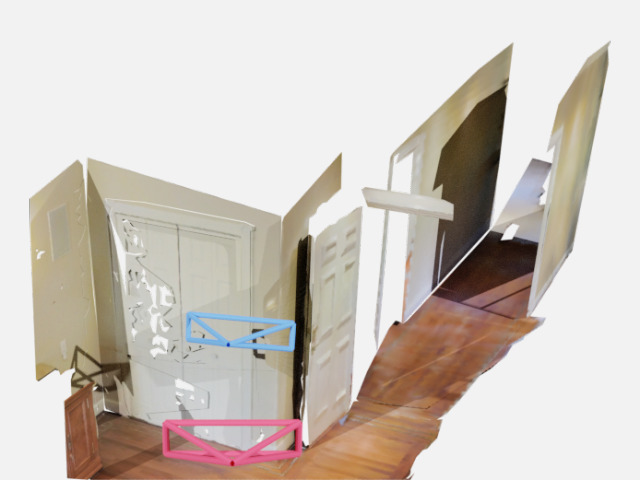}}
    & \frame{\includegraphics[width=0.148\textwidth]{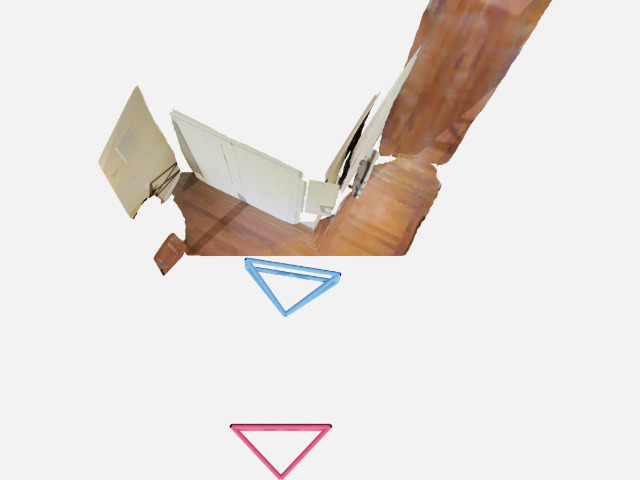}}
    & \frame{\includegraphics[width=0.148\textwidth]{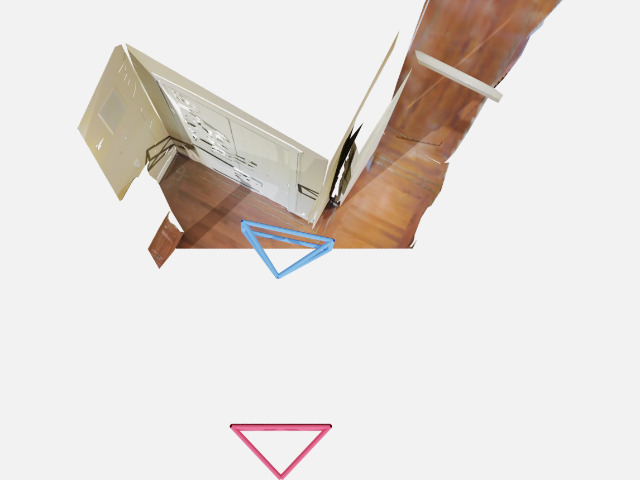}}\\

   \frame{\includegraphics[width=0.148\textwidth]{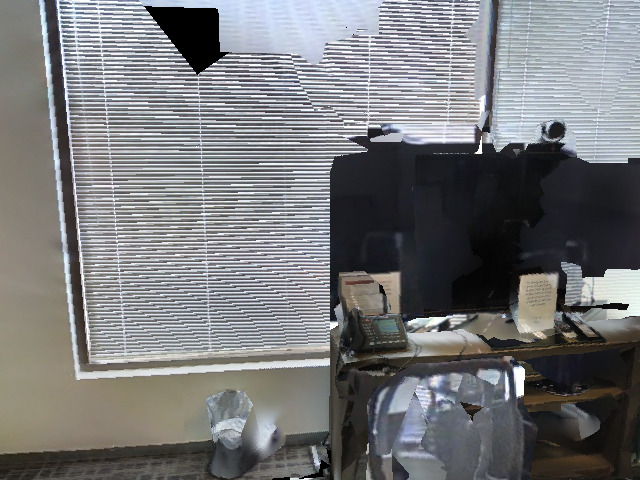}}
   & \frame{\includegraphics[width=0.148\textwidth]{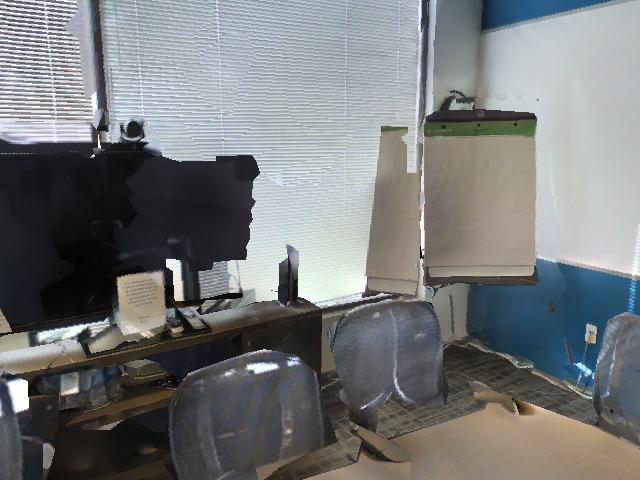}}
   & \frame{\includegraphics[width=0.148\textwidth]{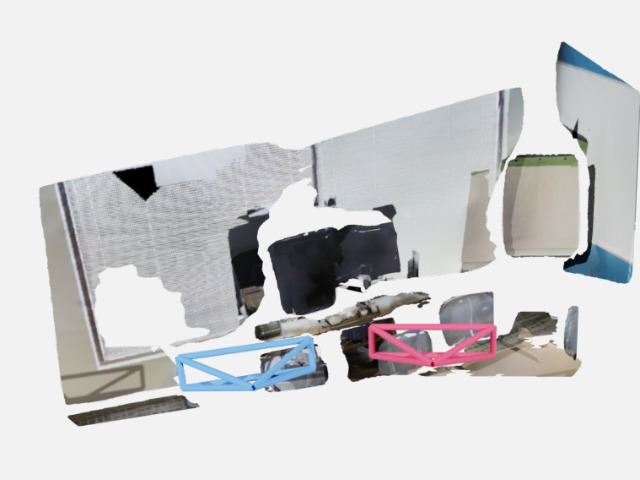}}
   & \frame{\includegraphics[width=0.148\textwidth]{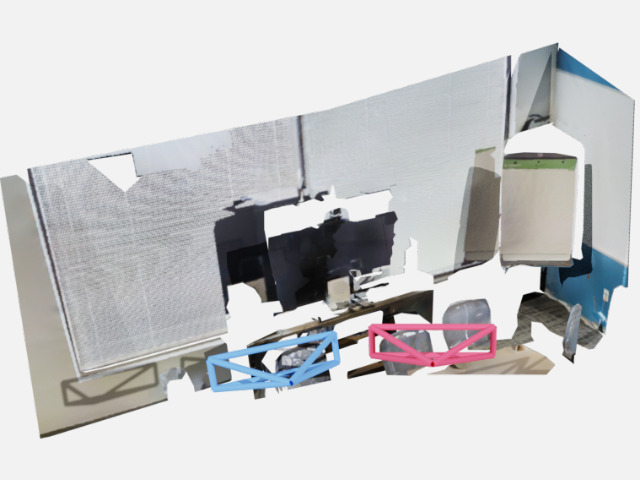}}
   & \frame{\includegraphics[width=0.148\textwidth]{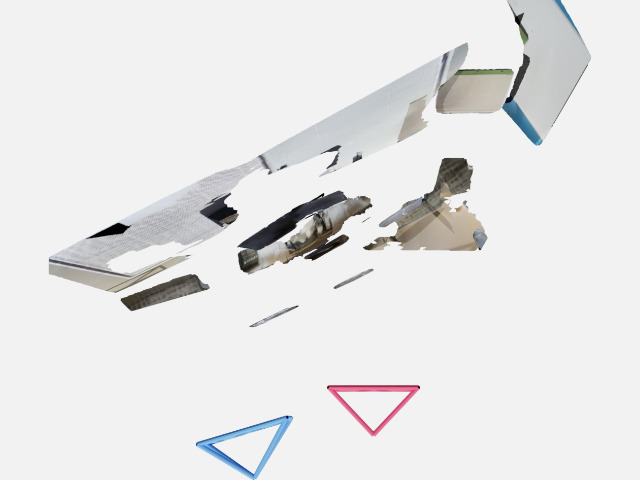}}
   & \frame{\includegraphics[width=0.148\textwidth]{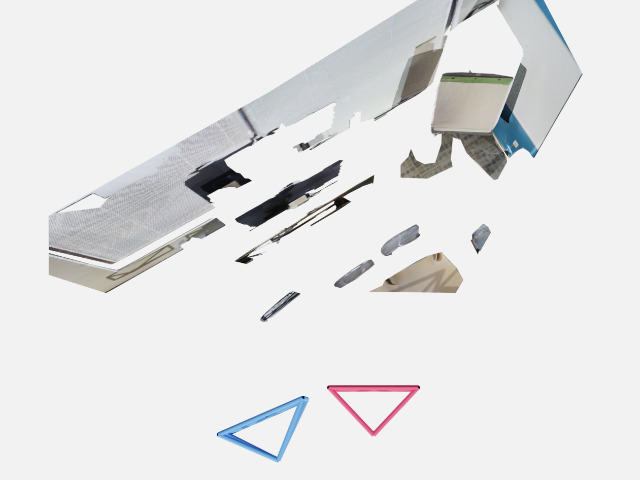}}\\

    \frame{\includegraphics[width=0.148\textwidth]{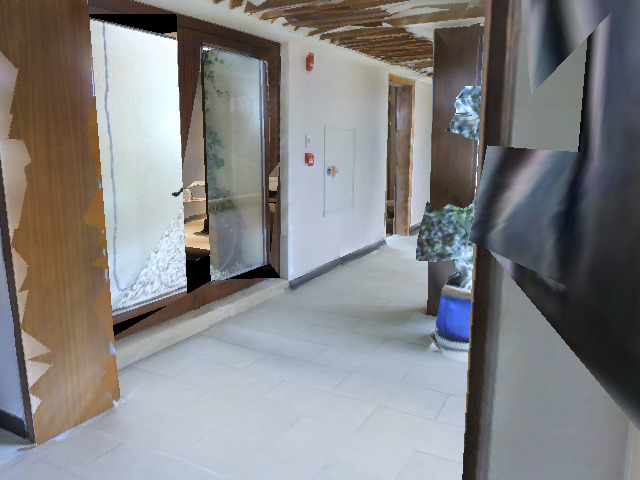}}
    & \frame{\includegraphics[width=0.148\textwidth]{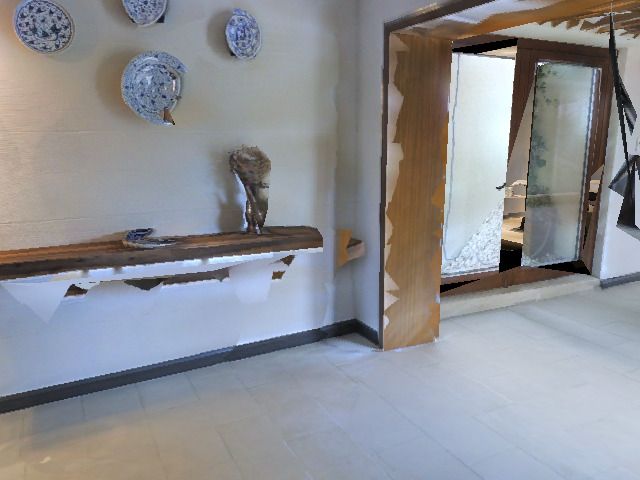}}
    & \frame{\includegraphics[width=0.148\textwidth]{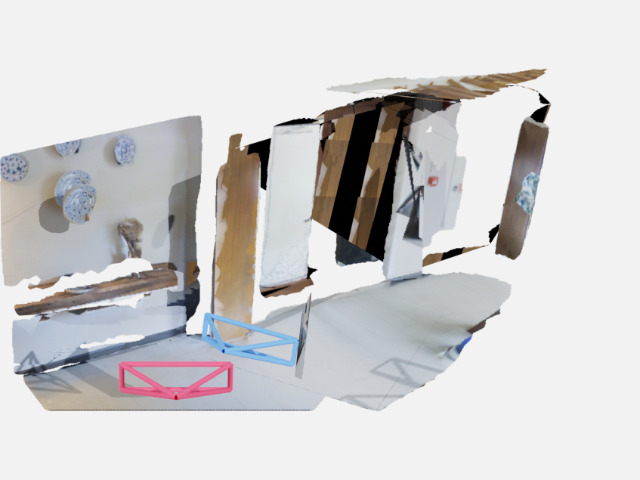}}
    & \frame{\includegraphics[width=0.148\textwidth]{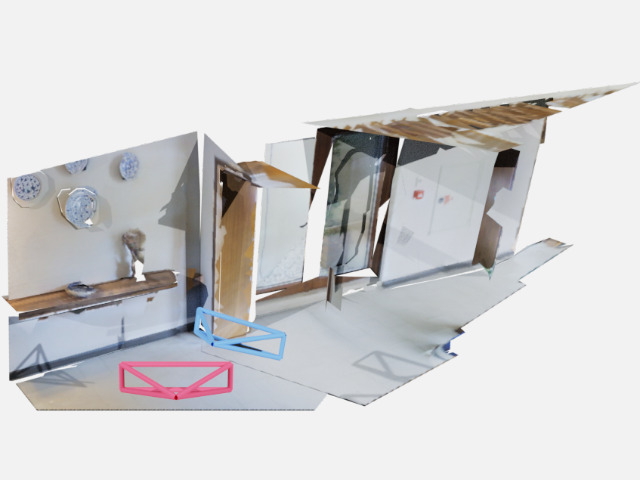}}
    & \frame{\includegraphics[width=0.148\textwidth]{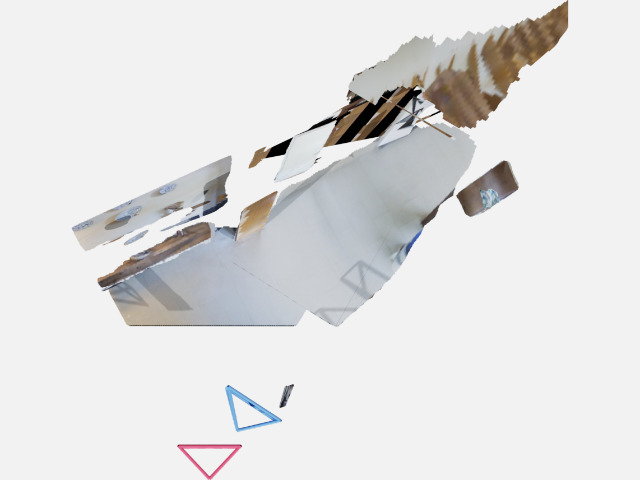}}
    & \frame{\includegraphics[width=0.148\textwidth]{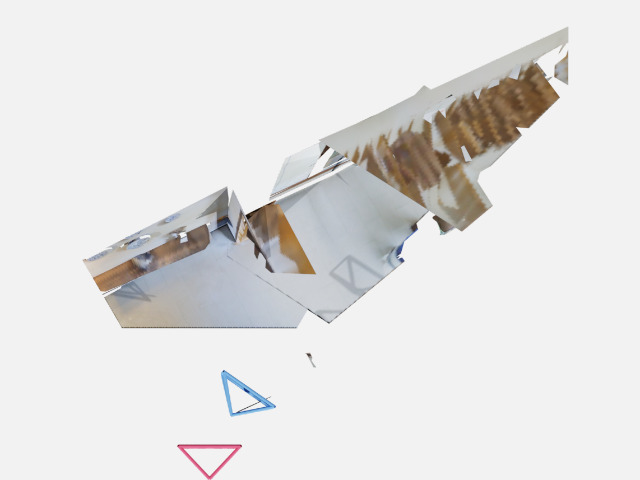}}\\

    \bottomrule
    \end{tabular}
    \caption{More results on Matterport3D test set, extending Figure~\ref{fig:example-wall} in our paper,
    \textbf{\textcolor{AccessibleBlue}{Blue}} and \textbf{\textcolor{AccessibleRed}{Red}} frustums show
    cameras for image 1 and 2.
    }
    \label{fig:supp-example-wall}
\end{figure*}

\begin{figure*}[!t]
    \centering
    \scriptsize
    \begin{tabular}{c@{\hskip4pt}c@{\hskip4pt}c@{\hskip4pt}c@{\hskip4pt}c@{\hskip4pt}c}
    \toprule

    Image 1 & Image 2 & Prediction & Ground Truth  & Prediction & Ground Truth \\
    \midrule
    \frame{\includegraphics[width=0.148\textwidth]{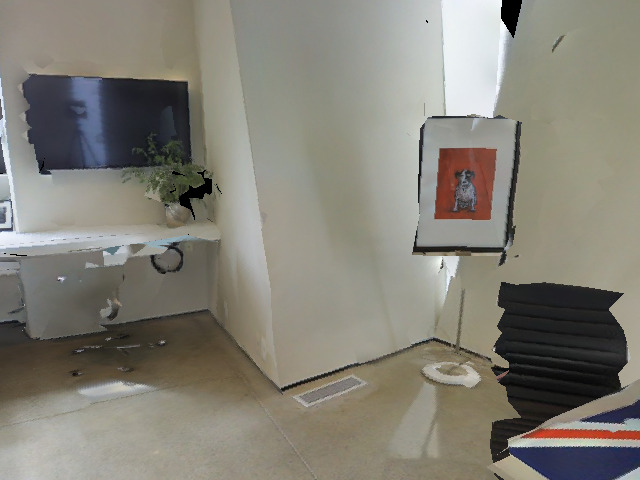}}
    & \frame{\includegraphics[width=0.148\textwidth]{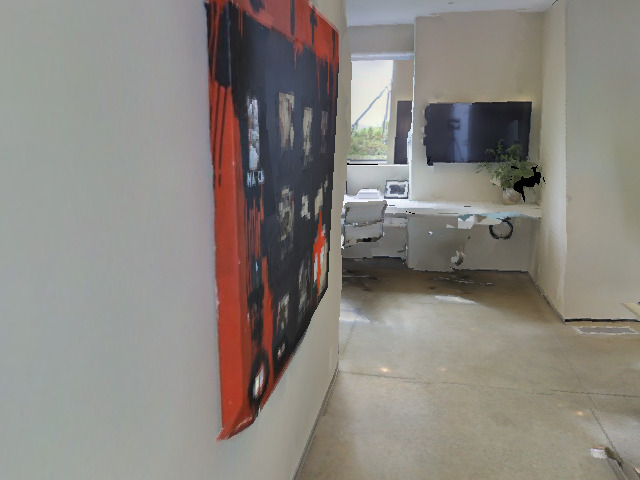}}
    & \frame{\includegraphics[width=0.148\textwidth]{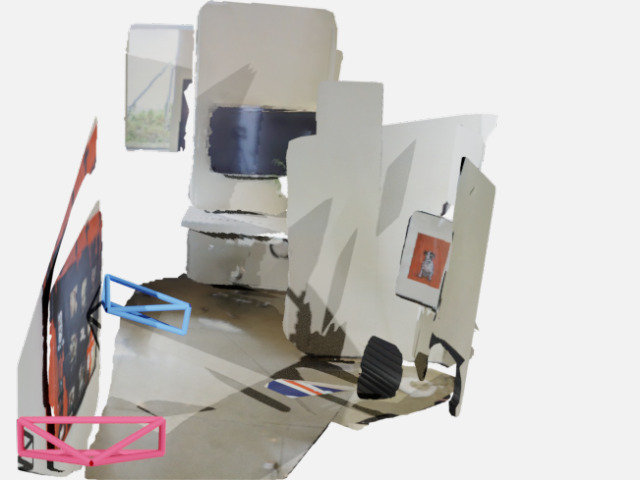}}
    & \frame{\includegraphics[width=0.148\textwidth]{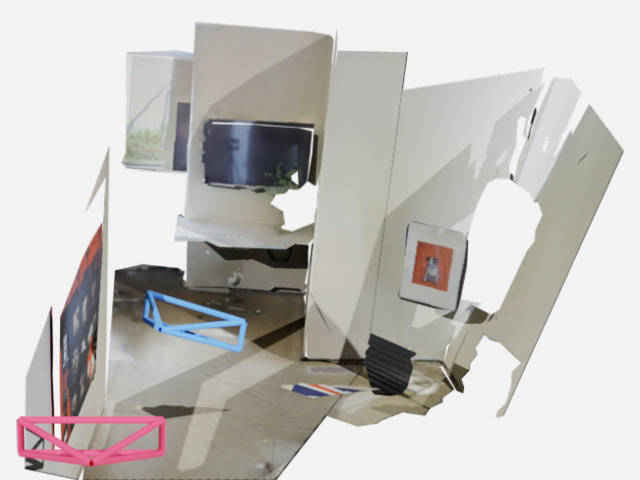}}
    & \frame{\includegraphics[width=0.148\textwidth]{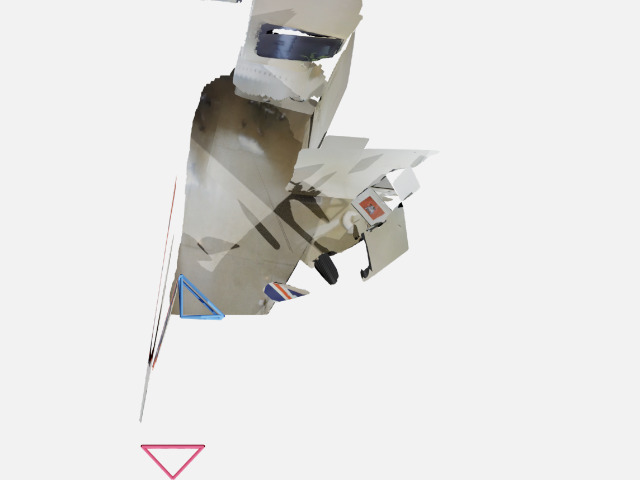}}
    & \frame{\includegraphics[width=0.148\textwidth]{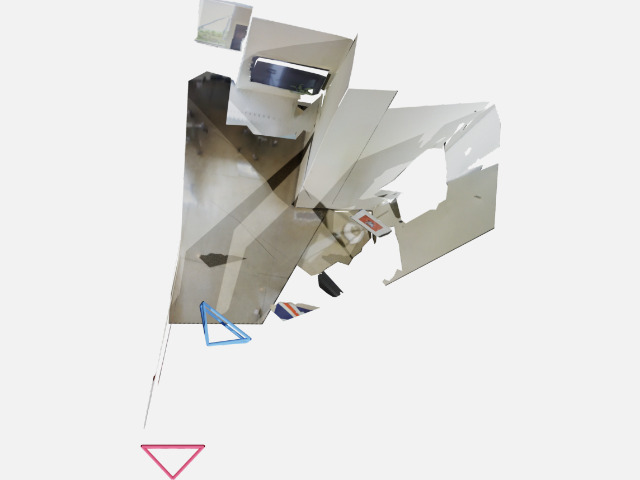}}\\

    \frame{\includegraphics[width=0.148\textwidth]{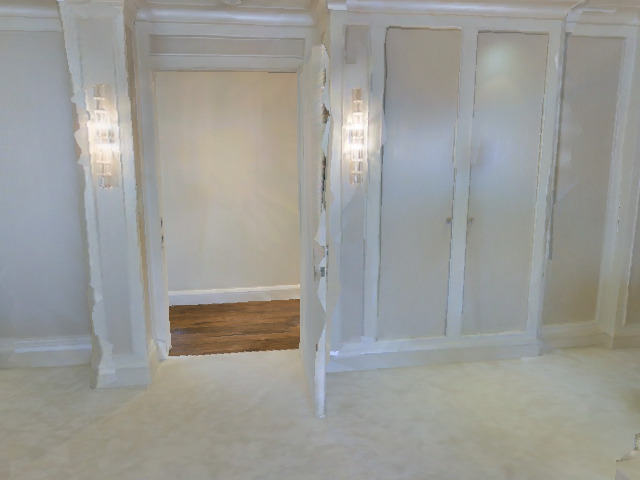}}
    & \frame{\includegraphics[width=0.148\textwidth]{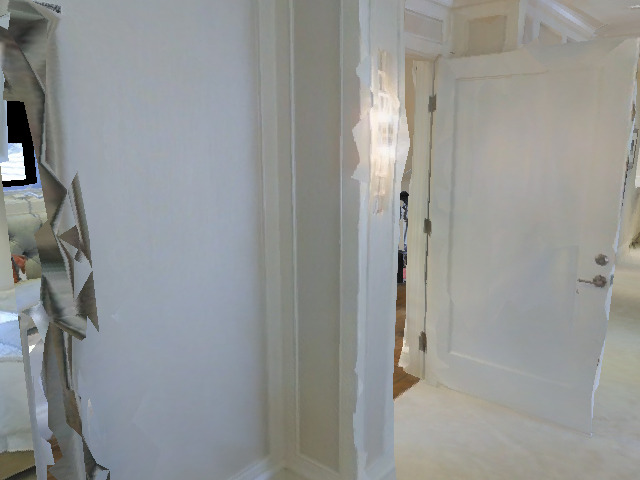}}
    & \frame{\includegraphics[width=0.148\textwidth]{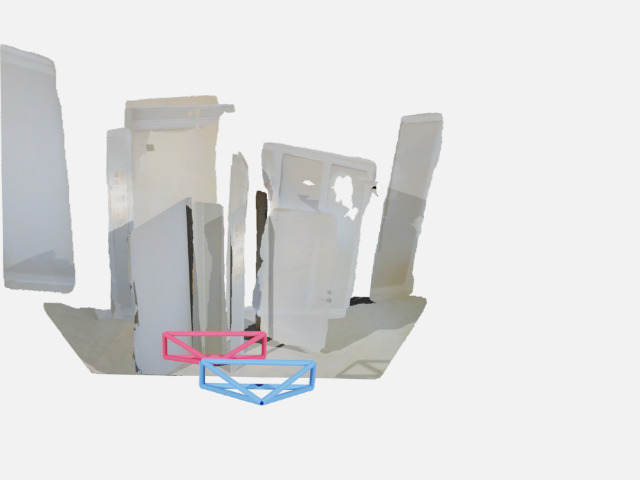}}
    & \frame{\includegraphics[width=0.148\textwidth]{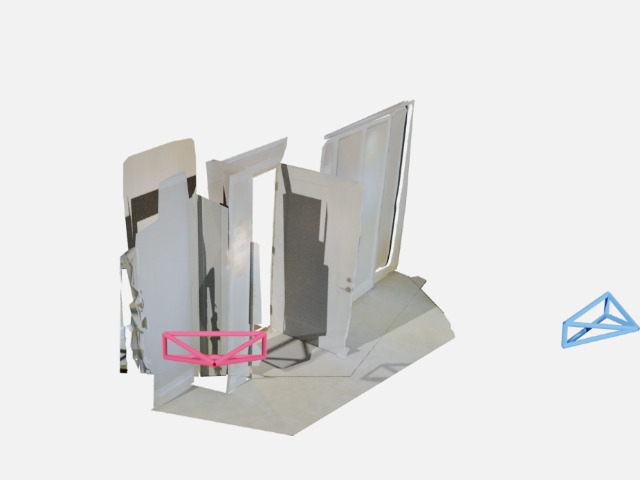}}
    & \frame{\includegraphics[width=0.148\textwidth]{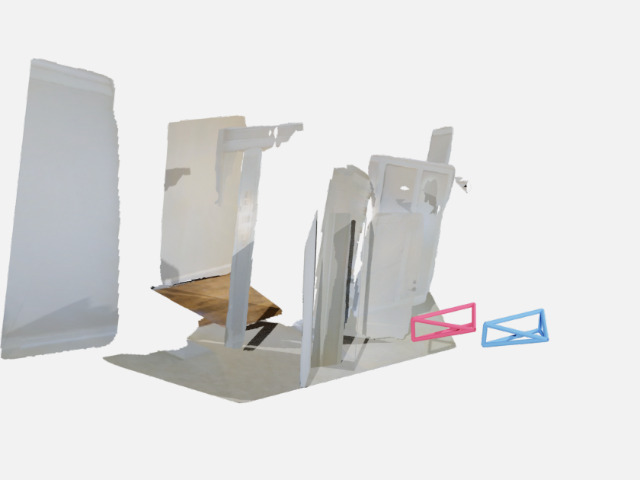}}
    & \frame{\includegraphics[width=0.148\textwidth]{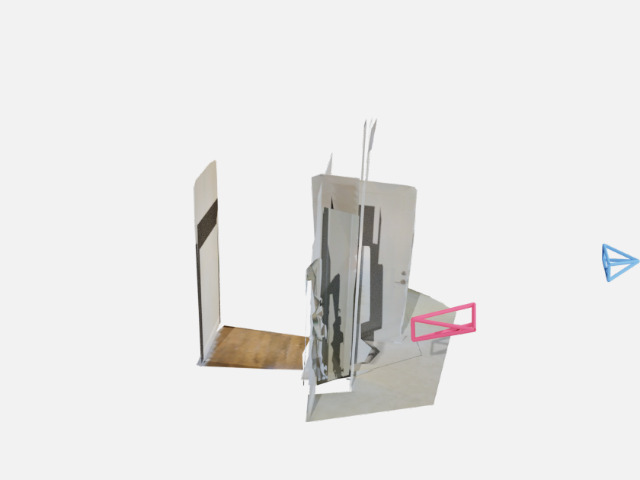}}\\

    \frame{\includegraphics[width=0.148\textwidth]{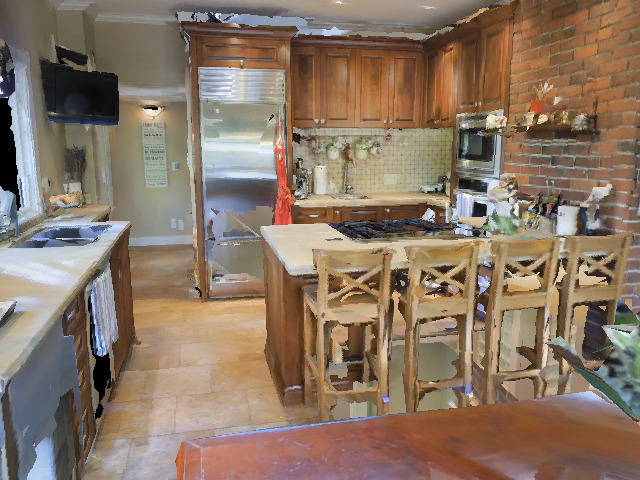}}
    & \frame{\includegraphics[width=0.148\textwidth]{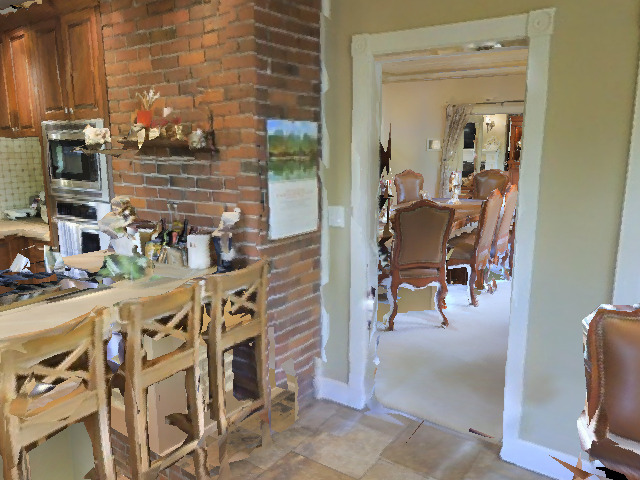}}
    & \frame{\includegraphics[width=0.148\textwidth]{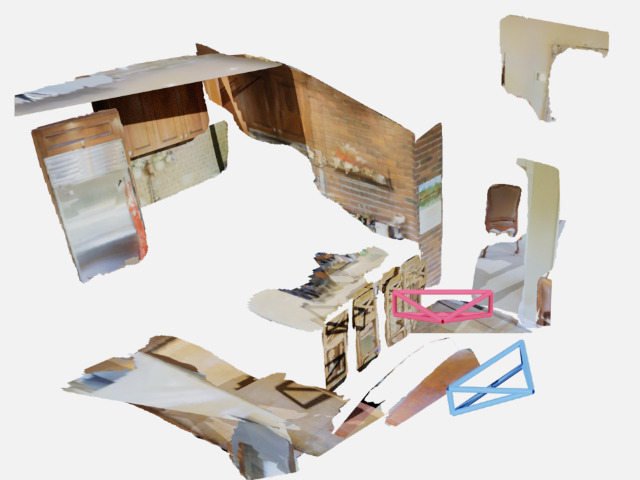}}
    & \frame{\includegraphics[width=0.148\textwidth]{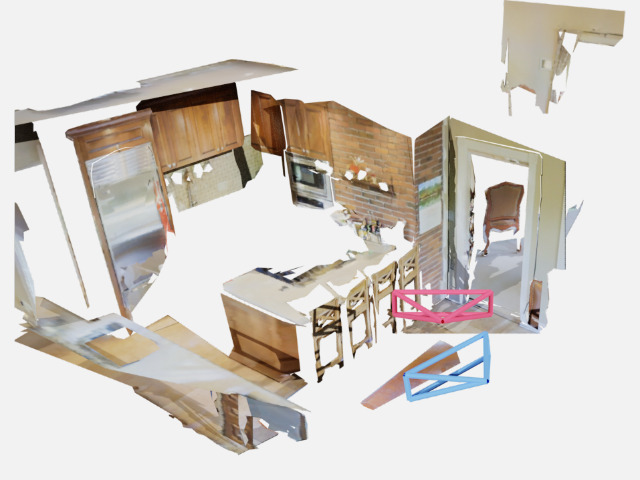}}
    & \frame{\includegraphics[width=0.148\textwidth]{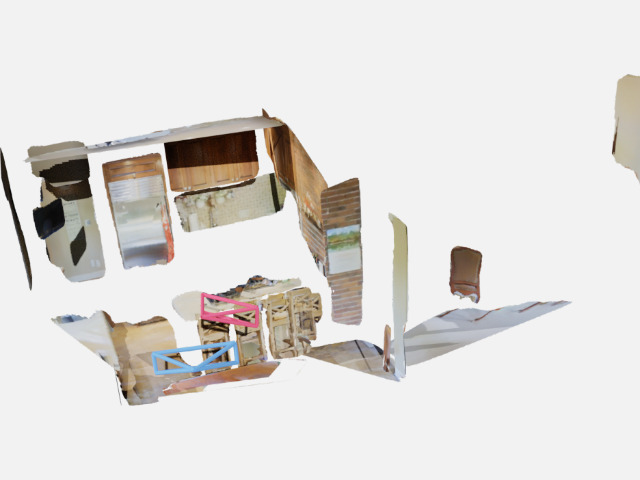}}
    & \frame{\includegraphics[width=0.148\textwidth]{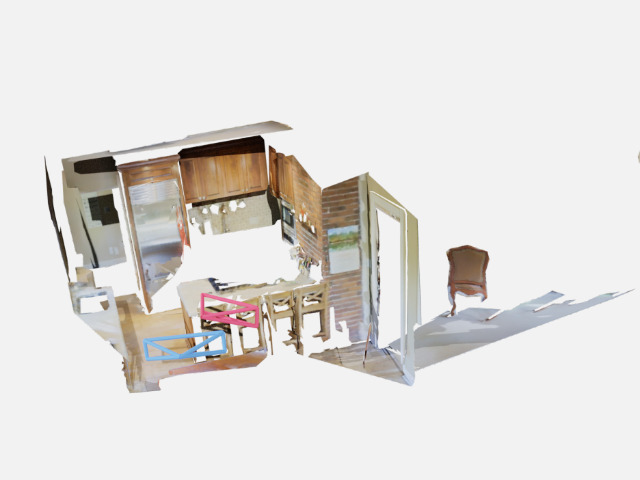}}\\

    \frame{\includegraphics[width=0.148\textwidth]{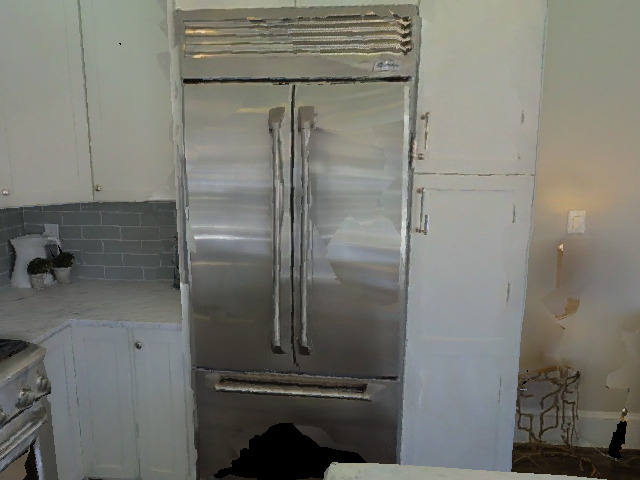}}
    & \frame{\includegraphics[width=0.148\textwidth]{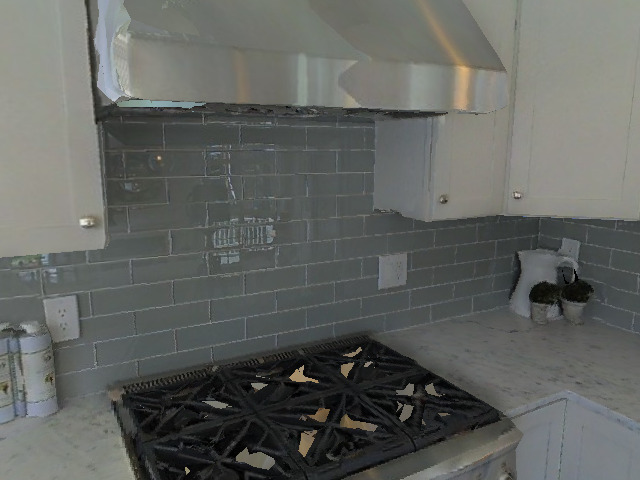}}
    & \frame{\includegraphics[width=0.148\textwidth]{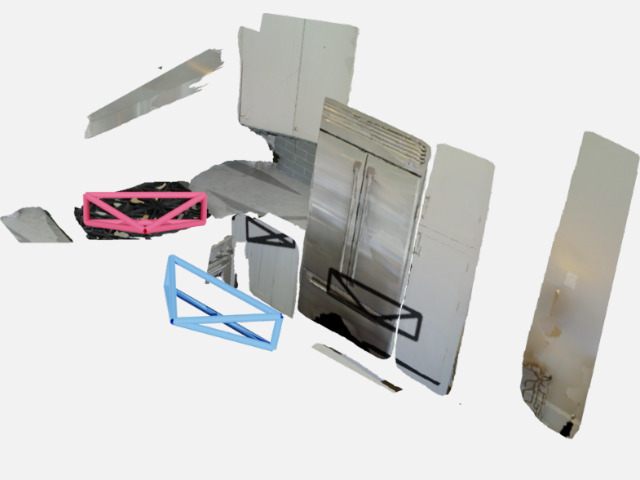}}
    & \frame{\includegraphics[width=0.148\textwidth]{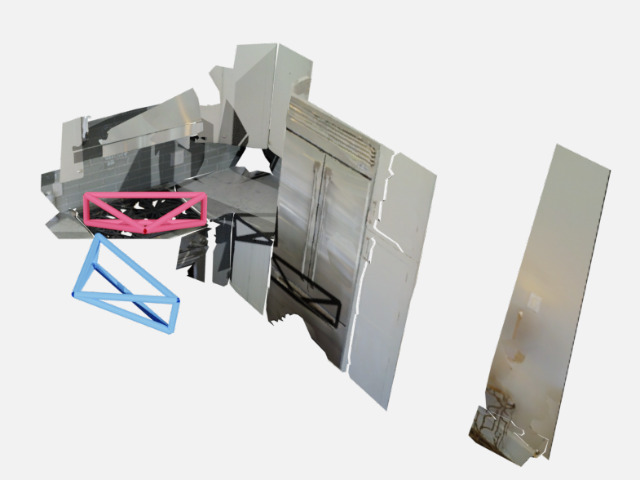}}
    & \frame{\includegraphics[width=0.148\textwidth]{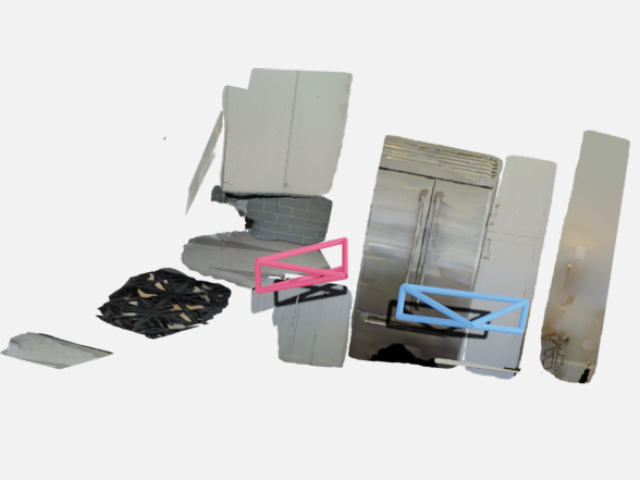}}
    & \frame{\includegraphics[width=0.148\textwidth]{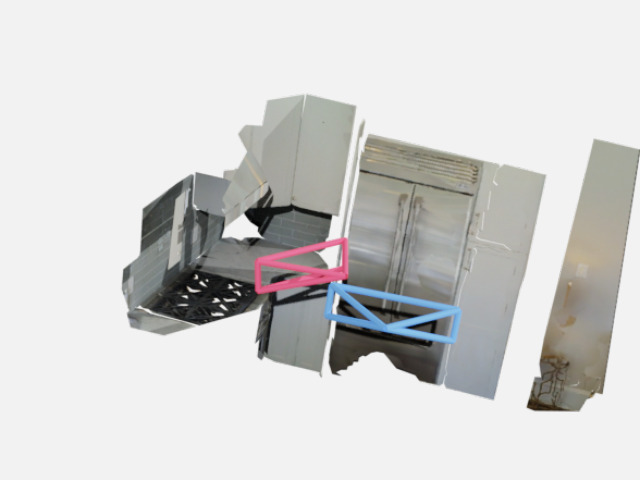}}\\

    \frame{\includegraphics[width=0.148\textwidth]{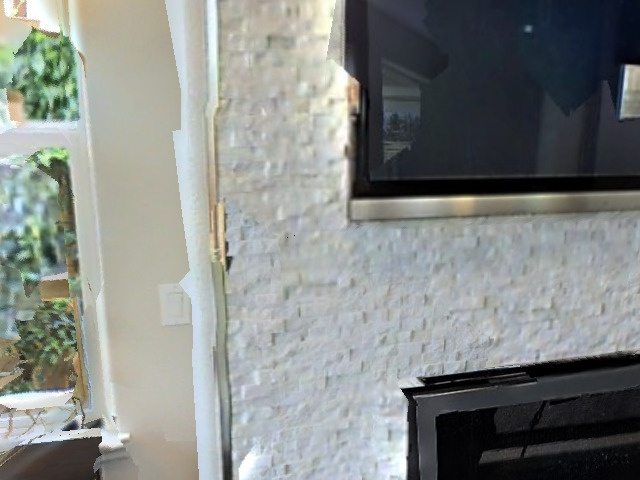}}
    & \frame{\includegraphics[width=0.148\textwidth]{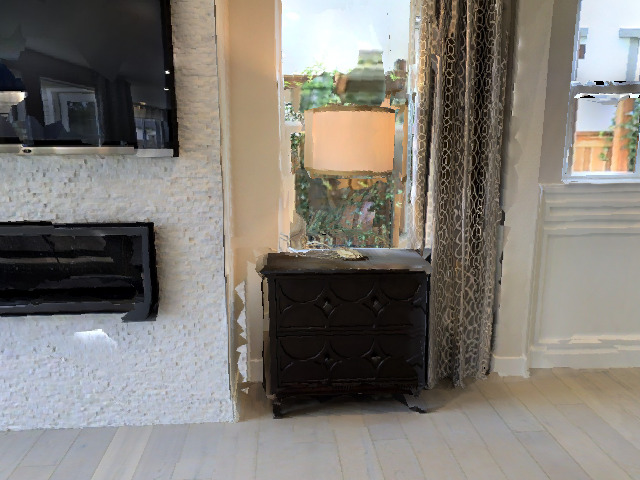}}
    & \frame{\includegraphics[width=0.148\textwidth]{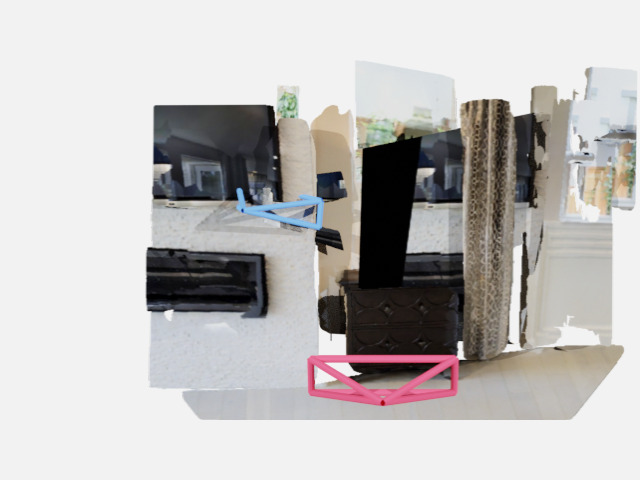}}
    & \frame{\includegraphics[width=0.148\textwidth]{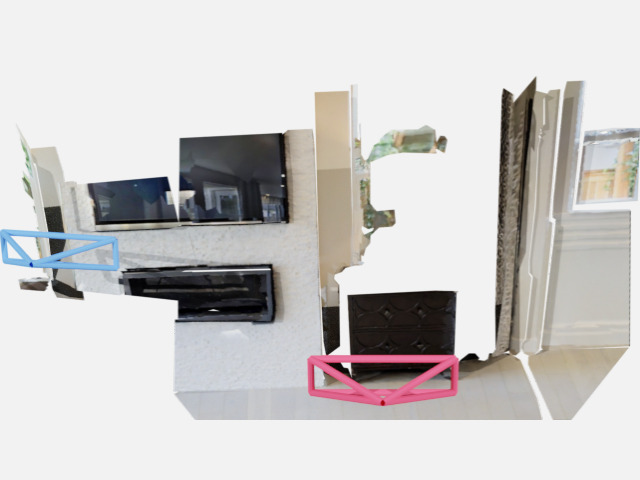}}
    & \frame{\includegraphics[width=0.148\textwidth]{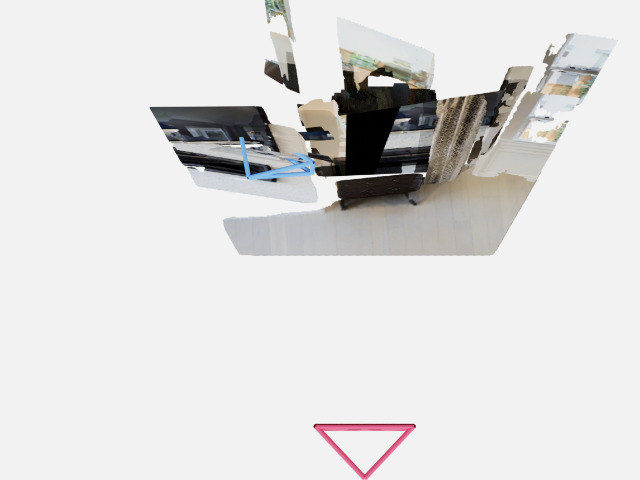}}
    & \frame{\includegraphics[width=0.148\textwidth]{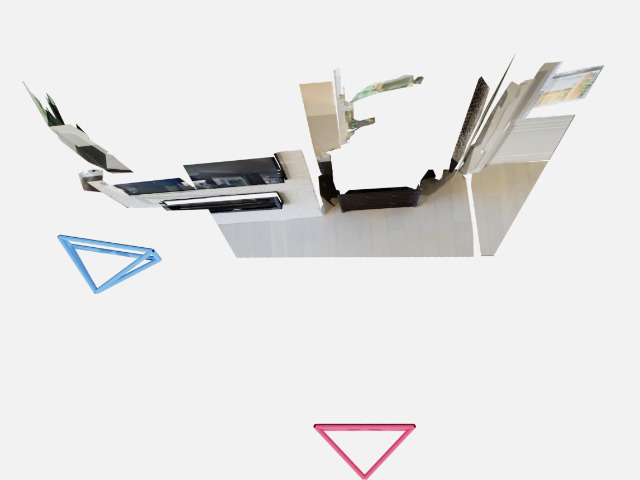}}\\

    \frame{\includegraphics[width=0.148\textwidth]{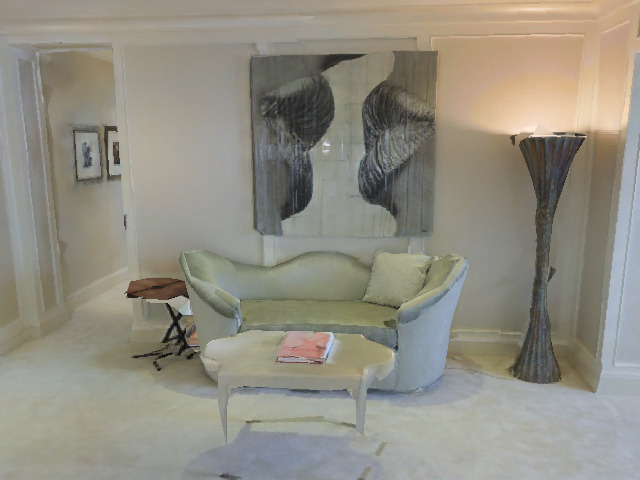}}
    & \frame{\includegraphics[width=0.148\textwidth]{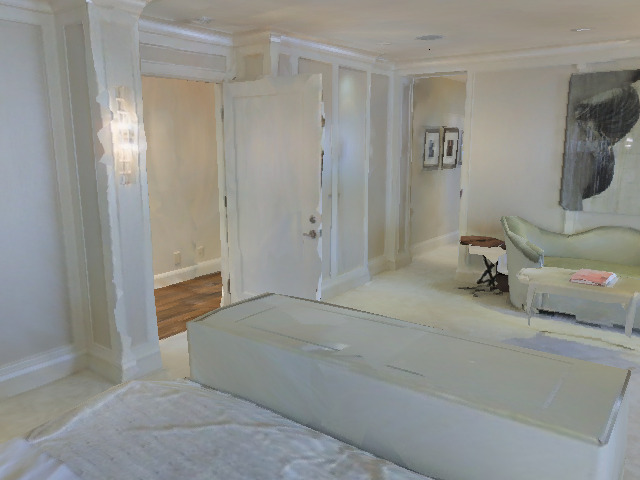}}
    & \frame{\includegraphics[width=0.148\textwidth]{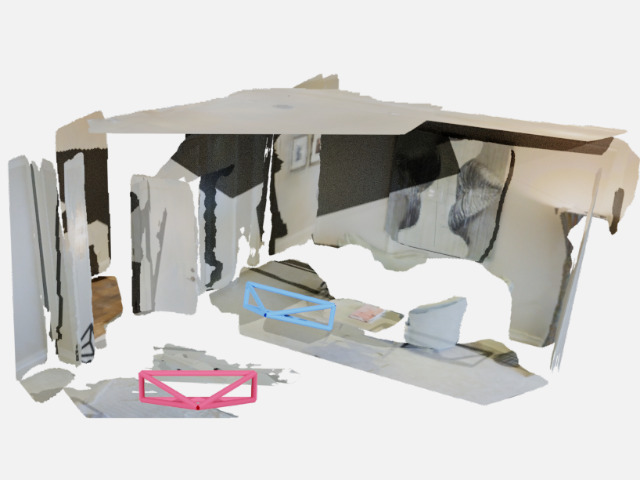}}
    & \frame{\includegraphics[width=0.148\textwidth]{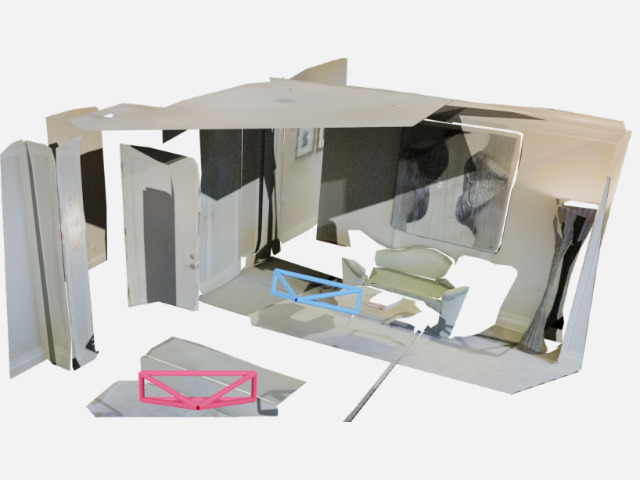}}
    & \frame{\includegraphics[width=0.148\textwidth]{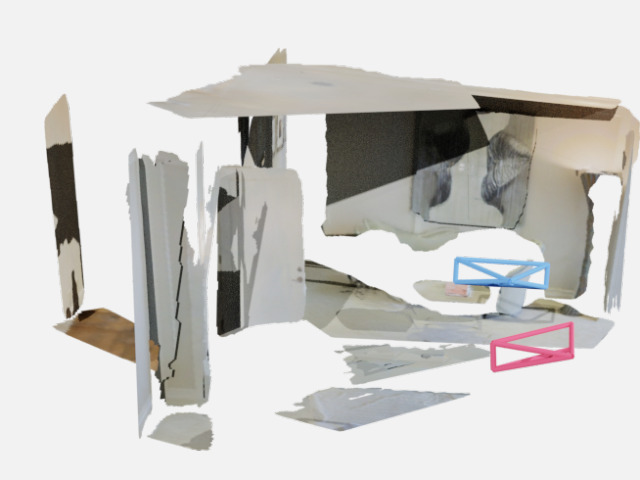}}
    & \frame{\includegraphics[width=0.148\textwidth]{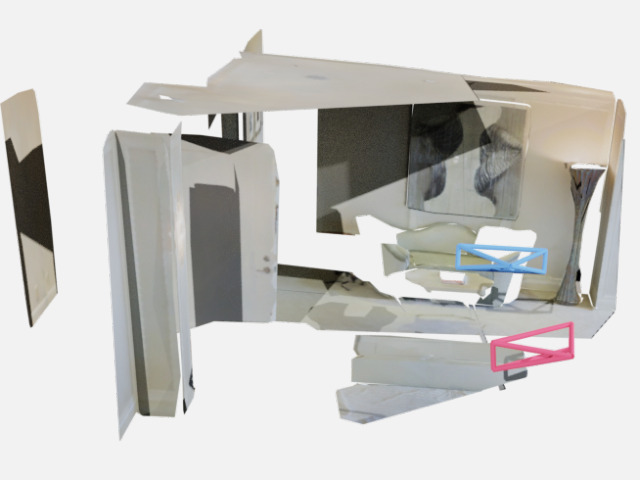}}\\

    \frame{\includegraphics[width=0.148\textwidth]{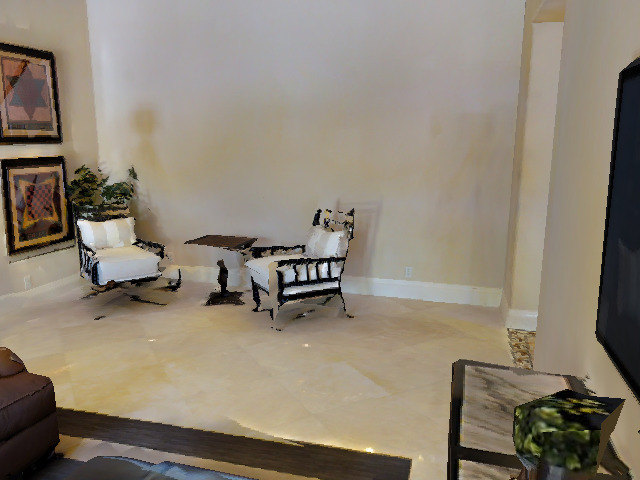}}
    & \frame{\includegraphics[width=0.148\textwidth]{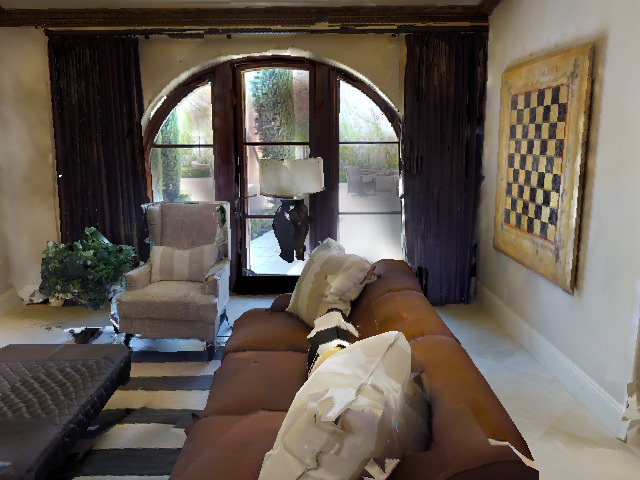}}
    & \frame{\includegraphics[width=0.148\textwidth]{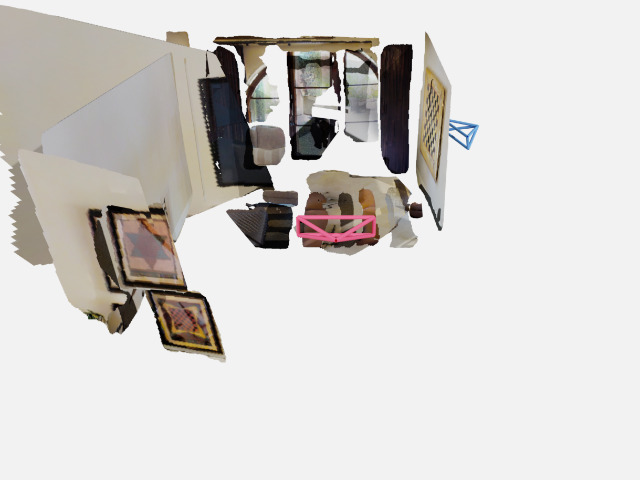}}
    & \frame{\includegraphics[width=0.148\textwidth]{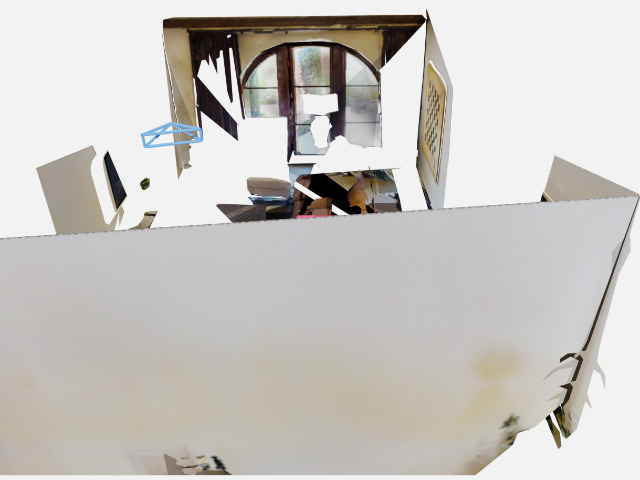}}
    & \frame{\includegraphics[width=0.148\textwidth]{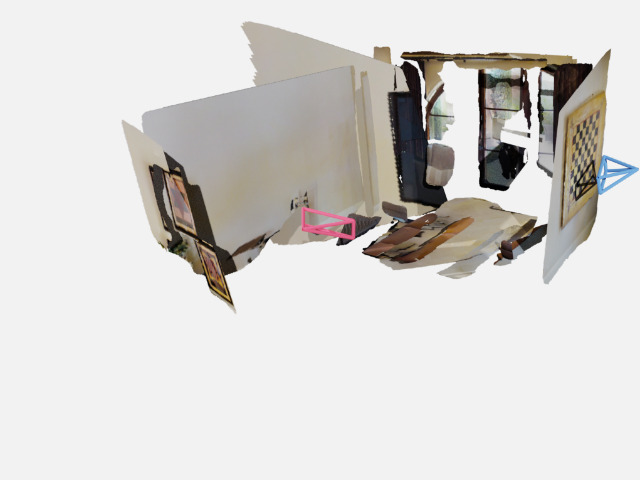}}
    & \frame{\includegraphics[width=0.148\textwidth]{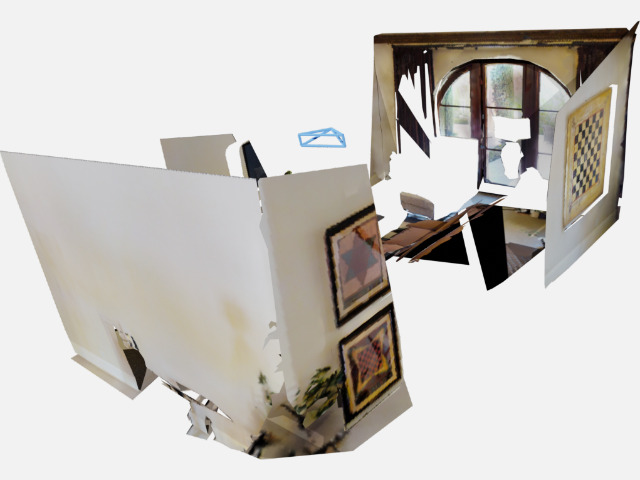}}\\

    \frame{\includegraphics[width=0.148\textwidth]{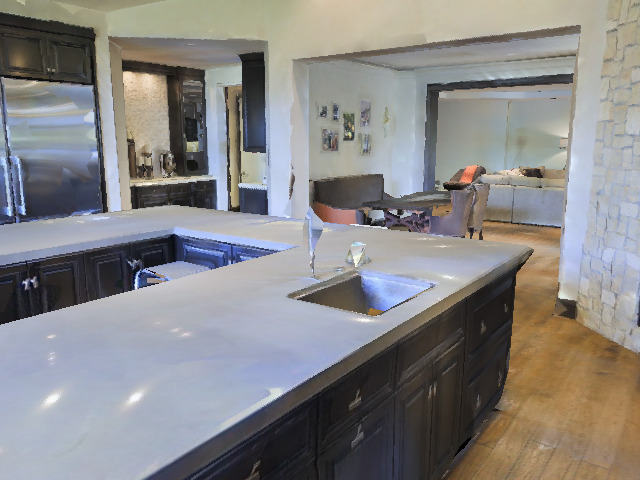}}
    & \frame{\includegraphics[width=0.148\textwidth]{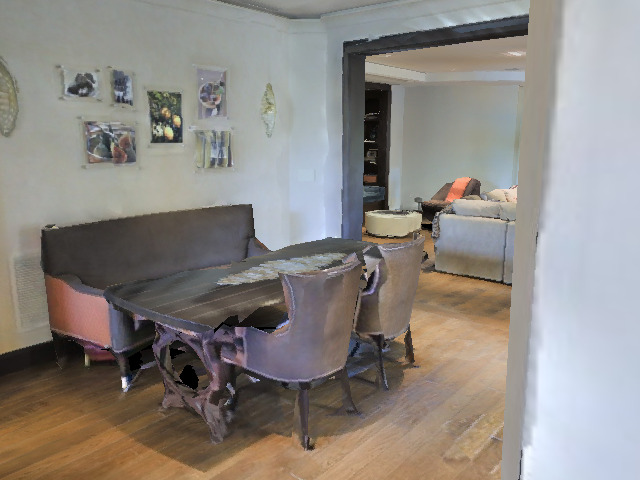}}
    & \frame{\includegraphics[width=0.148\textwidth]{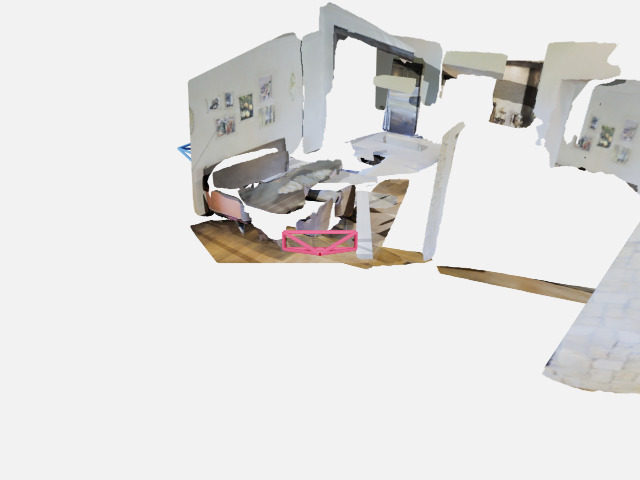}}
    & \frame{\includegraphics[width=0.148\textwidth]{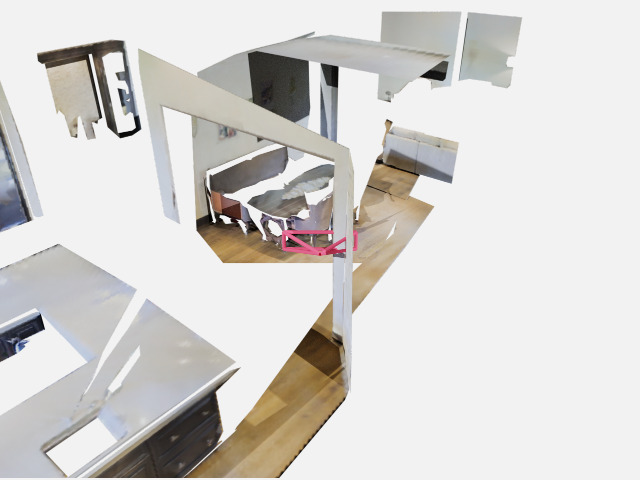}}
    & \frame{\includegraphics[width=0.148\textwidth]{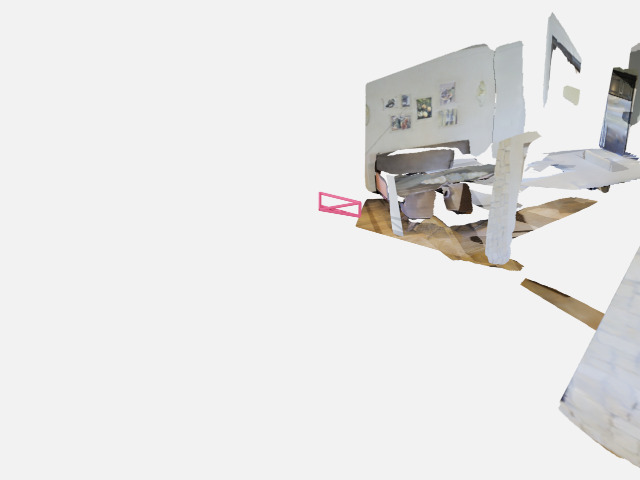}}
    & \frame{\includegraphics[width=0.148\textwidth]{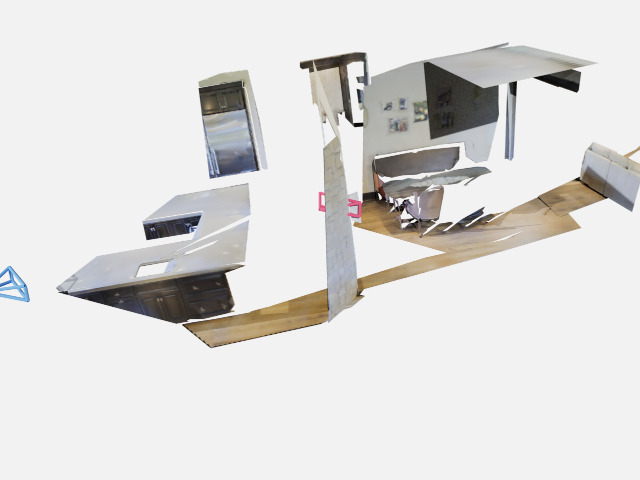}}\\

    \frame{\includegraphics[width=0.148\textwidth]{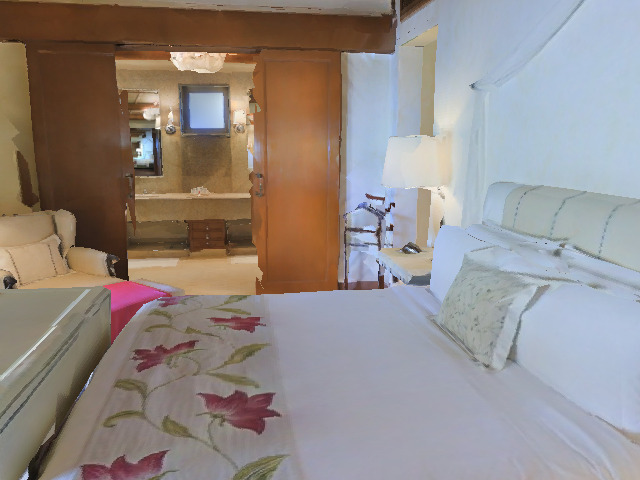}}
    & \frame{\includegraphics[width=0.148\textwidth]{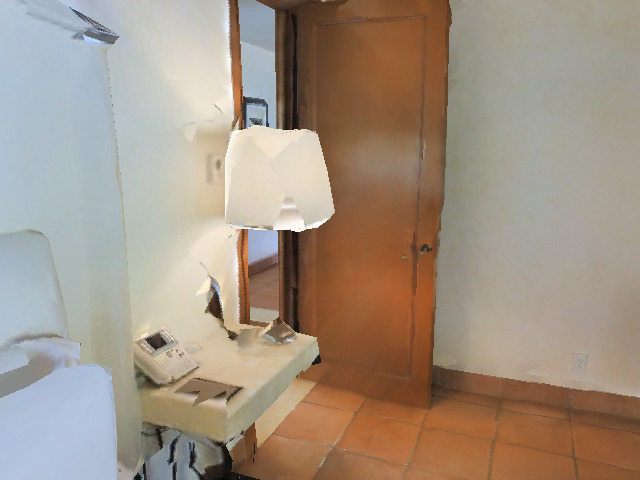}}
    & \frame{\includegraphics[width=0.148\textwidth]{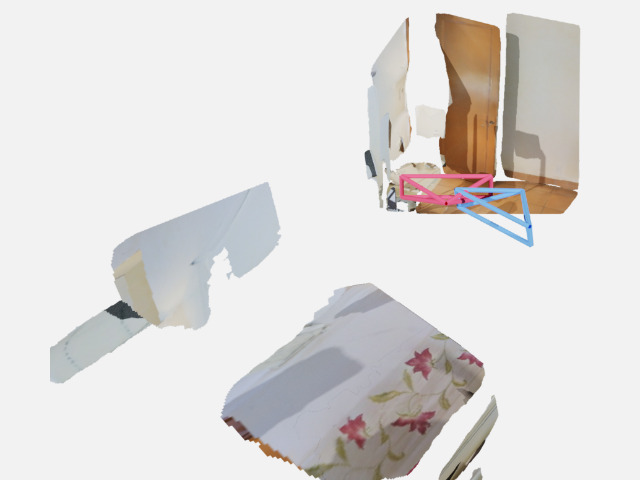}}
    & \frame{\includegraphics[width=0.148\textwidth]{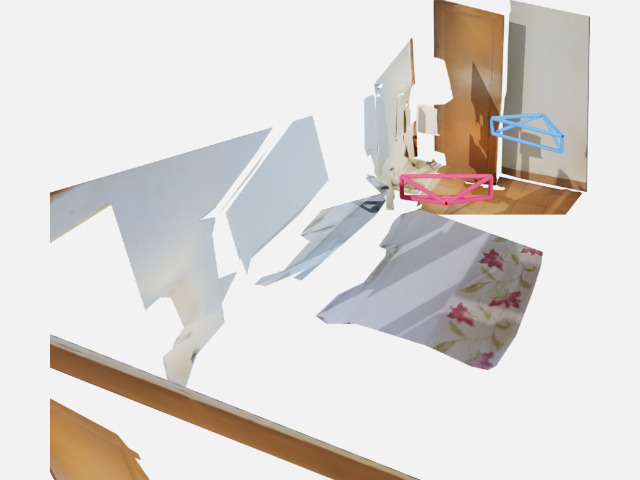}}
    & \frame{\includegraphics[width=0.148\textwidth]{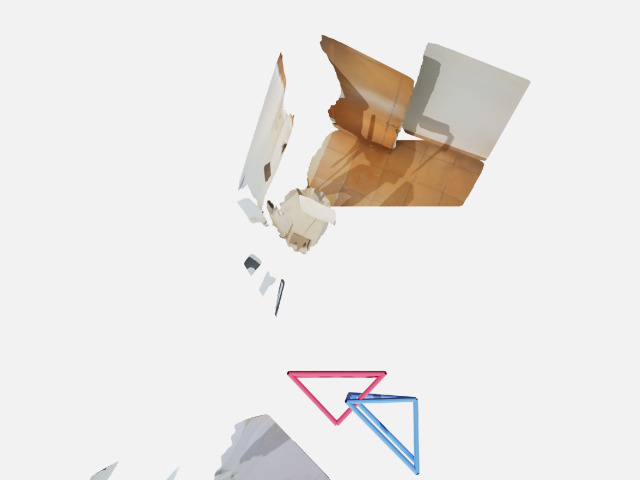}}
    & \frame{\includegraphics[width=0.148\textwidth]{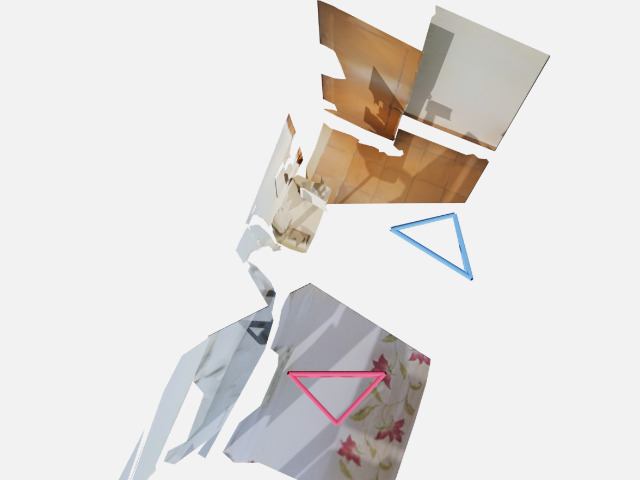}}\\

    \frame{\includegraphics[width=0.148\textwidth]{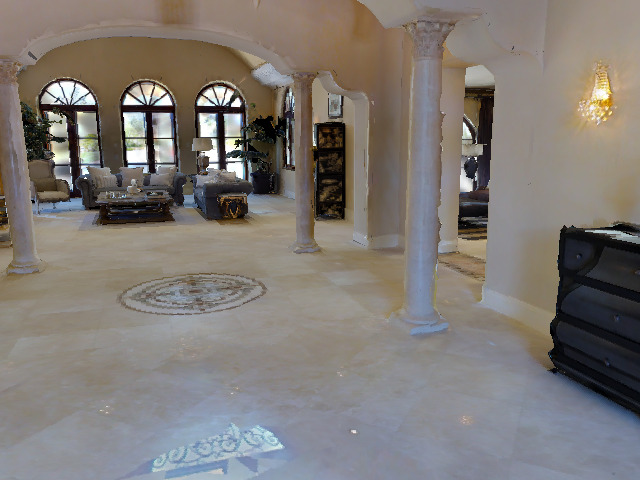}}
    & \frame{\includegraphics[width=0.148\textwidth]{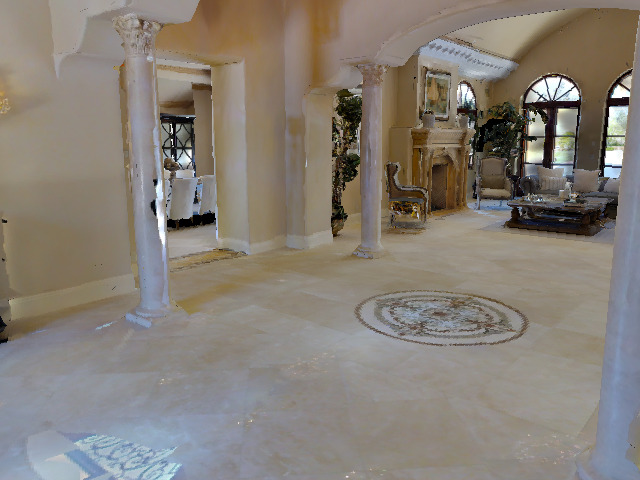}}
    & \frame{\includegraphics[width=0.148\textwidth]{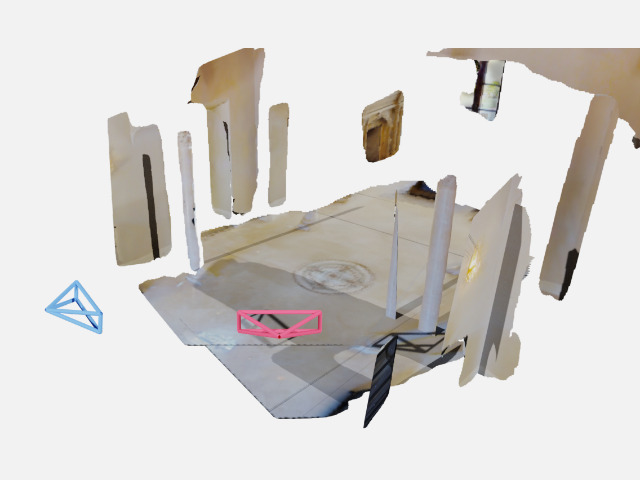}}
    & \frame{\includegraphics[width=0.148\textwidth]{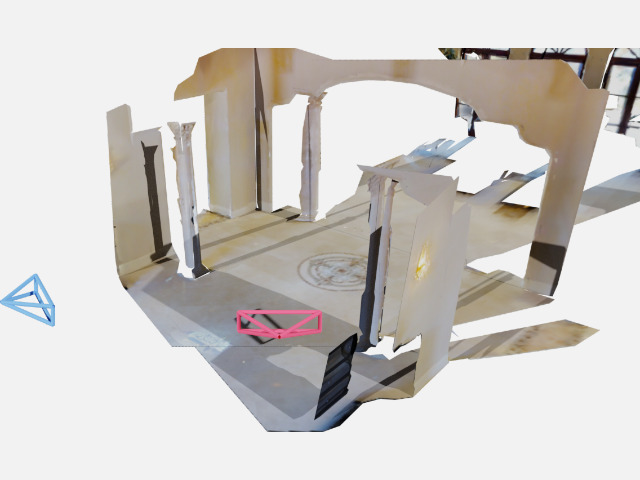}}
    & \frame{\includegraphics[width=0.148\textwidth]{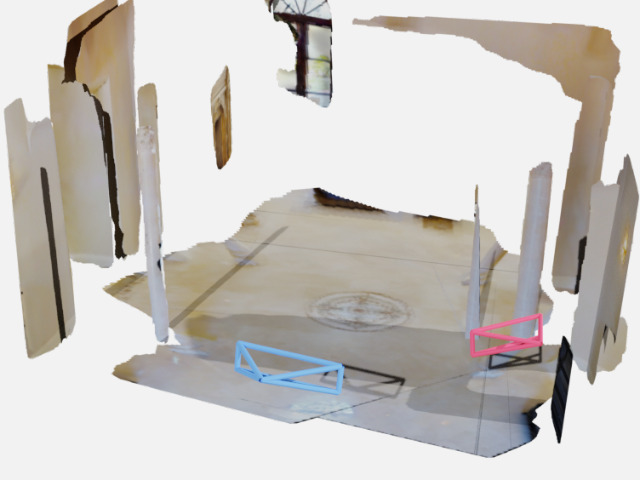}}
    & \frame{\includegraphics[width=0.148\textwidth]{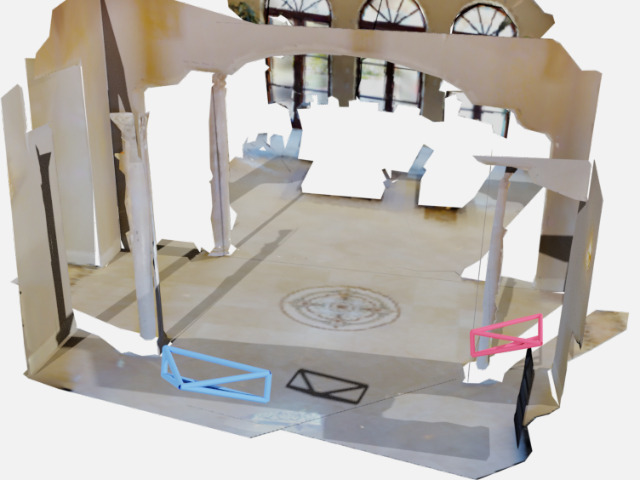}}\\

    \bottomrule
    \end{tabular}
    \caption{Random results on Matterport3D test set, automatically evenly spaced according to single-view AP. The first row has the highest single-view AP.
    \textbf{\textcolor{AccessibleBlue}{Blue}} and \textbf{\textcolor{AccessibleRed}{Red}} frustums show
    cameras for image 1 and 2.
    }
    \label{fig:supp-example-uniform}
\end{figure*}

We first present our reconstruction results on selected examples in Figure~\ref{fig:supp-example-wall}, extending Figure~\ref{fig:example-wall} in our paper. 
We show our prediction and ground truth from two novel views to see all planes in the whole scene -- a slightly raised view and a top down view.
We then present results automatically evenly spaced in the test set in Figure~\ref{fig:supp-example-uniform}, according to single-view AP.
As we quantitatively show in Section \ref{sec:single-view}, higher single-view AP means much better results.
Therefore, we hope the evenly spaced results can represent the overall performance of our approach better.
When single-view plane prediction is accurate, our reconstructions from sparse views are typically reasonable.

\begin{figure}[!t]
    \centering
    \scriptsize
    \begin{tabular}{c@{\hskip4pt}c@{\hskip4pt}c@{\hskip4pt}c@{\hskip4pt}c}
    \toprule
    
    Inputs & Appearance Only & ASNet~\cite{cai2020messytable} & Associative3D~\cite{Qian2020} & \textbf{Proposed} \\
    \midrule 
    \frame{\includegraphics[width=0.08\textwidth]{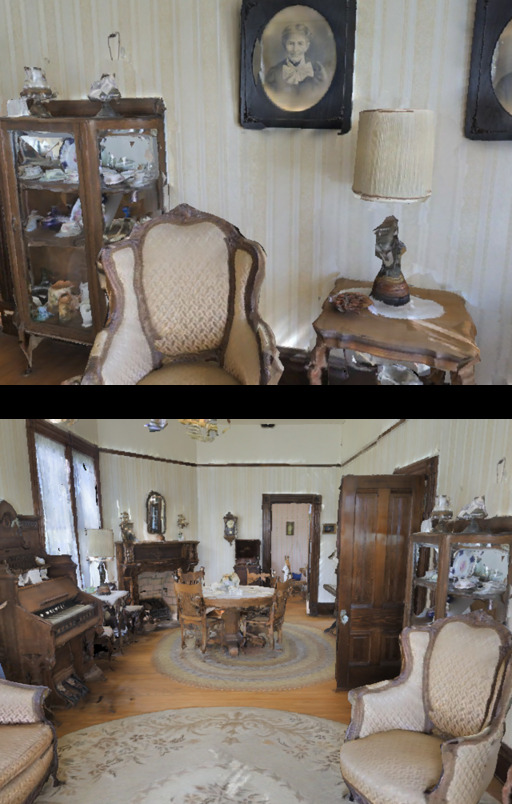}}
    & \frame{\includegraphics[width=0.08\textwidth]{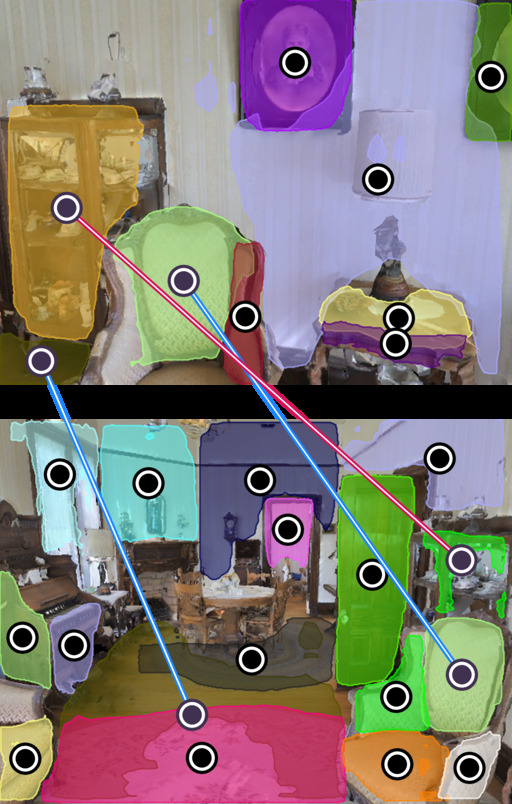}}
    & \frame{\includegraphics[width=0.08\textwidth]{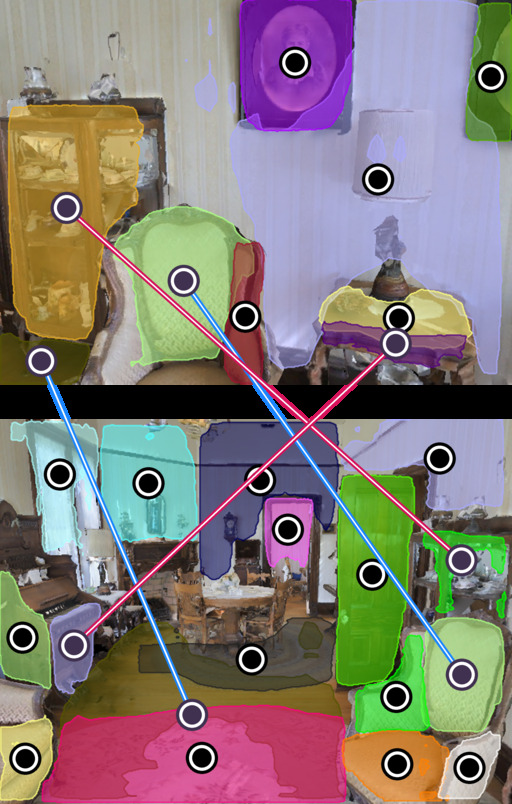}}
    & \frame{\includegraphics[width=0.08\textwidth]{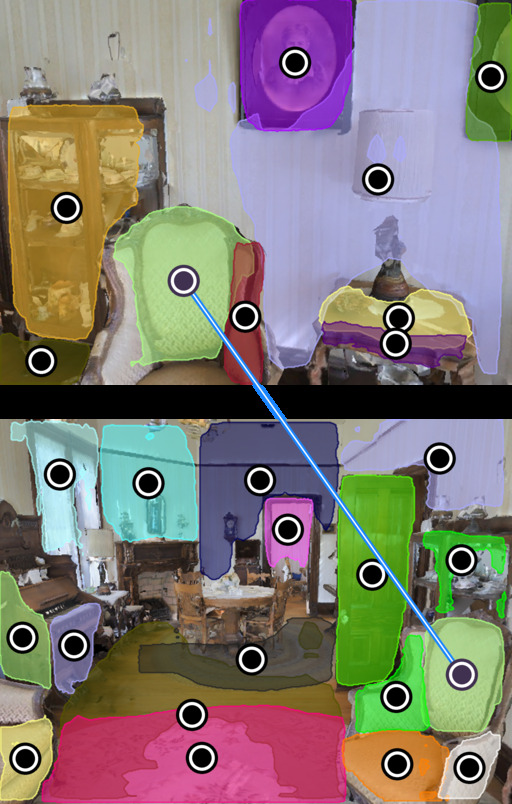}}
    & \frame{\includegraphics[width=0.08\textwidth]{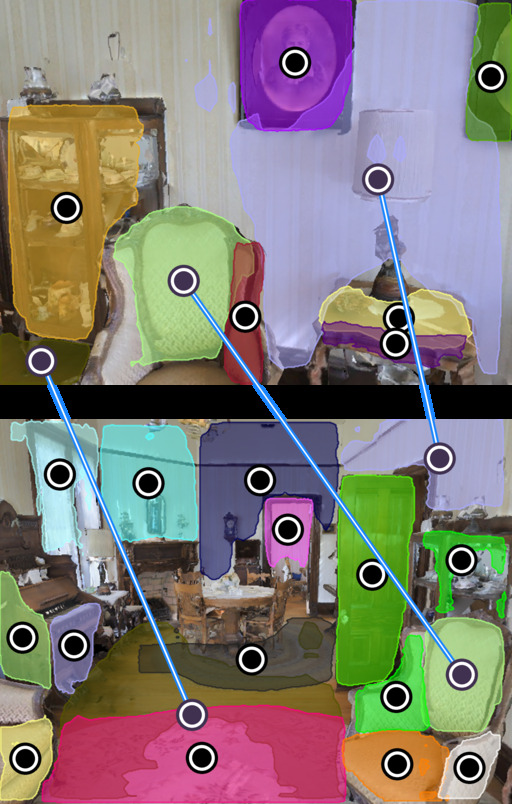}}\\

    \frame{\includegraphics[width=0.08\textwidth]{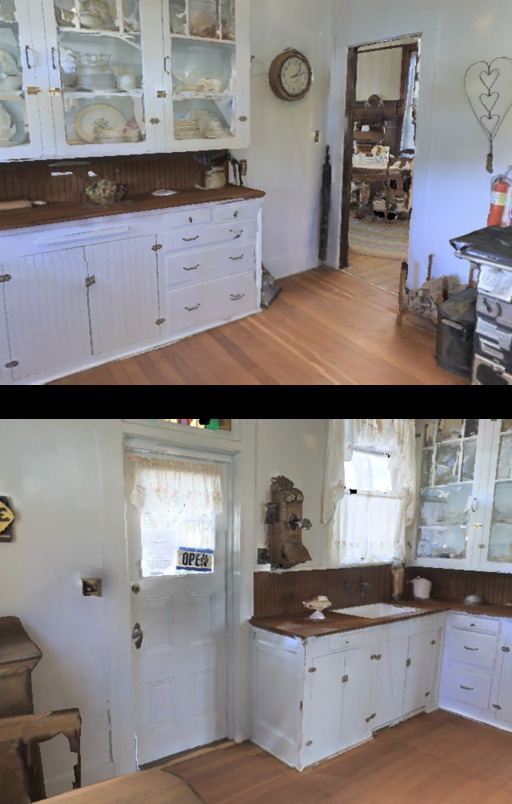}}
    & \frame{\includegraphics[width=0.08\textwidth]{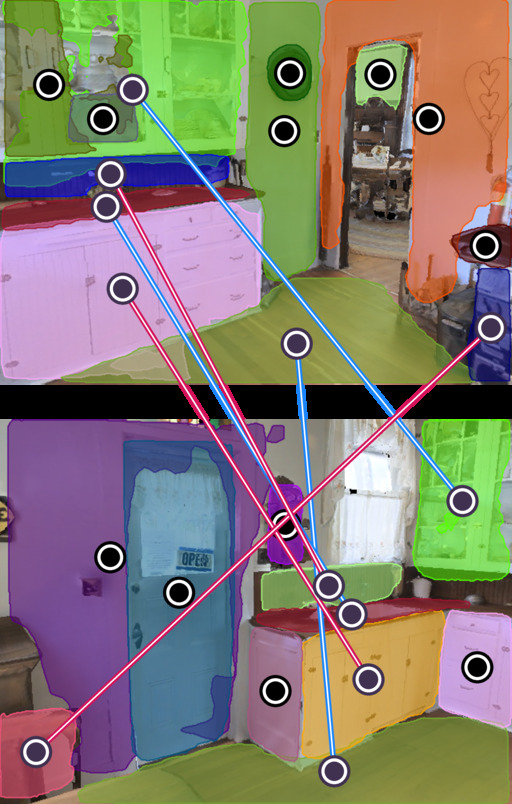}}
    & \frame{\includegraphics[width=0.08\textwidth]{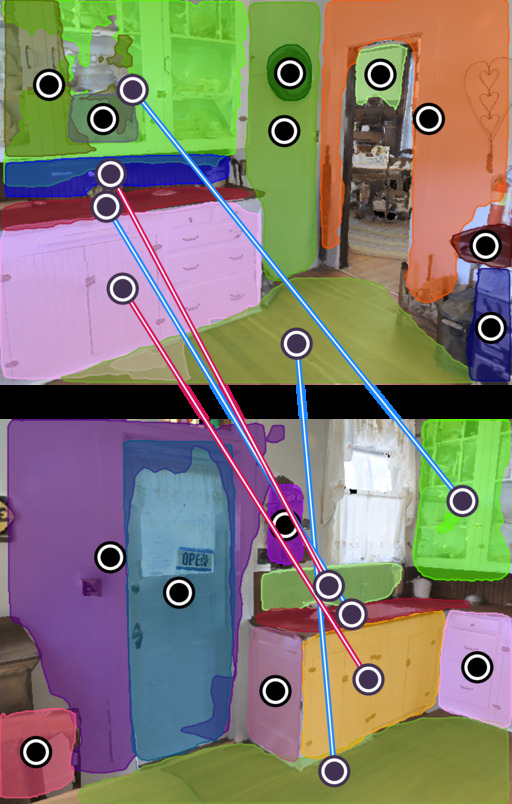}}
    & \frame{\includegraphics[width=0.08\textwidth]{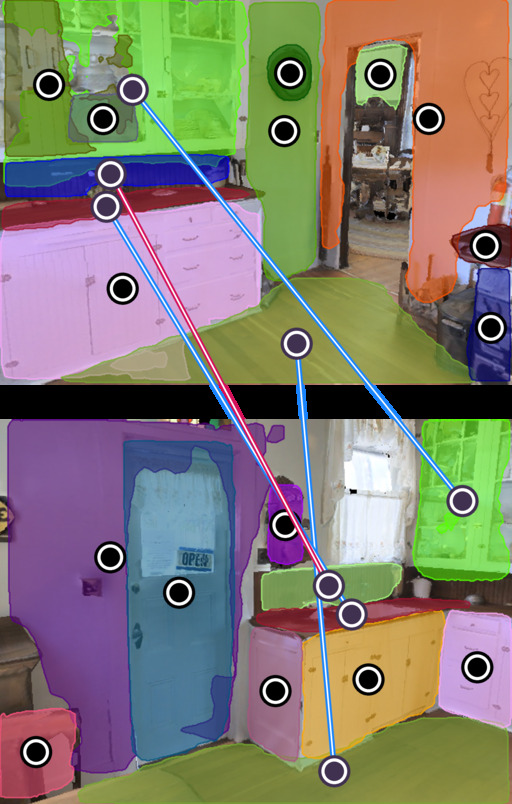}}
    & \frame{\includegraphics[width=0.08\textwidth]{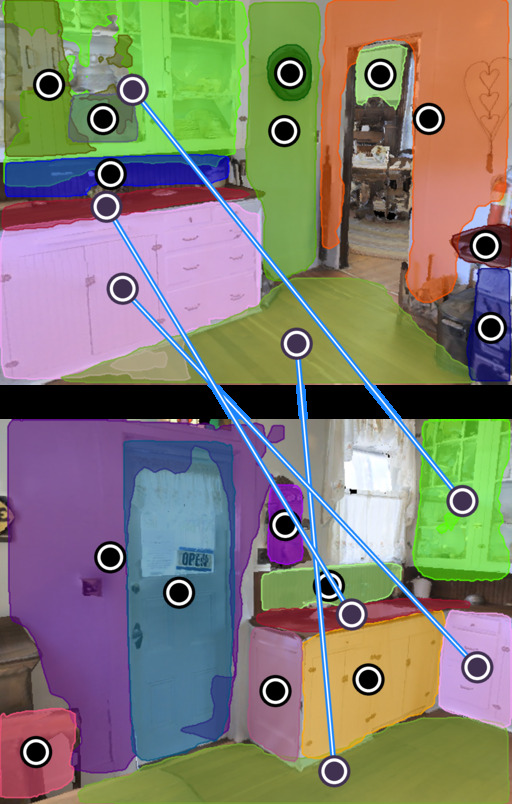}}\\ 

    \frame{\includegraphics[width=0.08\textwidth]{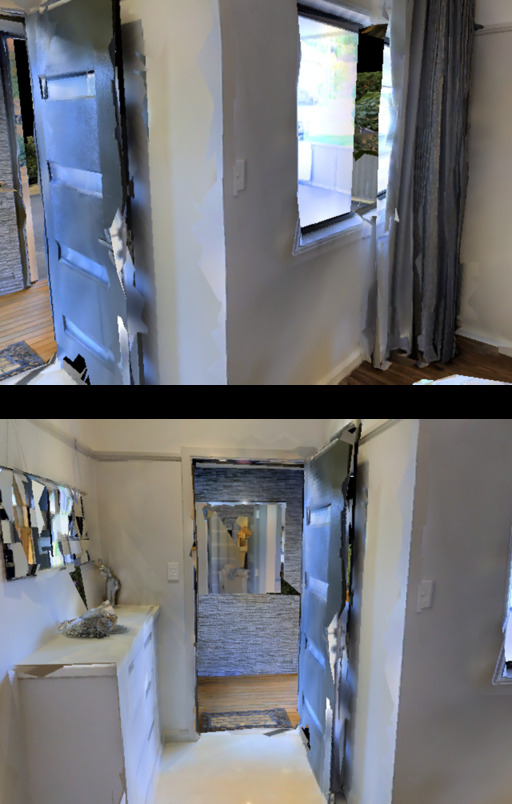}}
    & \frame{\includegraphics[width=0.08\textwidth]{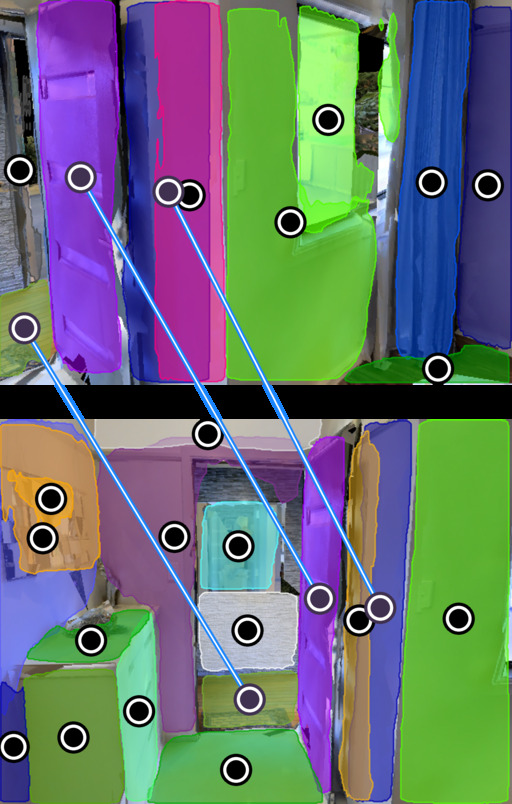}}
    & \frame{\includegraphics[width=0.08\textwidth]{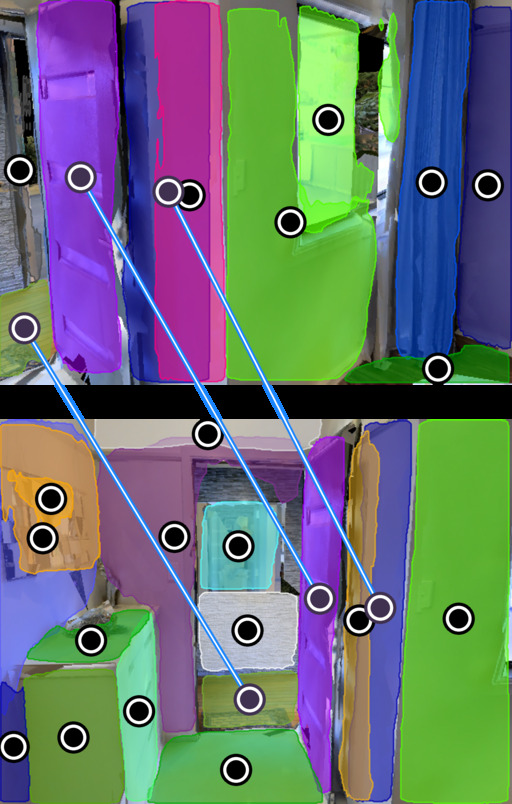}}
    & \frame{\includegraphics[width=0.08\textwidth]{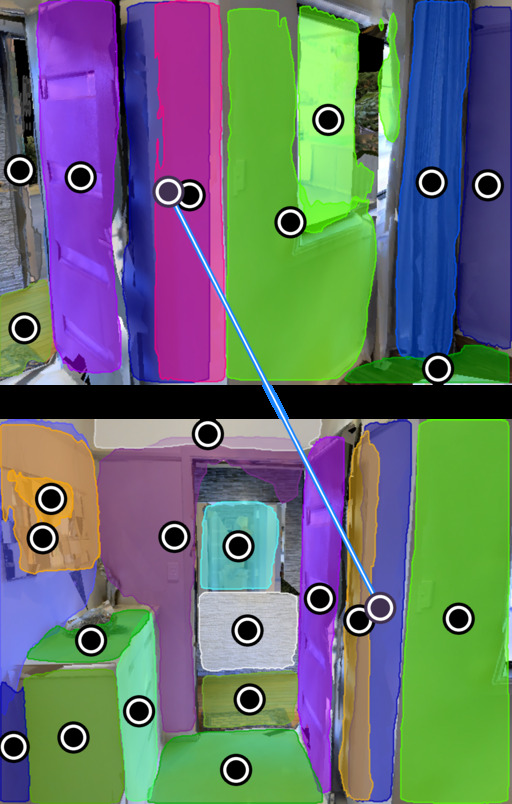}}
    & \frame{\includegraphics[width=0.08\textwidth]{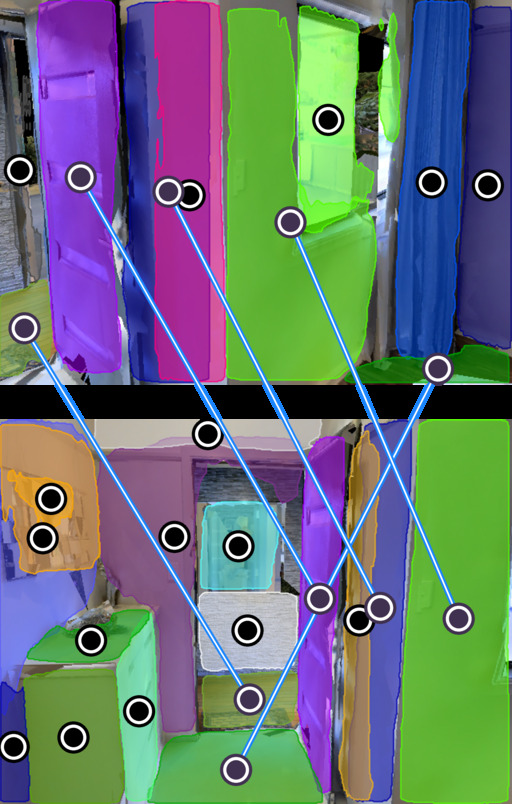}}\\

    \frame{\includegraphics[width=0.08\textwidth]{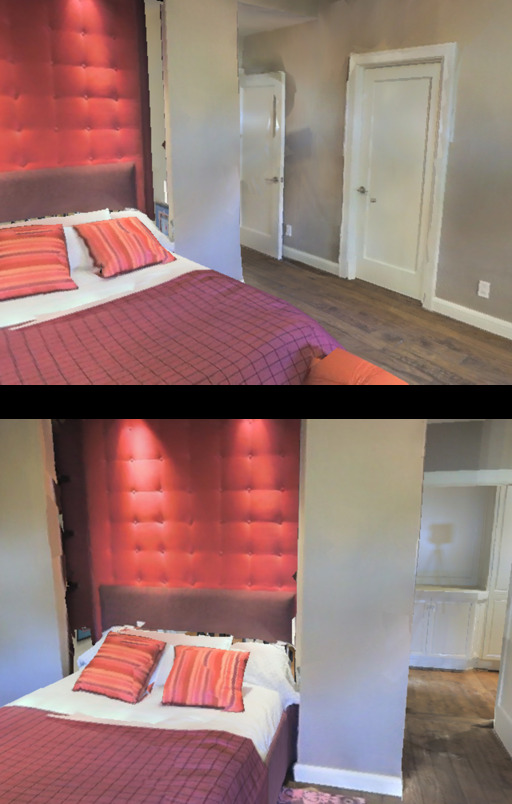}}
    & \frame{\includegraphics[width=0.08\textwidth]{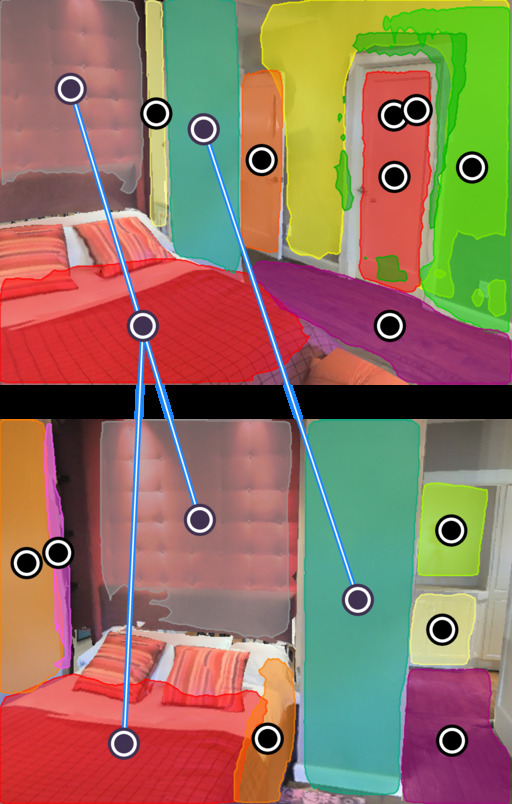}}
    & \frame{\includegraphics[width=0.08\textwidth]{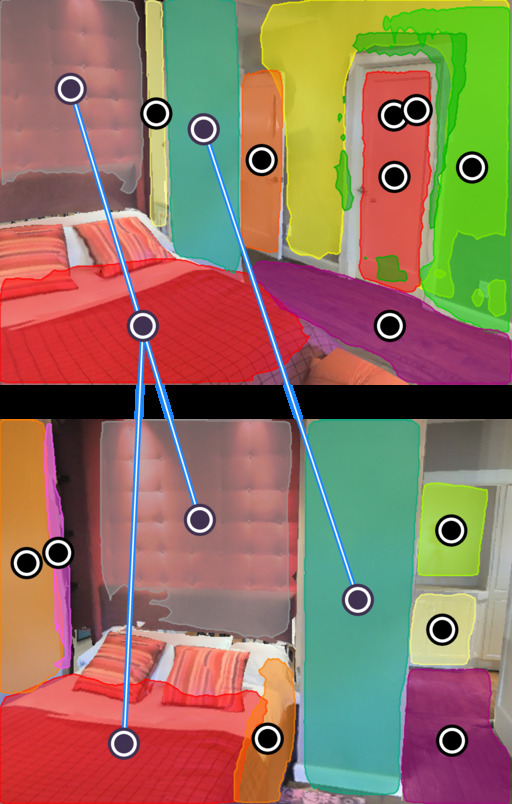}}
    & \frame{\includegraphics[width=0.08\textwidth]{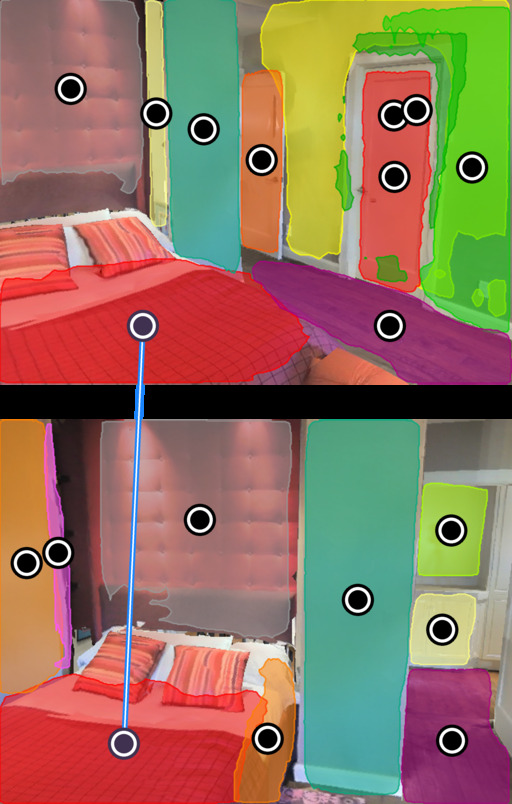}}
    & \frame{\includegraphics[width=0.08\textwidth]{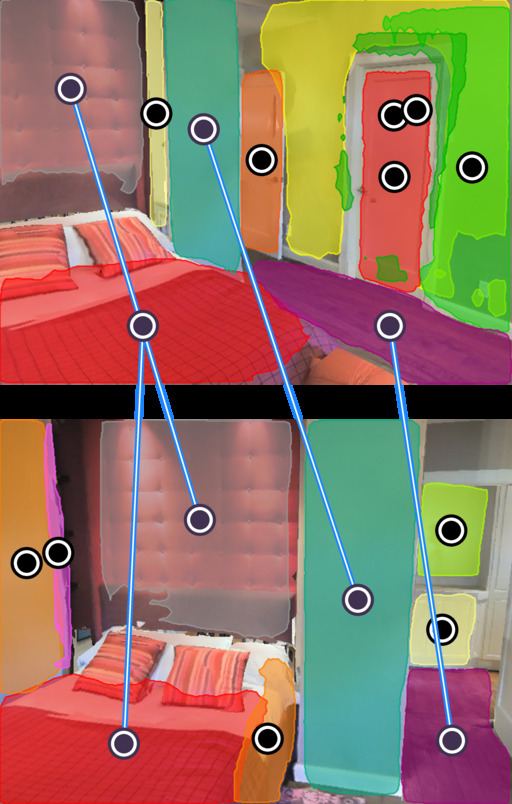}}\\

    \frame{\includegraphics[width=0.08\textwidth]{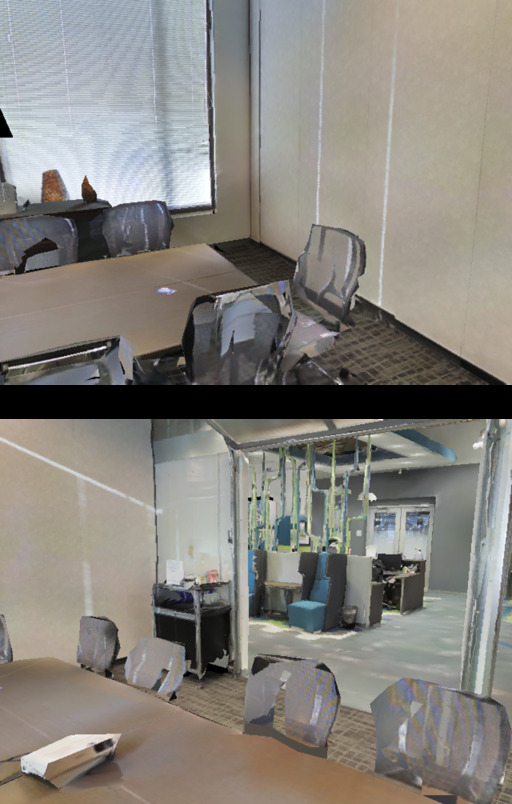}}
    & \frame{\includegraphics[width=0.08\textwidth]{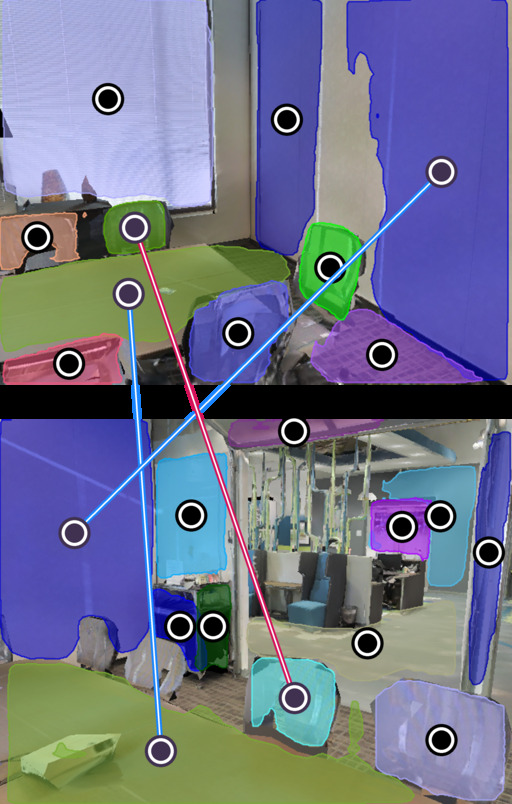}}
    & \frame{\includegraphics[width=0.08\textwidth]{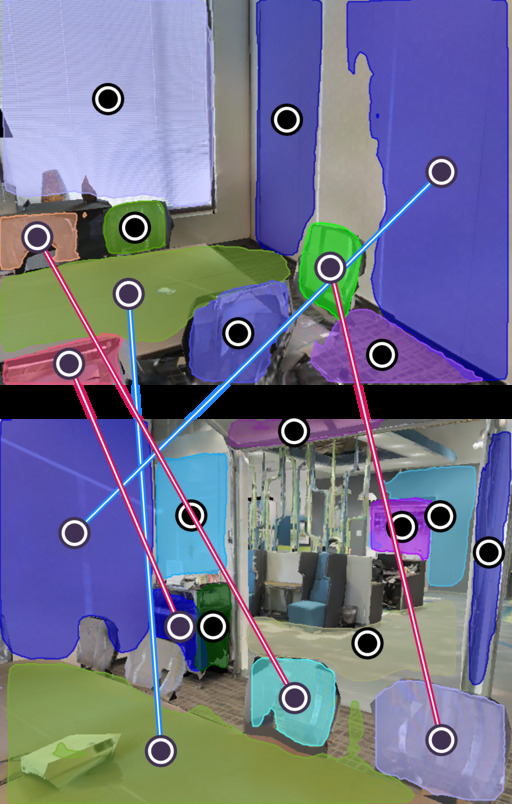}}
    & \frame{\includegraphics[width=0.08\textwidth]{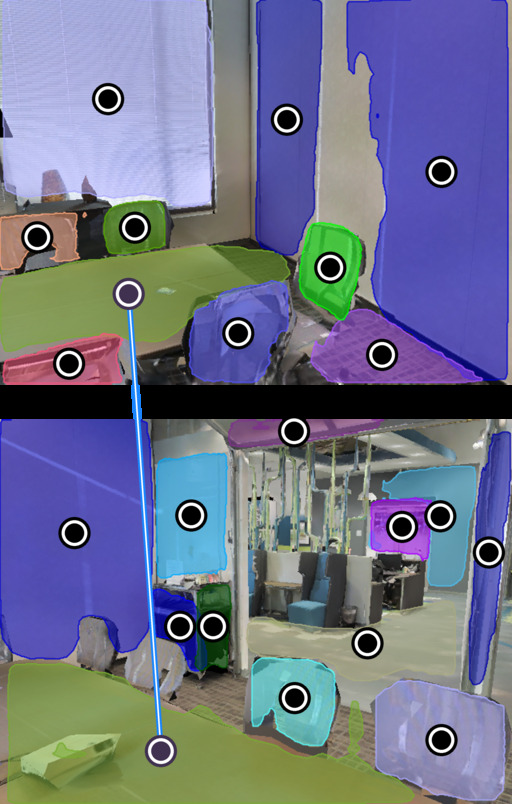}}
    & \frame{\includegraphics[width=0.08\textwidth]{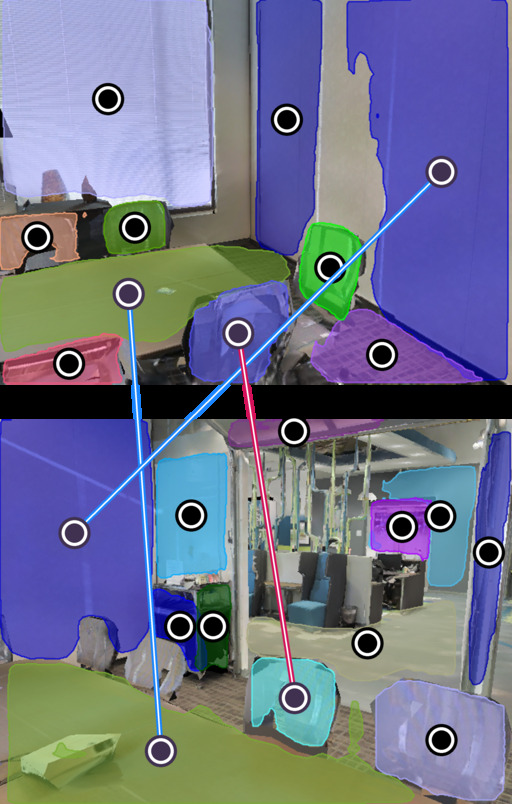}}\\

    \frame{\includegraphics[width=0.08\textwidth]{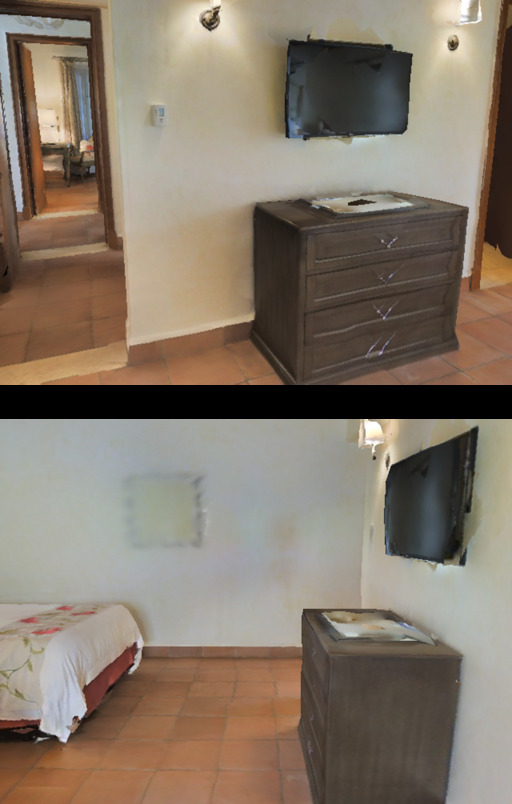}}
    & \frame{\includegraphics[width=0.08\textwidth]{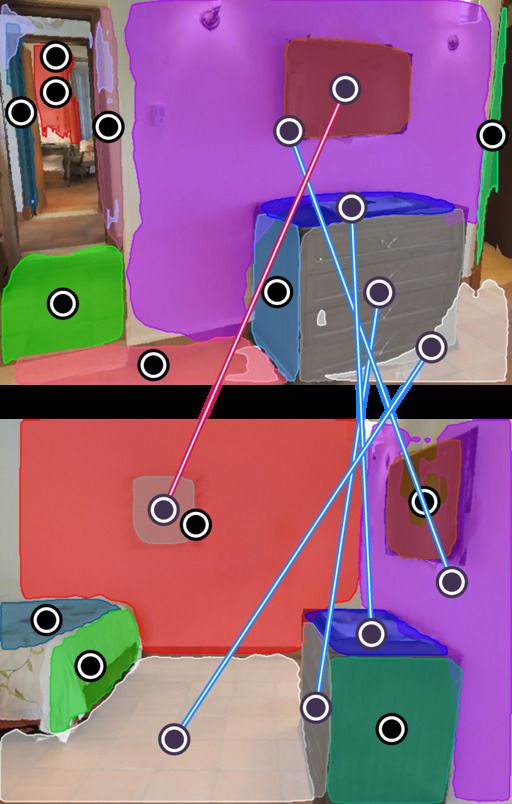}}
    & \frame{\includegraphics[width=0.08\textwidth]{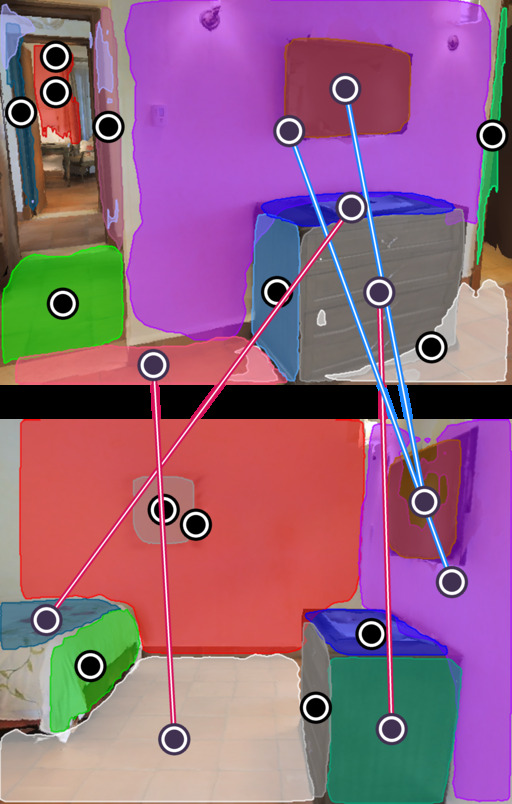}}
    & \frame{\includegraphics[width=0.08\textwidth]{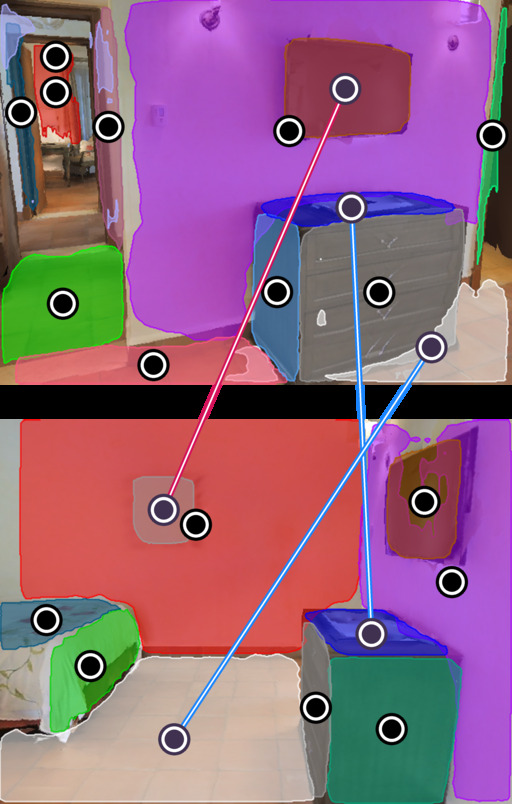}}
    & \frame{\includegraphics[width=0.08\textwidth]{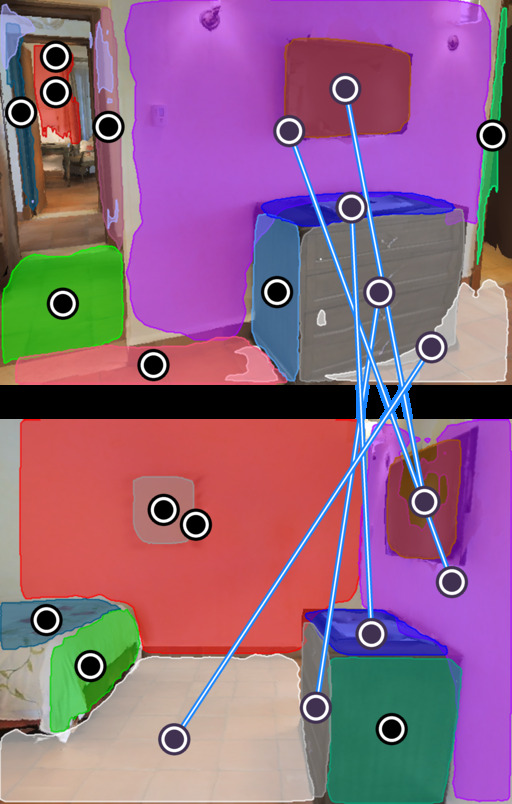}}\\

    \frame{\includegraphics[width=0.08\textwidth]{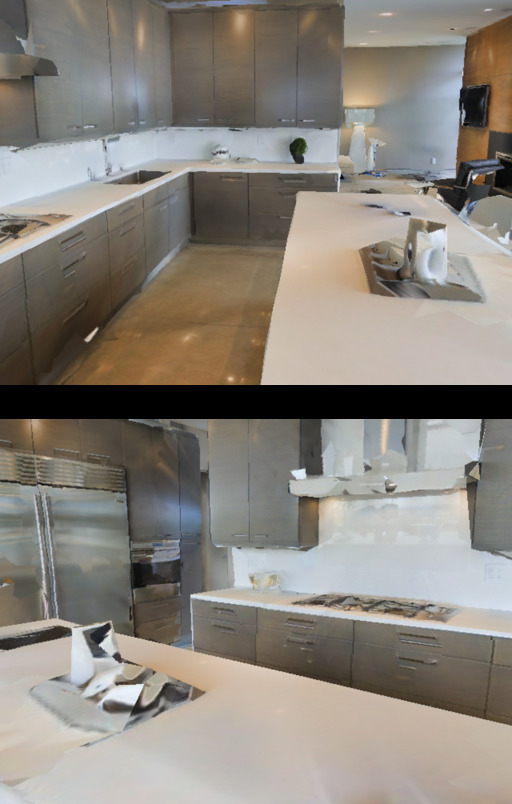}}
    & \frame{\includegraphics[width=0.08\textwidth]{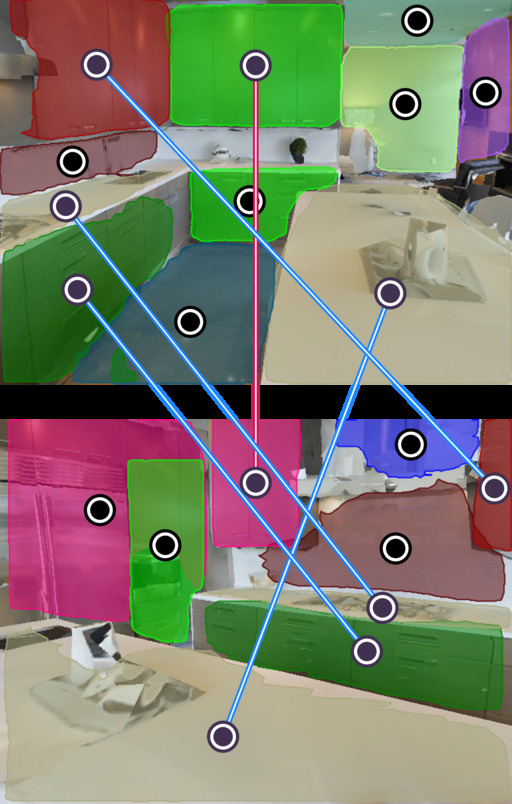}}
    & \frame{\includegraphics[width=0.08\textwidth]{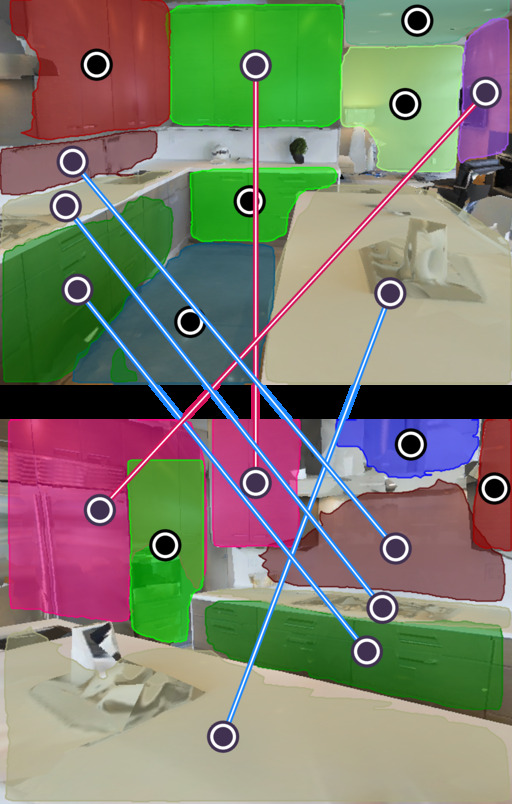}}
    & \frame{\includegraphics[width=0.08\textwidth]{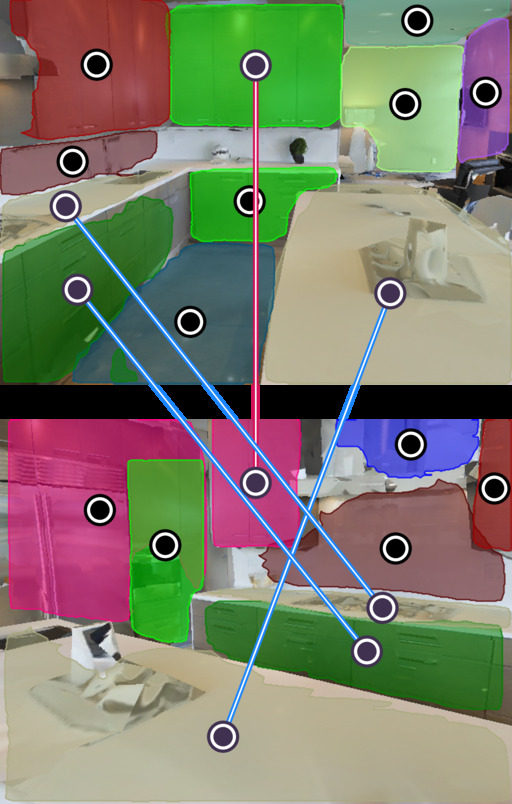}}
    & \frame{\includegraphics[width=0.08\textwidth]{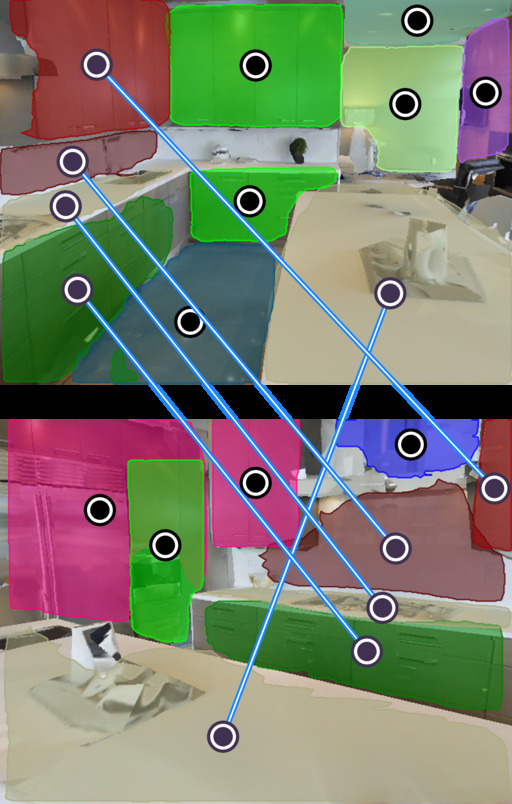}}\\
    
    \frame{\includegraphics[width=0.08\textwidth]{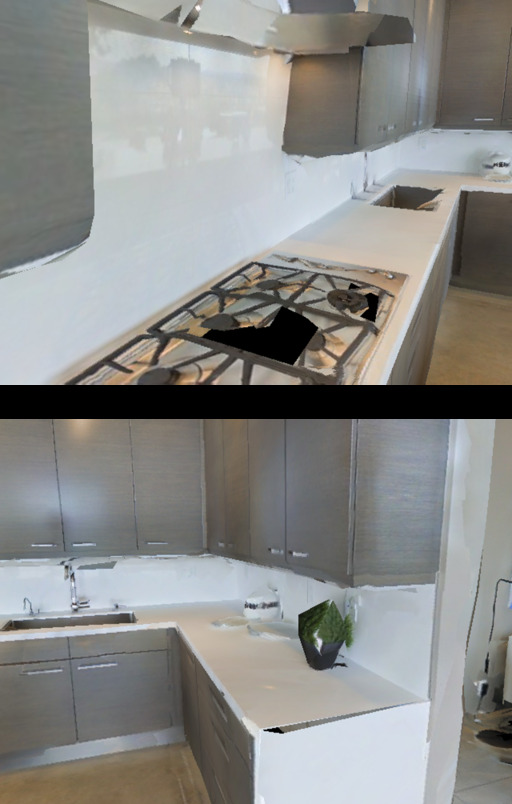}}
    & \frame{\includegraphics[width=0.08\textwidth]{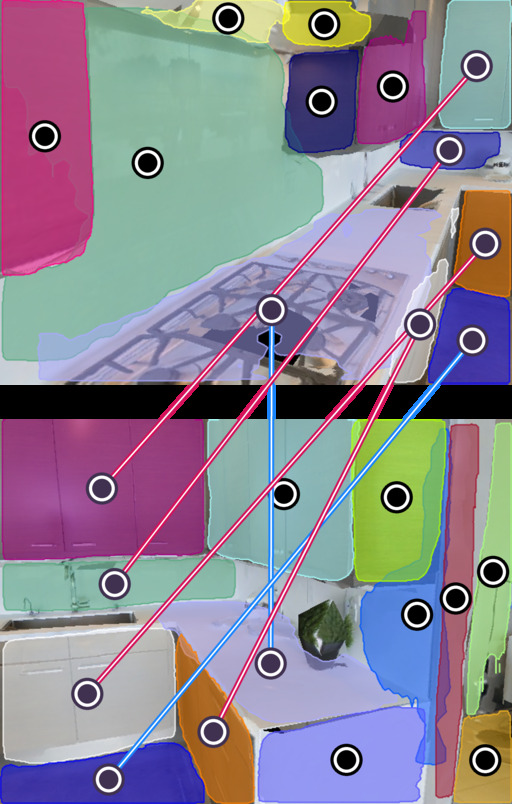}}
    & \frame{\includegraphics[width=0.08\textwidth]{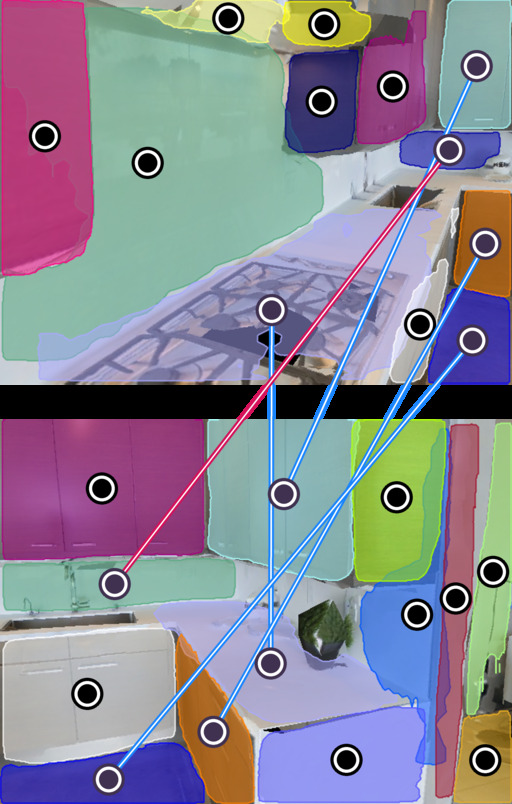}}
    & \frame{\includegraphics[width=0.08\textwidth]{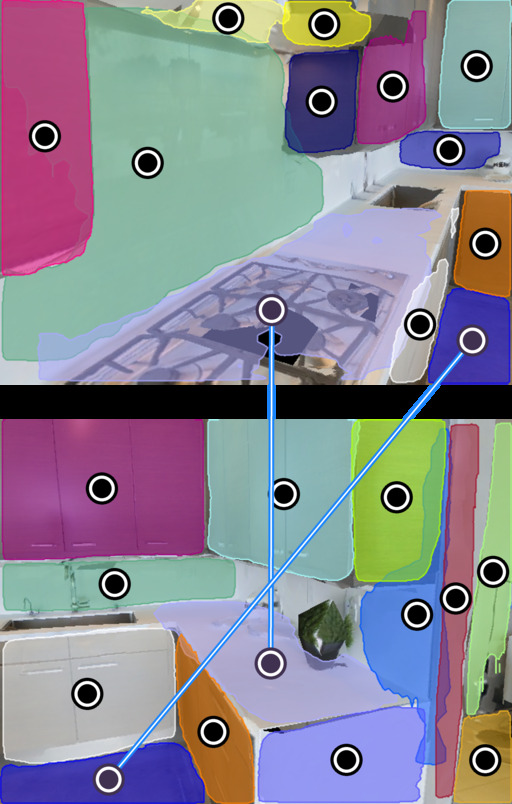}}
    & \frame{\includegraphics[width=0.08\textwidth]{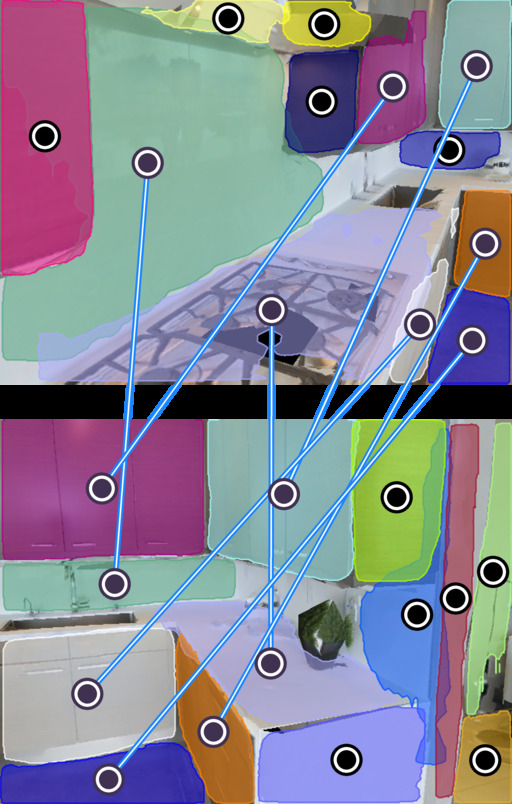}}\\

    \frame{\includegraphics[width=0.08\textwidth]{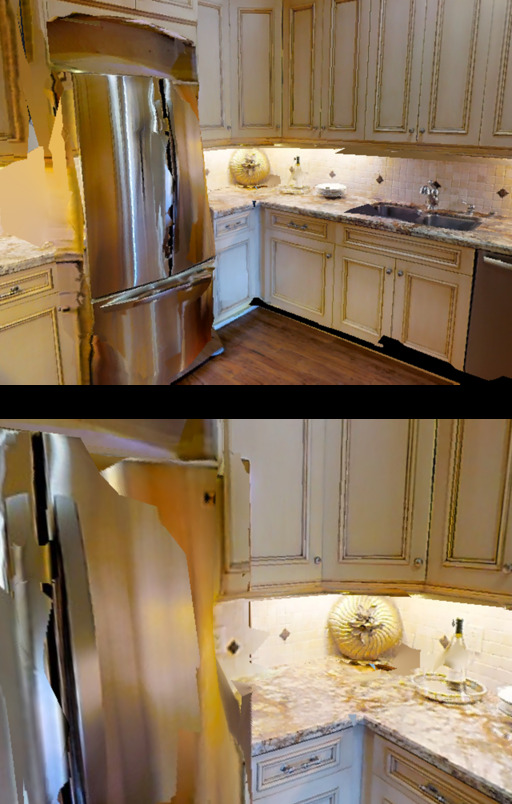}}
    & \frame{\includegraphics[width=0.08\textwidth]{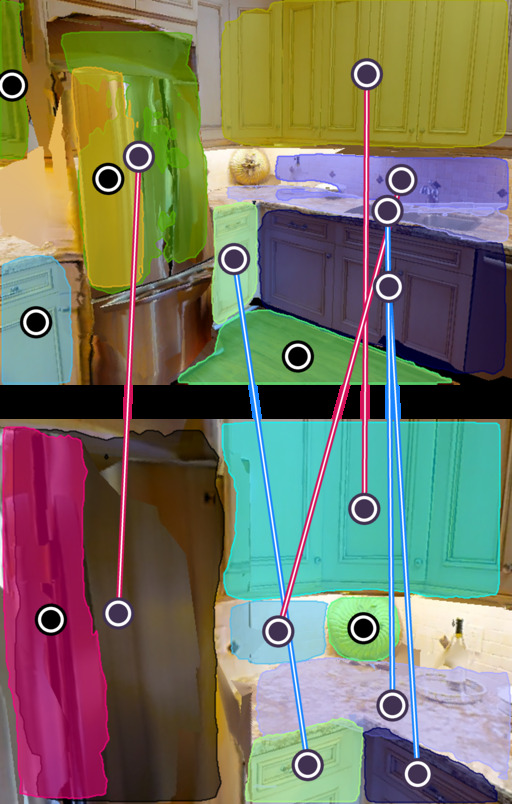}}
    & \frame{\includegraphics[width=0.08\textwidth]{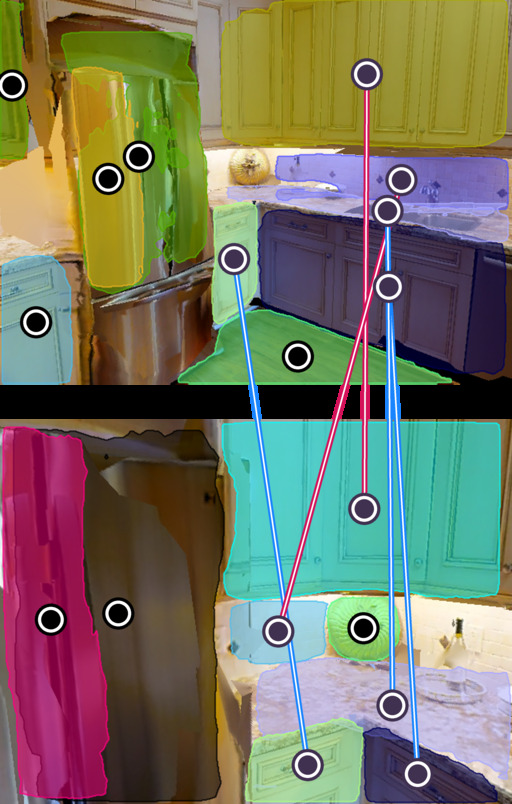}}
    & \frame{\includegraphics[width=0.08\textwidth]{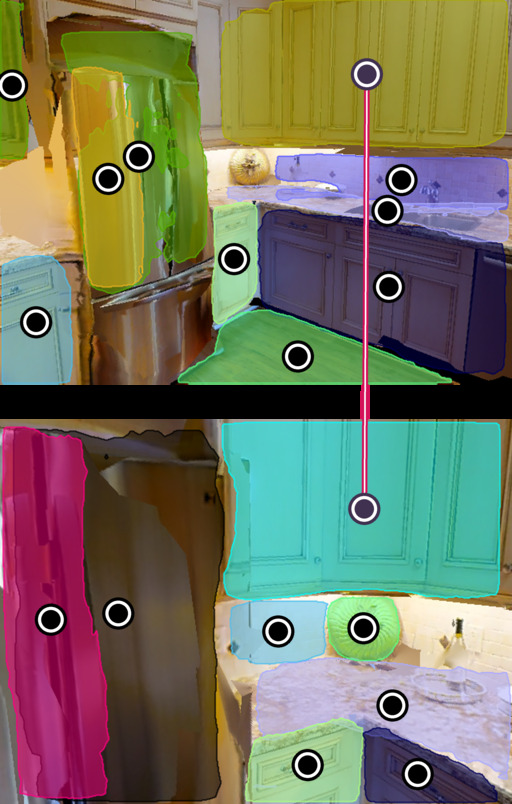}}
    & \frame{\includegraphics[width=0.08\textwidth]{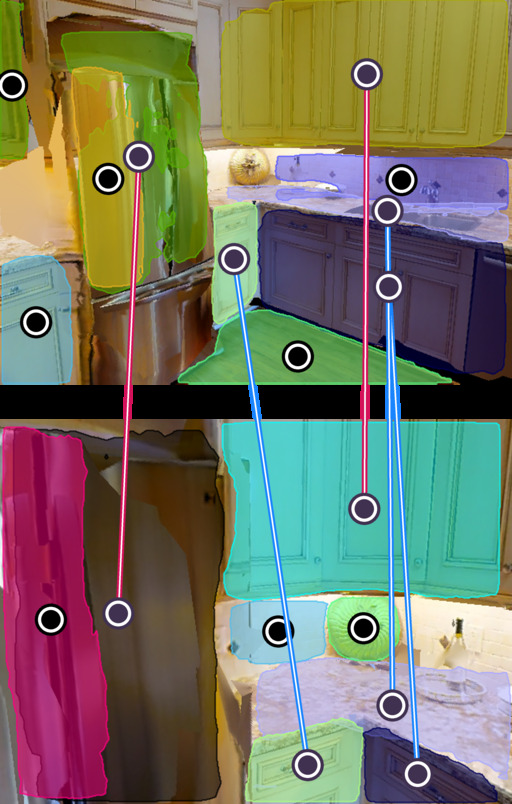}}\\

    \bottomrule
    \end{tabular}
	\caption{Selected correspondence predictions on {\em predicted} boxes, extending Figure~\ref{fig:corr_wall} in our paper, showing true positive matches in 
\textbf{\textcolor{AccessibleBlue}{Blue}} and false positives in 
\textbf{\textcolor{AccessibleRed}{Red}}. 
Correct matches are determined by assigning each ground truth box with a predicted box whose Mask IoU is greater than 0.5.
}
    \label{fig:supp-corr-wall}
\end{figure}

\begin{figure}[!t]
    \centering
    \scriptsize
    \begin{tabular}{c@{\hskip4pt}c@{\hskip4pt}c@{\hskip4pt}c@{\hskip4pt}c}
    \toprule
    
    Inputs & Appearance Only & ASNet~\cite{cai2020messytable} & Associative3D~\cite{Qian2020} & \textbf{Proposed} \\
    \midrule 
    \frame{\includegraphics[width=0.08\textwidth]{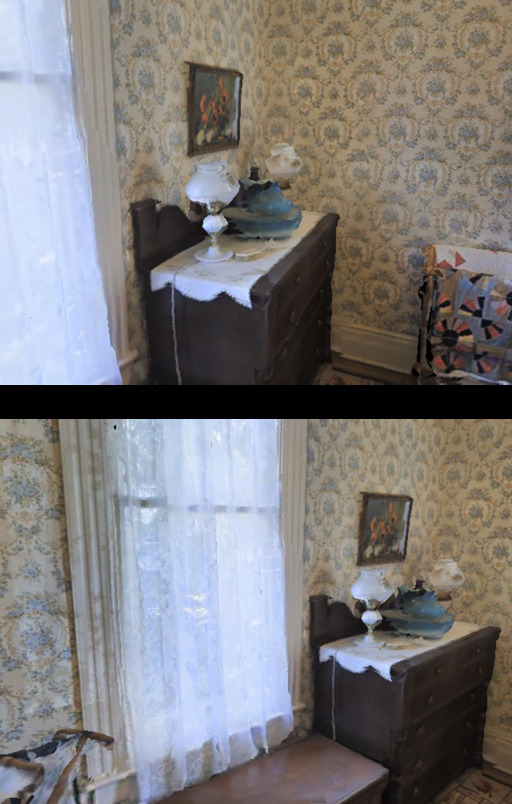}}
    & \frame{\includegraphics[width=0.08\textwidth]{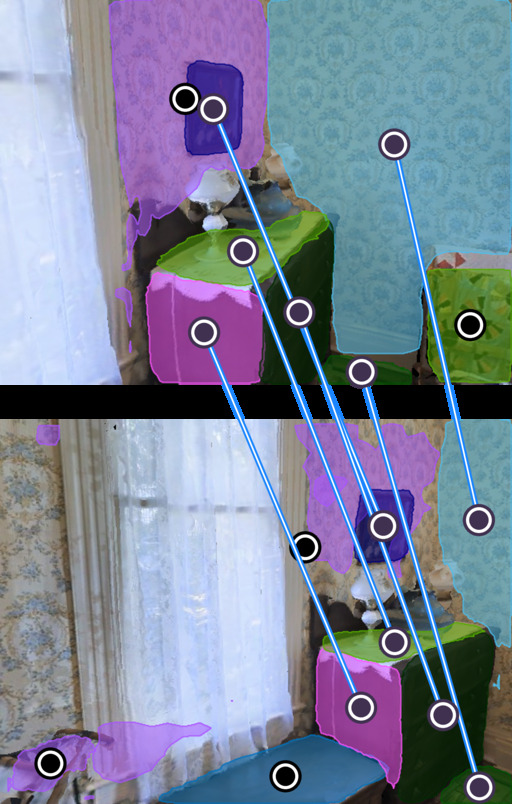}}
    & \frame{\includegraphics[width=0.08\textwidth]{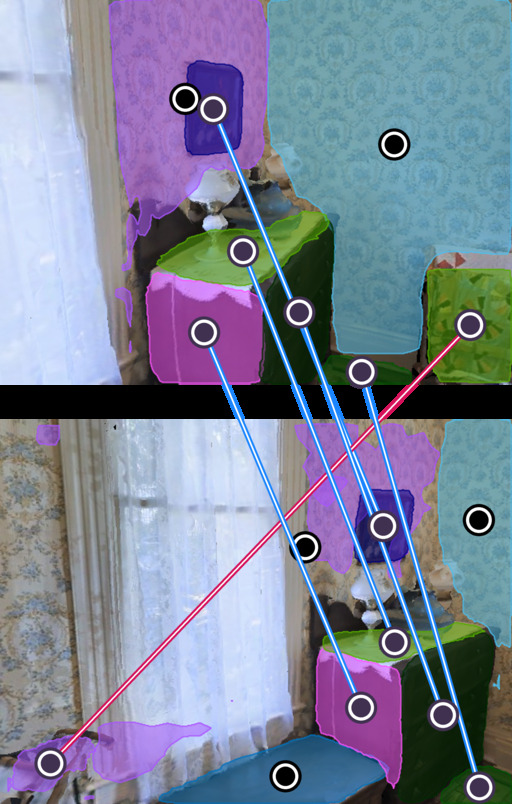}}
    & \frame{\includegraphics[width=0.08\textwidth]{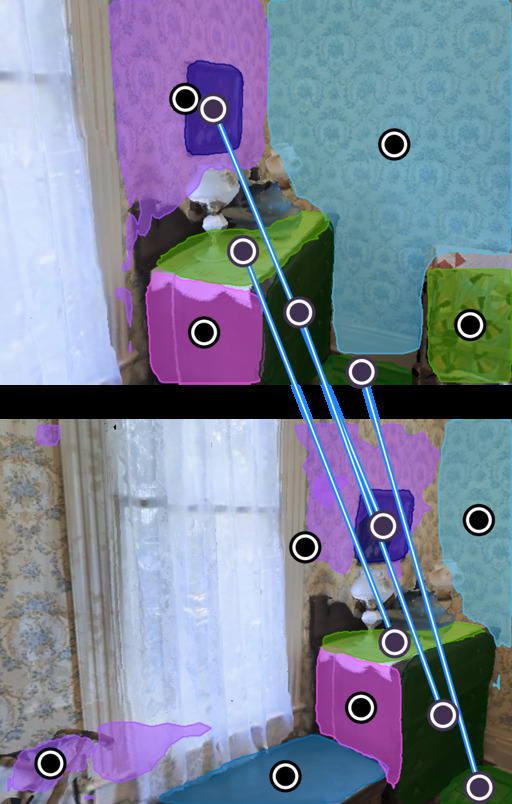}}
    & \frame{\includegraphics[width=0.08\textwidth]{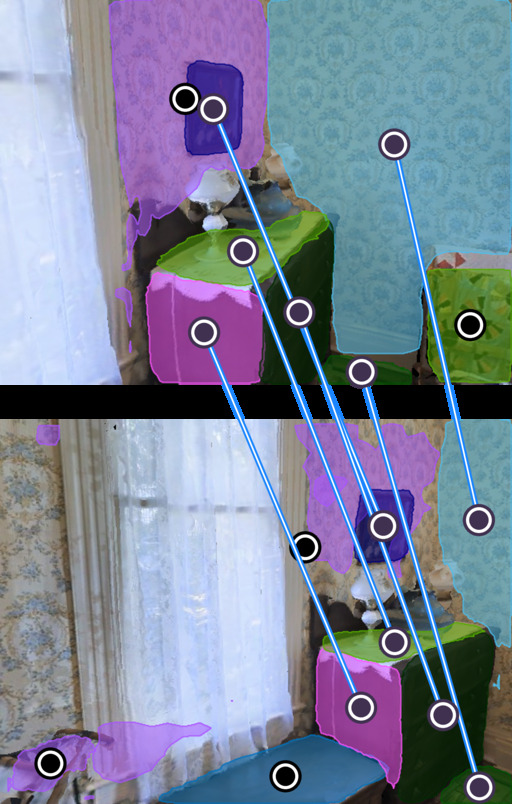}}\\

    \frame{\includegraphics[width=0.08\textwidth]{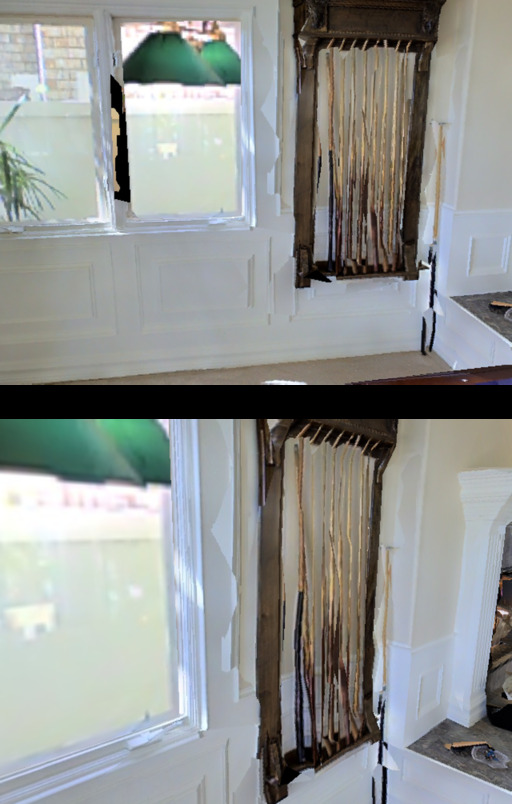}}
    & \frame{\includegraphics[width=0.08\textwidth]{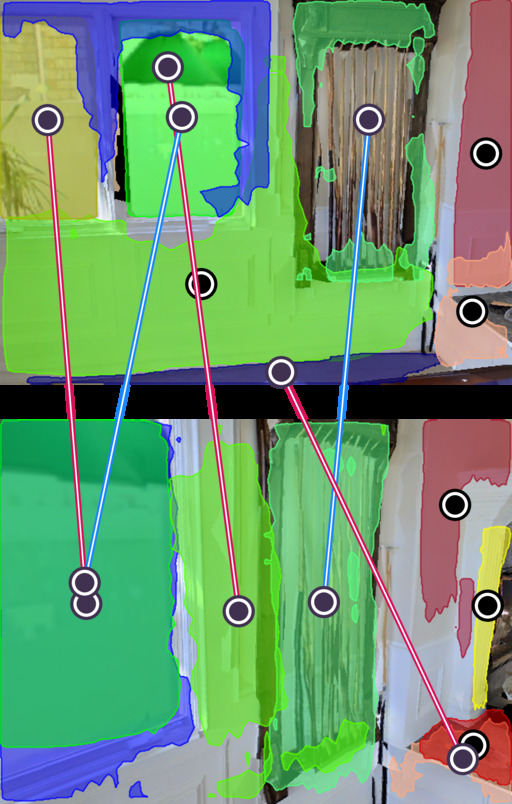}}
    & \frame{\includegraphics[width=0.08\textwidth]{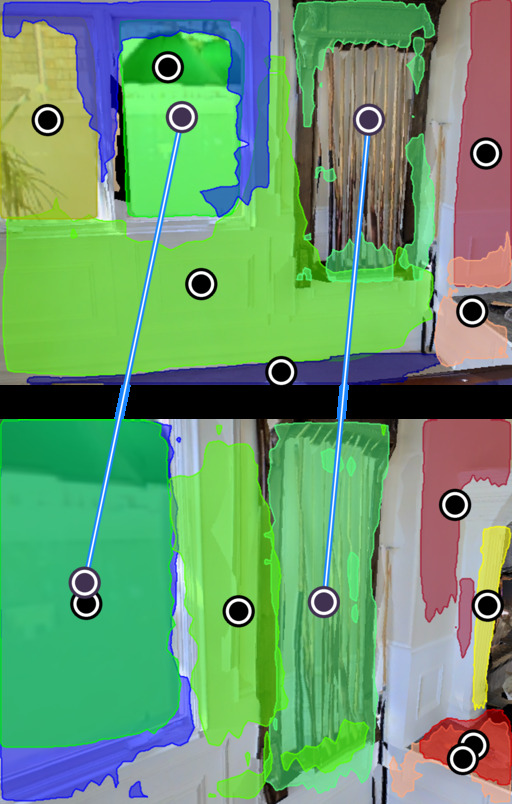}}
    & \frame{\includegraphics[width=0.08\textwidth]{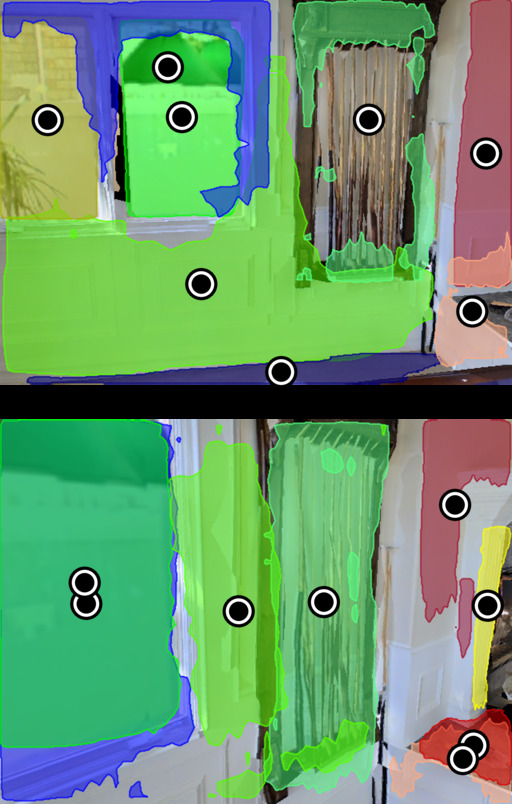}}
    & \frame{\includegraphics[width=0.08\textwidth]{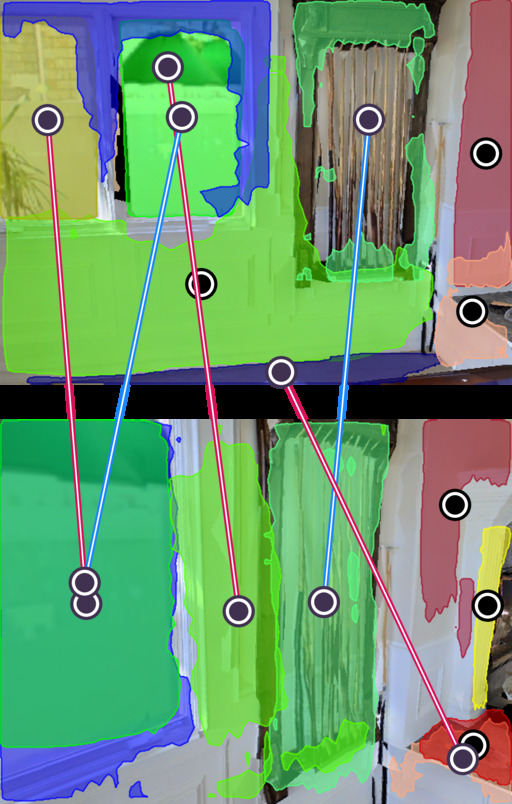}}\\ 

    \frame{\includegraphics[width=0.08\textwidth]{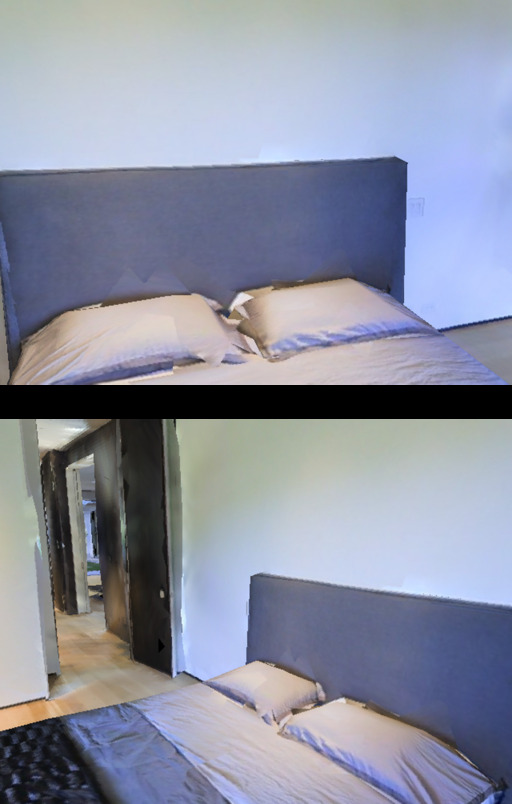}}
    & \frame{\includegraphics[width=0.08\textwidth]{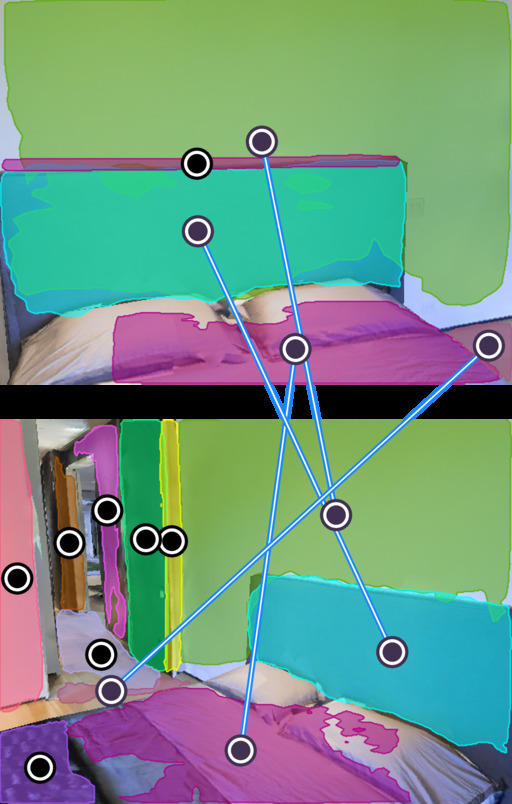}}
    & \frame{\includegraphics[width=0.08\textwidth]{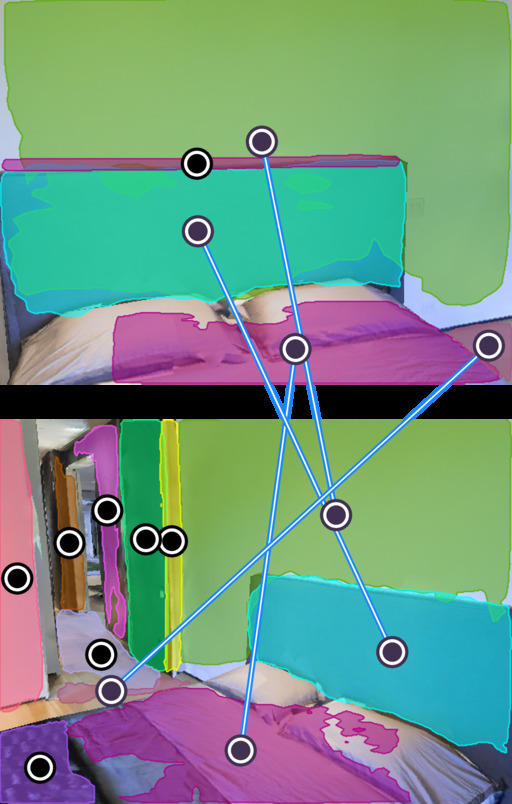}}
    & \frame{\includegraphics[width=0.08\textwidth]{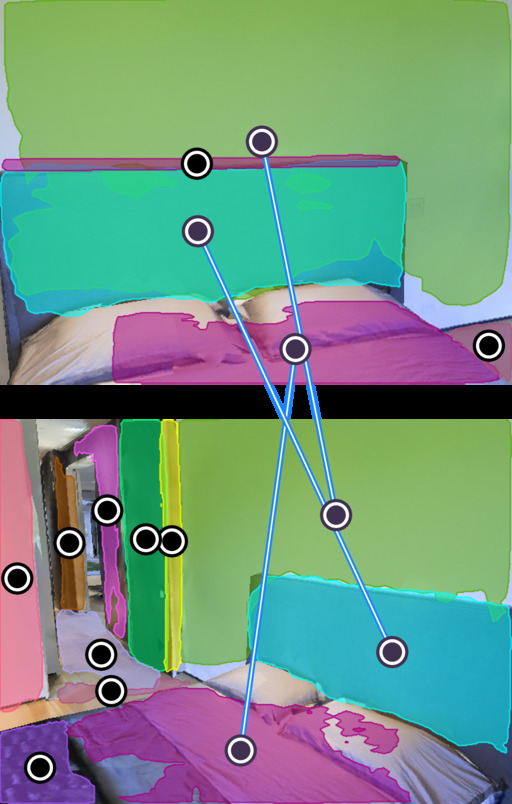}}
    & \frame{\includegraphics[width=0.08\textwidth]{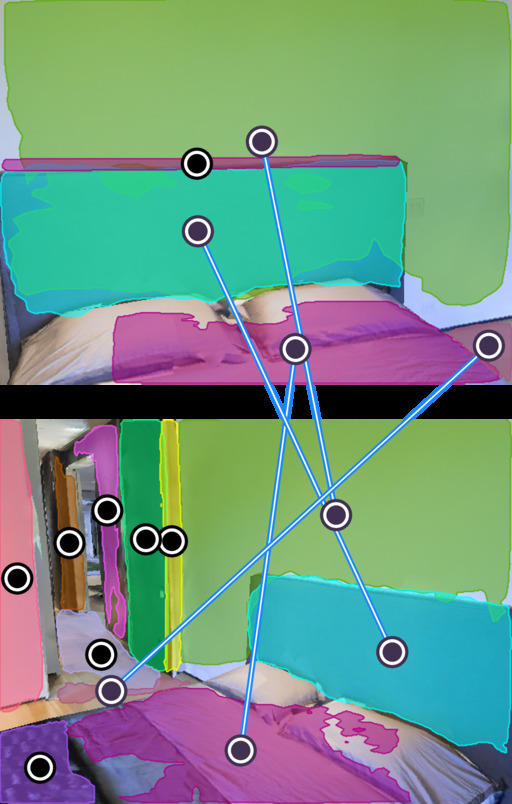}}\\

    \frame{\includegraphics[width=0.08\textwidth]{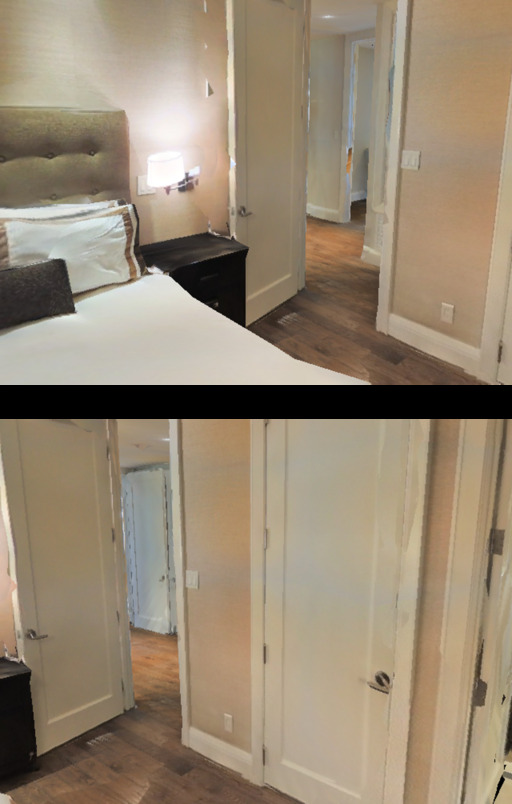}}
    & \frame{\includegraphics[width=0.08\textwidth]{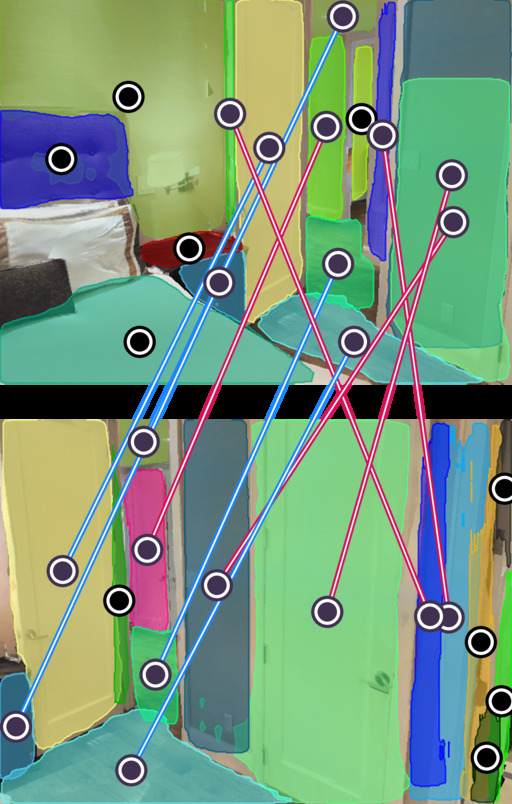}}
    & \frame{\includegraphics[width=0.08\textwidth]{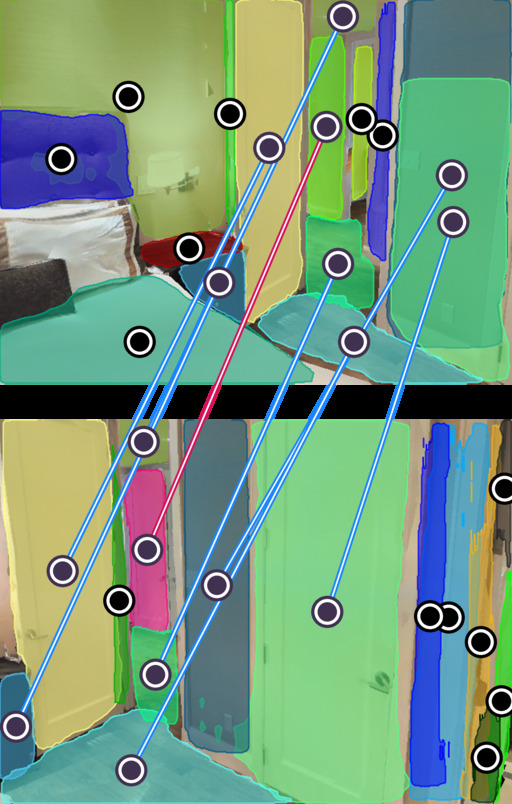}}
    & \frame{\includegraphics[width=0.08\textwidth]{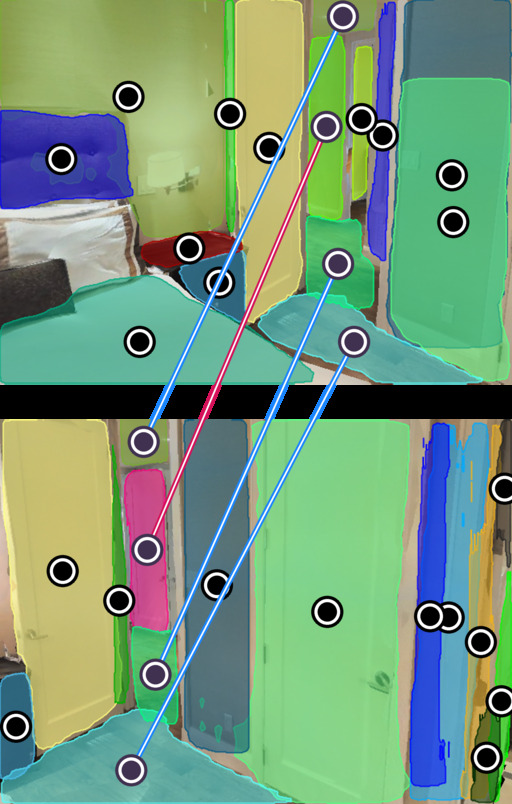}}
    & \frame{\includegraphics[width=0.08\textwidth]{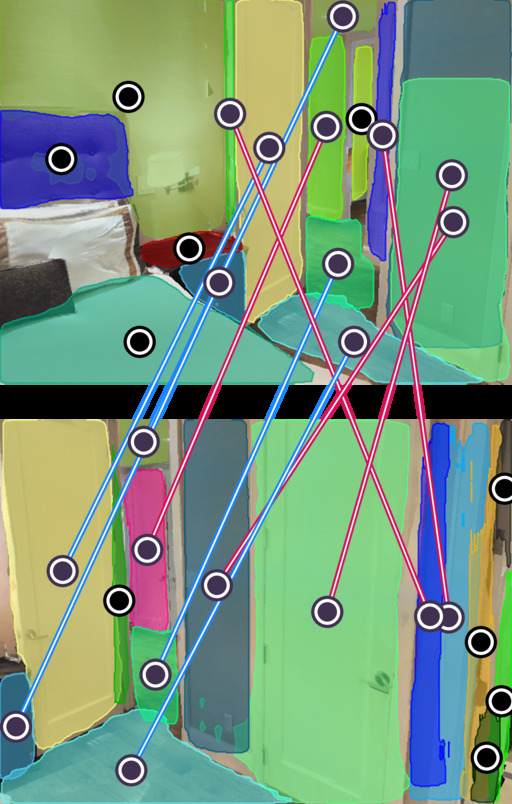}}\\

    \frame{\includegraphics[width=0.08\textwidth]{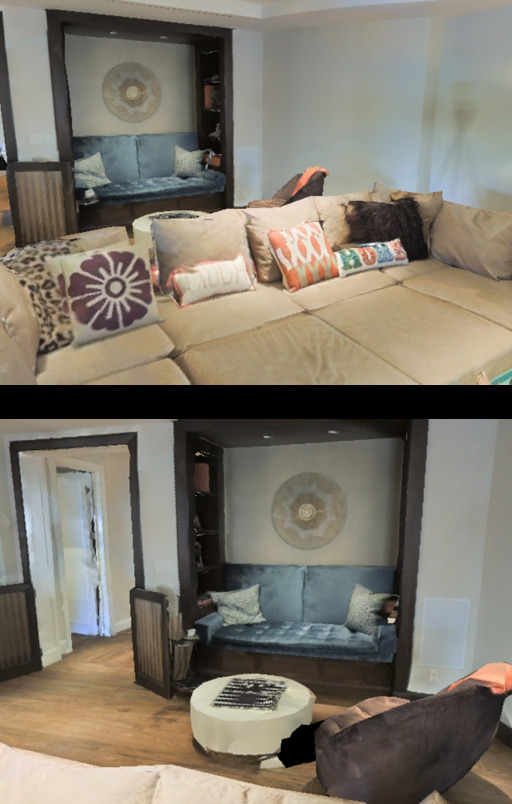}}
    & \frame{\includegraphics[width=0.08\textwidth]{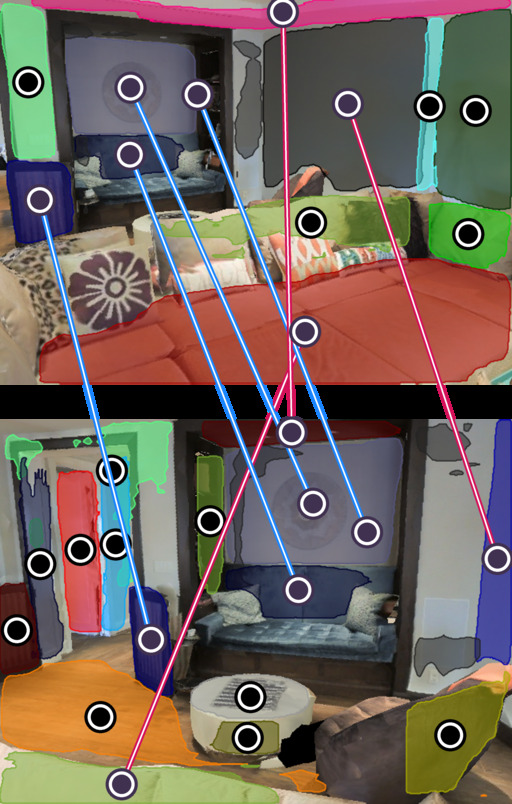}}
    & \frame{\includegraphics[width=0.08\textwidth]{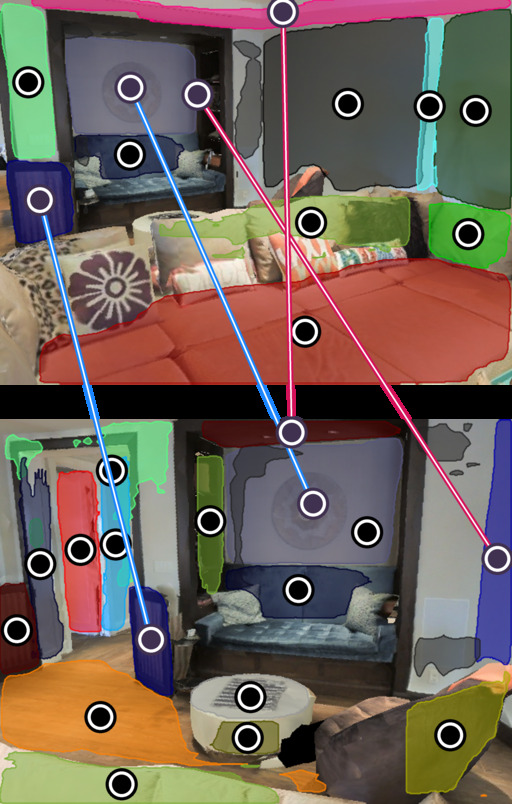}}
    & \frame{\includegraphics[width=0.08\textwidth]{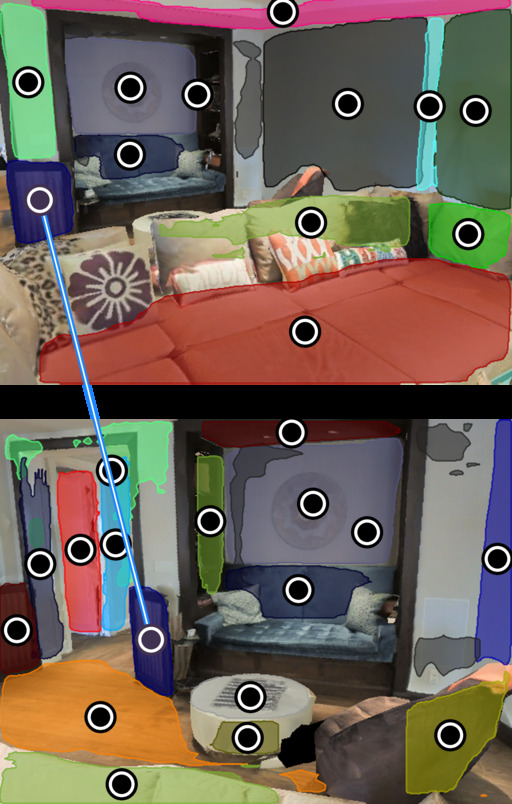}}
    & \frame{\includegraphics[width=0.08\textwidth]{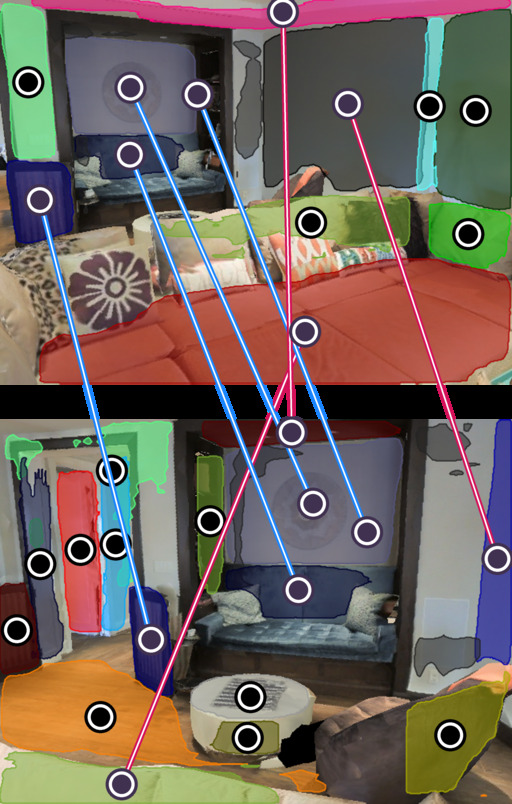}}\\

    \frame{\includegraphics[width=0.08\textwidth]{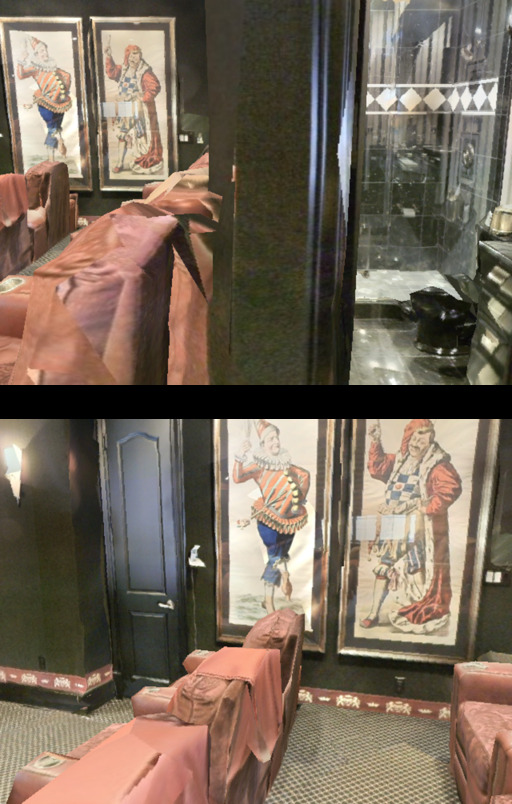}}
    & \frame{\includegraphics[width=0.08\textwidth]{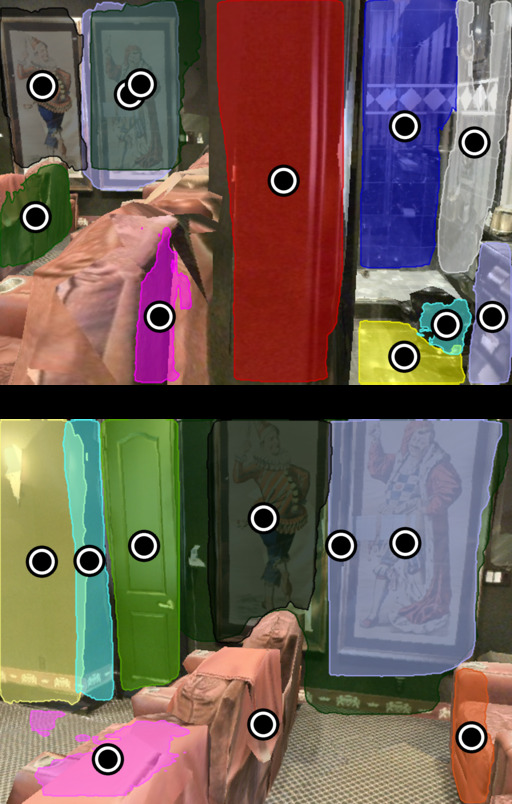}}
    & \frame{\includegraphics[width=0.08\textwidth]{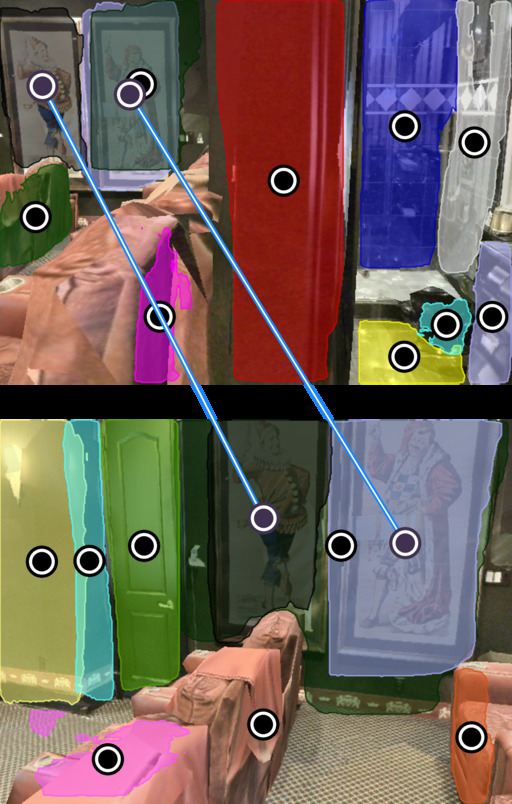}}
    & \frame{\includegraphics[width=0.08\textwidth]{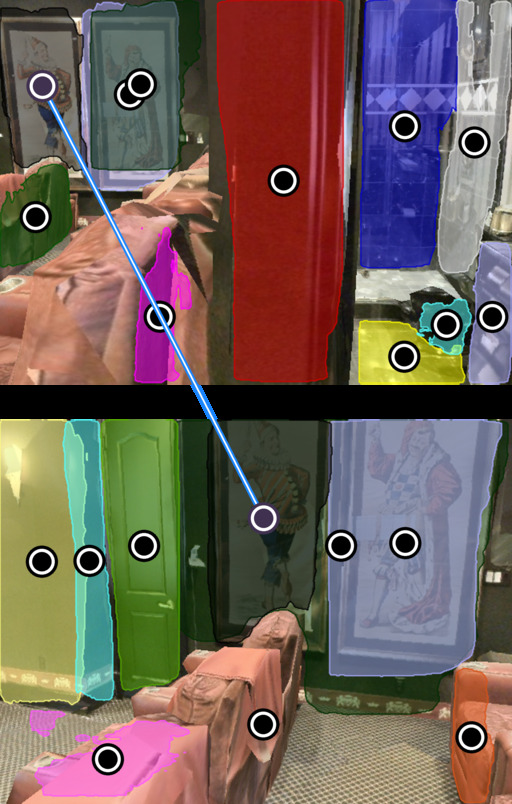}}
    & \frame{\includegraphics[width=0.08\textwidth]{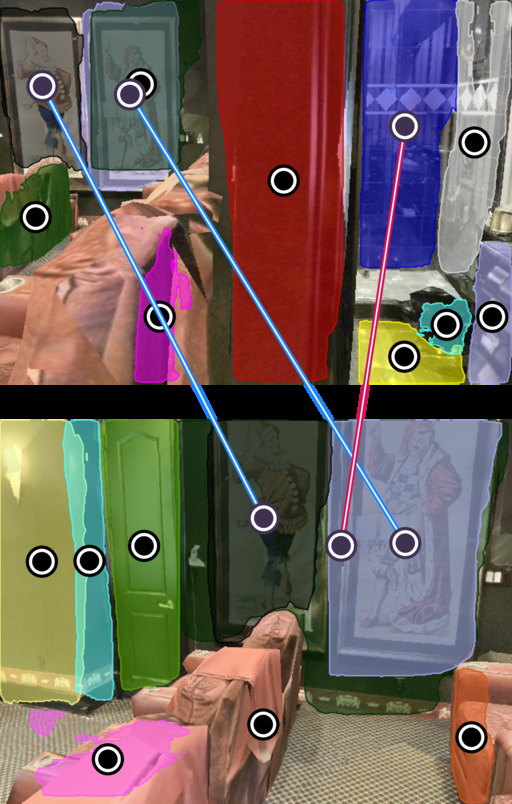}}\\

    \frame{\includegraphics[width=0.08\textwidth]{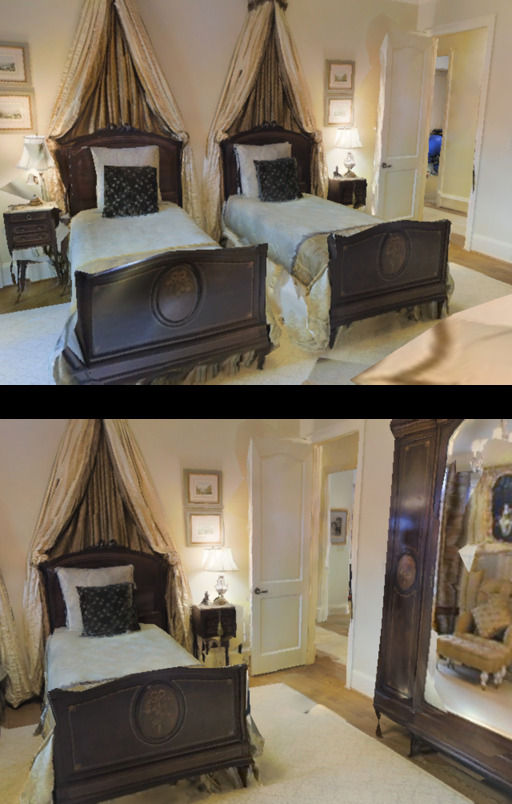}}
    & \frame{\includegraphics[width=0.08\textwidth]{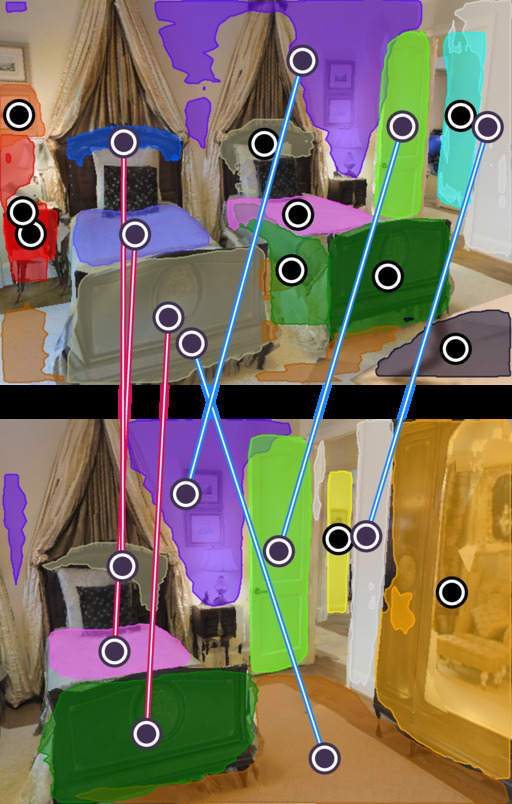}}
    & \frame{\includegraphics[width=0.08\textwidth]{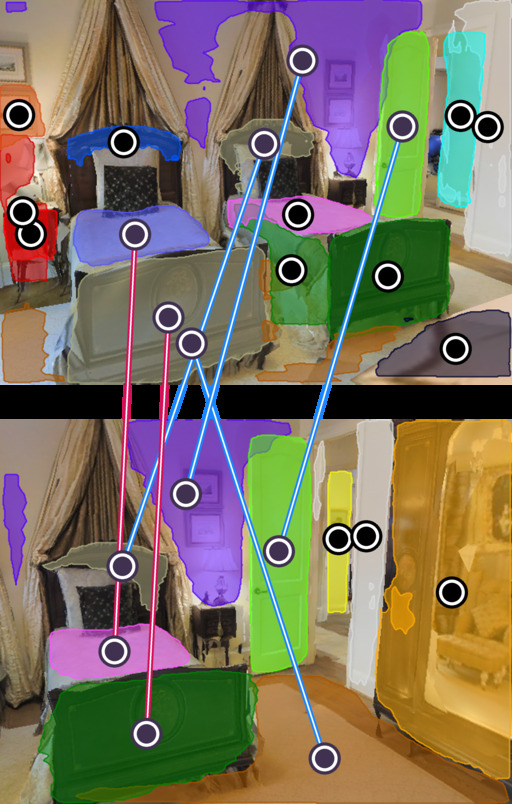}}
    & \frame{\includegraphics[width=0.08\textwidth]{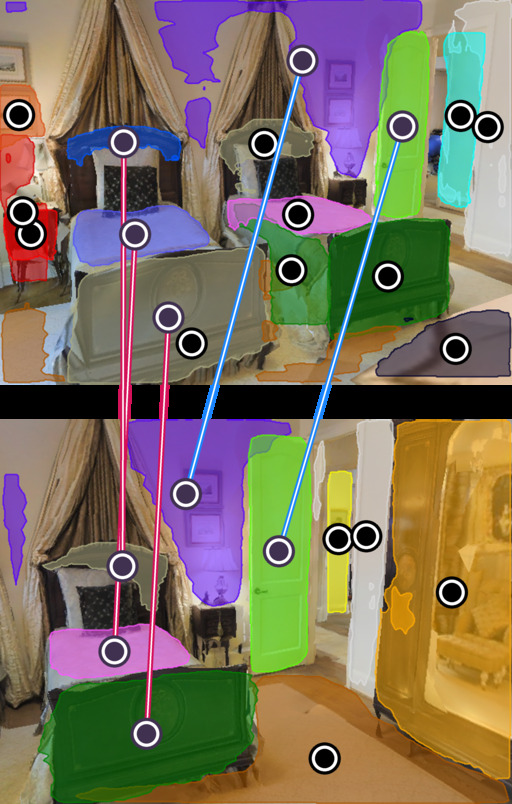}}
    & \frame{\includegraphics[width=0.08\textwidth]{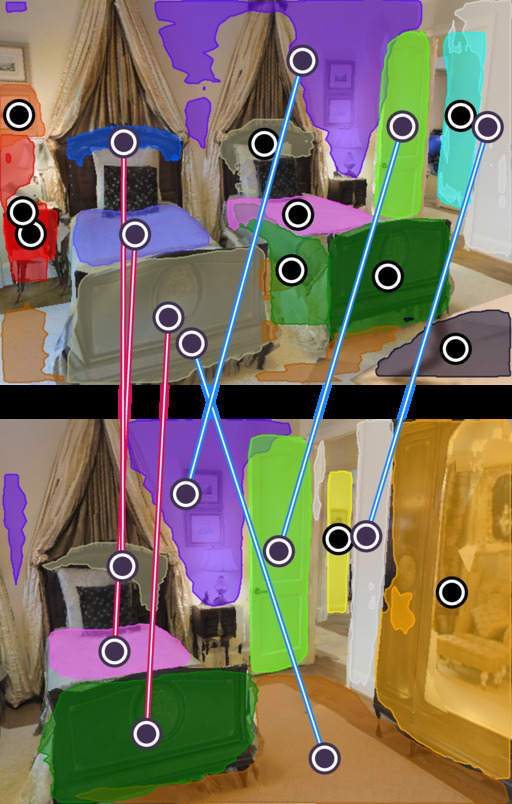}}\\
    
    \frame{\includegraphics[width=0.08\textwidth]{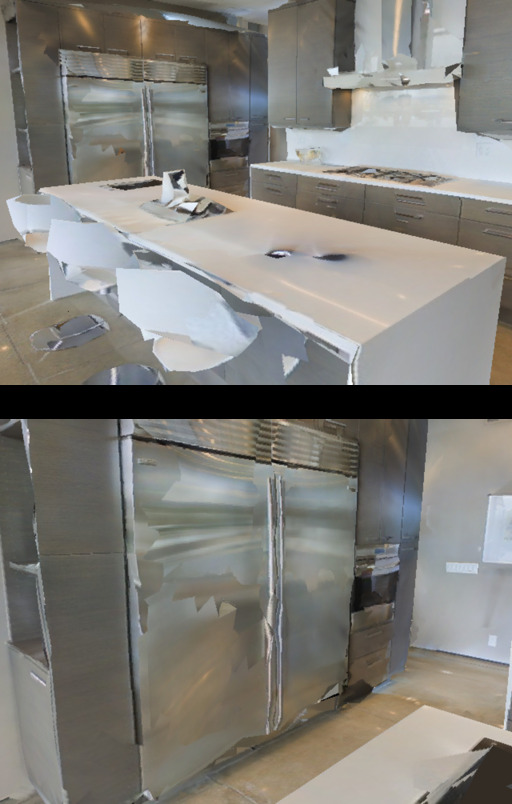}}
    & \frame{\includegraphics[width=0.08\textwidth]{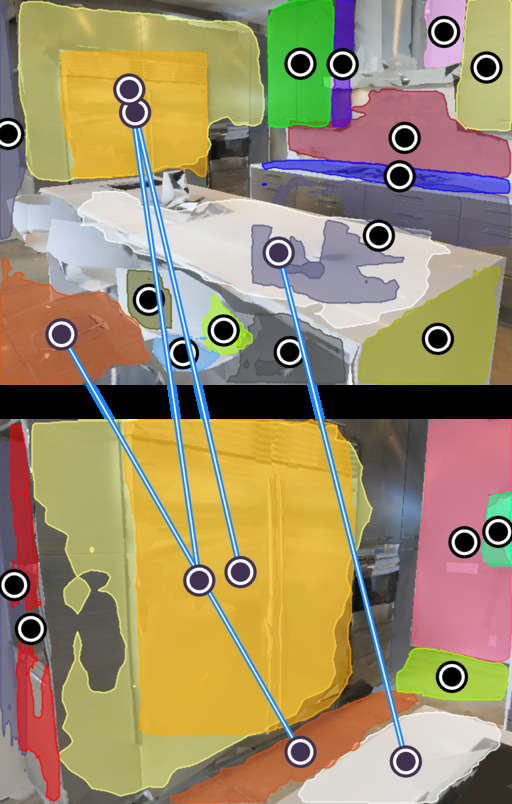}}
    & \frame{\includegraphics[width=0.08\textwidth]{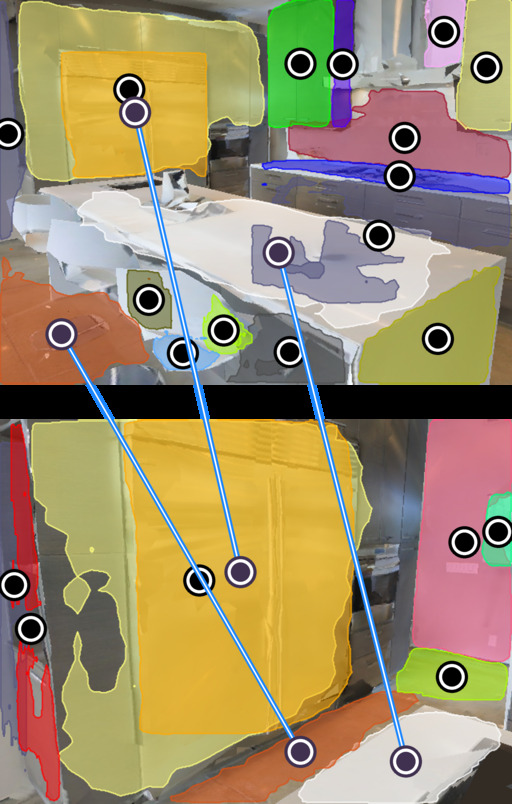}}
    & \frame{\includegraphics[width=0.08\textwidth]{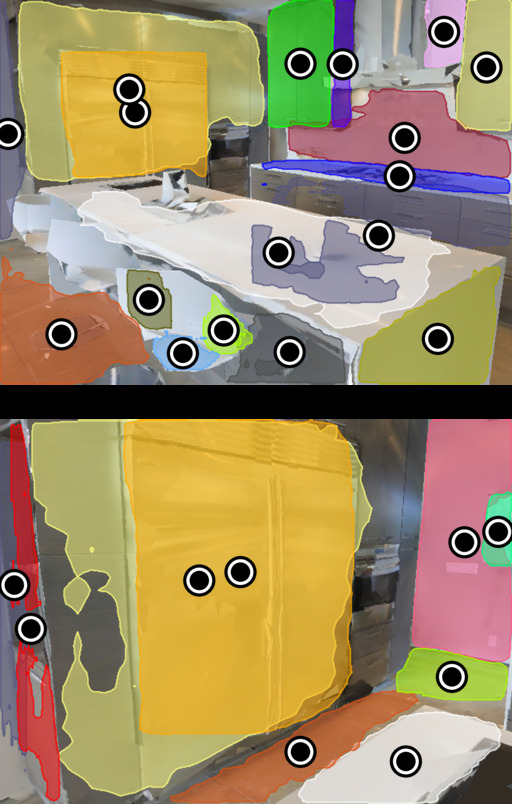}}
    & \frame{\includegraphics[width=0.08\textwidth]{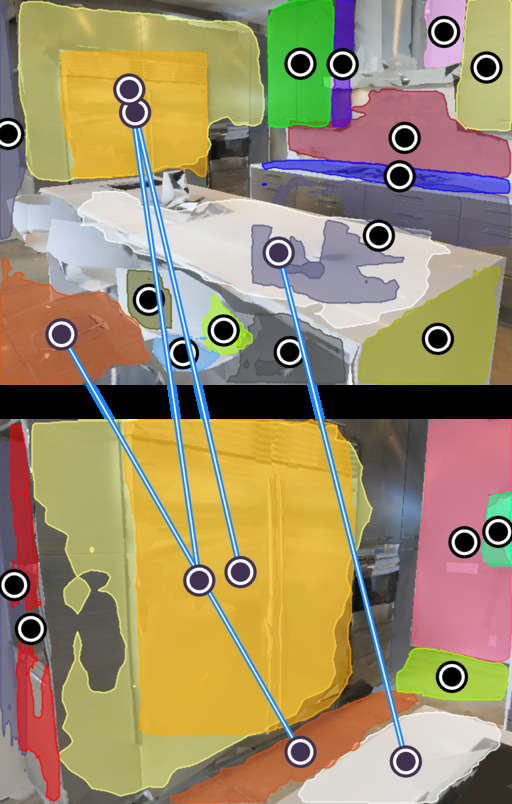}}\\

    \frame{\includegraphics[width=0.08\textwidth]{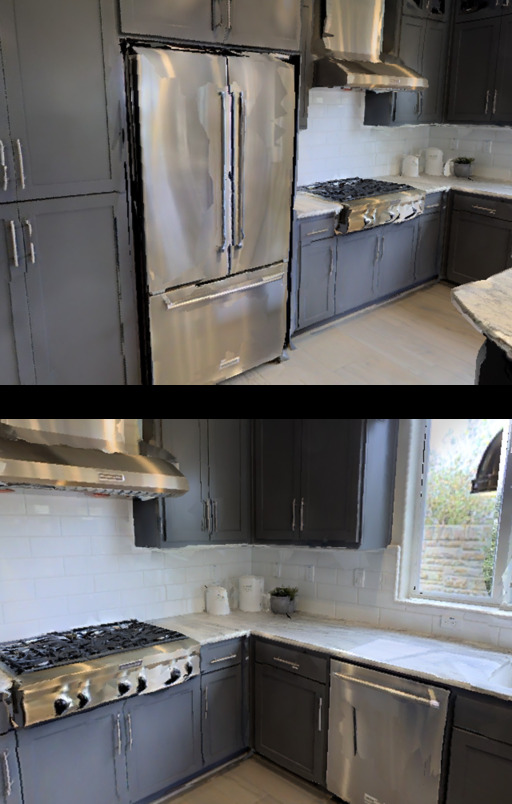}}
    & \frame{\includegraphics[width=0.08\textwidth]{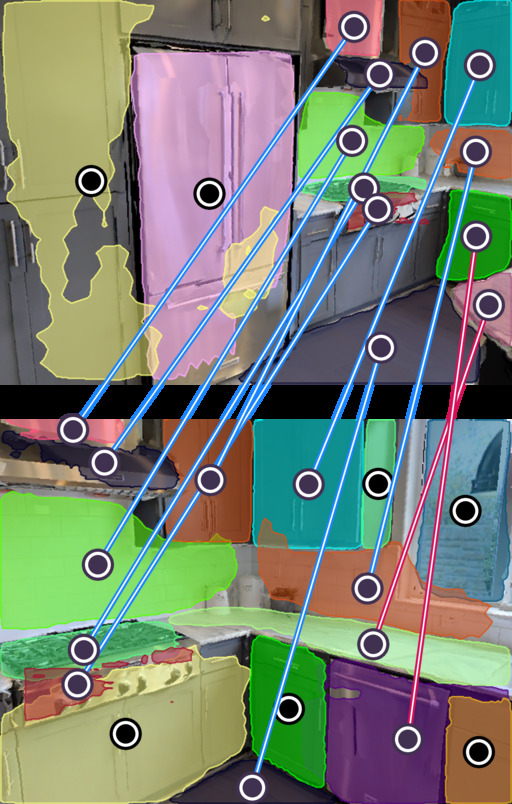}}
    & \frame{\includegraphics[width=0.08\textwidth]{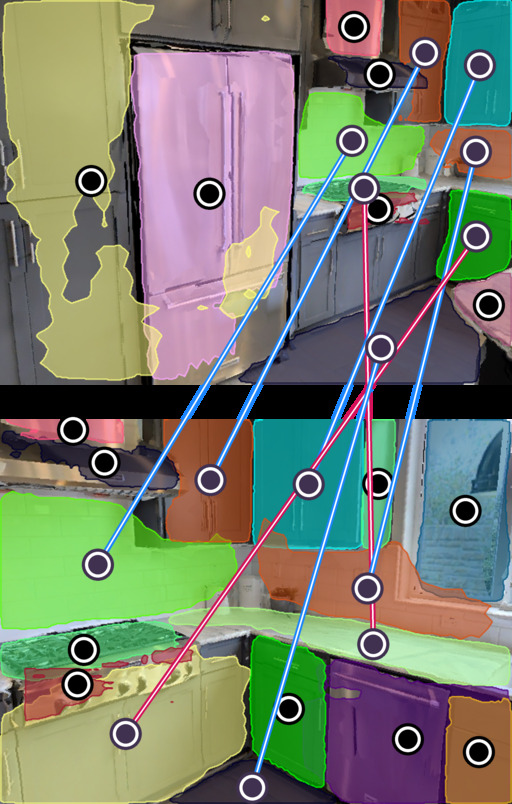}}
    & \frame{\includegraphics[width=0.08\textwidth]{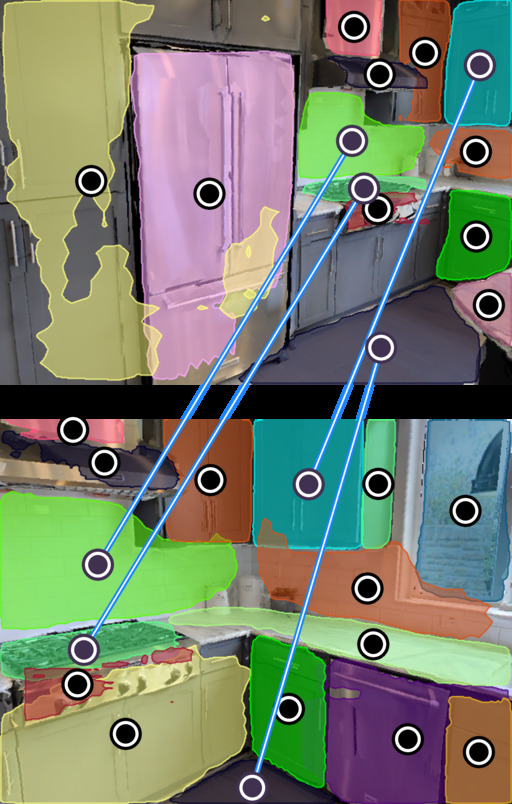}}
    & \frame{\includegraphics[width=0.08\textwidth]{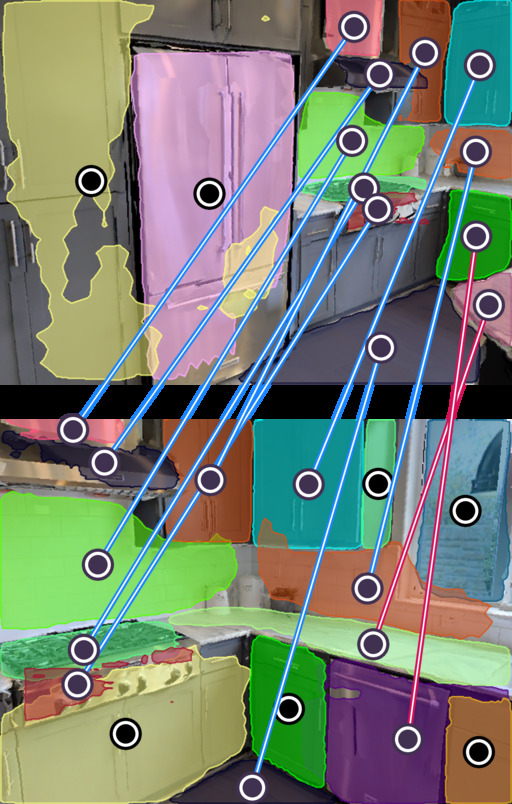}}\\

    \bottomrule
    \end{tabular}
	\caption{Random correspondence predictions on {\em ground truth} boxes, showing true positive matches in 
\textbf{\textcolor{AccessibleBlue}{Blue}} and false positives in 
\textbf{\textcolor{AccessibleRed}{Red}}.}
    \label{fig:supp-corr-wall-random}
\end{figure}

We also show more selected qualitative examples on plane correspondence prediction in Figure~\ref{fig:supp-corr-wall}, extending Figure~\ref{fig:corr_wall} in our paper.
Those examples use {\em predicted} bounding boxes. To determine the ground truth correspondence, we assign each ground truth box with a predicted box
whose mask IoU is greater than 0.5. We also randomly choose examples in Figure~\ref{fig:supp-corr-wall-random} to reflect the overall performance of our 
approach and baselines. Those examples use {\em ground truth} boxes which are the same as the ones used in correspondence evaluation. As shown in the
random results, there are many false positives and false negatives in challenging cases; therefore, there is still much space to improve in 
predicting plane correspondence.

\subsection{Generalization to other data.}
\begin{figure}[!t]
    \centering
    \scriptsize
    \begin{tabular}{c@{\hskip4pt}c@{\hskip4pt}c@{\hskip4pt}c}
    \toprule
    
    View 1 & View 2 & Pred Correspondences & Pred 3D \\
    \midrule
    \frame{\includegraphics[width=0.09\textwidth]{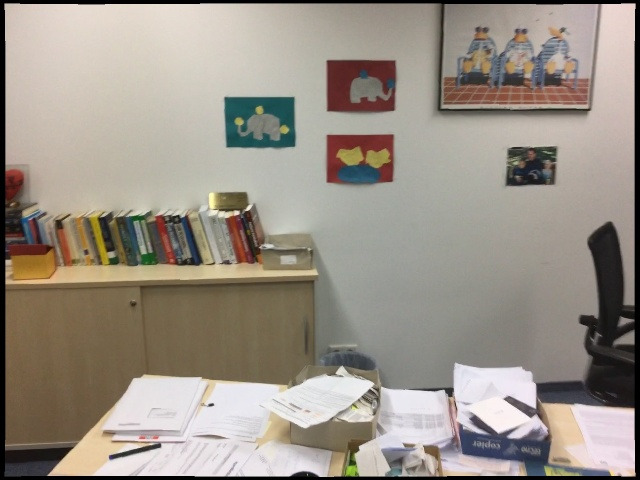}}
    & \frame{\includegraphics[width=0.09\textwidth]{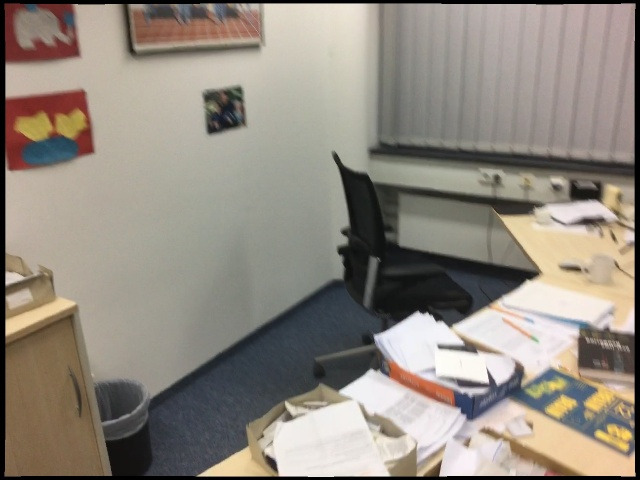}}
    & \frame{\includegraphics[width=0.18\textwidth]{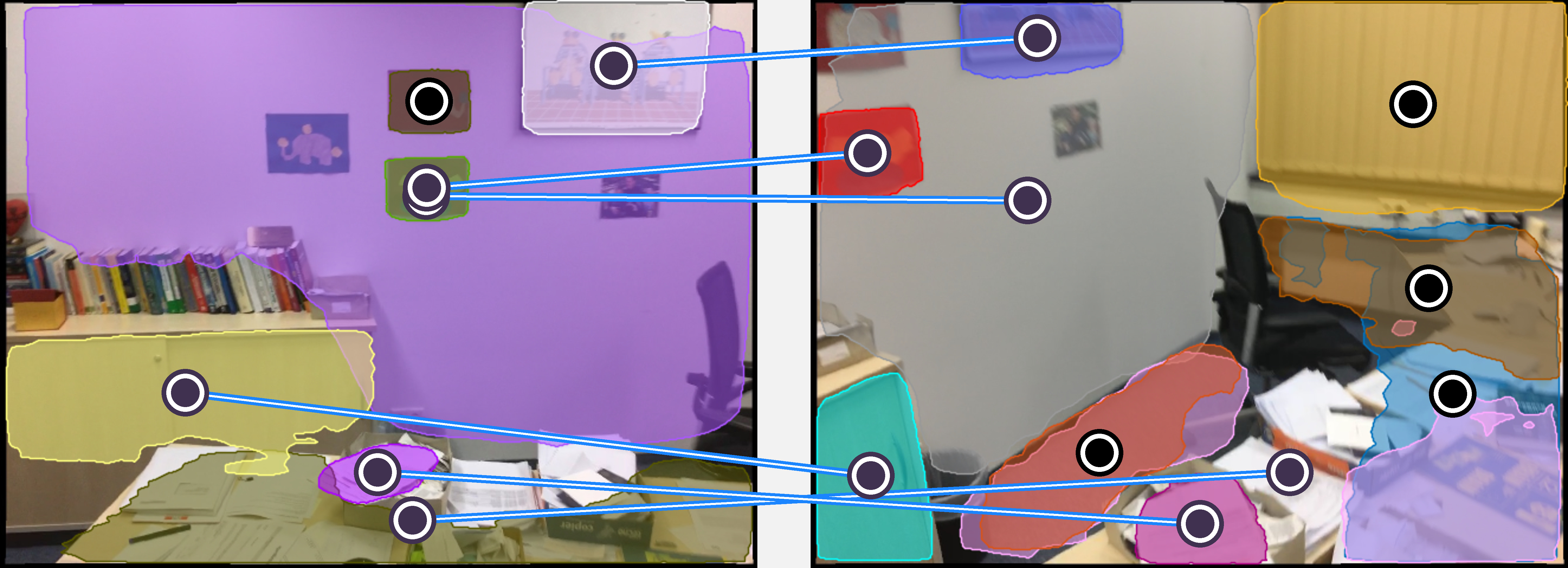}}
    & \frame{\includegraphics[width=0.09\textwidth]{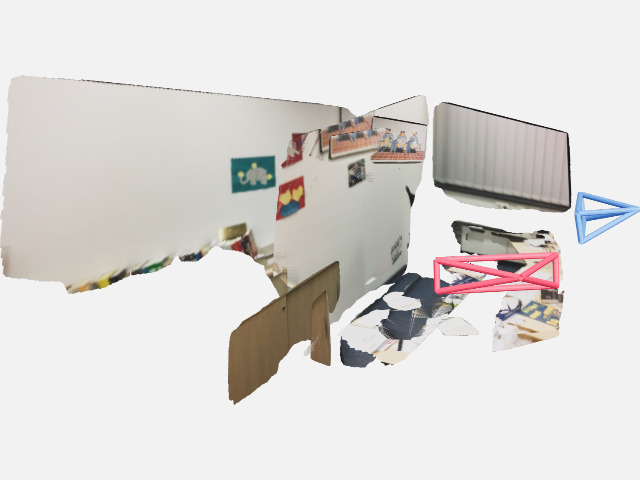}}\\
    
    \frame{\includegraphics[width=0.09\textwidth]{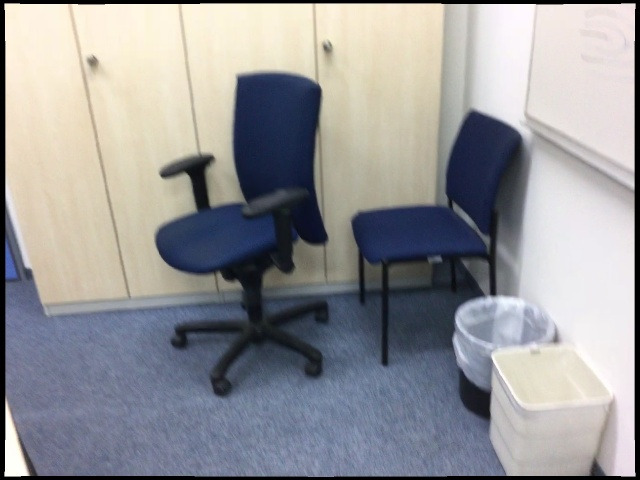}}
    & \frame{\includegraphics[width=0.09\textwidth]{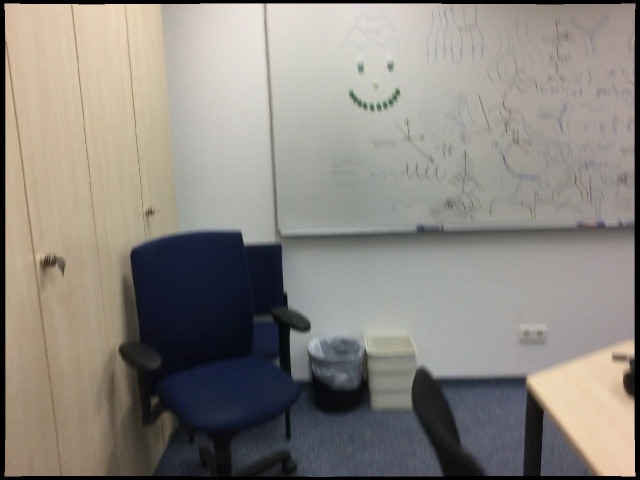}}
    & \frame{\includegraphics[width=0.18\textwidth]{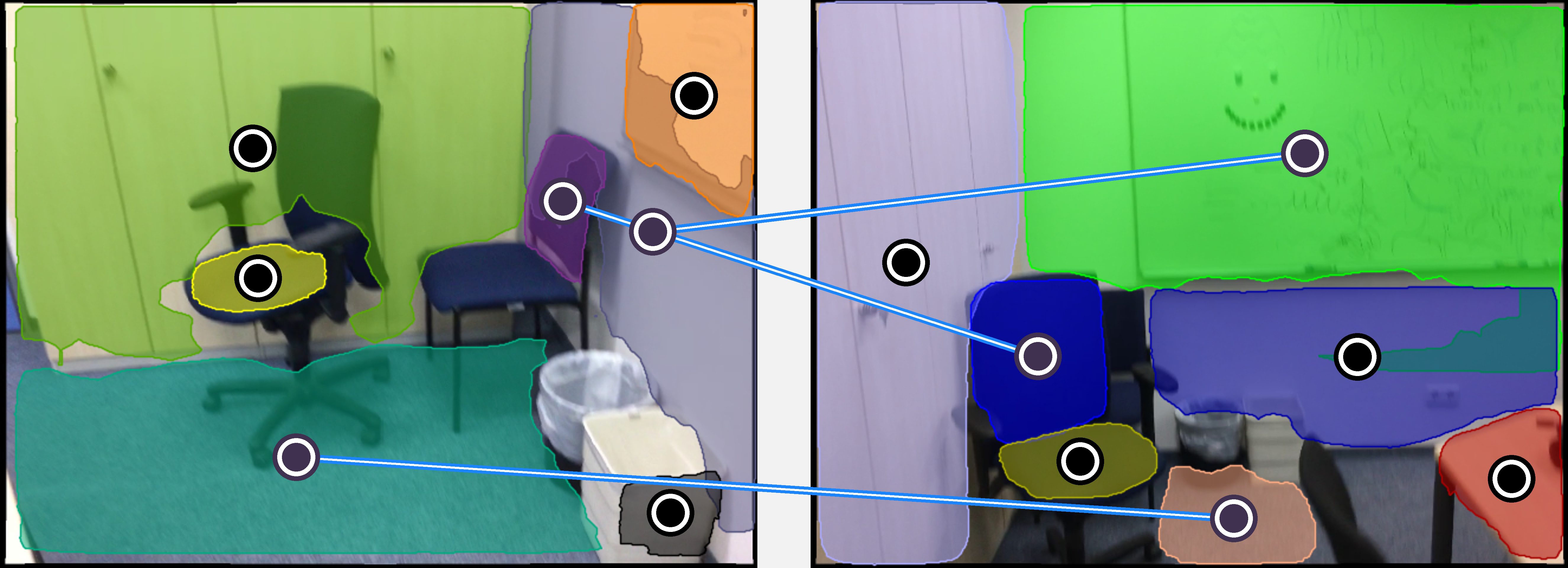}}
    & \frame{\includegraphics[width=0.09\textwidth]{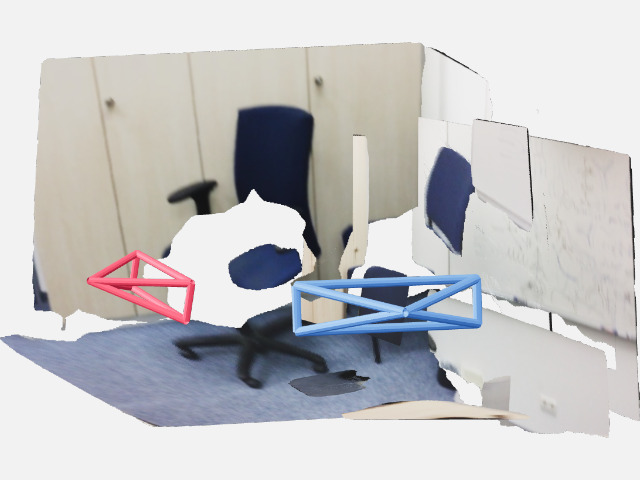}}\\
    
    \frame{\includegraphics[width=0.09\textwidth]{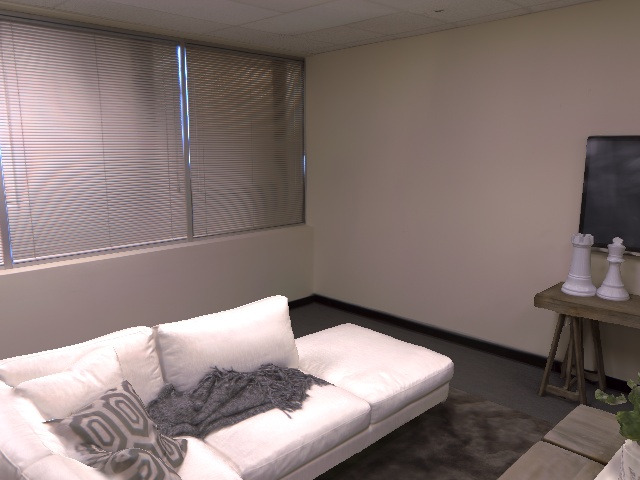}}
    & \frame{\includegraphics[width=0.09\textwidth]{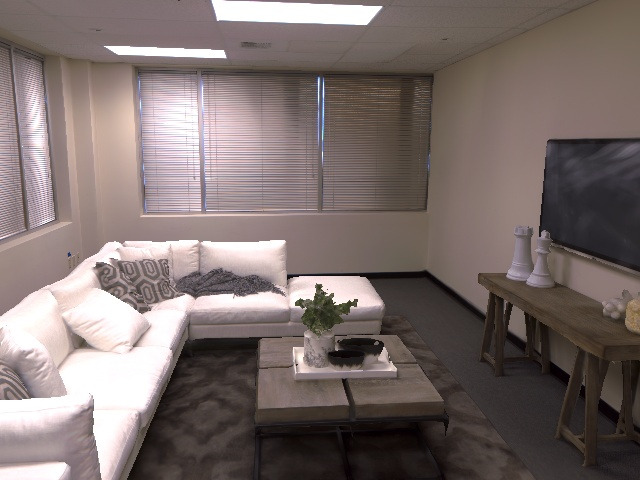}}
    & \frame{\includegraphics[width=0.18\textwidth]{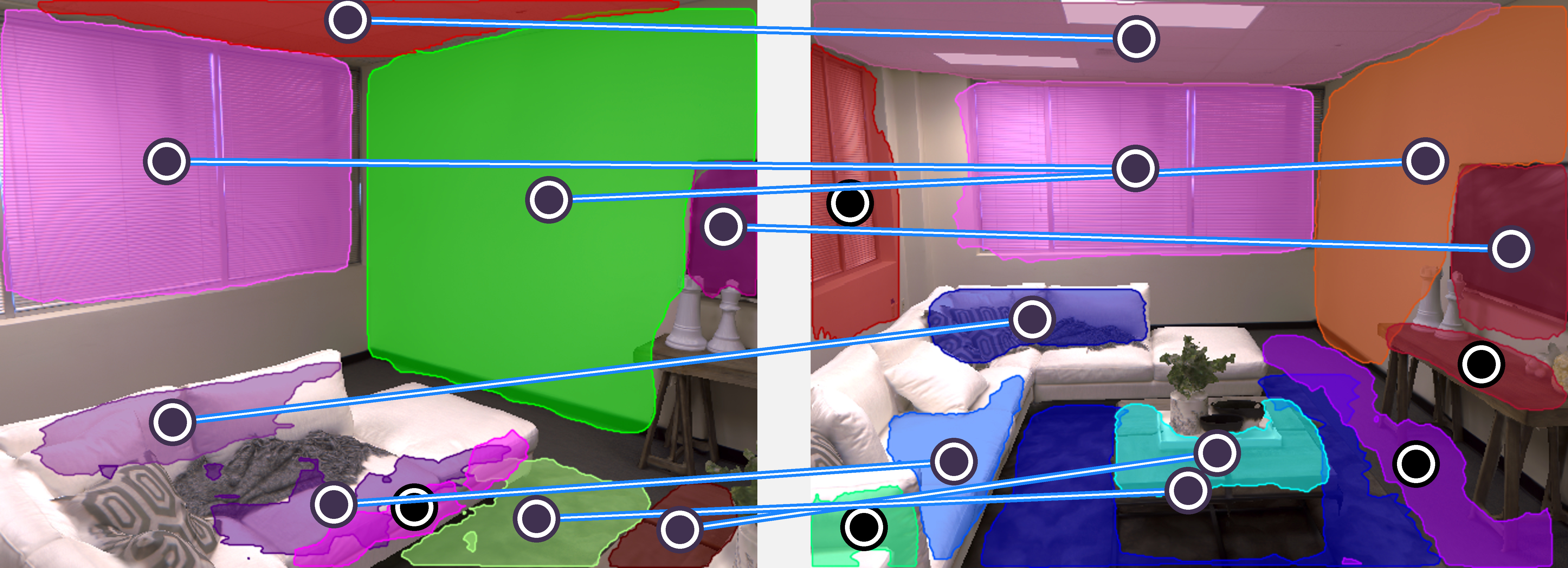}}
    & \frame{\includegraphics[width=0.09\textwidth]{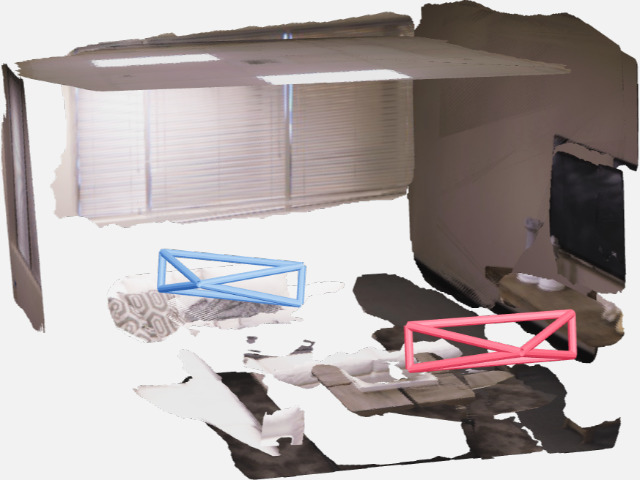}}\\
 
    \frame{\includegraphics[width=0.09\textwidth]{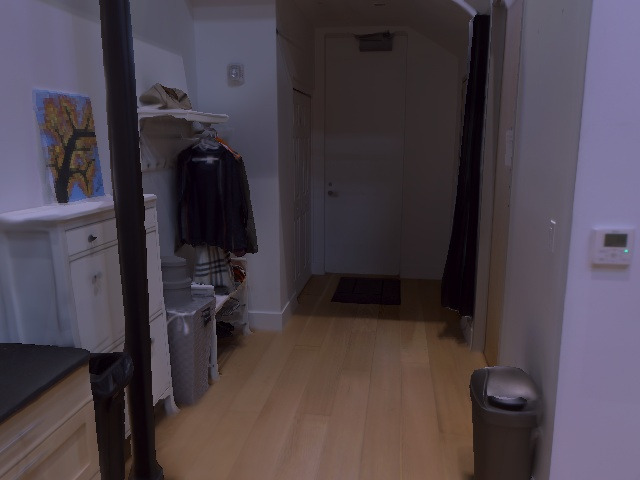}}
    & \frame{\includegraphics[width=0.09\textwidth]{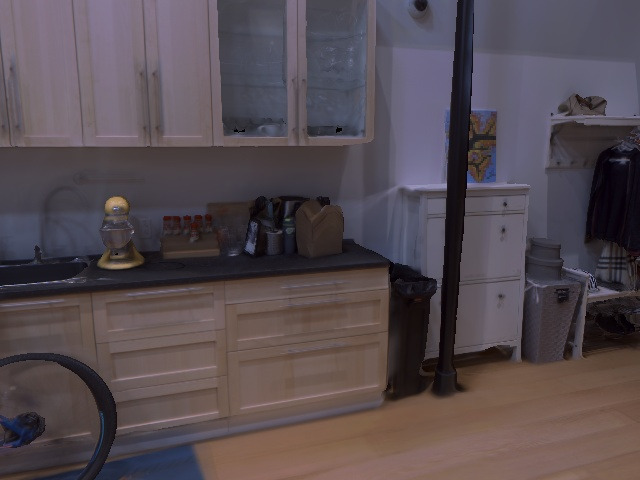}}
    & \frame{\includegraphics[width=0.18\textwidth]{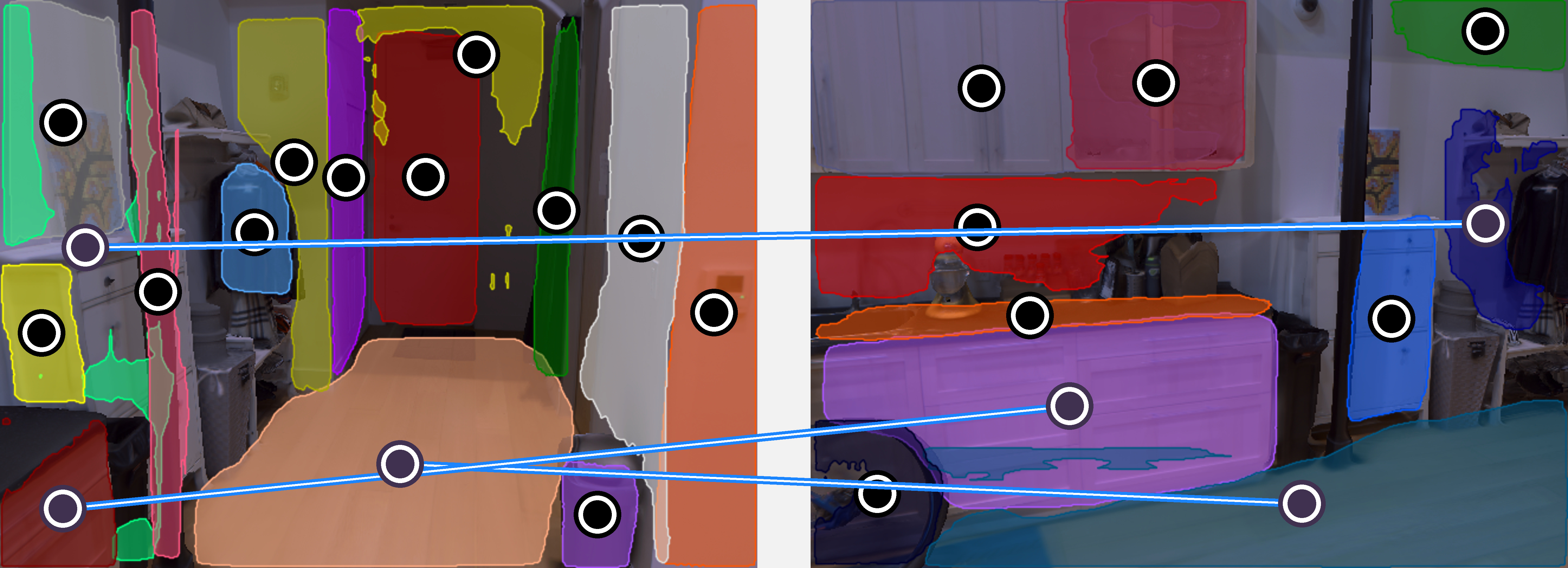}}
    & \frame{\includegraphics[width=0.09\textwidth]{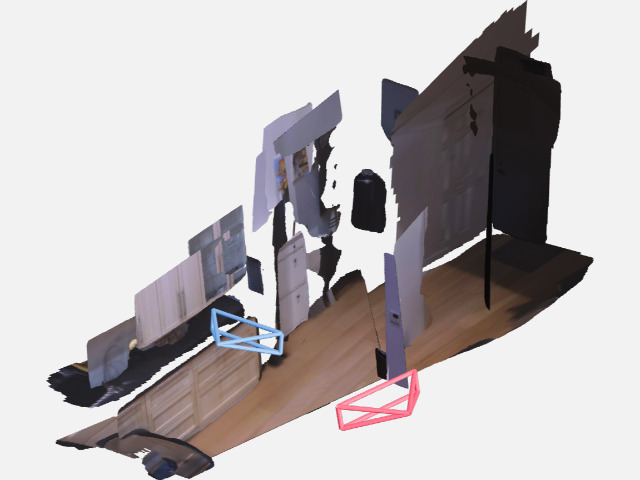}}\\

    \bottomrule
    \end{tabular}
    \caption{Qualitative results of generalization on ScanNet (row 1-2) and Replica Dataset (row 3-4).}
    \label{fig:generalization}
 \end{figure}

To test generalization, we also test our approach on images from ScanNet~\cite{dai2017scannet} and Replica~\cite{straub2019replica}. 
Figure~\ref{fig:generalization} shows results from our model trained on Matterport3D. 
Our model can obtain a reasonable interpretation.

\clearpage

\clearpage

\section{Implementation}
\subsection{Dataset Preprocessing}
\begin{figure}[t]
    \centering
    \includegraphics[width=\columnwidth]{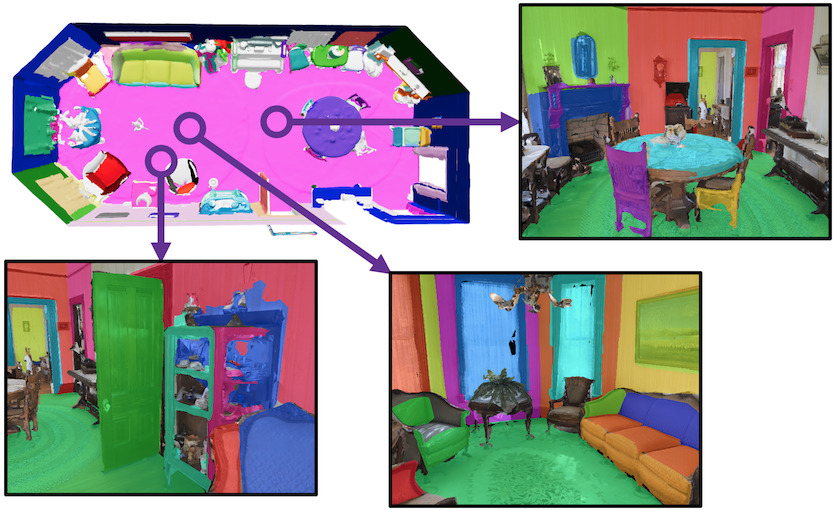}
    \caption{Example plane annotations of our dataset.}
    \label{fig:fitplane}
    \vspace{-0.5em}
    \end{figure}

We augment Matterport3D~\cite{chang2017matterport3d} with plane segmentation annotations on the original mesh.
This enables consistent plane instance ID and plane parameters through rendering, 
and automatically establishes plane correspondences across different images. 
We fit planes using RANSAC, following~\cite{liu2018planenet}, on mesh vertices within the same object instance (using the annotation provided by 
the original Matterport3D dataset). 
We then render the per-pixel plane instance ID, along with the RGB-D images using AI Habitat~\cite{savva2019habitat}.
Since the original instance segmentation mesh has ``ghost" objects and holes due to artifacts,
we further filter out bad plane annotations by comparing the depth information and the plane parameters. 
Figure~\ref{fig:fitplane} shows examples of plane annotations on the mesh and rendered plane segmentations.
Small plane masks with area less than 1\% of the total image pixels are removed. 

We follow~\cite{zhang2017physically} to generate random camera poses. To include more data, we keep all valid cameras in each horizontal 
sector.
The camera is of a random height 1.5-1.6m above the floor and a downward tilt angle of
11 degrees to simulate human’s view. The same camera intrinsic are used to render all the images.
To generate image pairs, we randomly sample cameras within 
each room, and enforce that there are at least three common planes and at least three unique planes in each image. 
Our ground truth data includes a depthmap and plane segmentation for each image, as well as plane correspondences and camera transformations for the image pair.

\subsection{Network Architectures}
We use Detectron2~\cite{wu2019detectron2} to implement our network. The overall architecture of our network is shown in Table~\ref{tab:full_arch}. The backbone, RPN, box branch and mask branch are identical to 
 Mask R-CNN~\cite{he2017mask}. The depth branch is the same as the depthmap decoder in~\cite{liu2019planercnn}. Table~\ref{tab:camera_cc} shows the 
 exact architecture of the camera pose module in our model. 
 \par \noindent {\bf Training details.} The plane detection backbone uses ResNet50-FPN pretrained on COCO~\cite{lin2014microsoft}. 
 We first train the plane prediction module without the plane embedding. 
 Then we freeze the network and train the embedding head using the predicted bounding boxes. 
 Finally, we freeze the whole network and train the camera pose module.
 We put in as large of a batch as possible to fit in the available GPUs. 
 The number of iterations are picked based on evaluation results on the validation set.
 We use batch size 8, 16, 32 on 4, 4, and 2 GPUs for 37k, 36k and 35k iterations to train the 
 plane prediction module, embedding head and camera pose module respectively. We used 2080Ti GPUs.
 We use SGD with momentum of 0.9, a weight decay of 0.0001, multistep learning rate with base $\mathrm{lr}=0.001$, 
 multiplied by 0.1 after 30k iterations, and warmup across all the training jobs.

 \begin{algorithm*}
    \caption{Discrete optimization. It takes as input the plane embeddings and parameters as well as the camera poses and probabilities, 
    it outputs the best binary correspondence $\CB$ and the best camera pose $\{\tB_{\hat{k}}, \RB_{\hat{k}}\}$. \texttt{cam2world} 
    converts plane parameters from the camera frame to the world frame as described in Eqn.~\ref{eqn:cam2world}.
    $\SB_k$ is the same term in Eqn.~\ref{eqn:costmatrix} of the main paper, terms in view 2 are denoted as $\eB', [\nB', o']$.}
    \label{alg:discrete}
    \begin{algorithmic}[1]
    \Require{$\eB, \eB', \nB, \nB', o, o', \{p_{\tB_k}, p_{\RB_k},\tB_k, \RB_k\}_{k=0,\cdots,K}$}
    \Ensure{$\CB, \{\tB_{\hat{k}}, \RB_{\hat{k}}\}$}
    \State $\mathtt{min\_cost} \gets \mathtt{Inf}$
    \State $\hat{k} \gets -1$
    \State $\Delta\eB_{ij} = ||\eB_i-\eB_j'||$
    \For{$k\in\{1,\cdots,K\}$}
        \State $\nB^w, o^w \gets \texttt{cam2world}(\nB, o, \tB_k, \RB_k)$
        \State ${\nB'}^{w}, {o'}^{w} \gets \texttt{cam2world}(\nB', o', \mathbf{0}, \mathbf{I}(3))$
        \State $\Delta\nB_{ij}\gets\mathrm{acos}(|\nB_i^{w\intercal}{\nB_j'}^w|)/\pi$
        \State $\Delta\oB_{ij}\gets\mathrm{min}(|o_i^w-{o_j'}^w|/O_{\mathrm{clamp}}, 1)$
        \State $\SB_k\gets\lambda_e\Delta\eB+\lambda_n\Delta\nB+\lambda_o\Delta\oB$
        \State $\CB_k\gets\texttt{Hungarian}(\SB_k)$
        \State $\CB_k(\SB_k<\mathtt{threshold})=0$
        \State $\mathtt{cost}\gets \lambda_h\sum_{i,j} (\CB_k\circ\SB_k)_{i,j} - \lambda_t\mathrm{log}(p_{\tB_k})-\lambda_R\mathrm{log}(p_{\RB_k}) - \lambda_N\sum_{i,j}\CB_{k,i,j}$
        \If {$\mathtt{cost}<\mathtt{min\_cost}$}
            \State $\mathtt{min\_cost}\gets\mathtt{cost}$
            \State $\hat{k}=k$
        \EndIf
    \EndFor
    \State \textbf{return} $\CB_{\hat{k}}$, $\{\tB_{\hat{k}}, \RB_{\hat{k}}\}$
    \end{algorithmic}
    \end{algorithm*}

 \subsection{Optimization}
 \par \noindent {\bf Associative3D~\cite{Qian2020} optimization.} 
 We re-build the heuristic optimization 
 on the same network branches our approach uses to ensure apples-to-apples comparisons; we also optimize their search to accommodate for differences in plane-vs-object matching
 by using Eqn.~\ref{eqn:discrete objective} as the objective.
 \vspace{0.75em}
 \par \noindent {\bf Discrete optimization.} Our discrete optimization searches through all $K$ camera options, ($K=32\times 32=1024$), and outputs 
 the best camera $\{\tB_{\hat{k}}, \RB_{\hat{k}}\}$ as well as a binary correspondence matrix $\CB$ for planes
 across images using the Hungarian algorithm, shown in Algorithm~\ref{alg:discrete}.
 \texttt{cam2world} converts plane parameters $[\nB, o]$ from the camera frame to the world frame, 
 \begin{equation}
\begin{split}
    \PB &= \left(1+\frac{\tB_k\cdot \RB_k (o\cdot\nB)}{||\RB_k (o\cdot\nB)||^2}\right)\RB_k (o\cdot\nB),\\
    \nB^w &= \PB / ||\PB||,\\
    o^w &= ||\PB||.\\
\end{split}
\label{eqn:cam2world}
 \end{equation}
 In practice, in Eqn.~\ref{eqn:discrete objective} of the paper, we normalize the offset error and normal error. 
 Since the offset error is unbounded, we clamp it by $O_{\mathrm{clamp}}$.
 We use $\mathtt{threshold}$ after Hungarian algorithm to reject matches with large distance error.
 The hyperparameters are determined using
 random search on the validation set: 
 $\mathtt{threshold}=0.7$, $\lambda_e=0.47$, $\lambda_n=0.25$, $\lambda_o=0.28$, $O_{\mathrm{clamp}}=4$, 
 $\lambda_N=0.311$, $\lambda_h=0.432$, $\lambda_t=0.166$, $\lambda_R=0.092$.

\vspace{0.75em}
\par \noindent {\bf Continuous optimization.} Continuous optimization further outputs refined camera $\{\hat{\tB}, \hat{\RB}\}$ 
and plane parameters $\hat{\piB}, \hat{\piB}'$ based on the discrete optimization results
by using non-linear optimization to minimize both the geometric distance between the corresponding planes and 
We use the trust-region reflective minimizer with default parameters implemented in scipy.optimize.least\_squares. 
For the pixel alignment error, we measure the distance between back-projected corresponding pixels on the corresponding planes. 
We first warp the texture of each plane segment to viewpoint normalized camera frame using the predicted plane parameters.
We then extract SIFT~\cite{lowe2004distinctive} features and match keypoints between corresponding planes. 
Incorrect matches are filtered out using RANSAC with an affine transformation.  
Theoretically the predicted keypoints of the two plane segments in a correctly normalized camera frame should satisfy a stronger relationship 
(2D rotation+translation) if plane parameters are accurate. 
However, since our plane parameters are not perfect, we relax to an affine transformation.
Finally, the corresponding pixels are back-projected to 3D using the camera intrinsics and extrinsics to 
calculate point-wise Euclidean distance. We use Eqn.~\ref{eqn:continuous} in our paper as the objective. We set $d_{\mathrm{cam}}$ to be the geodesic rotation distance
between $\hat{\RB}$ and $\RB_{\hat{k}}$ to regulate the deviation from the selected camera bin.

 \begin{table*}[t]
    \centering
    \caption{Overall architecture for our proposed network. The backbone, RPN, box, and mask branches are identical to Mask R-CNN. 
    The RPN predicts a bounding box for each of $A$ anchors in the input feature map. $C$ is the number of categories (here = 1 because we only have one ``plane" class).
    TConv is a transpose convolution with stride 2. ReLU is used between all Linear, Conv and TConv operations. Outputs of the normal branch and the embedding branch are normalized.
    Depth branch uses Conv and Deconv layers to generate a depthmap with the same resolution as the input image.
    The camera pose module is detailed in Table~\ref{tab:camera_cc}, it takes features from both images and outputs 32 logits for translation and rotation bins respectively.
    }
    \resizebox{\textwidth}{!}{
        \begin{tabular}{cclc}
            \toprule
            Index & Inputs & Operation & Output shape \\
            \midrule
            (1)	&	Inputs	&	Input Image	&	$H\times W\times3$	\\
            (2)	&	(1)	&	Backbone: ResNet50-FPN	&	$h\times w\times 256$	\\
            (3)	&	(2)	&	RPN	&	$h\times w\times A\times4$	\\
            (4)	&	(2),(3)	&	\texttt{RoIAlign}	&	$14\times14\times256$	\\
            (5)	&	(4)	&	Box: 2$\times$downsample, Flatten, $\mathrm{Linear}(7\times 7\times 256\rightarrow 1024)$, $\mathrm{Linear}(1024\rightarrow5C)$	&	$C\times5$	\\
            (6)	&	(4)	&	Mask: $4\times\mathrm{Conv}(256\rightarrow 256,3\times 3)$, $\mathrm{TConv}(256\rightarrow 256,2\times 2,2)$, $\mathrm{Conv}(256\rightarrow C, 1\times 1)$	&	$28\times 28\times C$	\\
            (7)	&	(4)	&	Normal: $4\times\mathrm{Conv}(256\rightarrow 256,3\times3)$, $\mathrm{Linear}(14\times14\times256\rightarrow1024)$, $\mathrm{Linear}(1024\rightarrow3)$	&	$C\times3$	\\
            (8)	&	(4)	&	Embedding: $4\times\mathrm{Conv}(256\rightarrow 256,3\times3)$, $\mathrm{Linear}(14\times14\times256\rightarrow1024)$, $\mathrm{Linear}(1024\rightarrow128)$	&	$C\times128$	\\
            (9)	&	(2)	&	Depth	&	$H\times W\times 1$	\\
            (10)	&	(2), (2)'	&	Camera pose module	&	$1\times32+1\times32$	\\
            \bottomrule
        \end{tabular}
    }
    \label{tab:full_arch}
\end{table*}

\begin{table}[t]
    \centering
    \caption{The architecture for \textbf{Proposed} camera pose module. It takes as input the $P_3$ features of the ResNet50-FPN backbone from both images ((2) in Table~\ref{tab:full_arch}). 
    Then six Conv layers are used to learn an appropriate image features for matching ((2)-(4)). \texttt{Maxpool} is used to reduce the feature size.
    We then compute the attention of the image features and reshape the result to $300\times 15\times 20$. 
    After six Conv layers (stride alternates between 1 and 2 to reduce feature size) and two Linear layers, we predict two multinomial distributions 
    over 32 bins for translation and rotation respectively. ReLU is used between all Linear, Conv layers.}
    \resizebox{\columnwidth}{!}{
    \begin{tabular}{cclc}
        \toprule
        Index & Inputs & Operation & Output shape \\
        \midrule
        (1)	&	Input	& $P_3$ features from ResNet50-FPN	&	$2\times 60 \times 80 \times 256$	\\
        (2)	&	(1)	&	$2\times\mathrm{Conv}(256\rightarrow 256,3\times 3)$, \texttt{Maxpool}	& $2\times 30\times 40\times 256$\\
        (3)	&	(2)	&	$2\times\mathrm{Conv}(256\rightarrow 256,3\times 3)$, \texttt{Maxpool}	& $2\times 15\times 20\times 256$	\\
        (4)	&	(3)	&	$\mathrm{Conv}(256\rightarrow 256,3\times 3)$, $\mathrm{Conv}(256\rightarrow 512,3\times 3)$	&	$2\times 15\times 20\times 512$	\\
        (5)	&	(4)	&	\texttt{Attention} &	$300\times 15\times 20$	\\
        (6)	&	(5)	&	$\mathrm{Conv}(300\rightarrow 128,3\times 3)$, $5\times\mathrm{Conv}(128\rightarrow 128,3\times 3)$,	&	$128\times 2\times 3$	\\
        (7)	&	(6)	&	$\mathrm{Linear}(128\times 2\times 3\rightarrow 64)$	&	$1\times64$	\\
        (8)	&	(7)	&	Translation: $\mathrm{Linear}(64\rightarrow 32)$	&	$1\times 32$	\\
        (9)	&	(7)	&	Rotation: $\mathrm{Linear}(64\rightarrow 32)$	&	$1\times 32$	\\
        \bottomrule
    \end{tabular}
    }
    \label{tab:camera_cc}
\end{table}

\begin{table}[t]
    \centering
    \caption{The architecture for benchmark {\em Associative3D~\cite{Qian2020} camera branch} in Section 4.4 of our paper. 
    Note this is a larger network compared to the original paper because we add as many nonlinearities as our architecture.
    It takes as input the features of the ResNet50 backbone from both images, then average pools the features and 
    passes through eight Linear layers to predict two multinomial distributions 
    over 32 bins for translation and rotation respectively.}
    \label{tab:resnet50_fc}
    \resizebox{\columnwidth}{!}{
    \begin{tabular}{cclc}
        \toprule
        Index & Inputs & Operation & Output shape \\
        \midrule
        (1)	&	Input	&	Image features	&	$2\times 15 \times 20 \times 2048$	\\
        (2)	&	(1)	&	\texttt{Avgpool}	&	$2\times 1 \times 1 \times 2048$	\\
        (3)	&	(2)	&	\texttt{Concat}	&	$1\times 1\times 4096$	\\
        (4)	&	(3)	&	$\mathrm{Linear}(4096\rightarrow 512), 5\times\mathrm{Linear}(512\rightarrow 512), \mathrm{Linear}(512\rightarrow 64)$	&	$1\times 64$	\\
        (5)	&	(4)	&	Translation: $\mathrm{Linear}(64\rightarrow 32)$	&	$1\times 32$	\\
        (6)	&	(4)	&	Rotation: $\mathrm{Linear}(64\rightarrow 32)$	&	$1\times 32$	\\
        \bottomrule
    \end{tabular}
    }
\end{table}
\begin{table}[t]
    \centering
    \caption{The architecture for benchmark {\em ResNet50-CatConv} camera branch in Section 4.4 of our paper. 
    It takes as input the features of the ResNet50 backbone from both images, concatenates the image features, 
    and then uses six Conv layers (stride alternates between 1 and 2 to reduce feature size) 
    and two Linear layers to predict two multinomial distributions 
    over 32 bins for translation and rotation respectively.}
    \label{tab:resnet50_catconv}
    \resizebox{\columnwidth}{!}{
    \begin{tabular}{cclc}
        \toprule
        Index & Inputs & Operation & Output shape \\
        \midrule
        (1)	&	Input	&	Image features	&	$2\times 15 \times 20 \times 2048$	\\
        (2)	&	(1)	&	\texttt{Concat}	&	$15\times 20\times 4096$	\\
        (3)	&	(2)	&	$\mathrm{Conv}(4096\rightarrow 128,3\times 3)$, $5\times\mathrm{Conv}(128\rightarrow 128,3\times 3)$,	&	$128\times 2\times 3$	\\
        (4)	&	(3)	&	$\mathrm{Linear}(128\times 2\times 3\rightarrow 64)$	&	$1\times 64$	\\
        (5)	&	(4)	&	Translation: $\mathrm{Linear}(64\rightarrow 32)$	&	$1\times 32$	\\
        (6)	&	(4)	&	Rotation: $\mathrm{Linear}(64\rightarrow 32)$	&	$1\times 32$	\\
        \bottomrule
    \end{tabular}
    }
\end{table}
\begin{table}[t]
    \centering
    \caption{The architecture for benchmark {\em ResNet50-Attention} camera branch in Section 4.4 of our paper. 
    It takes as input the features of the ResNet50 backbone from both images, 
    computes the attention of the image features and reshapes the result to $300\times 15\times 20$. 
    After six Conv layers (stride alternates between 1 and 2 to reduce feature size) and two Linear layers, 
    it predicts two multinomial distributions over 32 bins for translation and rotation respectively.}
    \label{tab:resnet50_cc}
    \resizebox{\columnwidth}{!}{
    \begin{tabular}{cclc}
        \toprule
        Index & Inputs & Operation & Output shape \\
        \midrule
        (1)	&	Input	&	Image features	&	$2\times 15 \times 20 \times 2048$	\\
        (2)	&	(1)	&	\texttt{Attention}	&	$300\times 15\times 20$	\\
        (3)	&	(2)	&	$\mathrm{Conv}(300\rightarrow 128,3\times 3)$, $5\times\mathrm{Conv}(128\rightarrow 128,3\times 3)$,	&	$128\times 2\times 3$	\\
        (4)	&	(3)	&	$\mathrm{Linear}(128\times 2\times 3\rightarrow 64)$	&	$1\times 64$	\\
        (5)	&	(4)	&	Translation: $\mathrm{Linear}(64\rightarrow 32)$	&	$1\times 32$	\\
        (6)	&	(4)	&	Rotation: $\mathrm{Linear}(64\rightarrow 32)$	&	$1\times 32$	\\
        \bottomrule
    \end{tabular}
    }
\end{table}

\subsection {ASNet~\cite{cai2020messytable} Baseline.}
Following the Appearance-Surrounding Network in~\cite{cai2020messytable}, 
in addition to the appearance branch, we add a surrounding branch for feature extraction to augment our embedding head.
We combine both appearance features and surrounding features to determine plane correspondences.
Instead of cropping the original image which causes one inference per detection, 
we increase the receptive field of RoIAlign by a factor of 2 to extract larger ROI features.
After RoIAlign, the features are cropped and passed to a surrounding extractor and an appearance extractor.
Both the extractors use the same number and size of Conv layers and linear layers as our embedding head uses (4 Conv layers and 1 FC layer). 
We train using cosine similarity weighted loss as described in Equation 1 and 2 of \cite{cai2020messytable}.
We ensure apples-to-apples comparisons by re-training the embedding head on the same freezed backbone our approach uses.

\end{document}